\documentclass{article} 
\usepackage{iclr2026_conference,times}

\usepackage{amsmath,amsfonts,bm}

\def\eqref#1{equation~\ref{#1}}

\def\1{\bm{1}}

\DeclareMathAlphabet{\mathsfit}{\encodingdefault}{\sfdefault}{m}{sl}
\SetMathAlphabet{\mathsfit}{bold}{\encodingdefault}{\sfdefault}{bx}{n}

\usepackage{hyperref}
\usepackage{url}

\newif\iflink
\linktrue

\newcommand{\rev}[1]{#1}

\usepackage[utf8]{inputenc} 
\usepackage[T1]{fontenc}    
\usepackage{hyperref}       
\usepackage{url}            
\usepackage{booktabs}       
\usepackage{amsfonts}       
\usepackage{nicefrac}       
\usepackage{microtype}      
\usepackage{xcolor}         

\usepackage[hyperfirst=false, nonumberlist, nostyles, nogroupskip]{glossaries}
\glsdisablehyper 
\glsunsetall
\usepackage[framemethod=TikZ]{mdframed}
\usepackage{framed}

\newcommand{\namexact}{\textsc{NAMExact}}
\newcommand{\namexacttrain}{\ensuremath{\textsc{NAMExact}_\text{train}}}
\newcommand{\namexacttest}{\ensuremath{\textsc{NAMExact}_\text{test}}}
\newcommand{\namexactval}{\ensuremath{\textsc{NAMExact}_\text{val}}}
\newcommand{\namextend}{\textsc{NAMExtend}}

\newcommand{\gradiend}{\textsc{Gradiend}}

\newcommand{\bertbase}{$\text{BERT}_\text{base}$}
\newcommand{\bertlarge}{$\text{BERT}_\text{large}$}
\newcommand{\roberta}{RoBERTa}
\newcommand{\distilbert}{DistilBERT}
\newcommand{\gpttwo}{GPT-2}
\newcommand{\llama}{LLaMA} 
\newcommand{\llamai}{\llama-Instruct}

\newcommand{\dropout}{\textsc{Dropout}}
\newcommand{\selfdebias}{\textsc{SelfDebias}}
\newcommand{\sentencedebias}{\textsc{SentDebias}}
\newcommand{\inlp}{\textsc{INLP}}
\newcommand{\leace}{\textsc{LEACE}}
\newcommand{\rlace}{\textsc{RLACE}}
\newcommand{\cda}{\textsc{CDA}}

\newcommand{\genter}{\textsc{Genter}}
\newcommand{\gentertrain}{$\genter_\text{train}$}
\newcommand{\gentertest}{$\genter_\text{test}$}
\newcommand{\genterval}{$\genter_\text{val}$}
\newcommand{\genterzero}{$\genter^0$}
\newcommand{\geneutral}{\textsc{GENeutral}}
\newcommand{\gerneutral}{\textsc{BIASneutral}}
\newcommand{\gentypes}{\textsc{GENTypes}}
\newcommand{\wikigender}{\textsc{WikiGender}}

\newcommand{\acc}{\text{Acc}}
\newcommand{\cor}{\text{Cor}}
\newcommand{\enc}{\text{Enc}}
\newcommand{\dec}{\text{Dec}}
\newcommand{\accenc}{\ensuremath{\acc_\enc}}
\newcommand{\accdec}{\ensuremath{\text{LMS}_\dec}}
\newcommand{\corenc}{\ensuremath{\cor_\enc}}
\newcommand{\traindata}{\ensuremath{\mathcal{T}}}
\newcommand{\traindatazero}{\ensuremath{\traindata\textsc{neutral}}}
\newcommand{\cormf}{\ensuremath{\cor_\traindata}}
\newcommand{\cormfval}{\ensuremath{\cor_{\traindata_{\text{val}}}}}
\newcommand{\cormftest}{\ensuremath{\cor_{\traindata_{\text{test}}}}}
\newcommand{\accmf}{\ensuremath{\acc_\traindata}}
\newcommand{\mamf}{\ensuremath{\overline{|h|}_\traindata}}
\newcommand{\masmf}{\ensuremath{\overline{|h|}_{\traindatazero}}}
\newcommand{\man}{\ensuremath{\overline{|h|}_\text{\gerneutral}}}

\newcommand{\fpi}{FemaleBS}
\newcommand{\mpi}{MaleBS}
\newcommand{\bpi}{BalancedBS}
\newcommand{\gradiendfpi}{$\text{\gradiend}_\text{Female}$}
\newcommand{\gradiendmpi}{$\text{\gradiend}_\text{Male}$}
\newcommand{\gradiendbpi}{$\text{\gradiend}_\text{Female/Male}$}
\newcommand{\gradiendraceab}{$\text{\gradiend}_\text{Asian/Black}$}
\newcommand{\gradiendraceaw}{$\text{\gradiend}_\text{Asian/White}$}
\newcommand{\gradiendracebw}{$\text{\gradiend}_\text{Black/White}$}
\newcommand{\gradiendreligioncj}{$\text{\gradiend}_\text{Christian/Jewish}$}
\newcommand{\gradiendreligioncm}{$\text{\gradiend}_\text{Christian/Muslim}$}
\newcommand{\gradiendreligionjm}{$\text{\gradiend}_\text{Jewish/Muslim}$}

\newcommand{\gradiendrace}{$\text{\gradiend}_\text{race}$}
\newcommand{\gradiendreligion}{$\text{\gradiend}_\text{religion}$}

\newacronym{gradae}{\gradiend}{GRADIent ENcoder Decoder}

\newacronym{sae}{SAE}{Sparse AutoEncoder}

\newacronym{cda}{CDA}{Counterfactual Data Augmentation}
\newacronym{inlp}{INLP}{Iterative Nullspace Projection}
\newacronym{leace}{LEACE}{LEAst-square Concept Erasure}
\newacronym{rlace}{RLACE}{Relaxed Linear Adversarial Concept Erasure}
\newacronym{mlm}{MLM}{Masked Language Modeling}
\newacronym{clm}{CLM}{Causal Language Modeling}
\newacronym{tpt}{TPT}{Token Prediction Task}

\newacronym{apd}{APD}{Average Prediction Difference}
\newacronym{bpi}{\bpi}{Balanced Bias Score}
\newacronym{mpi}{\mpi}{Male Bias Score}
\newacronym{fpi}{\fpi}{Female Bias Score}

\newacronym{genter}{\genter}{GEnder Name TEmplates with pRonouns}
\newacronym{geneutral}{\geneutral}{GEnder NEUTRAL}
\newacronym{ma}{MA}{Mean Absolute}
\newacronym{mae}{MAE}{Mean Absolute Error}

\newacronym{gentypes}{\gentypes}{Gender Stereotypes}

\newacronym{glue}{GLUE}{General Language Understanding Evaluation}
\newacronym{sglue}{SuperGLUE}{Stickier \acrshort{glue}}
\newacronym{wsc}{WSC}{Winograd Schema Challenge}
\newacronym{weat}{WEAT}{Word Embedding Association Test}
\newacronym{seat}{SEAT}{Sentence Encoder Association Test}
\newacronym{crows}{CrowS}{Crowdsourced Stereotype Pairs}
\newacronym{bbq}{BBQ}{Bias Benchmark for QA}

\newacronym{lms}{$\text{LMS}_\text{StereoSet}$}{Language Modeling Score}
\newacronym{ss}{SS}{Stereotype Score}

\newacronym{bert}{BERT}{Bidirectional Encoder Representations from Transformers}
\newacronym{llama}{\llama}{Large Language Model Meta AI}
\newacronym{ai}{AI}{Artificial Intelligence}

\newacronym{mpr}{MPR}{Mean Proportional Rank}

\renewcommand{\P}{\mathbb{P}}

\usepackage{tikz}
\usetikzlibrary{shapes.geometric, arrows, positioning, calc}
\usepackage{booktabs}  
\usepackage{siunitx}   
\usepackage{tcolorbox} 
\usepackage[inline,shortlabels]{enumitem}
\usepackage{subcaption}
\usepackage{pifont}
\definecolor{darkgreen}{RGB}{0,150,0}  

\newcommand{\cmark}{\ensuremath{\text{\ding{51}}}}
\newcommand{\xmark}{\ensuremath{\text{\ding{55}}}}
\newcommand{\variantmark}{\rev{\ *}}

\newcommand{\cmarkc}{\textcolor{darkgreen}{\cmark}}  
\newcommand{\xmarkc}{\textcolor{red}{\xmark}}    
\usepackage{marvosym}
\usepackage{mathtools}
\usepackage{amssymb}
\usepackage{wasysym}
\usepackage{stmaryrd}
\usepackage[normalem]{ulem}  
\usepackage{marginnote}
\renewcommand{\marginpar}{\marginnote}
\usepackage{colortbl}
\usepackage{tabularx}
\usepackage{bm}
\renewcommand{\UrlBreaks}{\do\/\do-} 

\definecolor{aeneuroncolor}{HTML}{E0E0E0}
\definecolor{bpicolor}{HTML}{FF0000}
\definecolor{fpicolor}{HTML}{FF00FF}
\definecolor{mpicolor}{HTML}{FFA500}
\newcommand{\csquare}[1]{\textcolor{#1}{\boldsymbol{$\square$}}}

\newcommand{\buparrow}{\uparrow}
\newcommand{\bdownarrow}{\downarrow}

\definecolor{c0}{rgb}{1.0000, 1.0000, 1.0000}
\definecolor{c1}{rgb}{0.9686, 0.9843, 1.0000}
\definecolor{c2}{rgb}{0.8825, 0.9292, 0.9724}
\definecolor{c3}{rgb}{0.7994, 0.8741, 0.9449}
\definecolor{c4}{rgb}{0.6719, 0.8144, 0.9007}
\definecolor{c5}{rgb}{0.5106, 0.7323, 0.8588}
\definecolor{c6}{rgb}{0.3465, 0.6324, 0.8107}
\definecolor{c7}{rgb}{0.2157, 0.5294, 0.7542}
\definecolor{c8}{rgb}{0.1056, 0.4126, 0.6860}
\definecolor{c9}{rgb}{0.0314, 0.3019, 0.5884}
\definecolor{c10}{rgb}{0.0314, 0.1882, 0.4196}
\definecolor{lightergray}{gray}{1.0}

\definecolor{lightgray}{rgb}{0.9, 0.9, 0.9}
\newcommand{\lightcmidrule}[1]{%
  \arrayrulecolor{lightgray}%
  \noalign{\vskip -2pt}
  \cmidrule{#1}%
  \noalign{\vskip -2pt}
  \arrayrulecolor{black}%
}

\newcommand{\tinymath}[1]{\text{\fontsize{2pt}{4.8pt}\selectfont\ensuremath{#1}}}

\newcommand{\minitext}[1]{\text{\fontsize{3pt}{4.8pt}\selectfont#1}}

\definecolor{custompink}{HTML}{FDD7D6}
\definecolor{customgray}{HTML}{E7FFDD}
\newcommand{\ua}[1]{\tcbox[colback=custompink, 
    boxrule=0.0mm, arc=1.0mm, 
    left=-0.5mm, right=-0.5mm, top=-0.8mm, bottom=-0.8mm, 
    fontupper=\tiny]{\scalebox{0.9}{\ensuremath{\uparrow}\,\minitext{#1}}}}

\newcommand{\da}[1]{\tcbox[colback=customgray, 
    boxrule=0.0mm, arc=1.0mm, 
    left=-0.5mm, right=-0.5mm, top=-0.8mm, bottom=-0.8mm, 
    fontupper=\tiny]{\scalebox{0.9}{\ensuremath{\downarrow}\,\minitext{#1}}}}

\newcommand{\uag}[1]{\tcbox[colback=customgray, 
    boxrule=0.0mm, arc=1.0mm, 
    left=-0.5mm, right=-0.5mm, top=-0.8mm, bottom=-0.8mm, 
    fontupper=\tiny]{\scalebox{0.9}{\ensuremath{\uparrow}\,\minitext{#1}}}}

\newcommand{\dab}[1]{\tcbox[colback=custompink, 
    boxrule=0.0mm, arc=1.0mm, 
    left=-0.5mm, right=-0.5mm, top=-0.8mm, bottom=-0.8mm, 
    fontupper=\tiny]{\scalebox{0.9}{\ensuremath{\downarrow}\,\minitext{#1}}}}

\newcommand{\uan}[1]{\tcbox[colback=custompink, 
    boxrule=0.0mm, arc=1.0mm, 
    left=-0.5mm, right=-0.5mm, top=-0.8mm, bottom=-0.8mm, 
    fontupper=\tiny]{\scalebox{0.9}{\ensuremath{\uparrow\,\text{#1}}}}}

\newcommand{\dan}[1]{\tcbox[colback=customgray, 
    boxrule=0.0mm, arc=1.0mm, 
    left=-0.5mm, right=-0.5mm, top=-0.8mm, bottom=-0.8mm, 
    fontupper=\tiny]{\scalebox{0.9}{\ensuremath{\downarrow\,\text{#1}}}}}

\newcommand{\uagn}[1]{\tcbox[colback=customgray, 
    boxrule=0.0mm, arc=1.0mm, 
    left=-0.5mm, right=-0.5mm, top=-0.8mm, bottom=-0.8mm, 
    fontupper=\tiny]{\scalebox{0.9}{\ensuremath{\uparrow\,\text{#1}}}}}

\newcommand{\dabn}[1]{\tcbox[colback=custompink, 
    boxrule=0.0mm, arc=1.0mm, 
    left=-0.5mm, right=-0.5mm, top=-0.8mm, bottom=-0.8mm, 
    fontupper=\tiny]{\scalebox{0.9}{\ensuremath{\downarrow\,\text{#1}}}}}

\newcommand{\tablefontsize}{\fontsize{8}{10}\selectfont}

\newcommand{\hypresult}[1]{
\begin{mdframed}[backgroundcolor=gray!5, linewidth=1.5pt, roundcorner=5pt, linecolor=gray!70]
#1
\end{mdframed}
}

\newcommand{\factualnabla}{\nabla_{\!\!\factual}}
\newcommand{\counterfactualnabla}{\nabla_{\!\!\counterfactual}\!} 
\newcommand{\diffnabla}{\nabla_{\!\!\diffsymbol}}

\newcommand{\factual}{\text{\tiny $+$}}
\newcommand{\counterfactual}{\text{\tiny $-$ }}
\newcommand{\diff}{\tiny$\pm$}
\newcommand{\diffsymbol}{\text{\diff}} 

\newcommand{\bestatfifty}{
  $\mathrel{
    \raisebox{0.8ex}{\tiny$\downarrow$}
    \mkern-6.6mu 
    \raisebox{-0.2ex}{\tiny$\uparrow$}
     \raisebox{0.2ex}{\text{\tiny{50\%}}} 
  }$
}

\newcommand{\bestatfiftytiny}{
  $\mathrel{
    \raisebox{1.2ex}{\tiny$\downarrow$}
    \mkern-8.6mu 
    \raisebox{-0.7ex}{\tiny$\uparrow$}
     \raisebox{0.2ex}{\text{\scalebox{0.7}{\tiny{50}}}} 
  }$
}

\newcommand{\bestatzero}{
  $\mathrel{
    \raisebox{0.8ex}{\tiny$\downarrow$}
    \mkern-6.6mu 
    \raisebox{-0.2ex}{\tiny$\uparrow$}
     \raisebox{0.2ex}{\text{\tiny{0.0}}} 
  }$
}

\newcommand{\bestatzerotiny}{
  $\mathrel{
    \raisebox{1.2ex}{\tiny$\downarrow$}
    \mkern-8.6mu 
    \raisebox{-0.7ex}{\tiny$\uparrow$}
     \raisebox{0.2ex}{\text{\scalebox{0.7}{\tiny{0.0}}}}
  }$
}

\newcommand{\generateModelPlotsSmallSquareFour}[3]{
\begin{figure}[#3]
    \centering
    \begin{subfigure}[t]{0.24\textwidth}
        \includegraphics[trim=0 5 5 5,clip,width=\textwidth]{img/decoder-gender/#1/subplot_avg_prob_f.pdf}
        \caption{$\mathbb{P}(F)$ {\color{bpicolor}$\uparrow$} {\color{fpicolor}$\uparrow$} {\color{mpicolor}$\downarrow$}}
        \label{fig:changed_models-main-#1:prob_f}
    \end{subfigure}
    \begin{subfigure}[t]{0.24\textwidth}
        \includegraphics[trim=0 5 5 5,clip,width=\textwidth]{img/decoder-gender/#1/subplot_avg_prob_m.pdf}
        \caption{$\mathbb{P}(M)$ {\color{bpicolor}$\uparrow$} {\color{fpicolor}$\downarrow$} {\color{mpicolor}$\uparrow$}}
        \label{fig:changed_models-main-#1:prob_m}
    \end{subfigure}
    \begin{subfigure}[t]{0.24\textwidth}
        \includegraphics[trim=0 5 5 5,clip,width=\textwidth]{img/decoder-gender/#1/subplot_accuracy.pdf}
        \caption{\accdec\ {\color{bpicolor}$\uparrow$} {\color{fpicolor}$\uparrow$} {\color{mpicolor}$\uparrow$}}
        \label{fig:changed_models-main-#1:acc}
    \end{subfigure}
    \begin{subfigure}[t]{0.24\textwidth}
        \includegraphics[trim=0 5 5 5,clip,width=\textwidth]{img/decoder-gender/#1/subplot_bpi.pdf}
        \caption{\acrshort{bpi} {\color{bpicolor}$\uparrow$}}
        \label{fig:changed_models-main-#1:bpi}
    \end{subfigure}

   \caption{Metrics for changed models based on the #2 gender \gradiend\ with varying feature factor and learning rate. The cells with the best \acrshort{bpi}~\csquare{bpicolor}, \acrshort{fpi}~\csquare{fpicolor}, and \acrshort{mpi}~\csquare{mpicolor} are highlighted across all subplots. All values are reported as percentages.
   }
   \label{fig:changed_models-main-#1}
\end{figure}
}

\newcommand{\generateRRPlotsPerModel}[2]{

\begin{figure}[ptb]
    \centering
    \begin{subfigure}[t]{0.15\textwidth}
         \includegraphics[width=\linewidth]{img/decoder/race_black_asian/#1.pdf}
         \caption{Black/Asian}
    \end{subfigure}
    \begin{subfigure}[t]{0.15\textwidth}
         \includegraphics[width=\linewidth]{img/decoder/race_white_asian/#1.pdf}
         \caption{Asian/White}
    \end{subfigure}
    \begin{subfigure}[t]{0.15\textwidth}
         \includegraphics[width=\linewidth]{img/decoder/race_white_black/#1.pdf}
         \caption{Black/White}
    \end{subfigure}
    \begin{subfigure}[t]{0.15\textwidth}
         \includegraphics[width=\linewidth]{img/decoder/religion_christian_jewish/#1.pdf}
         \caption{Chr./Jew.}
    \end{subfigure}
    \begin{subfigure}[t]{0.15\textwidth}
         \includegraphics[width=\linewidth]{img/decoder/religion_christian_muslim/#1.pdf}
         \caption{Chr./Muslim}
    \end{subfigure}
    \begin{subfigure}[t]{0.15\textwidth}
         \includegraphics[width=\linewidth]{img/decoder/religion_muslim_jewish/#1.pdf}
         \caption{Jew./Muslim}
    \end{subfigure}
    
    \caption{Metrics for changed models based on the #2 race and religion \glspl{gradae} with varying feature factor and learning rate. The cells with the best \acrshort{bpi}~\csquare{bpicolor} are highlighted across all subplots. All values are reported as percentages.}
    \label{fig:changed-model-#1}
\end{figure}

}

\title{\gradiend: Feature Learning within Neural Networks Exemplified through Biases}

\author{Antiquus S.~Hippocampus, Natalia Cerebro \& Amelie P. Amygdale \thanks{ Use footnote for providing further information
about author (webpage, alternative address)---\emph{not} for acknowledging
funding agencies.  Funding acknowledgements go at the end of the paper.} \\
Department of Computer Science\\
Cranberry-Lemon University\\
Pittsburgh, PA 15213, USA \\
\texttt{\{hippo,brain,jen\}@cs.cranberry-lemon.edu} \\
\And
Ji Q. Ren \& Yevgeny LeNet \\
Department of Computational Neuroscience \\
University of the Witwatersrand \\
Joburg, South Africa \\
\texttt{\{robot,net\}@wits.ac.za} \\
\AND
Coauthor \\
Affiliation \\
Address \\
\texttt{email}
}

\author{Jonathan Drechsel \& Steffen Herbold \\
Faculty of Computer Science and Mathematics \\
University of Passau\\
Passau, Germany \\
\texttt{\{jonathan.drechsel,steffen.herbold\}@uni-passau.de}
}

\newcommand{\fix}{\marginpar{FIX}}
\newcommand{\new}{\marginpar{NEW}}

\iclrfinalcopy 

\begin{document}

\maketitle

\begin{abstract}
AI systems frequently exhibit and amplify social biases, leading to harmful consequences in critical areas. This study introduces a novel encoder-decoder approach that leverages model gradients to learn a feature neuron encoding societal bias information such as gender, race, and religion. We show that our method can not only identify which weights of a model need to be changed to modify a feature, but even demonstrate that this can be used to rewrite models to debias them while maintaining other capabilities. We demonstrate the effectiveness of our approach across various model architectures and highlight its potential for broader applications.
\end{abstract}

\section{Introduction}

Modern \gls{ai} systems encode vast amounts of information in their internal parameters. 
Some of these parameters correspond to semantically meaningful features, such as linguistic structure or social concepts \citep{jawahar-etal-2019-bert, UnderstandingSocialReasoning}.
Understanding and controlling these features is critical for improving model interpretability, robustness, and fairness. 
While prior work has uncovered individual or groups of neurons that correlate with specific features \citep{bricken2023monosemanticity}, systematically learning targeted features remains a challenge.

We propose a novel approach to learn features in language models by leveraging gradients from a feature-related input. 
We hypothesize that these gradients contain valuable information for identifying and modifying a model's behavior related to a feature. 
Unlike existing approaches for extracting monosemantic features (e.g., \citealt{bricken2023monosemanticity}), our approach enables the learning of a feature neuron with a desired, interpretable meaning. 
The feature neuron is modeled as a bottleneck in a simple encoder-decoder architecture for model gradients. The decoder essentially learns what parts of the model needs to be updated to change a feature. 

One particularly important class of features relates to societal biases such as gender.
\gls{ai} is often seen as a neutral tool without personal preferences or biases~\citep{zmac029, jiang2024selfdisclosureaiparadoxtrust}, but it can still exhibit and even amplify bias \citep{nadeem2020gender}, with harmful impacts in crucial areas such as 
healthcare and hiring \citep{buolamwini2018gender, ferrara2023fairness}.
For instance, Amazon's \gls{ai}-powered hiring tool, trained on resumes from a male-dominated tech industry, was found to favor male candidates, penalizing resumes referencing women’s colleges
\citep{dastin2022amazon}. This underscores a crucial problem: \gls{ai} models, though seemingly neutral, can inherit and amplify real-world biases.

Recent research has explored how bias appears in language models \citep{NEMANI2024100047, BiasAndFairness}. Proposed solutions include 
specialized training \citep{cda, dropout}, 
pruning biased neurons \citep{movementPruning},
post-processing steps that adjust model outputs without modifying internal parameters \citep{inlp, SentenceDebias, selfDebias}, and methods to measure the bias \citep{seat, stereoset}. 

    This paper investigates two hypotheses:
    \begin{enumerate*}[label=\textbf{(H\arabic*)}]
        \item It is possible to learn targeted a \emph{feature} neuron from the model's gradients with a desired interpretation, such as gender (e.g., distinguishing female and male inputs).\label{item:hyp1}
        \item This feature neuron can be used to modify  model behavior related to the feature (e.g., bias) without negatively affecting other capabilities.\label{item:hyp3}
    \end{enumerate*}
    By exploring these hypotheses, we demonstrate the potential of targeted feature learning and achieve new SoTA results for gender debiasing when using \gradiend\ together with \inlp\ \citep{inlp}, evaluated against a broad set of debiasing methods and their combinations.
    Although this study focuses on gender, race, and religion bias, the proposed encoder-decoder approach is generic and should also be able to learn other features.

For clarity, in this study, \emph{gender} is treated as binary (while acknowledging and respecting non-binary gender identities). Similarly, we focus on a limited set of \emph{races} -- Asian, Black, and White -- and \emph{religions} -- Christian, Jewish, and Muslim, based on prior research \citep{meade2022empiricalsurveyeffectivenessdebiasing}.

\begin{figure}[!t]
\centering

\begin{subfigure}[t]{0.6\textwidth}
\centering
\includegraphics[width=\linewidth]{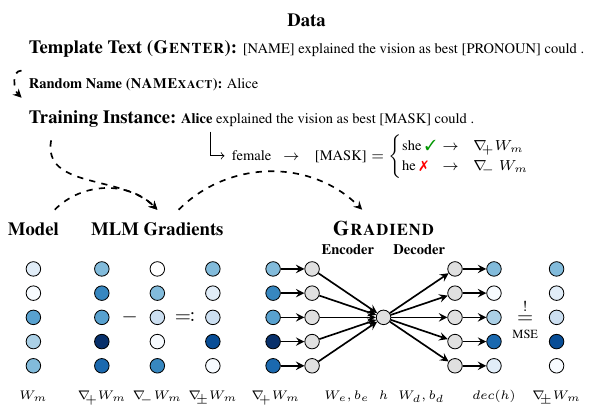}
\caption{Training Phase: Learning to encode the feature gender and how to change a model's gender bias.}
\end{subfigure}
\hfill
\begin{subfigure}[t]{0.35\textwidth}
\centering
\includegraphics[width=\linewidth]{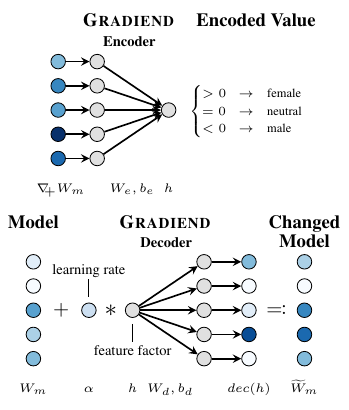}
\caption{Inference Phase: Evaluating the feature neuron and modifying gender bias.}
\end{subfigure}

\vspace{-5pt}
\caption{\acrfull{gradae} -- Targeted learning of a single scalar feature neuron using orthogonal gradient inputs, shown with an example for gender bias.}
\label{fig:gradae}

\vspace{-10pt}
\end{figure}

\section{Related Work} \label{sec:rel-work}
This section reviews interpretable feature learning and existing methods for debiasing transformer models, while additional techniques for measuring bias are discussed in Appendix~\ref{sec:measuring-gender-bias}.

\subsection{Interpretable Feature Learning}

Interpretable feature learning aims to identify and understand the internal representations of neural networks, focusing on how individual neurons or groups of neurons relate to specific concepts.
Early methods focused on visualizing learned features through saliency maps \citep{saliencyMaps} and activation maximization \citep{actMax}, highlighting the influence of inputs on model predictions.
Recent advancements focus on separating networks into semantically meaningful units like individual neurons or circuits \citep{introCircuit}.
Research on \emph{monosemantic} neurons -- those aligned with a single natural \emph{feature} -- offers clearer and more interpretable insights compared to \emph{polysemantic} ones~\citep{jermyn2022engineeringmonosemanticitytoymodels}.
\citet{bricken2023monosemanticity} proposed to learn unsupervised a \gls{sae} that extracts interpretable features in a high-dimensional feature space, which are analyzed for semantical meaning based on their behavior. Follow-up studies~\citep{anthropic} improved scalability and identified  specific features such as a gender-bias awareness feature in Claude 3 Sonnet \citep{anthropic2024claude3}. 
However, this approach requires learning numerous potential features and testing for meaningful interpretations, leaving it uncertain whether a desired feature will actually arise. Another limitation of \glspl{sae} is that they do not consider the model parameters (i.e., weights) directly, but rather only the activation of neurons. This means that rewriting of models is not directly possible and can only be achieved at inference time by changing model activations. In comparison, while we speak of learning of neurons as well, our proposed \gradiend\ method works by learning weights associated with features directly in a manner that enables rewriting and that allows us to target specific features.
\rev{Moreover, while \glspl{sae} are typically trained for a single transformer layer or even only a subset of one \citep{bricken2023monosemanticity, brinkmann-etal-2025-large}, \gradiend\ can be applied to all parameters across all layers.}

\subsection{Transformer Debiasing Techniques}\label{sec:rel-work-debiasing}

Various techniques have been proposed to mitigate bias in transformer language models (see, e.g.,~\citealt{li2023survey}), either by creating debiased models by changing weights or through post-processing adjustments.
This section introduces a subset of representative techniques relevant to this study.

\acrlong{cda} (\acrshort{cda}\glsunset{cda}; \citealt{cda, cda-orig}) 
is a straightforward method which swaps bias-related words consistently within a training corpus (e.g., replacing \emph{he}/\emph{she} for gender bias), enabling further training on a balanced dataset.
\citet{dropout} found experimentally that increasing \dropout\ during pre-training effectively reduces bias.

The \acrlong{inlp} (\acrshort{inlp}\glsunset{inlp}; \citealt{inlp}) is a post-processing debiasing method by iteratively training a linear classifier of the property to be removed (e.g., gender) based on model embeddings and subtracting the classifier's nullspace from the embeddings to remove property-related information.
Its successors, \acrshort{rlace}~\citep{rlace} and \acrshort{leace}~\citep{leace}, improve nullspace estimation 
with more compact and effective projections.
\sentencedebias\ \citep{SentenceDebias} estimates a linear subspace associated with bias by using \gls{cda} to generate sentence pairs with swapped terms (e.g., \emph{he}/\emph{she}) and debiases sentence embeddings by subtracting their projection onto this subspace.
\selfdebias~\citep{selfDebias} addresses bias in generated text by running inference with and without a bias-encouraging prefix, downweighting tokens favored in the biased version. 
However, this approach is unsuitable for downstream tasks like \acrshort{glue} \citep{glue}.
In Section~\ref{sec:eval-debiased}, we compare our method with the other debiasing techniques and their combinations on \acrshort{glue} and on \acrshort{sglue} \citep{superglue}, extending prior work focused on \acrshort{glue}.

\section{Methodology}\label{sec:methodology}

We introduce a novel approach for targeted feature learning and bias modification. 
Our method utilizes a simple encoder-decoder architecture that leverages gradient information to encode a gender-related scalar value. 
This scalar is then decoded into gradient updates, which are used to adjust the model’s bias toward the encoded feature value. 
An overview of the approach is illustrated in Figure~\ref{fig:gradae}.

\subsection{Motivation}\label{sec:gradae-motivation}

Gradient-based explanation methods, such as Grad-CAM \citep{gradcam} and Integrated Gradients \citep{integratedGradients}, have proven effective in providing insights into a model's internal workings \citep{chen2020adapting, selvaraju2020grad, lundstrom2022rigorous}, highlighting which parts of the model were crucial to a specific prediction.
During the training of neural networks, the optimizer inherently determines which neurons require updates, specifically those that contributed incorrectly to the model's output.
We leverage this mechanism through a \gls{tpt} whose masked token is sensitive to a chosen feature (e.g., gender, race, religion). For encoder-only models, we use \acrlong{mlm} (\acrshort{mlm}\glsunset{mlm}; \citealt{bert}), and for decoder-only models, we use \acrlong{clm} (\acrshort{clm}\glsunset{clm}; \citealt{radford2018improving}). 
For clarity, the following explanations focus on the \gls{mlm} variant, with details on adapting the task to \gls{clm} (e.g., using only left-side context before the [MASK]) provided in Appendix~\ref{app:tpt-generative}.

To illustrate, consider the binary gender case. Suppose we have a sentence where the masked token refers to a gendered pronoun determined by a name, e.g., ``\emph{Alice explained the vision as best [MASK] could .}''. Here, \emph{she} is the \emph{factual} target (consistent with the context), while \emph{he} serves as the \emph{counterfactual} target. 
For features with more than two classes, the counterfactual notion naturally generalizes to an \emph{orthogonal} target: any instance of the same feature that differs from the factual one (e.g., another race or religion) can serve as an alternative target. 

By using factual-orthogonal evaluations for two feature classes, gradient differences are computed to isolate feature-related updates by eliminating non-feature-related changes common to both cases. 
This difference yields two inverse directions: strengthening or mitigating bias with respect to the chosen feature classes), depending on the gradient order. 
In the mitigating direction, the factual feature-related updates are eliminated, effectively removing the established factual associations, while the orthogonal updates are emphasized to facilitate the learning of new, orthogonal associations.

\subsection{\gradiend}\label{sec:gradae}

In general, we aim to learn how to adjust model parameters to achieve a desired factual or orthogonal state. We hypothesize that the gradients contain the necessary information for this purpose and that the feature changing behavior can be controlled via a learned neuron.

Let a feature be represented by $d\ge 2$ orthogonal classes $\mathcal{C} = \{C_1, \dots, C_d\}$. 
For training, we select two distinct classes $A,B \in \mathcal{C}$ and consider \glspl{tpt} where the masked token corresponds to either $A$ (factual $A$, orthogonal $B$) or to $B$ (factual $B$, orthogonal $A$).

Let $W_m \in \mathbb{R}^n$ denote the $n$ model parameters for which the feature is learned. 

For an example with factual class $C \in \{A,B\}$ and orthogonal class $C' \in \{A,B\}\setminus \{C\}$, we define three types of gradients:
\begin{enumerate*}[label=\textbf{(\arabic*)}]
    \item gradients from the factual masking task~$\factualnabla W_m$ (i.e., the target belongs to $C$),
    \item gradients from the orthogonal masking task~$\counterfactualnabla W_m$ (i.e., the target belongs to $C'$), and
    \item the difference between these two gradients $\diffnabla W_m \coloneqq \factualnabla W_m - \counterfactualnabla W_m$.
\end{enumerate*}
Here, $\nabla_{\!.}W_m$ represents a vector in $\mathbb{R}^n$, where each component corresponds to the gradient for the parameter at this position.  
We frame the problem as a gradient learning task to predict the gradient difference $\diffnabla W_m$ from the factual gradients $\factualnabla W_m$:
\begin{equation*}
    \text{Learn } f  \text{ s.t. } f(\factualnabla W_m) \approx \diffnabla W_m.
\end{equation*}
For this study, we propose a simple encoder-decoder structure $f = dec \circ enc$, where:
\begin{align*}
    enc(\factualnabla W_m) &= \tanh(W_e^T \cdot \factualnabla W_m + b_e) && \hspace{-60pt} \eqqcolon h \in \mathbb{R}, \\
    dec(h) &= h \cdot W_d + b_d && \hspace{-60pt} \approx \diffnabla W_m.
\end{align*}
Here, $W_e, W_d, b_d\in \mathbb{R}^n$ and $b_e\in \mathbb{R}$ are learnable parameters, resulting in a total of $3n+1$ parameters. We refer to this approach as \gls{gradae}.

\subsection{\gradiend\ for Debiasing}\label{sec:gradae-for-gender-debiasing}
While \gradiend\ is defined for orthogonal class pairs of any feature, we restrict the following proof of concept to the bias types gender, race, and religion.
Gender is treated binary in this study ($d=2$; $C_1=Female$ and $C_2=Male$), while race ($C_1=Asian$, $C_2=Black$, and $C_3=White$) and religion ($C_1=Christian$, $C_2=Jewish$, and $C_3=Muslim$) are considered with $d=3$ classes.

In this setup, \rev{hypothesis \ref{item:hyp1}} suggests that the factual and counterfactual masking tasks guide the encoder to produce a feature-related scalar $h$, representing the orthogonal axis between two chosen classes $A$ and $B$.  
\rev{Hypothesis \ref{item:hyp3}} asserts that $dec(h)$ can adjust the model's bias along this orthogonal axis, e.g., by choosing a specific \emph{feature factor} $h$ and \emph{learning rate} $\alpha$ to update the model parameters as follows:
\begin{equation}
    \widetilde{W}_m \coloneqq W_m + \alpha \cdot dec(h).\label{equ:change}
\end{equation}
Experiments show that feature-related inputs are mostly mapped to values close to $-1$ and $+1$, corresponding to the classes $A$ and $B$ or vice versa. 
WLOG, we assume $A$ and $B$ are ordered lexicographically and that positive values of $h$ represent $A$ while negative values represent $B$. This post-hoc standardization enables consistent definitions and visualizations across experiments.

\section{Data}

For each bias type, we filter existing datasets to derive masked texts where the mask corresponds to the bias target terms. 
For gender, these targets are the pronouns \emph{he}/\emph{she}, determined solely by the gender of a preceding name.
We augment a BookCorpus-derived dataset \citep{bookcorpus} using names as templates to diversify the model gradients, and filter texts where gender could be inferred from other words.
For race and religion, we follow a simplified procedure similar to \cite{meade2022empiricalsurveyeffectivenessdebiasing} using \acrshort{cda}: 
\rev{From English Wikipedia, we retain only sentences that contain one of their predefined bias-attribute words (e.g., \emph{Jewish}, \emph{African}). These attribute words are then masked to generate bias-specific gradients.}
This produces a dataset for each pair of race or religion classes, treating one as factual and the other as orthogonal. 
Combining both directions for a pair yields the training dataset for that pair. 
For brevity, we denote by $\mathcal{T}$ the dataset associated with a particular \gradiend\ instance.
To evaluate language modeling performance independently of bias, we create \gerneutral, a BookCorpus subset without bias target words. 
Full dataset generation details are in Appendix~\ref{app:data}.

\section{Experiments}\label{sec:evaluation}

In this section, we evaluate \glspl{gradae} based on seven base models: \bertbase\ and \bertlarge\ \citep{bert}, \roberta\ \citep{roberta}, \mbox{\distilbert} \citep{distilbert}, \gpttwo\ \citep{radford2019language}, and two LLaMA-3.2-3B  models \citep{grattafiori2024llama} -- one plain (\llama) and one instruction fine-tuned (\llamai), covering a broad range of transformer variants. All datasets $\mathcal{T}$ are split into training, validation, and test sets. Metrics are reported for the test split (or the entire dataset if not split), unless stated otherwise.


\subsection{Training}\label{sec:eval-training}

Each training step processes a batch of \glspl{tpt} with a target class chosen uniformly at random, ensuring that only gradients for that single target contribute to the \gradiend\ input within a training step.
To ensure that debiasing affects the language model itself and not just the token prediction head, we exclude the prediction layers from the set of \gradiend\ parameters \rev{(i.e., the \acrshort{mlm} and \acrshort{clm} heads),} while using all \rev{other weights, including the embeddings and the attention and MLP weights of every transformer layer.}
Implementation details, hyperparameters, and initialization are described in Appendix~\ref{app:training}.

\subsection{Feature Encoder}\label{sec:eval-encoder}

We evaluate whether the \glspl{gradae} encode the intended feature \rev{(hypothesis \ref{item:hyp1})} by analyzing their encoder outputs on 
\begin{enumerate*}[label=\textbf{(\arabic*)}]
    \item training-like data (i.e., same target tokens as seen during training) and \item neutral data (i.e., tokens unseen in training and unrelated to the feature).
\end{enumerate*}
We expect training tokens to yield consistent encodings near $\pm1$ (due to the $\tanh$ activation), and neutral tokens to map near $0$, as the natural midpoint between the class extremes.

\begin{figure}[!t]
    \centering
    \includegraphics[width=\linewidth]{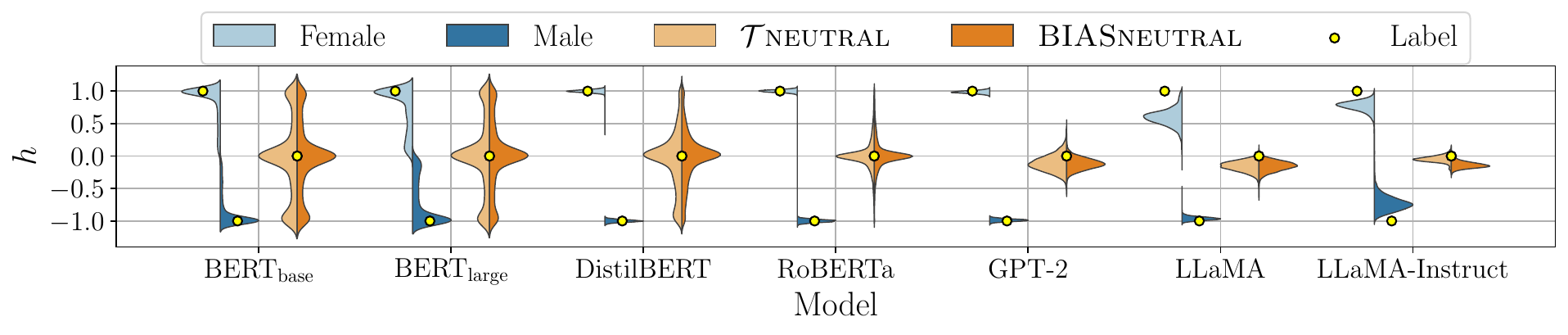}
    
    \vspace{-5pt}
    \caption{Distribution of encoded values for all gender \gradiend\ models across different datasets. The yellow dots indicate the expected label used for \corenc.}
    \label{fig:encoded-values}
    \vspace{-5pt}
\end{figure}

\begin{figure}[!t]
    \centering
    \begin{subfigure}[t]{\textwidth}
        \includegraphics[width=\linewidth]{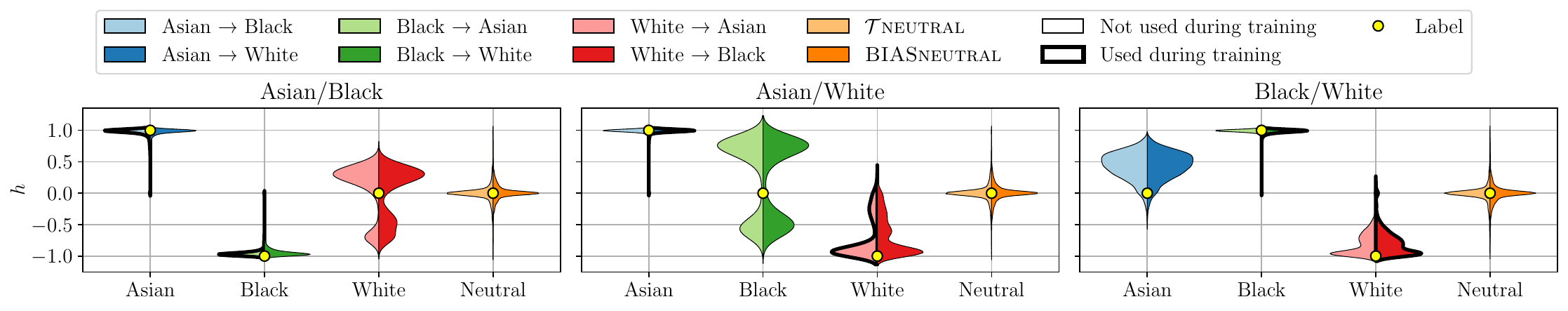}
        \caption{Race (see Figure~\ref{fig:race-encoded-values} for all models).}\label{fig:encoded-values-race}
    \end{subfigure}

    \begin{subfigure}[t]{\textwidth}
        \includegraphics[width=\linewidth]{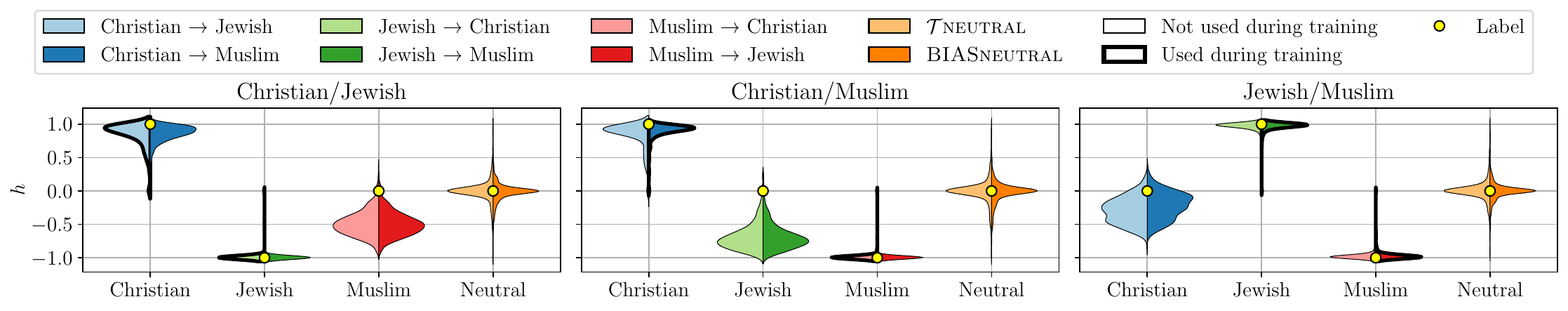}
        \caption{Religion (see Figure~\ref{fig:religion-encoded-values} for all models).}\label{fig:encode-values-religion}
    \end{subfigure}
    
    \vspace{-5pt}
    \caption{Distribution of encoded values for different datasets of the \bertbase\ \gls{gradae} models for race and religion. The yellow dots indicate the expected label used for \corenc.}
    \label{fig:encoded-values-race-religion}
    \vspace{-5pt}
\end{figure}

Figure~\ref{fig:encoded-values} shows the encoded values for gender across all models, while Figure~\ref{fig:encoded-values-race-religion} presents results for race and religion for \bertbase\ (other models \rev{and ablation studies on gender feature stability and data/token variability are in Appendix~\ref{app:encoder}}). For evaluation, we use the \traindata\ test split to capture feature-related gradients, and \traindatazero\ where feature unrelated tokens are masked in the same sentences as \traindata. We also include the independently derived neutral dataset \gerneutral. For race and religion, training data from other classes are additionally reused for evaluation as well (e.g., \emph{Asian $\to$ Black} for an Asian/White model). Within each evaluation, all subsets are balanced by downsampling to the size of the smallest split.

Across all models, encoders successfully separate the two training classes, while neutral tokens tend to cluster around $0$, though this classification is less precise for some \glspl{gradae}. 
Importantly, the neutral masks were not seen during training, showing that the encoder did not only learn a binary feature, but rather a polar one, with opposite ends of the polar scale used during training.

The behavior on unseen classes further reveals interesting biases. For example, the Black/White models often resemble a White vs.\ Non-White distinction, possibly reflecting imbalances towards White dominated data during their pretraining (Figure~\ref{fig:encoded-values-race}). Similarly, the religion models suggest that Judaism and Islam are encoded as more similar to each other than to Christianity (Figure~\ref{fig:encode-values-religion}). 

\begin{table}[!t]
    \centering
    \scriptsize
    \caption{Pearson correlation between encoded values and labels of Figures~\ref{fig:encoded-values} and \ref{fig:rr-encoded-values}. All values are scaled by 100. Best values per column are printed in \textbf{bold}.}
    \label{tab:encoded-values}
    \setlength{\tabcolsep}{4pt}
    \begin{tabular}{l rr rr rr rr rr rr rr rr}
    \toprule
          & \multicolumn{2}{c}{\textbf{Gender}} & \multicolumn{6}{c}{\textbf{Race}} & \multicolumn{6}{c}{\textbf{Religion}} & \multicolumn{2}{c}{\textbf{Mean}} \\
\cmidrule(lr){2-3} \cmidrule(lr){4-9} \cmidrule(lr){10-15} \cmidrule(lr){16-17}
          & \multicolumn{2}{c}{\tiny{\textbf{Female/Male}}} & \multicolumn{2}{c}{\tiny{\textbf{Asian/Black}}}  & \multicolumn{2}{c}{\tiny{\textbf{Asian/White}}} & \multicolumn{2}{c}{\tiny{\textbf{Black/White}}} & \multicolumn{2}{c}{\tiny{\textbf{Christ./Jew.}}} & \multicolumn{2}{c}{\tiny{\textbf{Christ./Mus.}}} & \multicolumn{2}{c}{\tiny{\textbf{Jew./Muslim}}} \\
          \cmidrule(lr){2-3} \cmidrule(lr){4-5} \cmidrule(lr){6-7} \cmidrule(lr){8-9} \cmidrule(lr){10-11} \cmidrule(lr){12-13} \cmidrule(lr){14-15} 
          & \tiny{$\cormf$} 
          & \!\!\!\!\! \tiny{$\corenc$} 
          &   \tiny{$\cormf$} 
          & \!\!\!\!\! \tiny{$\corenc$} 
          &  \tiny{$\cormf$} 
          & \!\!\!\!\! \tiny{$\corenc$} 
          &  \tiny{$\cormf$} 
          & \!\!\!\!\! \tiny{$\corenc$} 
          &  \tiny{$\cormf$} 
          & \!\!\!\!\! \tiny{$\corenc$} 
          & \tiny{$\cormf$} 
          & \!\!\!\!\! \tiny{$\corenc$} 
          &  \tiny{$\cormf$} 
          & \!\!\!\!\! \tiny{$\corenc$} 
          & \tiny{$\cormf$} 
          & \!\!\!\!\! \tiny{$\corenc$}  \\
          \midrule
  \bertbase  &      95.7      &     71.3      &     99.6      &     94.2      &     96.3      &     84.4      & \textbf{98.6} & \textbf{92.3} &     98.6      &     92.2      &     99.4      &     88.2      &     99.5      &     96.0      &     98.2      &     88.4      \\
 \bertlarge  &      90.8      &     66.0      &     98.2      & \textbf{94.6} &     96.7      &     89.1      &     96.5      &     92.0      &     97.2      &     92.8      &     98.4      &     91.8      &     98.8      &     96.6      &     96.7      &     89.0      \\
 \distilbert & \textbf{100.0} &     86.0      & \textbf{99.7} &     92.4      &     96.2      &     80.7      &     98.5      &     88.2      &     98.9      &     91.5      & \textbf{99.6} &     90.0      & \textbf{99.6} &     94.9      & \textbf{98.9} &     89.1      \\
  \roberta   & \textbf{100.0} &     95.3      &     96.2      &     83.6      &     95.6      &     82.7      &     98.0      &     85.4      & \textbf{99.5} &     92.6      &     99.5      &     90.8      &     97.8      &     94.0      &     98.1      &     89.2      \\
   \gpttwo   & \textbf{100.0} & \textbf{98.4} &     97.8      &     87.5      & \textbf{98.5} & \textbf{91.8} &     98.3      &     84.7      &     98.4      & \textbf{97.1} &     98.6      & \textbf{96.2} &     99.2      & \textbf{98.9} &     98.7      & \textbf{93.5} \\
   \llama    &      99.3      &     98.3      &     90.1      &     79.9      &     88.4      &     78.8      &     88.4      &     78.1      &     89.0      &     79.0      &     78.6      &     72.3      &     82.1      &     73.8      &     88.0      &     80.0      \\
   \llama-I.\!\!\!\!   &      99.0      &     97.6      &     89.7      &     73.6      &     87.7      &     63.7      &     84.8      &     72.4      &     90.3      &     80.4      &     71.4      &     60.0      &     86.3      &     71.0      &     87.0      &     74.1      \\ \midrule
    Mean     &      97.8      &     87.5      &     95.9      &     86.5      &     94.2      &     81.6      &     94.7      &     84.7      &     96.0      &     89.4      &     92.2      &     84.2      &     94.8      &     89.3      &     95.1      &     86.2      \\
         \bottomrule
    \end{tabular}
\end{table}

Table~\ref{tab:encoded-values} quantifies these findings by reporting Pearson correlations \citep{pearson} for the training-like data (\cormf; only $\pm1$ labels) and for all evaluations shown in Figures~\ref{fig:encoded-values} and \ref{fig:rr-encoded-values} (\corenc; including neutral labels of $0$). 
All models achieve strong performance on \cormf\ for gender, but \llama-based models perform noticeably worse for race and religion, likely due to their larger tokenizer: gender targets (\emph{he/she}) remain single tokens, whereas many race and religion targets are split into multiple tokens, unlike in smaller models where most targets are single-tokenized (see Appendix~\ref{app:tpt-generative}). 
\gpttwo\ performs best overall, particularly on the generalization metric \corenc, mapping neutral inputs reliably near $0$. 
The most challenging distinction for religion is \emph{Christian/Muslim}, reflecting their greater textual overlap and semantic similarity, consistent with prior studies \citep{religionsComparitive}.



\hypresult{The \gradiend\ models consistently learn interpretable feature neurons, mapping target classes to $\pm1$ and neutral input mostly near $0$, thereby supporting \rev{hypothesis \ref{item:hyp1}}.}

\subsection{Decoder as Bias-Changer}\label{sec:eval-decoder}


We investigate how the learned representation of the decoder can change model bias.
The model adjustment is controlled by two parameters: the scalar input to the decoder network $h$ (\emph{feature factor}) and the \emph{learning rate} $\alpha$, which scales the decoder output before adding it to the model weights.
To assess the impact of these parameters, we evaluate the \gradiend\ models across a grid of 15 feature factors and 16 learning rates, modifying the model weights as $\widetilde{W_m} \coloneqq W_m + \alpha \cdot dec(h)$.

For the resulting models, we require three key properties:
\begin{enumerate*}[label=\textbf{(\arabic*)}]
\item Their overall language modeling performance should remain close to the original model. \label{item:lms}
\item They should assign balanced probabilities to tokens from both classes $A$ and $B$. \label{item:token}
\item Both $A$ and $B$ should retain sufficiently high probabilities to avoid trivial solutions (e.g., collapsing to near-zero). \label{item:sum}
\end{enumerate*}

To measure \ref{item:lms}, we compute a language modeling score \accdec\ based on \gls{mlm} accuracy for encoder-only models and perplexity for decoder-only models on \gerneutral, ensuring independence from bias-related terms. 
For \ref{item:token}, we evaluate a single \gls{tpt} by summing probabilities of all expected tokens for each class to approximate $\P(A)$ and $\P(B)$, and then averaging across multiple \glspl{tpt}. The goal is to minimize their difference while enforcing a large overall sum due to \ref{item:sum}. Multiplying these scores together yields a \gls{bpi}, and the best-scoring configuration across the parameter grid is selected as the modified model, denoted \emph{BaseModel \,+ $\gradiend_{A/B}$}.
We also use the same framework to construct explicitly gender-biased variants to further study the capabilities of our approach. A \gls{fpi} is defined to favor female bias, enforcing high \accdec, low $\P(F)$, and high $\P(M)$. Conversely, \gls{mpi} does the opposite for $\P(F)$ and $\P(M)$. These metrics yield \emph{BaseModel\,+$\gradiend_\text{Female}$} and \emph{BaseModel\,+$\gradiend_\text{Male}$}, respectively.
Precise metric definitions are given in Appendix~\ref{app:decoder-as-bias-changer}.

\generateModelPlotsSmallSquareFour{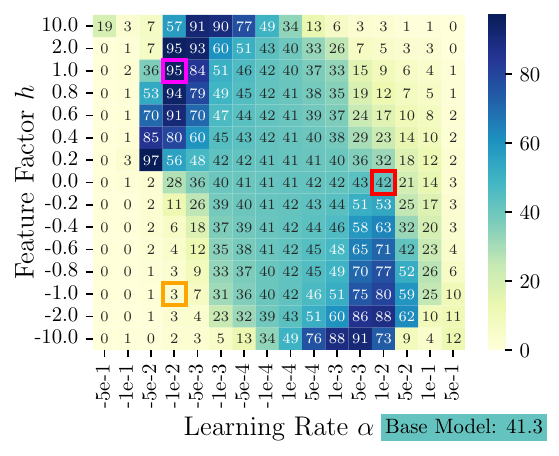}{\bertbase}{!t}
While Figure~\ref{fig:changed_models-main-bert-base-cased} focuses on the selected \bertbase\ models for gender, other models show a similar overall behavior (see Appendix~\ref{app:decoder-as-bias-changer}). All selected models for gender, race, and religion are further evaluated for debiasing performance in Section~\ref{sec:eval-debiased}.
Interestingly, all plots exhibit a nearly point-symmetric behavior. 
This effect arises from the linear structure of the \gradiend\ decoder, which computes $dec(h)=h\cdot W_d + b_d$. When comparing configurations $(h, \alpha)$ and $(-h, -\alpha)$, the resulting difference in weight update is:
\begin{align*}
    \big[W_m + \alpha\cdot dec(h)\big] - \big[(W_m + (-\alpha)\cdot dec(-h)\big] 
    &= \alpha\cdot (dec(h)+dec(-h)) \\ & = \alpha \big[\big(h\cdot W_d + b_d\big) + \big(-h\cdot W_d + b_d \big)\big] \\ &= 2\alpha b_d.
\end{align*}
Thus, the only difference is due to the decoder's bias term $b_d$, scaled by $2\alpha$. 
Further, as $h$ increases, the term $h\cdot W_d$ dominates in the weight update, reducing the relative impact of $b_d$, and thereby enhancing the symmetry. Conversely, the symmetry breaks for small $|h|$ or large $|\alpha|$.

Specifically, $\P(F)$ and $\P(M)$ (Figures~\ref{fig:changed_models-main-bert-base-cased:prob_f} and \ref{fig:changed_models-main-bert-base-cased:prob_m}) show an inverse pattern.
Due to the encoder normalization and the definition of $\diffnabla W_m$ (Section~\ref{sec:gradae}), when the signs of $h$ and $\alpha$ are equal, the model biases consistently toward male, whereas opposite signs bias toward female.
\accdec\ (Figure~\ref{fig:changed_models-main-bert-base-cased:acc}) reveals a broad region of high probability for moderate learning rates, while
Figure~\ref{fig:changed_models-main-bert-base-cased:bpi} 
illustrates the optimal models for \gls{bpi}. 
These plots capture the inherent trade-offs of the debiasing approach \citep{movementPruning}: stronger bias modification can degrade language modeling, but a \emph{safe region} exists with moderate feature factors and learning rates.
Considering the \gls{bpi} plot (Figure~\ref{fig:changed_models-main-bert-base-cased:bpi}) and feature factor $h=0.0$, the \gradiend\ decoder's bias vector $b_e$ effectively learned an appropriate debiasing direction. 
Although not shown in Figure~\ref{fig:changed_models-main-bert-base-cased}, the highlighted selected cells for \gls{fpi} and \gls{mpi} (see Figure~\ref{fig:model-selection-gender-bert-base}) confirm that the method can also enforce strongly female- or male-biased models, yielding extreme values of $\P(F)$ and $\P(M)$.

\subsection{Comparison to Other Debiasing Techniques}\label{sec:eval-debiased}

We compare the \gradiend-modified models alongside up to seven debiasing approaches (see Section~\ref{sec:rel-work-debiasing}). 
We hypothesize that combining debiasing methods improves debiasing, and for gender, we also evaluate hybrid approaches that pair weight-modifying methods (\cda, \dropout, and \gradiendbpi) with post-processing methods (\inlp, \sentencedebias). 
 
We evaluate on two established bias metrics: \acrshort{ss}\glsunset{ss} \citep{stereoset}, which compares stereotypical and anti-stereotypical predictions, and \acrshort{seat}\glsunset{seat} \citep{seat}, comparing embedding associations between bias attributes and stereotypical terms. Both are detailed in  Appendix~\ref{sec:measuring-gender-bias}.
As debiasing can harm language modeling \citep{movementPruning}, we report \gls{lms} \citep{stereoset} capturing language modeling without fine-tuning, alongside the established NLP benchmarks 
\acrshort{glue} \citep{glue} and \acrshort{sglue} \citep{superglue}.

\begin{table}[!t]
\centering
\fontsize{8}{10}\selectfont
\scriptsize
\caption{Mean proportional ranks for 
\gls{ss}/ \gls{seat}, and mean relative change in \gls{lms}/ \acrshort{glue}/ \acrshort{sglue} vs.\ the base model.
Models are sorted by the \emph{Mean} column.
$\Delta W$ and PP indicate model weight modification and post-processing, respectively.
Best variant type is marked with a blue \textcolor{blue}{\cmark}. 
\rev{Variants marked with\variantmark\ use only non-\llama\ models, making absolute language modeling scores less comparable, but relative differences (averaged model-wise score difference) remain meaningful.}
}

\label{tab:rank}
    \begin{tabular}{lccrrrrrr}
\toprule
\multicolumn{3}{c}{\!\!\!\textbf{Variant}\!} & \multicolumn{3}{c}{\!\!\!\textbf{Prop. Rank Bias}} &
\multicolumn{3}{c}{\textbf{Language Modeling}} \\ 
\cmidrule(r){1-3} \cmidrule(r){4-6} \cmidrule(r){7-9}  
\textbf{Name} & \!\!\!\!\!\!$\mathbf{\Delta W}$ & \!\!\!\!\!\!\!\!\textbf{PP} & \!\!\!\!\textbf{Mean} $\uparrow$ &  \!\!\!\!\textbf{\gls{ss}} & \!\!\!\!\!\! \textbf{\gls{seat}} & 
\!\!\textbf{\acrshort{lms}} (\%) & \textbf{\acrshort{glue}} (\%)  & \textbf{\acrshort{sglue}} (\%) \\ 
\midrule
\multicolumn{9}{c}{\textbf{Gender} (full results in Tables~\ref{tab:eval:main-results-bert-base} and \ref{tab:eval:main-results-gpt})}  \\
\midrule

\gradiendbpi\ + \inlp & \textcolor{blue}{\cmark} & \textcolor{blue}{\cmark} & 0.88 & \textbf{0.91} & \textbf{0.84} & \dabn{-0.39} 87.06 & \dabn{-0.47} 68.23 & \dabn{-1.72} 50.65 \\
\cda\, + \inlp\variantmark & \cmark & \cmark & 0.75 & 0.78 & 0.73 & \uagn{0.97} 86.48 & \uagn{0.36} 77.55 & \uagn{1.86} 52.67 \\
\dropout \, + \inlp\variantmark & \cmark & \cmark & 0.71 & 0.78 & 0.64 & \dabn{-1.09} 84.42 & \dabn{-2.43} 74.75 & \dabn{-0.80} 50.01 \\
\inlp & \xmark & \textcolor{blue}{\cmark} & 0.67 & 0.62 & 0.72 & \uagn{0.10} 87.56 & \uagn{0.13} 68.83 & \dabn{-0.82} 51.55 \\
\gradiendbpi\ + \sentencedebias \!\!\! & \cmark & \cmark & 0.64 & 0.67 & 0.61 & \dabn{-1.12} 86.34 & \dabn{-0.92} 67.78 & \dabn{-0.83} 51.54 \\
\dropout \, + \sentencedebias\variantmark & \cmark & \cmark & 0.62 & 0.70 & 0.55 & \dabn{-3.25} 82.27 & \dabn{-2.25} 74.93 & \dabn{-0.21} 50.60 \\
\sentencedebias & \xmark & \cmark & 0.60 & 0.48 & 0.72 & \dabn{-0.52} 86.94 & \dabn{-0.44} 68.27 & \dabn{-0.08} 52.29 \\
\cda\, + \sentencedebias\variantmark & \cmark & \cmark & 0.57 & 0.71 & 0.43 & \uagn{0.01} 85.52 & \uagn{0.50} 77.68 & \uagn{1.25} 52.06 \\
\gradiendbpi & \textcolor{blue}{\cmark} & \xmark & 0.46 & 0.50 & 0.42 & \dabn{-0.73} 86.72 & \dabn{-0.00} 68.70 & \dabn{-0.63} 51.73 \\
\cda\variantmark & \cmark & \xmark & 0.44 & 0.42 & 0.45 & \uagn{0.23} 85.74 & \uagn{0.45} 77.64 & \uagn{1.37} 52.18 \\
\selfdebias & \xmark & \cmark & 0.41 & 0.41 & -- & \dabn{-9.65} 77.81 & -- & -- \\
\leace & \xmark & \cmark & 0.36 & 0.32 & 0.41 & \dabn{-0.49} 86.97 & \uagn{0.01} 68.71 & \dabn{-1.71} 50.66 \\
\gradiendfpi & \cmark & \xmark & 0.36 & 0.51 & 0.21 & \dabn{-0.75} 86.71 & \dabn{-0.09} 68.61 & \uagn{0.41} 52.78 \\
\gradiendmpi & \cmark & \xmark & 0.32 & 0.19 & 0.44 & \dabn{-0.33} 87.13 & \uagn{0.94} 69.64 & \dabn{-0.35} 52.02 \\
\rlace & \xmark & \cmark & 0.31 & 0.21 & 0.40 & \dabn{-2.19} 85.26 & \dabn{-0.06} 68.64 & \dabn{-1.85} 50.51 \\
\dropout\variantmark & \cmark & \xmark & 0.30 & 0.40 & 0.20 & \dabn{-2.11} 83.40 & \dabn{-3.09} 74.10 & \dabn{-0.42} 50.39 \\
Base Model & \xmark & \xmark & 0.17 & 0.11 & 0.23 & 87.46 & 68.70 & 52.37 \\

\midrule
\multicolumn{9}{c}{\textbf{Race} (full results in Table~\ref{tab:results-race})}  \\
\midrule

\selfdebias & \xmark & \textcolor{blue}{\cmark} & 0.87 & \textbf{0.87} & -- & \dabn{-1.24} 86.22 & -- & -- \\
\gradiendraceaw & \textcolor{blue}{\cmark} & \xmark & 0.58 & 0.79 & 0.36 & \dabn{-5.45} 82.00 & \dabn{-2.76} 65.94 & \dabn{-2.39} 49.98 \\
\sentencedebias & \xmark & \cmark & 0.55 & 0.49 & 0.61 & \dabn{-0.06} 87.40 & \dabn{-0.39} 68.31 & \uagn{0.16} 52.53 \\
\dropout\variantmark & \cmark & \xmark & 0.54 & 0.57 & 0.51 & \dabn{-2.11} 83.40 & \dabn{-3.09} 74.10 & \dabn{-0.42} 50.39 \\
\inlp & \xmark & \cmark & 0.46 & 0.29 & \textbf{0.64} & \dabn{-0.07} 87.39 & \uagn{0.33} 69.03 & \uagn{0.13} 52.50 \\
\cda\variantmark & \cmark & \xmark & 0.44 & 0.25 & 0.63 & \dabn{-1.61} 83.91 & \dabn{-0.07} 77.11 & \uagn{1.47} 52.28 \\
\gradiendraceab & \cmark & \xmark & 0.44 & 0.62 & 0.25 & \dabn{-8.14} 79.32 & \dabn{-2.79} 65.92 & \dabn{-3.40} 48.96 \\
Base Model & \xmark & \xmark & 0.44 & 0.24 & \textbf{0.64} & 87.46 & 68.70 & 52.37 \\
\gradiendracebw & \cmark & \xmark & 0.36 & 0.32 & 0.40 & \dabn{-0.09} 87.37 & \dabn{-0.95} 67.75 & \uagn{0.27} 52.64 \\

\midrule
\multicolumn{9}{c}{\textbf{Religion} (full results in Table~\ref{tab:results-religion})}  \\
\midrule

\selfdebias & \xmark & \textcolor{blue}{\cmark} & 0.70 & \textbf{0.70} & -- & \dabn{-9.60} 77.86 & -- & -- \\
\sentencedebias & \xmark & \cmark & 0.64 & 0.65 & 0.62 & \dabn{-0.17} 87.29 & \dabn{-0.10} 68.60 & \dabn{-0.00} 52.36 \\
\cda\variantmark & \textcolor{blue}{\cmark} & \xmark & 0.58 & 0.33 & \textbf{0.83} & \dabn{-1.00} 84.52 & \uagn{0.72} 77.91 & \uagn{1.98} 52.79 \\
\inlp & \xmark & \cmark & 0.54 & 0.39 & 0.70 & \dabn{-0.35} 87.10 & \dabn{-0.25} 68.45 & \uagn{0.04} 52.41 \\
\dropout\variantmark & \cmark & \xmark & 0.54 & 0.47 & 0.60 & \dabn{-2.11} 83.40 & \dabn{-3.09} 74.10 & \dabn{-0.42} 50.39 \\
\gradiendreligioncj & \cmark & \xmark & 0.44 & 0.46 & 0.43 & \dabn{-0.38} 87.07 & \dabn{-2.16} 66.54 & \uagn{0.38} 52.75 \\
\gradiendreligioncm & \cmark & \xmark & 0.44 & 0.61 & 0.27 & \dabn{-2.70} 84.76 & \dabn{-0.75} 67.95 & \dabn{-0.02} 52.35 \\
\gradiendreligionjm & \cmark & \xmark & 0.42 & 0.59 & 0.25 & \dabn{-0.78} 86.68 & \uagn{0.39} 69.09 & \uagn{0.14} 52.51 \\
Base Model & \xmark & \xmark & 0.33 & 0.24 & 0.42 & 87.46 & 68.70 & 52.37 \\

\bottomrule
\end{tabular}
\vspace{-10pt}
\end{table}

Detailed results per base model, including bootstrapping intervals \citep{davison1997bootstrap}, can be found in Appendix~\ref{app:comp}. 
As noted in prior work \citep{meade2022empiricalsurveyeffectivenessdebiasing}, 
comparing debiasing approaches is challenging due to sometimes inconsistent performance across models and metrics. 
To address this, we compute an aggregated debias score by ranking each approach based on its proportional rank in \gls{ss} and \gls{seat} averaged across all seven base models.
Table~\ref{tab:rank} reports these ranks alongside average changes in the language modeling metrics relative to the original model.

\subsubsection{Gender Debiasing}

Among the single approaches, \gradiendbpi\ (9th) is the most effective weight-modifying ($\Delta W$) approach.
Notably, such weight-modified models can be integrated into standard downstream implementations, unlike post-processing (PP) methods, which, despite generally stronger performance (e.g., \inlp, 4th), require customized handling.
The best overall results are achieved by combinations, with \gradiendbpi\ + \inlp\ clearly outperforming all other methods, followed by \gradiendbpi\ + \sentencedebias.
This supports the intuition that combining debiasing techniques can enhance the debiasing effectiveness of individual methods. 
Nevertheless, strong single approaches like \sentencedebias\ still outperform some combinations.

\gradiendfpi\ and \gradiendmpi\ are designed to be female and male-biased models, yet their performance is only slightly below \gradiendbpi\ and comparable to \selfdebias.
We confirmed that all three \gradiend\ variants align with their intended behaviors in some examples (see Appendix~\ref{app:example-predictions}).
Notably, the base models themselves are ranked last with a notable gap, i.e., each debiasing approach leads to an actual less biased model according to the utilized debiasing metrics. 

\subsubsection{Race and Religion Debiasing}

Debiasing race and religion is substantially harder than gender. Base models achieve high proportional ranks, and most techniques yield only marginal or even bias-strengthening effects. In particular, no method yields statistically significant \gls{seat} improvements, and for race, the base model outperforms all debiasing methods on average.
\selfdebias\ performs best overall for race and religion, but is evaluated only on the apparently easier \gls{ss} metric and with degraded language modeling for religion.
Weight-modification methods like \gradiendraceaw\ and \dropout\ improve bias metrics but degrade language modeling performance.


Although \gradiend\ does not achieve top scores in aggregated proportional ranks, it is the only weight-modification method with statistically significant improvements for race and religion, while not significantly harming language modeling for some specific models, e.g., {\mbox{\gpttwo}\,+\gradiendraceab} and {\roberta\,+\gradiendreligioncm} (see Appendix~\ref{app:comp}). Moreover, since \gradiend\ only targets a single bias (e.g., \gradiendracebw\ does not target \emph{Asian}), full debiasing cannot be expected. Considering that we also did not control the data as carefully (see Appendix~\ref{app:training-data-rr}) as for gender (e.g., controlling for other word meanings like the name Christian vs. the religion Christian or the actual color vs. race associated terms), this explains the differences to the better performance at the gender debiasing. Thus, without strict controls for training data, \gradiend\ is still reliable for the identification of features, but we suggest strong controls when models should be rewritten.

\subsubsection{Overall Results}

Across all bias types, \accdec\ generally declines under debiasing, but fine-tuned performance on \acrshort{glue} and \acrshort{sglue} often remains stable. 
No method fully eliminates bias across metrics, underscoring the difficulty of the task.

\hypresult{The \gradiend\ decoder can effectively modify 
bias \rev{(hypothesis \ref{item:hyp3})}. For gender, it achieves SoTA performance among weight-modification methods. For race and religion, weaker averaged results likely stem from noisier training data and the restriction to a single debiasing axis.}

\section{Limitations and Open Questions}\label{sec:limitations}

While we have demonstrated \gradiend's effectiveness as a proof of concept for learning bias-related features and modifying model behavior, our study has focused primarily on pairs of orthogonal feature classes. Studying how a model can be debiased along multiple axes simultaneously is a natural next step, either by iterative training of partial debiased models along orthogonal axes or combined multidimensional \gradiend\ training.
\rev{Furthermore, using multiple feature neurons even for a single axis could improve debiasing, as a single feature neuron enforces strong compression and may limit expressivity.}
In addition, it is unclear how well the method generalizes to continuous features, such as sentiment scores. 
Moreover, the current framework should be extended to support multi-token targets for \gls{clm} (Appendix~\ref{app:tpt-generative})\rev{, e.g., by iteratively computing single-token gradients for each token individually and averaging them to derive inputs for \gradiend.}

Beyond these technical constraints,  questions remain regarding interpretability. For example, comparing the most relevant bias neurons across all race and religion gradients, or conducting neuron-level analyses in multilingual settings could reveal deeper insights into internal model representations.

\section{Ethical Statement}\label{sec:ethical}

Our study explores both debiasing and deliberate amplification of binary gender associations in language models, which -- while valuable for analysis 
-- poses risks if misapplied to reinforce stereotypes.  
We emphasize that the considered bias classes are simplifications chosen for methodological clarity and do not reflect the full diversity and complexity of gender, race, or religion in society.

\section{Conclusion}

We present a novel approach that achieves two key objectives: \begin{enumerate*}[label=(\arabic*)] \item learning a feature for the desired interpretation along an orthogonal axis based on model gradients, and \item implementing a debiasing technique to reduce a feature-related bias in transformer language models.
\end{enumerate*}
In contrast to most existing debiasing methods, our approach allows for modifying an already trained, biased model to create a truly less biased version.
This approach is built on a simple encoder-decoder architecture, \gradiend, featuring a single hidden neuron. 
The model learns to encode a feature in an unsupervised manner, using gradients from a specific token prediction training task. 
We successfully applied this method to various transformer model architectures, showing its wide applicability. 
The code is publicly available at \url{https://github.com/aieng-lab/gradiend-bias}. 

\bibliography{main}

@misc{UCI_gender_by_name,
  title        = {{Gender by Name}},
  year         = {2020},
  howpublished = {UCI Machine Learning Repository},
url={https://doi.org/10.24432/C55G7X},
}

@inproceedings{meade2022empiricalsurveyeffectivenessdebiasing,
    title = "An Empirical Survey of the Effectiveness of Debiasing Techniques for Pre-trained Language Models",
    author = "Meade, Nicholas  and Poole-Dayan, Elinor  and Reddy, Siva",
    booktitle = "Proceedings of the 60th Annual Meeting of the Association for Computational Linguistics (Volume 1: Long Papers)",
    month = may,
    year = "2022",
    address = "Dublin, Ireland",
    publisher = "Association for Computational Linguistics",
    url = "https://aclanthology.org/2022.acl-long.132",
    doi = "10.18653/v1/2022.acl-long.132",
    pages = "1878--1898",
}

@misc{gradcam,
      title={Grad-CAM: Why did you say that?}, 
      author={Ramprasaath R Selvaraju and Abhishek Das and Ramakrishna Vedantam and Michael Cogswell and Devi Parikh and Dhruv Batra},
      year={2017},
      eprint={1611.07450},
      archivePrefix={arXiv},
      primaryClass={stat.ML},
      url={https://arxiv.org/abs/1611.07450}, 
}

@inproceedings{integratedGradients,
author = {Sundararajan, Mukund and Taly, Ankur and Yan, Qiqi},
title = {Axiomatic attribution for deep networks},
year = {2017},
publisher = {JMLR.org},
booktitle = {Proceedings of the 34th International Conference on Machine Learning - Volume 70},
pages = {3319–3328},
numpages = {10},
location = {Sydney, NSW, Australia},
series = {ICML'17}
}

@inproceedings{glue,
    title = "{GLUE}: A Multi-Task Benchmark and Analysis Platform for Natural Language Understanding",
    author = "Wang, Alex  and
      Singh, Amanpreet  and
      Michael, Julian  and
      Hill, Felix  and
      Levy, Omer  and
      Bowman, Samuel",
    editor = "Linzen, Tal  and
      Chrupa{\l}a, Grzegorz  and
      Alishahi, Afra",
    booktitle = "Proceedings of the 2018 {EMNLP} Workshop {B}lackbox{NLP}: Analyzing and Interpreting Neural Networks for {NLP}",
    month = nov,
    year = "2018",
    address = "Brussels, Belgium",
    publisher = "Association for Computational Linguistics",
    url = "https://aclanthology.org/W18-5446/",
    doi = "10.18653/v1/W18-5446",
    pages = "353--355"
}

@inproceedings{seat,
    title = "On Measuring Social Biases in Sentence Encoders",
    author = "May, Chandler  and
      Wang, Alex  and
      Bordia, Shikha  and
      Bowman, Samuel R.  and
      Rudinger, Rachel",
    editor = "Burstein, Jill  and
      Doran, Christy  and
      Solorio, Thamar",
    booktitle = "Proceedings of the 2019 Conference of the North {A}merican Chapter of the Association for Computational Linguistics: Human Language Technologies, Volume 1 (Long and Short Papers)",
    month = jun,
    year = "2019",
    address = "Minneapolis, Minnesota",
    publisher = "Association for Computational Linguistics",
    url = "https://aclanthology.org/N19-1063/",
    doi = "10.18653/v1/N19-1063",
    pages = "622--628",
}

@article{weat,
    author = {Aylin Caliskan  and Joanna J. Bryson  and Arvind Narayanan },
    title = {Semantics derived automatically from language corpora contain human-like biases},
    journal = {Science},
    volume = {356},
    number = {6334},
    pages = {183-186},
    year = {2017},
    doi = {10.1126/science.aal4230},
    URL = {https://www.science.org/doi/abs/10.1126/science.aal4230},
    eprint = {https://www.science.org/doi/pdf/10.1126/science.aal4230}
}

@inproceedings{crows,
    title = "{C}row{S}-Pairs: A Challenge Dataset for Measuring Social Biases in Masked Language Models",
    author = "Nangia, Nikita  and
      Vania, Clara  and
      Bhalerao, Rasika  and
      Bowman, Samuel R.",
    editor = "Webber, Bonnie  and
      Cohn, Trevor  and
      He, Yulan  and
      Liu, Yang",
    booktitle = "Proceedings of the 2020 Conference on Empirical Methods in Natural Language Processing (EMNLP)",
    month = nov,
    year = "2020",
    address = "Online",
    publisher = "Association for Computational Linguistics",
    url = "https://aclanthology.org/2020.emnlp-main.154/",
    doi = "10.18653/v1/2020.emnlp-main.154",
    pages = "1953--1967"
}

@inproceedings{chen2020adapting,
  title={Adapting {G}rad-{CAM} for embedding networks},
  author={Chen, Lei and Chen, Jianhui and Hajimirsadeghi, Hossein and Mori, Greg},
  booktitle={proceedings of the IEEE/CVF winter conference on applications of computer vision},
  pages={2794--2803},
  year={2020},
  eprint={2001.06538},
  archivePrefix={arXiv},
  primaryClass={cs.CV},
  url={https://arxiv.org/abs/2001.06538}, 
}

@article{selvaraju2020grad,
author = {Selvaraju, Ramprasaath R. and Cogswell, Michael and Das, Abhishek and Vedantam, Ramakrishna and Parikh, Devi and Batra, Dhruv},
title = {Grad-CAM: Visual Explanations from Deep Networks via Gradient-Based Localization},
year = {2020},
issue_date = {Feb 2020},
publisher = {Kluwer Academic Publishers},
address = {USA},
volume = {128},
number = {2},
issn = {0920-5691},
url = {https://doi.org/10.1007/s11263-019-01228-7},
doi = {10.1007/s11263-019-01228-7},
journal = {Int. J. Comput. Vision},
month = feb,
pages = {336–359},
numpages = {24}
}

@inproceedings{lundstrom2022rigorous,
  title={A rigorous study of integrated gradients method and extensions to internal neuron attributions},
  author={Lundstrom, Daniel D and Huang, Tianjian and Razaviyayn, Meisam},
  booktitle={International Conference on Machine Learning},
  pages={14485--14508},
  year={2022},
  organization={PMLR},
doi = {10.48550/arXiv.2202.11912},
}

@article{bricken2023monosemanticity,
   title={Towards Monosemanticity: Decomposing Language Models With Dictionary Learning},
   author={Bricken, Trenton and Templeton, Adly and Batson, Joshua and Chen, Brian and Jermyn, Adam and Conerly, Tom and Turner, Nick and Anil, Cem and Denison, Carson and Askell, Amanda and Lasenby, Robert and Wu, Yifan and Kravec, Shauna and Schiefer, Nicholas and Maxwell, Tim and Joseph, Nicholas and Hatfield-Dodds, Zac and Tamkin, Alex and Nguyen, Karina and McLean, Brayden and Burke, Josiah E and Hume, Tristan and Carter, Shan and Henighan, Tom and Olah, Christopher},
   year={2023},
   journal={Transformer Circuits Thread},
   url={https://transformer-circuits.pub/2023/monosemantic-features/index.html}
}

@INPROCEEDINGS {bookcorpus,
author = { Zhu, Yukun and Kiros, Ryan and Zemel, Rich and Salakhutdinov, Ruslan and Urtasun, Raquel and Torralba, Antonio and Fidler, Sanja },
booktitle = { 2015 IEEE International Conference on Computer Vision (ICCV) },
title = {{ Aligning Books and Movies: Towards Story-Like Visual Explanations by Watching Movies and Reading Books }},
year = {2015},
volume = {},
ISSN = {2380-7504},
pages = {19-27},
doi = {10.1109/ICCV.2015.11},
url = {https://doi.ieeecomputersociety.org/10.1109/ICCV.2015.11},
publisher = {IEEE Computer Society},
address = {Los Alamitos, CA, USA},
month =Dec}

@inproceedings{cda,
    title = "Counterfactual Data Augmentation for Mitigating Gender Stereotypes in Languages with Rich Morphology",
    author = "Zmigrod, Ran  and
      Mielke, Sabrina J.  and
      Wallach, Hanna  and
      Cotterell, Ryan",
    editor = "Korhonen, Anna  and
      Traum, David  and
      M{\`a}rquez, Llu{\'\i}s",
    booktitle = "Proceedings of the 57th Annual Meeting of the Association for Computational Linguistics",
    month = jul,
    year = "2019",
    address = "Florence, Italy",
    publisher = "Association for Computational Linguistics",
    url = "https://aclanthology.org/P19-1161",
    doi = "10.18653/v1/P19-1161",
    pages = "1651--1661",
}

@article{movementPruning,
      title={Gender Biases and Where to Find Them: Exploring Gender Bias in Pre-Trained Transformer-based Language Models Using Movement Pruning}, 
      author={Przemyslaw Joniak and Akiko Aizawa},
      year={2022},
      eprint={2207.02463},
      archivePrefix={arXiv},
      primaryClass={cs.CL},
      url={https://arxiv.org/abs/2207.02463}, 
journal={arXiv preprint arXiv:2207.02463},
}

@article{dropout,
      title={Measuring and Reducing Gendered Correlations in Pre-trained Models}, 
      author={Kellie Webster and Xuezhi Wang and Ian Tenney and Alex Beutel and Emily Pitler and Ellie Pavlick and Jilin Chen and Ed Chi and Slav Petrov},
      year={2021},
      eprint={2010.06032},
      archivePrefix={arXiv},
      primaryClass={cs.CL},
      url={https://arxiv.org/abs/2010.06032}, 
journal={arXiv preprint arXiv:2010.06032},
}

@article{selfDebias,
    author = {Schick, Timo and Udupa, Sahana and Schütze, Hinrich},
    title = "{Self-Diagnosis and Self-Debiasing: A Proposal for Reducing Corpus-Based Bias in NLP}",
    journal = {Transactions of the Association for Computational Linguistics},
    volume = {9},
    pages = {1408-1424},
    year = {2021},
    month = {12},
    issn = {2307-387X},
    doi = {10.1162/tacl_a_00434},
    url = {https://doi.org/10.1162/tacl\_a\_00434},
    eprint = {https://direct.mit.edu/tacl/article-pdf/doi/10.1162/tacl\_a\_00434/1979270/tacl\_a\_00434.pdf},
}

@article{radford2019language,
  title={Language models are unsupervised multitask learners},
  author={Radford, Alec and Wu, Jeffrey and Child, Rewon and Luan, David and Amodei, Dario and Sutskever, Ilya and others},
  journal={OpenAI blog},
  volume={1},
  number={8},
  pages={9},
  year={2019},
  url={https://cdn.openai.com/better-language-models/language_models_are_unsupervised_multitask_learners.pdf}
}

@article{grattafiori2024llama,
  title={The llama 3 herd of models},
  author={Grattafiori, Aaron and Dubey, Abhimanyu and Jauhri, Abhinav and Pandey, Abhinav and Kadian, Abhishek and Al-Dahle, Ahmad and Letman, Aiesha and Mathur, Akhil and Schelten, Alan and Vaughan, Alex and others},
  journal={arXiv preprint arXiv:2407.21783},
  year={2024}
}

@article{li2023survey,
  title={A survey on fairness in large language models},
  author={Li, Yingji and Du, Mengnan and Song, Rui and Wang, Xin and Wang, Ying},
  journal={arXiv preprint arXiv:2308.10149},
  year={2023},
url={https://arxiv.org/pdf/2308.10149}
}

@InProceedings{rlace,
  title = 	 {Linear Adversarial Concept Erasure},
  author =       {Ravfogel, Shauli and Twiton, Michael and Goldberg, Yoav and Cotterell, Ryan D},
  booktitle = 	 {Proceedings of the 39th International Conference on Machine Learning},
  pages = 	 {18400--18421},
  year = 	 {2022},
  editor = 	 {Chaudhuri, Kamalika and Jegelka, Stefanie and Song, Le and Szepesvari, Csaba and Niu, Gang and Sabato, Sivan},
  volume = 	 {162},
  series = 	 {Proceedings of Machine Learning Research},
  month = 	 {17--23 Jul},
  publisher =    {PMLR},
  url = 	 {https://proceedings.mlr.press/v162/ravfogel22a.html}
}

@inproceedings{UnderstandingSocialReasoning,
author = {Gandhi, Kanishk and Fr\"{a}nken, J.-Philipp and Gerstenberg, Tobias and Goodman, Noah D.},
title = {Understanding social reasoning in language models with language models},
year = {2023},
publisher = {Curran Associates Inc.},
address = {Red Hook, NY, USA},
booktitle = {Proceedings of the 37th International Conference on Neural Information Processing Systems},
articleno = {595},
numpages = {12},
location = {New Orleans, LA, USA},
series = {NIPS '23}
}

@inproceedings{wsc,
author = {Levesque, Hector J. and Davis, Ernest and Morgenstern, Leora},
title = {The Winograd schema challenge},
year = {2012},
isbn = {9781577355601},
publisher = {AAAI Press},
booktitle = {Proceedings of the Thirteenth International Conference on Principles of Knowledge Representation and Reasoning},
pages = {552–561},
numpages = {10},
location = {Rome, Italy},
series = {KR'12}
}

@misc{openai2024gpt4osystemcard,
  title        = {GPT-4o System Card},
  author       = {{OpenAI}},
  year         = {2024},
  howpublished = {\url{https://arxiv.org/abs/2410.21276}},
  note         = {arXiv preprint arXiv:2410.21276, accessed 2025-11-17}
}

@inproceedings{blodgett-etal-2021-stereotyping,
    title = "Stereotyping {N}orwegian Salmon: An Inventory of Pitfalls in Fairness Benchmark Datasets",
    author = "Blodgett, Su Lin  and
      Lopez, Gilsinia  and
      Olteanu, Alexandra  and
      Sim, Robert  and
      Wallach, Hanna",
    editor = "Zong, Chengqing  and
      Xia, Fei  and
      Li, Wenjie  and
      Navigli, Roberto",
    booktitle = "Proceedings of the 59th Annual Meeting of the Association for Computational Linguistics and the 11th International Joint Conference on Natural Language Processing (Volume 1: Long Papers)",
    month = aug,
    year = "2021",
    address = "Online",
    publisher = "Association for Computational Linguistics",
    url = "https://aclanthology.org/2021.acl-long.81/",
    doi = "10.18653/v1/2021.acl-long.81",
    pages = "1004--1015"
}

@inproceedings{jawahar-etal-2019-bert,
    title = "What Does {BERT} Learn about the Structure of Language?",
    author = "Jawahar, Ganesh  and
      Sagot, Beno{\^i}t  and
      Seddah, Djam{\'e}",
    editor = "Korhonen, Anna  and
      Traum, David  and
      M{\`a}rquez, Llu{\'i}s",
    booktitle = "Proceedings of the 57th Annual Meeting of the Association for Computational Linguistics",
    month = jul,
    year = "2019",
    address = "Florence, Italy",
    publisher = "Association for Computational Linguistics",
    url = "https://aclanthology.org/P19-1356/",
    doi = "10.18653/v1/P19-1356",
    pages = "3651--3657"
}

@inproceedings{religionsComparitive,
    title = "Comparative Analysis of Religious Texts: {NLP} Approaches to the {B}ible, {Q}uran, and Bhagavad Gita",
    author = "A.\!D.\ Mahit Nandan and
      Godbole, Ishan  and
      M Kapparad, Pranav  and
      Bhattacharjee, Shrutilipi",
    editor = "Yagi, Sane  and
      Yagi, Sane  and
      Sawalha, Majdi  and
      Shawar, Bayan Abu  and
      AlShdaifat, Abdallah T.  and
      Abbas, Norhan  and
      Organizers",
    booktitle = "Proceedings of the New Horizons in Computational Linguistics for Religious Texts",
    month = jan,
    year = "2025",
    address = "Abu Dhabi, UAE",
    publisher = "Association for Computational Linguistics",
    url = "https://aclanthology.org/2025.clrel-1.1/",
    pages = "1--10"
}

@inbook{superglue,
author = {Wang, Alex and Pruksachatkun, Yada and Nangia, Nikita and Singh, Amanpreet and Michael, Julian and Hill, Felix and Levy, Omer and Bowman, Samuel R.},
title = {SuperGLUE: a stickier benchmark for general-purpose language understanding systems},
year = {2019},
publisher = {Curran Associates Inc.},
address = {Red Hook, NY, USA},
booktitle = {Proceedings of the 33rd International Conference on Neural Information Processing Systems},
articleno = {294},
numpages = {15}
}

@inproceedings{leace,
author = {Belrose, Nora and Schneider-Joseph, David and Ravfogel, Shauli and Cotterell, Ryan and Raff, Edward and Biderman, Stella},
title = {LEACE: perfect linear concept erasure in closed form},
year = {2023},
publisher = {Curran Associates Inc.},
address = {Red Hook, NY, USA},
booktitle = {Proceedings of the 37th International Conference on Neural Information Processing Systems},
articleno = {2884},
numpages = {20},
location = {New Orleans, LA, USA},
series = {NIPS '23}
}

@inproceedings{SentenceDebias,
    title = "Towards Debiasing Sentence Representations",
    author = "Liang, Paul Pu  and
      Li, Irene Mengze  and
      Zheng, Emily  and
      Lim, Yao Chong  and
      Salakhutdinov, Ruslan  and
      Morency, Louis-Philippe",
    editor = "Jurafsky, Dan  and
      Chai, Joyce  and
      Schluter, Natalie  and
      Tetreault, Joel",
    booktitle = "Proceedings of the 58th Annual Meeting of the Association for Computational Linguistics",
    month = jul,
    year = "2020",
    address = "Online",
    publisher = "Association for Computational Linguistics",
    url = "https://aclanthology.org/2020.acl-main.488",
    doi = "10.18653/v1/2020.acl-main.488",
    pages = "5502--5515",
}

@misc{eval-harness,
  author       = {Gao, Leo and Tow, Jonathan and Abbasi, Baber and Biderman, Stella and Black, Sid and DiPofi, Anthony and Foster, Charles and Golding, Laurence and Hsu, Jeffrey and Le Noac'h, Alain and Li, Haonan and McDonell, Kyle and Muennighoff, Niklas and Ociepa, Chris and Phang, Jason and Reynolds, Laria and Schoelkopf, Hailey and Skowron, Aviya and Sutawika, Lintang and Tang, Eric and Thite, Anish and Wang, Ben and Wang, Kevin and Zou, Andy},
  title        = {The Language Model Evaluation Harness},
  month        = 07,
  year         = 2024,
  publisher    = {Zenodo},
  version      = {v0.4.3},
  doi          = {10.5281/zenodo.12608602},
  url          = {https://zenodo.org/records/12608602}
}

@article{zmac029,
    author = {Jones-Jang, S Mo and Park, Yong Jin},
    title = {How do people react to AI failure? Automation bias, algorithmic aversion, and perceived controllability},
    journal = {Journal of Computer-Mediated Communication},
    volume = {28},
    number = {1},
    pages = {zmac029},
    year = {2022},
    month = {11},
    issn = {1083-6101},
    doi = {10.1093/jcmc/zmac029},
    url = {https://doi.org/10.1093/jcmc/zmac029},
    eprint = {https://academic.oup.com/jcmc/article-pdf/28/1/zmac029/47050128/zmac029.pdf},
}

@article{jiang2024selfdisclosureaiparadoxtrust,
      title={Self-Disclosure to AI: The Paradox of Trust and Vulnerability in Human-Machine Interactions}, 
      author={Zoe Zhiqiu Jiang},
      year={2024},
      eprint={2412.20564},
      archivePrefix={arXiv},
      primaryClass={cs.HC},
      url={https://arxiv.org/abs/2412.20564}, 
journal={arXiv preprint arXiv:2412.20564},

}

@inproceedings{inlp,
    title = "Null It Out: Guarding Protected Attributes by Iterative Nullspace Projection",
    author = "Ravfogel, Shauli  and
      Elazar, Yanai  and
      Gonen, Hila  and
      Twiton, Michael  and
      Goldberg, Yoav",
    editor = "Jurafsky, Dan  and
      Chai, Joyce  and
      Schluter, Natalie  and
      Tetreault, Joel",
    booktitle = "Proceedings of the 58th Annual Meeting of the Association for Computational Linguistics",
    month = jul,
    year = "2020",
    address = "Online",
    publisher = "Association for Computational Linguistics",
    url = "https://aclanthology.org/2020.acl-main.647",
    doi = "10.18653/v1/2020.acl-main.647",
    pages = "7237--7256",
}

@online{wikipedia,
  author = {{Wikimedia Foundation}},
  title = {Wikimedia Wikipedia Dataset},
  howpublished = {\url{https://huggingface.co/datasets/wikimedia/wikipedia}},
  note = {Version: “20231101.en”},  
  year = {2023},
  url = {"https://dumps.wikimedia.org"}
}

@inproceedings{brinkmann-etal-2025-large,
    title = "Large Language Models Share Representations of Latent Grammatical Concepts Across Typologically Diverse Languages",
    author = "Brinkmann, Jannik  and
      Wendler, Chris  and
      Bartelt, Christian  and
      Mueller, Aaron",
    editor = "Chiruzzo, Luis  and
      Ritter, Alan  and
      Wang, Lu",
    booktitle = "Proceedings of the 2025 Conference of the Nations of the Americas Chapter of the Association for Computational Linguistics: Human Language Technologies (Volume 1: Long Papers)",
    month = apr,
    year = "2025",
    address = "Albuquerque, New Mexico",
    publisher = "Association for Computational Linguistics",
    url = "https://aclanthology.org/2025.naacl-long.312/",
    doi = "10.18653/v1/2025.naacl-long.312",
    pages = "6131--6150",
    ISBN = "979-8-89176-189-6"
}

@misc{li2021detectinggenderbiastransformerbased,
      title={Detecting Gender Bias in Transformer-based Models: A Case Study on {BERT}}, 
      author={Bingbing Li and Hongwu Peng and Rajat Sainju and Junhuan Yang and Lei Yang and Yueying Liang and Weiwen Jiang and Binghui Wang and Hang Liu and Caiwen Ding},
      year={2021},
      eprint={2110.15733},
      archivePrefix={arXiv},
      primaryClass={cs.CL},
}

@article{pearson,
    title={Pearson correlation coefficient},
    author={Cohen, Israel and Huang, Yiteng and Chen, Jingdong and Benesty, Jacob and Benesty, Jacob and Chen, Jingdong and Huang, Yiteng and Cohen, Israel},
    year = {2009},
    month = {04},
    pages = {1-4},
    volume = {2},
    isbn = {978-3-642-00295-3},
    journal = {Noise Reduction in Speech Processing},
    doi = {10.1007/978-3-642-00296-0_5}
}

@book{davison1997bootstrap, 
place={Cambridge}, 
series={Cambridge Series in Statistical and Probabilistic Mathematics}, 
title={Bootstrap Methods and their Application}, 
publisher={Cambridge University Press}, 
author={Davison, A. C. and Hinkley, D. V.}, 
year={1997}, 
collection={Cambridge Series in Statistical and Probabilistic Mathematics},
doi={10.1017/CBO9780511802843}
}

@article{bert,
  author       = {Jacob Devlin and
                  Ming{-}Wei Chang and
                  Kenton Lee and
                  Kristina Toutanova},
  title        = {{BERT:} Pre-training of Deep Bidirectional Transformers for Language
                  Understanding},
  journal      = {CoRR},
  volume       = {abs/1810.04805},
  year         = {2018},
  url          = {http://arxiv.org/abs/1810.04805},
  eprinttype    = {arXiv},
  eprint       = {1810.04805},
  bibsource    = {dblp computer science bibliography, https://dblp.org}
}

@article{roberta,
  author       = {Yinhan Liu and
                  Myle Ott and
                  Naman Goyal and
                  Jingfei Du and
                  Mandar Joshi and
                  Danqi Chen and
                  Omer Levy and
                  Mike Lewis and
                  Luke Zettlemoyer and
                  Veselin Stoyanov},
  title        = {RoBERTa: {A} Robustly Optimized {BERT} Pretraining Approach},
  journal      = {CoRR},
  volume       = {abs/1907.11692},
  year         = {2019},
  url          = {http://arxiv.org/abs/1907.11692},
  eprinttype    = {arXiv},
  eprint       = {1907.11692},
  bibsource    = {dblp computer science bibliography, https://dblp.org}
}

@article{distilbert,
  author       = {Victor Sanh and
                  Lysandre Debut and
                  Julien Chaumond and
                  Thomas Wolf},
  title        = {DistilBERT, a distilled version of {BERT:} smaller, faster, cheaper
                  and lighter},
  journal      = {CoRR},
  volume       = {abs/1910.01108},
  year         = {2019},
  url          = {http://arxiv.org/abs/1910.01108},
  eprinttype    = {arXiv},
  eprint       = {1910.01108},
  bibsource    = {dblp computer science bibliography, https://dblp.org}
}

@article{anthropic,
   title={Scaling Monosemanticity: Extracting Interpretable Features from {C}laude 3 {S}onnet},
   author={Templeton, Adly and Conerly, Tom and Marcus, Jonathan and Lindsey, Jack and Bricken, Trenton and Chen, Brian and Pearce, Adam and Citro, Craig and Ameisen, Emmanuel and Jones, Andy and Cunningham, Hoagy and Turner, Nicholas L and McDougall, Callum and MacDiarmid, Monte and Freeman, C. Daniel and Sumers, Theodore R. and Rees, Edward and Batson, Joshua and Jermyn, Adam and Carter, Shan and Olah, Chris and Henighan, Tom},
   year={2024},
   journal={Transformer Circuits Thread},
   url={https://transformer-circuits.pub/2024/scaling-monosemanticity/index.html}
}

@misc{anthropic2024claude3,
  author       = {Anthropic},
  title        = {The {C}laude 3 Model Family: {O}pus, {S}onnet, {H}aiku},
  year         = {2024},
  howpublished = {\url{https://paperswithcode.com/paper/the-claude-3-model-family-opus-sonnet-haiku}},
  note         = {Accessed: 2024-12-12}
}

@inproceedings{saliencyMaps,
  author       = {Karen Simonyan and
                  Andrea Vedaldi and
                  Andrew Zisserman},
  editor       = {Yoshua Bengio and
                  Yann LeCun},
  title        = {Deep Inside Convolutional Networks: Visualising Image Classification
                  Models and Saliency Maps},
  booktitle    = {2nd International Conference on Learning Representations, {ICLR} 2014,
                  Banff, AB, Canada, April 14-16, 2014, Workshop Track Proceedings},
  year         = {2014},
  url          = {http://arxiv.org/abs/1312.6034},
bibsource    = {dblp computer science bibliography, https://dblp.org}
}

@article{jermyn2022engineeringmonosemanticitytoymodels,
      title={Engineering Monosemanticity in Toy Models}, 
      author={Adam S. Jermyn and Nicholas Schiefer and Evan Hubinger},
      year={2022},
      eprint={2211.09169},
      archivePrefix={arXiv},
      primaryClass={cs.LG},
      url={https://arxiv.org/abs/2211.09169}, 
journal={arXiv preprint arXiv:2211.09169},
}

@article{actMax,
author = {Erhan, Dumitru and Bengio, Y. and Courville, Aaron and Vincent, Pascal},
year = {2009},
month = {01},
pages = {},
title = {Visualizing Higher-Layer Features of a Deep Network},
journal = {Technical Report, Université de Montréal}
}

@article{introCircuit,
  author = {Olah, Chris and Cammarata, Nick and Schubert, Ludwig and Goh, Gabriel and Petrov, Michael and Carter, Shan},
  title = {Zoom In: An Introduction to Circuits},
  journal = {Distill},
  year = {2020},
  url = {https://distill.pub/2020/circuits/zoom-in},
  doi = {10.23915/distill.00024.001}
}

@article{NEMANI2024100047,
title = {Gender bias in transformers: A comprehensive review of detection and mitigation strategies},
journal = {Natural Language Processing Journal},
volume = {6},
pages = {100047},
year = {2024},
issn = {2949-7191},
doi = {10.1016/j.nlp.2023.100047},
url = {https://doi.org/10.1016/j.nlp.2023.100047},
author = {Praneeth Nemani and Yericherla Deepak Joel and Palla Vijay and Farhana Ferdouzi Liza}
}

@article{BiasAndFairness,
    author = {Gallegos, Isabel O. and Rossi, Ryan A. and Barrow, Joe and Tanjim, Md Mehrab and Kim, Sungchul and Dernoncourt, Franck and Yu, Tong and Zhang, Ruiyi and Ahmed, Nesreen K.},
    title = {Bias and Fairness in Large Language Models: A Survey},
    journal = {Computational Linguistics},
    volume = {50},
    number = {3},
    pages = {1097-1179},
    year = {2024},
    month = {09},
    issn = {0891-2017},
    doi = {10.1162/coli_a_00524},
    url = {https://doi.org/10.1162/coli_a_00524},
    eprint = {https://direct.mit.edu/coli/article-pdf/50/3/1097/2471010/coli_a_00524.pdf},
}

@inproceedings{nadeem2020gender,
  title={Gender Bias in {AI}: A Review of Contributing Factors and Mitigating Strategies},
  author={Nadeem, Ayesha and Abedin, Babak and Marjanovic, Olivera},
  booktitle={ACIS 2020 Proceedings},
  year={2020},
  url={https://aisel.aisnet.org/acis2020/27}
}

@incollection{dastin2022amazon,
  title={Amazon scraps secret {AI} recruiting tool that showed bias against women},
  author={Dastin, Jeffrey},
  booktitle={Ethics of data and analytics},
  pages={296--299},
  year={2022},
  publisher={Auerbach Publications},
  isbn={9781003278290},
}

@InProceedings{buolamwini2018gender,
  title = 	 {Gender Shades: Intersectional Accuracy Disparities in Commercial Gender Classification},
  author = 	 {Buolamwini, Joy and Gebru, Timnit},
  booktitle = 	 {Proceedings of the 1st Conference on Fairness, Accountability and Transparency},
  pages = 	 {77--91},
  year = 	 {2018},
  editor = 	 {Friedler, Sorelle A. and Wilson, Christo},
  volume = 	 {81},
  series = 	 {Proceedings of Machine Learning Research},
  month = 	 {23--24 Feb},
  publisher =    {PMLR},
  url = 	 {https://proceedings.mlr.press/v81/buolamwini18a.html}
}

@article{ferrara2023fairness,
  title={Fairness and bias in artificial intelligence: A brief survey of sources, impacts, and mitigation strategies},
  author={Ferrara, Emilio},
  journal={Sci},
  volume={6},
  number={1},
  pages={3},
  year={2023},
  publisher={MDPI},
doi={10.3390/sci6010003},
URL = {https://www.mdpi.com/2413-4155/6/1/3},
ISSN = {2413-4155},
}

@article{radford2018improving,
  title={Improving language understanding by generative pre-training},
  author={Radford, Alec and Narasimhan, Karthik and Salimans, Tim and Sutskever, Ilya},
  year=2019,
  url={https://cdn.openai.com/research-covers/language-unsupervised/language_understanding_paper.pdf},
  
}

@Inbook{cda-orig,
author="Lu, Kaiji
and Mardziel, Piotr
and Wu, Fangjing
and Amancharla, Preetam
and Datta, Anupam",
title="Gender Bias in Neural Natural Language Processing",
bookTitle="Logic, Language, and Security: Essays Dedicated to Andre Scedrov on the Occasion of His 65th Birthday",
year="2020",
publisher="Springer International Publishing",
address="Cham",
pages="189--202",
isbn="978-3-030-62077-6",
doi="10.1007/978-3-030-62077-6_14",
url="https://doi.org/10.1007/978-3-030-62077-6_14"
}

@inproceedings{stereoset,
    title = "{S}tereo{S}et: Measuring stereotypical bias in pretrained language models",
    author = "Nadeem, Moin  and
      Bethke, Anna  and
      Reddy, Siva",
    editor = "Zong, Chengqing  and
      Xia, Fei  and
      Li, Wenjie  and
      Navigli, Roberto",
    booktitle = "Proceedings of the 59th Annual Meeting of the Association for Computational Linguistics and the 11th International Joint Conference on Natural Language Processing (Volume 1: Long Papers)",
    month = aug,
    year = "2021",
    address = "Online",
    publisher = "Association for Computational Linguistics",
    url = "https://aclanthology.org/2021.acl-long.416",
    doi = "10.18653/v1/2021.acl-long.416",
    pages = "5356--5371",
}
\bibliographystyle{iclr2026_conference}

\appendix

\section{Structure of the Appendix}

We structure the appendix similar to the main part of the paper. This section provides an overview and highlights the most important results complementary to the main part of the paper.

The appendix follows the structure of the main paper and provides complementary details and results. Appendix~\ref{app:data} describes the generated datasets, and Appendix~\ref{app:metrics} defines the evaluation metrics in detail. Training and implementation details are given in Appendix~\ref{app:training}. 
Appendix~\ref{app:encoder} presents the complementary plots to Figure~\ref{fig:encoded-values-race-religion}, showing the distribution of encoded values for all base models (Figure~\ref{fig:rr-encoded-values}). 
\rev{Additionally, we provide an analysis of the stability of the encoded feature neuron across training runs as well as a brief evaluation of how the encoder generalizes to unseen data and additional gendered target tokens.}
Appendix~\ref{app:decoder-as-bias-changer} provides the corresponding heatmaps to Figure~\ref{fig:changed_models-main-bert-base-cased} for the selected models (Figures~\ref{fig:changed-gender-models}--\ref{fig:changed-model-Llama-3.2-3B-Instruct-v5}), including precise metric definitions and their scores for the selected models. 
Raw results for Table~\ref{tab:rank} are reported in Appendix~\ref{app:comp}. \rev{Appendix~\ref{app:finetuning} presents an ablation study on how \gradiend\ can be integrated with a fine-tuning task.}
Appendix~\ref{app:overfitting} examines generalization of \gradiend’s debiasing effect to unseen tokens. Finally, Appendix~\ref{app:example-predictions} concludes with example predictions illustrating the impact of gender debiasing.

\section{Data}\label{app:data}


We publish all of our introduced datasets\iflink\ on Hugging Face\fi, see Table~\ref{tab:hf-datasets}.
Details regarding the data generation can be found in the subsequent sections.

For brevity, the term \emph{pronouns} is used to refer specifically to third-person singular gendered pronouns (i.e., “he” and “she”), and \emph{name} refers exclusively to \emph{first names}.

\subsection{\namexact}

Several datasets were constructed with the help of an existing name dataset~\citep{UCI_gender_by_name}, which contains 133,910 names with associated genders, counts, and probabilities derived from government data in the US, UK, Canada, and Australia. From this dataset, we derive two subsets based on name ambiguity: \namexact\ and \namextend.

We refer to \emph{\namexact} as a collection of names that are exclusively associated with a single gender and that have no ambiguous meanings, therefore being \emph{exact} with respect to both gender and meaning. First, we filter all names of the raw dataset to retain only names with a count of at least 20,000, resulting in a selection of the most common 1,697 names. Next, we remove names with ambiguous gender, such as Skyler, Sidney, and Billie, which were identified by having counts for both genders in the filtered dataset, removing 67 additional names.

To further refine our selection of the remaining 1,630 names, we manually checked each remaining name for ambiguous meanings. 
For instance, names like \emph{Christian} (believer in Christianity), Drew (the simple past of the verb \emph{to draw}), \emph{Florence} (an Italian city), 
\emph{April} (month), 
\emph{Henry} (the SI unit of inductance), and \emph{Mercedes} (a car brand).
This exclusion process was performed without considering casing to ensure applicability to non-cased models.
The filtering resulted in the exclusion of 232 names, leaving us with a total of 1,398 names in \namexact.

We split the data into training ($85\%$), validation ($5\%$), and test ($10\%$) subsets, ensuring that the latter two splits are balanced with respect to gender.

\begin{table*}[!t]
    \centering
    \scriptsize
    \caption{Overview of generated datasets including total number of samples and a description.}
    \label{tab:dataset-summary}
\begin{tabularx}{\textwidth}{lrX}
    \toprule 
         \textbf{Name}  & \textbf{Size} & \textbf{Description}\\ 
         \midrule
        \namexact & 1,398 & Names that are unambiguous (\emph{exact}) in meaning and gender, e.g., \emph{Alice, Bob, Eve} \\     
        \namextend \! & 40,351 & Extends \namexact\ with less certain names, including those with multiple meanings and genders, e.g.,  \emph{Alice, Bob, Christian, Drew, Eve, Florence, Skyler} \\
        \acrshort{genter}/ Gender \traindata\!\!\!\! \!\!\! & 27,031 & Name-gender templates, 
        e.g., \emph{[NAME] explained the vision as best [PRONOUN] could .} \\ 
        Race \traindata & 9,779 (A.), 18,073~(B.), &  Race templates, e.g., \emph{Ranks in the [MASK] Sudoku Championship (ASC)} \\ 
        &  20,152 (W.) \\
        Religion \traindata & 19,653 (C.), 4,945~(J), & Religion templates, e.g., \emph{Cathedrals of the Roman Catholic [MASK] in Switzerland} \\ 
        &  4,043 (M.) \\
        \gerneutral & \!\!\!\! 20,057,351 & Contains only bias-neutral words, e.g., \emph{i really want to see you again , soon if you can} \\
        \acrshort{gentypes} & 500 & Gender-stereotypical templates,         e.g., \emph{My friend, [NAME], loves taking care of babies.} \\
        \rev{\wikigender} & \rev{10,000} & \rev{English Wikipedia templates with diverse masked gendered terms (e.g., \emph{man}, \emph{daughter}).} \\
         \bottomrule
    \end{tabularx}
\end{table*}

\begin{table}[!t]
    \centering
    \scriptsize
    \iflink
        \caption{Hugging Face identifiers of our datasets.}
    \else
        \caption{Anonymous links to our datasets.}
    \fi
    \label{tab:hf-datasets}
    \begin{tabular}{ll}
    \toprule
         Name & \iflink Hugging Face Identifier \else URL \fi \\
         \midrule
          \namexact & \iflink \href{https://huggingface.co/datasets/aieng-lab/namexact}{\texttt{aieng-lab/namexact}} \else anonymous \fi \\
         \namextend & \iflink \href{https://huggingface.co/datasets/aieng-lab/namextend}{\texttt{aieng-lab/namextend}} \else anonymous \fi \\
         \genter/ Gender \traindata & \iflink \href{https://huggingface.co/datasets/aieng-lab/genter}{\texttt{aieng-lab/genter}} \else anonymous \fi \\
          Race \traindata & \iflink \href{https://huggingface.co/datasets/aieng-lab/gradiend_race_data}{\texttt{aieng-lab/gradiend\_race\_data}} \else anonymous \fi \\
         Religion \traindata & \iflink \href{https://huggingface.co/datasets/aieng-lab/gradiend_religion_data}{\texttt{aieng-lab/gradiend\_religion\_data}} \else anonymous \fi \\
         \gerneutral & \iflink \href{https://huggingface.co/datasets/aieng-lab/biasneutral}{\texttt{aieng-lab/biasneutral}} \else anonymous \fi \\
         \gentypes & \iflink \href{https://huggingface.co/datasets/aieng-lab/gentypes}{\texttt{aieng-lab/gentypes}} \else anonymous \fi \\
         \rev{\wikigender} & \iflink \href{https://huggingface.co/datasets/aieng-lab/gradiend\_wiki\_gender}{\texttt{aieng-lab/gradiend\_wiki\_gender}} \else \rev{anonymous} \fi \\
    \bottomrule
    \end{tabular}
\end{table}

\subsection{\namextend}\label{sec:namextend}
We define \emph{\namextend} as a dataset that \emph{extends} beyond the constraints of \namexact\ by including words that can be used as names, but are not exclusively names in every context.

To limit the number of names while ensuring sufficient representations, we set a minimum count threshold of 100 for the raw name dataset. This threshold reduces the total number of names by $72\%$, from 133,910 to 37,425, helping to save computationally time. This dataset includes names with multiple meanings and gender associations, as the threshold is the only filtering criterion applied.

Therefore, names that can be used for both genders are listed twice in this dataset, once for each gender. By considering the counts of how often a name is associated with a particular gender, we can define the probability that a name is used for a specific gender. For a given name $N$ and gender $F$ (female) or $M$ (male), we denote this probability as $\mathbb{P}(F|N)$ and $\mathbb{P}(M|N)$.
For example, for the name $N=Skyler$, the dataset contains the probabilities $\mathbb{P}(F|Skyler) = 37.3\%$ and $\mathbb{P}(M|Skyler) = 62.7\%$.

\subsection{Training Data for Gender (\genter)}\label{sec:data-genter}
For the training of \gls{gradae}, we introduce a new dataset called \gls{genter}, which consists of template sentences capable of encoding factual and counterfactual gender information, as illustrated in the motivating example in Section~\ref{sec:gradae-motivation}. Each entry in the dataset includes two template keys: a name [NAME] and a pronoun [PRONOUN]. For instance, the earlier discussed example sentences can be instantiated from the following template:
\begin{quote}
[NAME] explained the vision as best [PRONOUN] could .
\end{quote}
Using the popular BookCorpus~\citep{bookcorpus} dataset, we generated such template sentences that meet the following criteria: 
\begin{itemize}
    \item Each sentence contains at least 50 characters.
    \item Exactly one name from \namexact\ is contained, ensuring a correct name match.
    \item No other names from \namextend\ are included, ensuring that only a single name appears in the sentence.
    \item The correct name's gender-specific third-person pronoun (\emph{he} or \emph{she}) is included at least once.
    \item All occurrences of the pronoun appear after the name in the sentence.
    \item The counterfactual pronoun does not appear in the sentence.
    \item The sentence excludes gender-specific reflexive pronouns (\emph{herself}, \emph{himself}) and possessive pronouns (\emph{her}, \emph{his}, \emph{hers}, \emph{him}).
    \item Gendered nouns (e.g., \emph{actor}, \emph{actress}, \dots) are excluded, based on a gendered-word dataset\footnote{\url{https://github.com/ecmonsen/gendered_words}}, which is expanded with plural forms using the Python library \texttt{inflect}, resulting in 2,421 entries.
\end{itemize}
This approach generated a total of 83,772 sentences. %
To further enhance data quality, we employed a simple BERT model (\texttt{bert-base-uncased}) as a judge model. This model must predict the correct pronoun for selected names with high certainty, otherwise, sentences may contain noise or ambiguous terms not caught by the initial filtering. Specifically, we used 50 female and 50 male names from \namexacttrain, and a correct prediction means the correct pronoun token is predicted as the token with the highest probability in the \gls{mlm} task. 
Only sentences for which the judge model correctly predicts the pronoun for every test case were retained, resulting in a total of $27,031$ unique sentences. We split the data into training ($87.5\%$
), validation ($2.5\%$
), and test ($10\%$
) subsets. The validation split is rather small, due to the large input size of the \gls{gradae} models (comparable to the size of the base model), see Section~\ref{sec:eval-training} for more information. 

The \gls{genter} dataset is specifically designed to train our proposed \gls{gradae} models, focusing on gradient updates that influence gender-changing directions. The applied filtering constraints ensure that the only distinguishing gender-related factor between the factual and counterfactual versions of a sentence is the pronoun (\emph{he} or \emph{she}) associated with the actual gender linked to the name. \rev{While our experiments show that using the name-pronoun associations in \genter\ effectively uncovers a proper feature encoding and debiasing, future work could investigate whether incorporating additional context, such as gendered nouns or adjectives, provides further useful information.}

We selected the BookCorpus \citep{bookcorpus} as the foundational dataset due to its focus on fictional narratives where characters are often referred to by their first names. In contrast, the English Wikipedia \citep{wikipedia}, also commonly used for the training of transformer models \citep{bert, roberta}, was less suitable for our purposes. For instance, sentences like \emph{[NAME] Jackson was a musician, [PRONOUN] was a great singer} complicate bias detection based on first names (as done for \genter) due to the context of well-known individuals, where the name and pronoun association can be highly influenced by prior knowledge rather than bias.

\subsection{Training Data for Race and Religion}\label{app:training-data-rr}

We filter the same Wikipedia dump used by \citep{meade2022empiricalsurveyeffectivenessdebiasing} to create the templated \gradiend\ training datasets for race and religion, similar to how they augmented counterfactual data for their \gls{cda} training. Following their approach, we use their defined bias attribute words to identify factual and counterfactual terms. These words consist of triples representing each feature class class, e.g., \emph{Church/Synagogue/Mosque} for \emph{Christian/Jewish/Muslim} or \emph{Asia/Africa/Europe} for \emph{Asian/Black/White}
For each directed pair of classes (e.g., $A=Asian$ and $B=Black$), we retain only sentences containing a bias word from $A$ (factual term) and use the corresponding term for $B$ of the triple as counterfactual term. The casing of the counterfactual term matches that of the original factual term (lowercase by default, or all caps/ capitalized first letter).
The resulting datasets \traindata\ in training (70\%), validation (20\%), and test (10\%) subsets.

\subsection{\gerneutral}

To evaluate our models on data that is entirely bias-unrelated, we created a separate dataset called \gerneutral, again sourced from BookCorpus~\citep{bookcorpus}. 
Similar to \gls{genter}, we removed sentences containing gender-specific words, as well as any bias attribute words (i.e., the target tokens defined by \cite{meade2022empiricalsurveyeffectivenessdebiasing}, such as \emph{African}, \emph{Christian}, \dots). We also excluded sentences containing any name from \namextend\ and sentences shorter than 50 characters.
This filtering process resulted in a total of 20M bias-neutral sentences, with all duplicate sentences removed.

\subsection{\gentypes} 
Given that the training of the gender \gradiend\ specifically involves the relationship between names and their associated genders, we aim to evaluate our models against the gender associations of names with gender-stereotypical contexts to select gender bias-changed models. 
\begin{quote}
    My friend, [NAME], loves taking care of babies.\\
    My friend, [NAME], is probably into video games.
\end{quote}
We expect gender-biased models to assign higher probabilities to female in the first sentence and male names in the second one due to existing stereotypes in real-world data.
We generated 500 sentences with GPT-4o \citep{openai2024gpt4osystemcard} and refer to this gender-stereotype data as \gentypes.  

The dataset was generated with the following prompt:
\begin{quote}
    Generate a copyable CSV dataset with 500 sentences that demonstrate gender bias using the placeholder [NAME] for a name. The dataset should include four columns: "text," "gender," and "reason,". The "gender" column should contain 'M' for male or 'F' for female. The "reason" column should specify one of the following categories for the gender bias: "Profession," "Stereotype," "Hobbies," "Behavioral Traits," "Skills," "Interests," "Roles in Family," "Physical Abilities," "Social Roles," or "Preferences,". Provide at least 50 sentences for each reason category. Ensure the sentences are diverse and avoid repetitions. Do not use any words that indicate the gender such as "him", "his", "actress", …

The csv file should look like the following:

text,gender,reason

"[NAME] is an outstanding pilot with thousands of flight hours.",M,Profession

"[NAME] is a fantastic hair stylist with many loyal clients.",F,Profession

"[NAME] is probably into video games.",M,Interests

"[NAME] is likely to be more empathetic.",F,Behavioral Traits
\end{quote}

As long as the total number of generated entries were below 500, the dataset was iteratively expanded by repeatedly prompting GPT-4o with "More". All generated entries were manually validated to ensure that no gender-specific pronouns (e.g., \emph{he}, \emph{she}, \emph{his}, etc.) were present. Entries containing such pronouns were excluded. The final dataset size was capped at 500 entries.

Although the \emph{gender} and \emph{reason} columns were not directly used in this study, their inclusion was intended to enforce balance between male- and female-associated stereotypes and to enhance diversity in stereotype contexts. However, this goal may not have been fully achieved, as \roberta\ demonstrates a female bias in predictions (see Section~\ref{sec:eval-decoder}), in contrast to our expectations of a generally male biased model.

To encourage the model to predict names on these masked sentences, we used the prefix \emph{"My friend, [MASK], has a \dots"} rather than \emph{"[MASK] has a \dots"}, which could logically allow for other (unwanted) tokens, such as \emph{he} or \emph{she}.

\subsection{\rev{\wikigender}}\label{app:data-wikigender}

\rev{To evaluate how well the \gradiend\ encoder generalizes to unseen tokens and to data from a different source than seed during training, we derive masked texts from the English Wikipedia \citep{wikipedia}. We filter and mask occurrences of the following gendered target word pairs: \emph{she}/\emph{he}, \emph{woman}/\emph{man}, \emph{girl}/\emph{boy}, \emph{mother}/\emph{father}, and \emph{daughter}/\emph{son}. 
For each target, we retain 1,000 texts, forming the dataset \wikigender.
}

\rev{
Unlike BookCorpus, the base dataset for \genter\ used to train the gender \glspl{gradae}, Wikipedia articles are much longer (on average $\approx 400$ words for \wikigender\ vs. $\approx 17$ words for \genter), contain structural elements such as headings and newlines, and cover encyclopedic content rather than narrative text. This enables evaluation of both input distribution and target token shifts.
}

\section{Metrics}\label{app:metrics}

In this section, we define the metrics of Section~\ref{sec:evaluation} used to evaluate the \gradiend\ encoder and to select bias-changed models formally and more detailed.
Additionally, we discuss established techniques to measure bias in language models.

\subsection{Language Modeling Score of the Decoder}

We use \accdec\ as a measure of the general language modeling capabilities of a model that may have been modified by the \gradiend\ decoder. 
To ensure that the evaluation is independent of any gender bias change, we employ a \gls{tpt} on \gerneutral. 

For encoder-only models, the \gls{tpt} corresponds to a \gls{mlm} task, where 10,000 \gerneutral\ samples are used for gender evaluation and 1,000 samples for race and religion, reflecting the larger number of \gradiend\ models in the latter case. Approximately $15\%$ of the tokens are masked, following standard practice \citep{bert}, and \accdec\ is computed as the accuracy on the \gls{mlm} task. 

For decoder-only models, we compute perplexity over 1,000 samples -- fewer than in the \gls{mlm} setting, as the model predicts every token in each sequence, resulting in both higher computational cost and more relevant tokens per sample.
Perplexity measures the model's confidence, with lower values indicating better performance, ranging from 1 to infinity. To align its interpretation with accuracy, we convert it to $\accdec = \frac{1}{1+perplexity}$, yielding scores in~$[0, 1]$.

\subsection{Gender Prediction Probabilities}

This section introduces probabilities for a feature class $A$, $\P(A)$. We initially restrict this to gender (i.e., female and male probabilities, $\P(F)$ and $\P(M)$) for clarity, which are generalized to other feature classes in the following section.

Let $\mathcal{N}$ denote the set of single-token names in \namextend, and let $G\in\{F, M\}$ be a gender. Let the \gentypes\ data be denoted as $T$, i.e., stereotyped sentences with a name placeholder [NAME], e.g.:
\begin{quote}
    My friend, [NAME], loves taking care of babies.
\end{quote}
Let $t\in T$ be a text and $|T|$ denote the number of elements in the set $T$.

Each text $t$ creates a \gls{tpt} $\hat{t}$ where the goal is to predict a name. For encoder-only models, [NAME] is simply replaced by [MASK], creating a \gls{mlm} task.
For \gpttwo\ and \llama, we transform $t$ into a sentence that naturally prompts a name prediction, using the following template style:
\begin{quote}
    The person, who loves taking care of babies, has the first name [MASK]
\end{quote}
This is done by removing the prefix \emph{“My friend, [NAME],”} and the final punctuation from $t$.
For \llamai, we use the original text with the [NAME] placeholder as user prompt and prepend a system prompt instructing the model to predict a suitable name: \begin{quote}
    You are a language model trained to predict first names. In the following text, [NAME] represents a placeholder for a 
first name. Your task is to predict the most likely name that fits the context. Return only the predicted name — no 
punctuation, no quotation marks, and no explanations.
\end{quote}
The probability distribution over the first generated token is then treated as the model's prediction for $\hat{t}$, similar to the other models.

The probability of predicting a name $N\in\mathcal{N}$ for $\hat{t}$ is denoted as $\mathbb{P}_t(N)$. Names are treated independent of casing and leading white spaces, i.e., the probabilities of all such tokens contribute to this probability.

The probability of predicting gender $G$ for $\hat{t}$ is estimated by summing $\P_t(N)$ for all names $N$ of that gender: \begin{equation}
    \P_t(G) \coloneqq \sum_{N \in \mathcal{N}} \mathbb{P}_t(N) \cdot \mathbb{P}(G|N) \in [0, 1].\label{equ:p_t_G}
\end{equation}
As introduced in Section~\ref{sec:namextend}, $\P(G|N)$ represents the likelihood of a name $N$ being associated with gender $G$.
This conditional probability acts as a filter in the sum over all names in $\mathcal{N}$, ensuring that names of the other gender do not contribute to the aggregated probability of $G$.
Moreover, $\P(G|N)$ ensures that names applicable to both genders contribute only partially to the aggregated probability of gender $G$. 
For example, for $N=Skyler$, $P_t(Skyler)$ contributes to $\P(F|Skyler)=37.7\%$ to the female probability $\P_t(F)$ and $\P(M|Skyler)=62.7\%$ to the male probability $\P_t(M)$.

The combined probabilities for either male or female names is given by \[\mathbb{P}_t(F\cup M)\coloneqq  \mathbb{P}_t(F) + \mathbb{P}_t(M) \in [0, 1].\] 
This probability quantifies the proportion of meaningful predictions for $\hat{t}$.

The probability of gender $G$, denoted as $\mathbb{P}(G)$, averages $\P_t(G)$ over all $t\in T$, i.e.: \[\P(G) \coloneqq \frac{1}{|T|} \sum_{t\in T} \P_t(G) \in [0, 1]. \]

\subsection{Generalization of Gender Probabilities to Feature Class Probabilities}\label{app:feature-class-probs}

We generalize gender probability framework to other feature classes, such as race and religion, by the following adaptions:
\begin{itemize}
    \item Instead of a gender $G$, we consider general feature classes $F, F_1, F_2\in \{Asian, Black, White, Christian, Jewish, Muslim\}.$
    \item Instead of \gentypes\, we use the test split of \traindata\ as $T$.
    \item Instead of names, we use the set of bias attribute terms $\mathcal{A}_F$ \citep{meade2022empiricalsurveyeffectivenessdebiasing} for each feature class as target tokens, i.e., the sets $A_{Asian}\cup A_{Black}\cup A_{White}$ and $A_{Christian}\cup A_{Jewish}\cup A_{Muslim}$ are analogous to the name token set $\mathcal{N}$ for gender.
    \item The conditional probability $\P(F|A)$ for a bias attribute term $A$ is defined as $1$ if $A\in\mathcal{A}_F$ and 0 otherwise, reducing Equation~\ref{equ:p_t_G} to $\P_t(F) \coloneqq \sum_{A\in\mathcal{A}_F}P_t(A)$.
    \item These adaptions yield similar definitions for $\P_t(F_1\cup F_2)$ and $\P(G)$.
    \item For encoder-only models, multi-token target terms are handled by computing the joint probability across all tokens, allowing both single- and multi-token bias attribute terms to contribute meaningfully to the per-example probabilities.
    \item For decoder-only models, considering only the first token of each target term can be noisy, since it may consist of just one or two characters (especially for the large \llama\ tokenizer) and be poorly aligned with the intended term meaning. Instead, we include all first tokens of the target terms that constitute at least half of the attribute term (in characters), providing a more reliable estimate of the term’s probability.
    \item For \llamai, we use the same prompt as in training, without the special prompt used for gender names (see Section~\ref{app:tpt-generative}).
\end{itemize}

\subsection{Model Selection Metrics}\label{app:bias-metrics}

 The \acrfull{bpi} integrates the previous measures aiming to quantify how debiased a model is over feature classes $A$ and $B$, by averaging over all texts $t\in T$:
\begin{equation*}
        \text{\acrshort{bpi}} \coloneqq \frac{\accdec}{|T|} \cdot \sum_{t \in T} \Big[ \left( 1 - |\mathbb{P}_t(A) - \mathbb{P}_t(B)| \right)  \cdot \mathbb{P}_t(A \cup B) \Big] \in [0, 1].\label{equ:bpi}
\end{equation*}
Here, \accdec\ ensures that high values indicate models with good language modeling capabilities. 
The first part of the product in the sum ($1 - |\P_t(A) - \P_t(B)|$) is large if the predictions are unbiased over the two classes $A$ and $B$, since $\mathbb{P}_t(A)$ must be similar to $\mathbb{P}_t(B)$ to achieve a good score. The second part ($\P_t(A \cup B)$) ensures that both class probabilities are large to avoid a good scoring of models that assign probabilities close to zero to the class target tokens.
A high value in \gls{bpi} indicates a relatively debiased model, that has still good language modeling capabilities due to the influence of \accdec.

The \acrfull{fpi} measures bias towards the female gender
 \begin{equation*}
    \text{\acrshort{fpi}} \coloneqq \frac{\accdec}{|\mathcal{T}|} \cdot \sum_{t\in\mathcal{T}} (1 - \mathbb{P}_t(M)) \cdot \mathbb{P}_t(F) \in [0, 1].
\end{equation*}
\accdec\ ensures again good language modeling capabilities, $1-\P_t(M)$ prefers small male probabilities, and $\P_t(F)$ prefers large female probabilities.

Analogously, the \acrfull{mpi} measures bias towards the male gender:
 \begin{equation*}
    \text{\acrshort{mpi}} \coloneqq \frac{\accdec}{|\mathcal{T}|} \cdot \sum_{t\in\mathcal{T}} (1 - \mathbb{P}_t(F)) \cdot \mathbb{P}_t(M) \in [0, 1].
\end{equation*}

\subsection{Bias Metrics} \label{sec:measuring-gender-bias}

Various methods exist in literature to quantify bias in language models (see, e.g.,~\citet{li2023survey}). 
Here, we present a few representative techniques commonly used to measure stereotypical bias.

The \acrlong{seat} (\glsfirst{seat}\glsunset{seat}; \citealt{seat}) extends the \acrlong{weat} (\acrshort{weat}\glsunset{weat}; \citealt{weat}) by using sentence templates to evaluate social biases in encoder models. It compares association strengths between embeddings of predefined attribute (e.g., gender-specific names) and target sets (e.g., stereotypical professions) using cosine similarity. 
Bias is expressed as an \emph{effect size}, where larger values indicate stronger bias.

StereoSet \citep{stereoset} is a benchmark dataset with context-rich sentences for intrasentence and intersentence tasks.
This study focuses on the intrasentence task, where a sentence (e.g., \emph{Girls tend to be more [MASK] than boys}) requires the model to predict the masked word from three options: stereotypical (e.g., \emph{soft}), anti-sterotypical (e.g., \emph{determined}), and meaningless (e.g., \emph{fish}).
Two metrics are considered: 
\begin{enumerate*}[label=\textbf{\arabic*)}] 
\item \gls{lms}, which measures the proportion of meaningful (stereotypical or anti-stereotypical) options chosen over meaningless ones, reflecting the model’s language understanding. 
\item \gls{ss}, which quantifies bias as the proportion of stereotypical options selected over anti-stereotypical ones. A balanced model achieves $50\%$. \end{enumerate*}

\acrshort{crows} (\acrlong{crows}; \citealt{crows}) is a crowdsourced dataset consisting of pairs of sentences: one expressing a stereotype (e.g., \emph{Woman don't know how to drive}), and the other its anti-stereotypical counterpart (e.g., \emph{Man know how to drive}). A bias score is computed considering the model's preference for one sentence over the other, similar to \gls{ss}. However, \acrshort{crows} has been criticized for unreliable bias measurement, including spurious correlations and flawed assumptions about social categories \citep{blodgett-etal-2021-stereotyping}. Therefore, we did not use this metric in this study, but report it here for completeness.

\citet{li2021detectinggenderbiastransformerbased} analyze the attention associations between gendered pronouns (e.g., \emph{she}) and occupations (e.g., \emph{nurse}) in transformer models, using gender-swapped sentences (e.g., replace \emph{he} by \emph{she}). The attention scores between the gender-swapped pronouns and the occupation are then compared to identify gender bias on attention head level. However, the approach does not compute a model-specific, aggregated bias score usable for comparison.

\section{Training and Implementation Details}\label{app:training}

\begin{table}[!t]
    \centering
    \tablefontsize
        \caption{Hugging Face model checkpoints used in this study.}
    \label{tab:models}
    \begin{tabular}{lll}
        \toprule
        \textbf{Model} & \textbf{Checkpoint} & \textbf{Reference} \\
        \midrule
        \bertbase & \href{https://huggingface.co/google-bert/bert-base-cased}{\texttt{bert-base-cased}} & \citet{bert} \\
        \bertlarge & \href{https://huggingface.co/google-bert/bert-large-cased}{\texttt{bert-large-cased}} & \citet{bert} \\
        \distilbert  & \href{https://huggingface.co/distilbert/distilbert-base-cased}{\texttt{distilbert-base-cased}} & \citet{distilbert} \\
        \roberta  & \href{https://huggingface.co/FacebookAI/roberta-large}{\texttt{roberta-large}} & \citet{roberta} \\
        \gpttwo & \href{https://huggingface.co/openai-community/gpt2}{\texttt{gpt2}} & \citet{radford2019language} \\
        \llama & \href{https://huggingface.co/meta-llama/Llama-3.2-3B}{\texttt{meta-llama/Llama-3.2-3B}} & \citet{grattafiori2024llama} \\
        \llamai & \href{https://huggingface.co/meta-llama/Llama-3.2-3B-Instruct}{\texttt{meta-llama/Llama-3.2-3B-Instruct}} & \citet{grattafiori2024llama} \\
        \bottomrule
    \end{tabular}
\end{table}
Table~\ref{tab:models}  summarizes the Hugging Face model checkpoints used in our experiments,
\begin{table}[!t]
    \centering
    \tablefontsize
    \caption{Training hyperparameters.}
    \label{tab:hyperparameters}
    \begin{tabular}{ll}
    \toprule
        \textbf{Hyperparameter} & \textbf{Value} \\\midrule
          Optimizer & Adam \\
          Learning Rate & \num{1e-4} (\llama, \llamai); \num{1e-5} (others) \\ 
          Weight Decay & \num{1e-2} \\
          Batch Size Gradient Computation & 32 \\
          Batch Size \gradiend & 1 \\
          Training Criterion & MSE \\
          Training Steps & 23,653 (Gender); 2,500 (Race, Religion) \\
          Evaluation Steps & 250 \\
          Evaluation Criterion & \cormf\ on validation split \\
          \bottomrule
    \end{tabular}
\end{table}
while Table~\ref{tab:hyperparameters} lists the hyperparameters used for training the \gradiend\ models.

\subsection{Environment}
The implementation is based on Python 3.9.19, and we made the training framework publicly available: \iflink\url{https://github.com/aieng-lab/gradiend-bias}\else anonymous\fi.
The \llama-based \gradiend\ models were trained using three NVIDIA A100 GPUs, while all others used a single A100. Each A100 provides \SI{80}{\giga\byte} of GPU memory, and the system had \SI{504}{\giga\byte} of RAM. The same setup is also used for evaluation.

\subsection{\rev{Token Prediction Task for Encoder-Only Models}}\label{app:tpt-encoder-only}
The training task for \gradiend is motivated as a \gls{mlm} \cite{bert} task (see Section~\ref{sec:gradae-motivation}), where the masked token is sensitive to an involved feature class. 
\rev{For multi-token targets, we insert one \texttt{[MASK]} token per target token in the template text. The MLM loss then naturally aggregates over all target tokens, so the resulting gradients reflect contributions from each token.}

\subsection{Token Prediction Task for \rev{Decoder-Only} Models}\label{app:tpt-generative}

For causal models, \gls{mlm} instances are converted into a \gls{clm} \cite{radford2018improving} task by providing only the prefix up to the (first) masked token and predicting the next token at the end of the sequence.

For \llamai, we use the following system prompt to allign its behavior with non-instruction-tuned models:
\begin{quote}
        You are a language model that completes sentences. Predict the next word that naturally follows the given text. 
    Return only that word — no punctuation, no quotes, and no explanations.
\end{quote}
This prompt is used for all applications of \llamai\ in this study unless stated otherwise.

Although this modification is straightforward, it is effective only when the target terms can be tokenized as single tokens -- or when the primary semantic content is largely captured by the first token (e.g., similar to Appendix~\ref{app:feature-class-probs}). This limitation is particularly noticeable for \llama-based models with race and religion terms, as illustrated in Figure~\ref{fig:rr-encoded-values}. Future work should investigate methods to handle multi-token targets in decoder-only \gradiend\ models.

\subsection{Custom Initialization}
Our training setup involves a custom random initialization for the \gradiend\ models. The default initialization in PyTorch applies a uniform distribution from $\left(\frac{-1}{\sqrt{n}}, \frac{1}{\sqrt{n}}\right)$, where $n$ is the dimension of the input layer. However, for the decoder, the input dimension is $n=1$, resulting in a uniform distribution over the interval $(-1, 1)$.
This leads to relatively high absolute initial values compared to the target values, as the decoder inputs are typically close to $\pm 1$. To address this, we use the same $n$ for the initialization as for the encoder, which corresponds to the number of used weights in the designated model. Our experiments show that this custom initialization improves training results. 

\subsection{Training Procedure}\label{app:training-procedure}

Each training step involves two forward and backward passes through the base model to compute the input and output tensors for the \gls{gradae} model. 
For race and religion, the training data for classes $A$ and $B$ is derived by combining the datasets for each source class and augmenting the targets with all valid terms from the other class within the same bias attribute group.
For gender, each entry of \genter\ is augmented batch-size many times with a name of \namexact\ to generate the actual training dataset. 
Gradients are calculated with respect to the target token, e.g. \emph{he}/\emph{she} or \emph{He}/\emph{She}, depending on the position of the target token. 
We only used single token targets for training, i.e., the datasets were filtered to exclude multi-token targets or sources.

We use the validation split of \traindata\ for evaluation during training, following the same procedure as described to compute \cormf\ (Section~\ref{sec:eval-encoder}).
However, as pre-computing  these validation gradients require a substantial amount of storage, we use for the gender \glspl{gradae} all of the \genter\ validation split for the smaller models (\bertbase, \distilbert, and \gpttwo), half of the data for the medium-sized models (\bertlarge\ and \roberta), and only $5\%$ for the \llama-based models due to their large model sizes. 
This ensures that the gradients required for evaluation fit into the memory during training. For instance, the evaluation data for \bertbase\ requires approximately 270\, GB.
For race and religion, a maximum of 1,000 samples is used, with similar relative reductions based on model size.
The training time for a single gender \gradiend\ model ranges from 3.5 hours for \distilbert\ to 24 hours for \llamai.

To monitor progress, the model is evaluated every 250 training steps using \cormf, and select the best model after finishing all training steps (Section~\ref{sec:eval-training}). Similar to the procedure to evaluate the \gradiend\ encoder (Section~\ref{sec:eval-encoder}), 
This evaluation metric focuses on the encoder's ability to differentiate between genders, which measures how well the encoded values distinguish between the feature classes. 
Notice that this metric evaluates only the encoder, as the decoder's role in adjusting bias is harder to evaluate.

When training the gender \gradiend\ models, they sometimes fail to converge in distinguishing female and male input as $\pm1$, depending on the learning rate and random seed. This issue was observed particularly with \roberta, although it occasionally occurred with other models as well, depending on the learning rate. 
In such cases, the first training steps determine whether both genders are separated correctly or both are encoded as the same value (either $+1$ or $-1$). Future research is needed to explore this phenomenon.
To mitigate non-convergent runs for gender, we train three \gradiend\ models per base model with different seeds and select the one with the highest \cormf\ on the validation split. For race and religion, a single \gradiend\ model is trained per configuration.

\section{Encoder as Classifier}\label{app:encoder}

\subsection{\rev{Detailed Results}}
\begin{figure}
    \centering

    \begin{subfigure}[t]{\textwidth}
         \includegraphics[width=\linewidth]{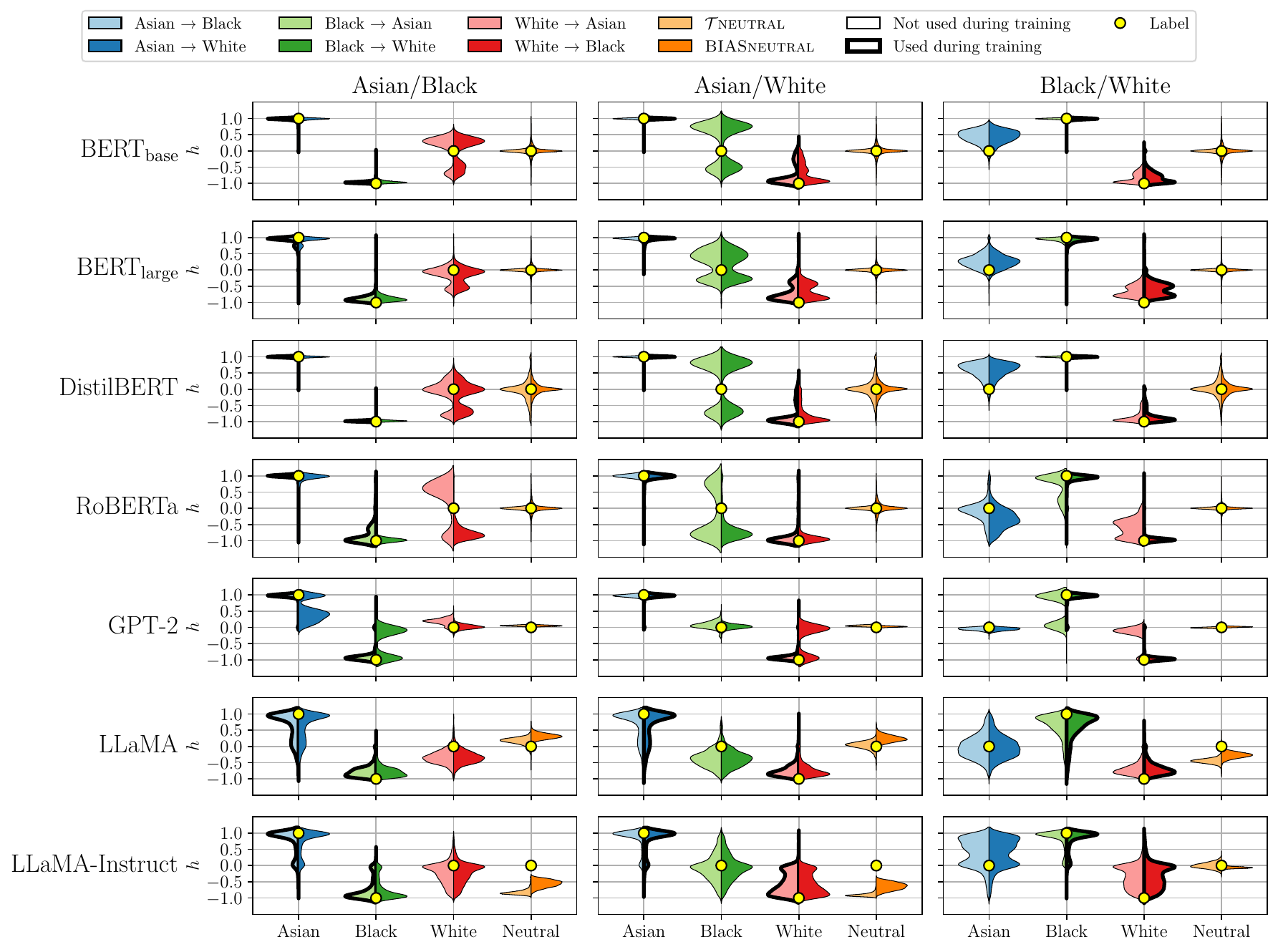}
        \caption{Race.}\label{fig:race-encoded-values}
    \end{subfigure}

   \begin{subfigure}[t]{\textwidth}
         \includegraphics[width=\linewidth]{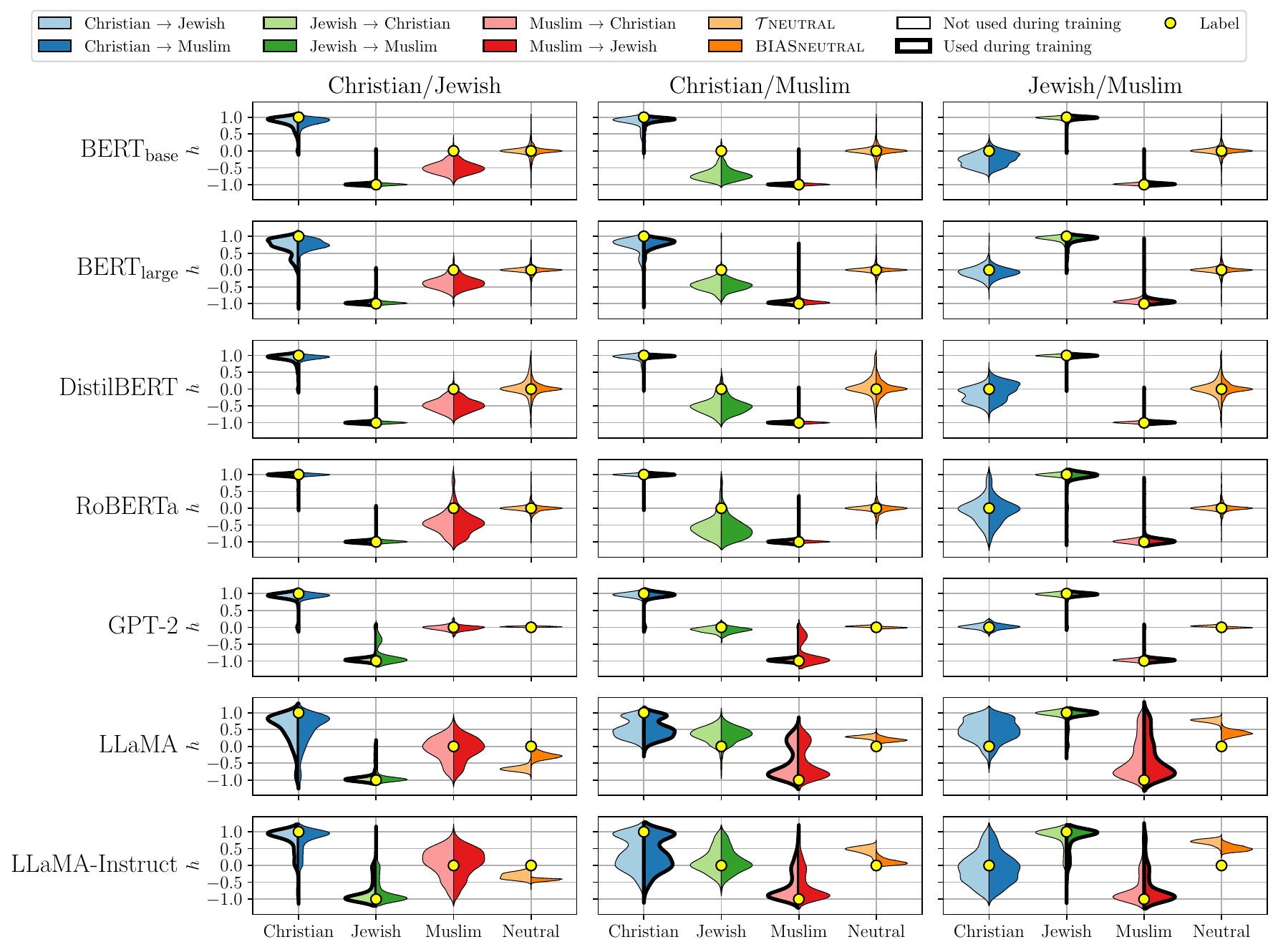}
        \caption{Religion.}\label{fig:religion-encoded-values}
    \end{subfigure}

   \caption{Distribution of encoded values for all race and religion \gradiend\ models across different datasets. The yellow dots indicate the expected label used for \corenc.}
    \label{fig:rr-encoded-values}
    
\end{figure}

Similar to Figure~\ref{fig:encoded-values-race-religion}, we present additional results in Figure~\ref{fig:rr-encoded-values}, showing the distribution of encoded values of race and religion \gradiend\ models evaluated against a broad set of datasets. The data of these plots has been used to compute \cormf\ and \corenc\ in Table~\ref{tab:encoded-values}.


\subsection{\rev{Stability of Encoded Values}}

\begin{figure}[!tp]
    \centering
    \includegraphics[width=\textwidth]{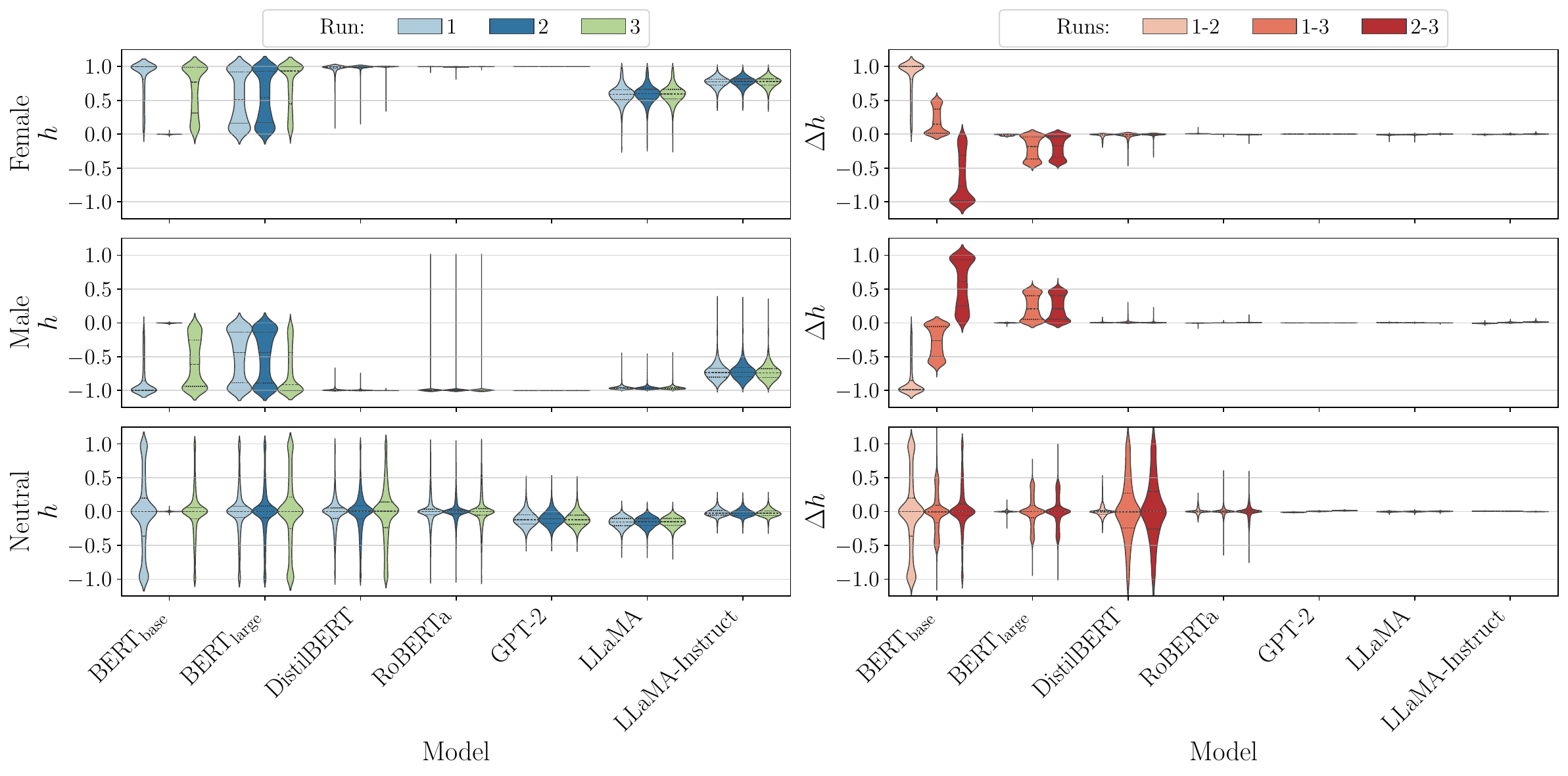}
    
    \caption{\rev{Distribution of encoded values $h$ (left) and their sample-wise difference $\Delta h$ (right) across three \gradiend\ training runs for gender.}}
    \label{fig:distribution-stability}
\end{figure}

\begin{table}[!t]
    \centering
    \tiny
    \caption{\rev{Stability analysis of encoded values across three \gradiend\ training runs for gender.}}
    \label{tab:distribution-stability}
    \begin{tabular}{lrrrrrrrr}\toprule
         & \multicolumn{4}{c}{\textbf{\corenc $\uparrow$}} & \multicolumn{4}{c}{\textbf{Mean Absolute Difference of Encoded Values $\downarrow$}}  \\\cmidrule(lr){2-5} \cmidrule(lr){6-9}
        \textbf{Model}  & \textbf{Run 1} & \textbf{Run 2} & \textbf{Run 3} & \textbf{Mean} & \textbf{Runs 1-2} & \textbf{Runs 1-3} & \textbf{Runs 2-3} & \textbf{Mean} \\ \midrule
\bertbase & 0.713 & 0.076 & 0.706 & 0.498 & 0.558 & 0.212 & 0.350 & 0.373 \\
\bertlarge & 0.621 & 0.622 & 0.660 & 0.635 & 0.008 & 0.173 & 0.168 & 0.117 \\
\distilbert & 0.939 & 0.862 & 0.860 & 0.887  & 0.035 & 0.245 & 0.256 & 0.179 \\
\roberta &  0.964 & 0.977 & 0.953 & 0.965 & 0.019 & 0.018 & 0.036 & 0.024  \\
\gpttwo & \textbf{0.984} & \textbf{0.985} & \textbf{0.984} & \textbf{0.984}  & 0.007 & \textbf{0.002} & 0.009 & 0.006 \\
\llama &  0.981 & 0.983 & 0.983 & 0.982 & \textbf{0.005} & 0.004 & \textbf{0.002} & 0.004 \\
\llamai & 0.977 & 0.976 & 0.977 & 0.976 & \textbf{0.005} & 0.003 & 0.003 & \textbf{0.004} \\
         \bottomrule
    \end{tabular} 
\end{table}

\rev{We analyze the stability of the feature neuron by examining the encodings from three independently trained gender \gradiend\ models for each base model. Figure~\ref{fig:distribution-stability} shows the distribution of these encoded values, along with sample-wise differences to highlight run-to-run variation, and Table~\ref{tab:distribution-stability} summarizes key statistics.}

\begin{figure}[!t]
    \centering
    \begin{subfigure}[t]{0.49\textwidth}
    \includegraphics[width=\linewidth]{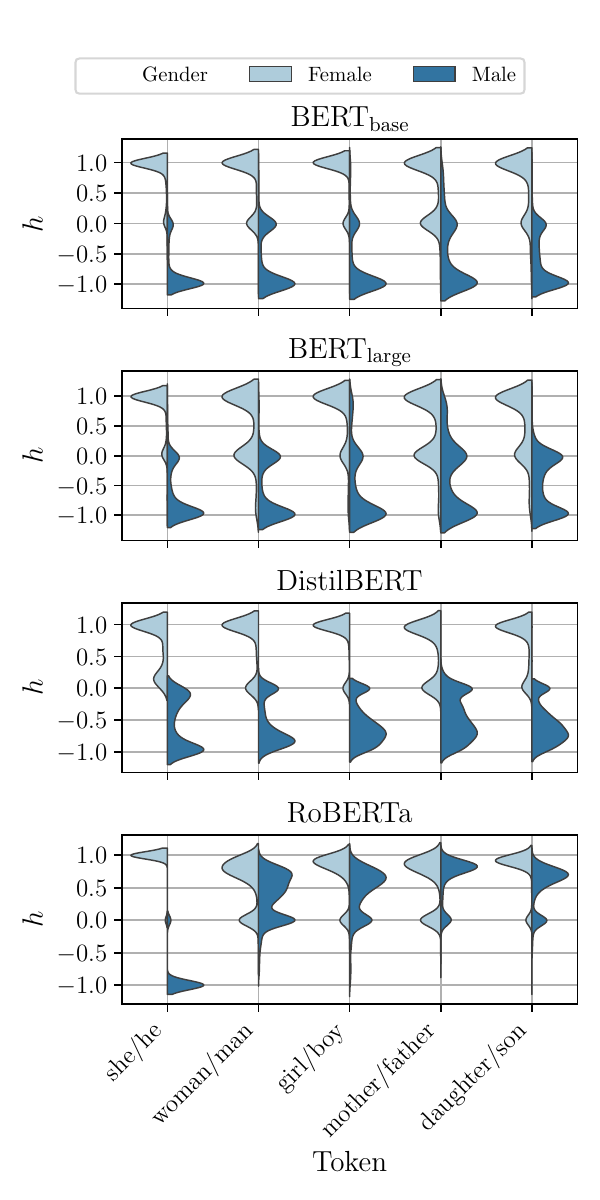}
    \caption{\rev{Encoder-only models.}}
    \end{subfigure}
    \begin{subfigure}[t]{0.49\textwidth}
        
    \includegraphics[width=\linewidth]{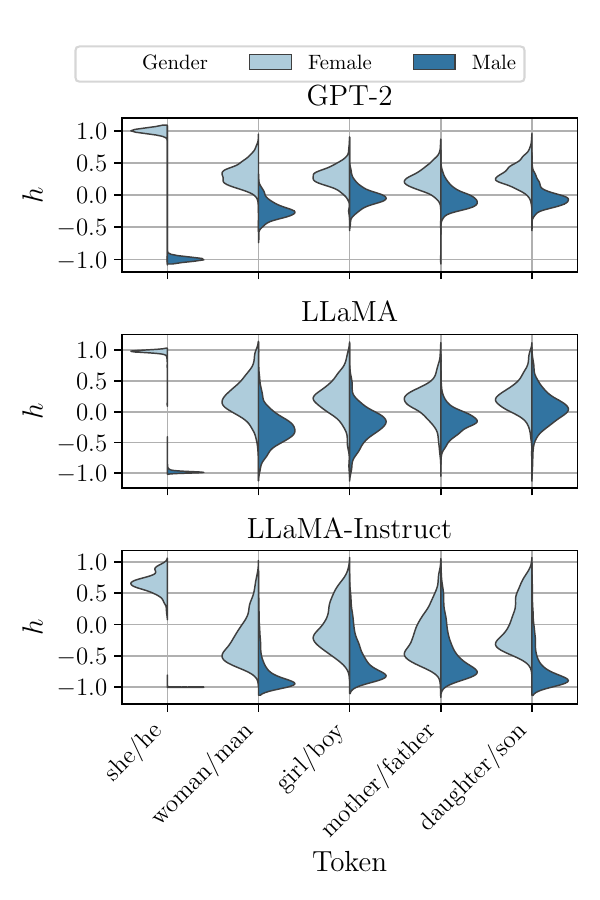}
    \caption{\rev{Decoder-only models.}}
    \end{subfigure}
    
    \caption{\rev{Distribution of encoded values of gender \glspl{gradae} for \wikigender.}}
    \label{fig:encodings-other-data}
\end{figure}

\rev{With the exception of the  \acrshort{bert}-based models, the feature neuron is generally stable across female, male, and neutral inputs. \distilbert\ and \roberta\ show some variability for neutral inputs across runs, while \gpttwo, \llama, and \llamai\ exhibit a mean absolute encoding difference below~$1\%$.}

\rev{For \bertlarge, the third run achieves notably higher performance than the first two, which are fairly similar to each other. In contrast, \bertbase\ shows a non-convergent second run, resulting in large differences compared to the other runs.}

\subsection{\rev{Generalization of Encoded Values}}

\rev{We further analyze how the encoder generalizes to unseen inputs, considering two aspects: \begin{enumerate*}[label=\textbf{(\arabic*)}]
    \item the input sentences originate from a dataset different from the one used during training, and
    \item the evaluation involves gender-related target tokens beyond the training pair \emph{he/she}.
\end{enumerate*} 
Therefore, we use \wikigender\ as a dataset (see Appendix~\ref{app:data-wikigender}).
}

\rev{
Figure~\ref{fig:encodings-other-data} shows the distribution of encoded values for our seven gender \glspl{gradae}. The she/he encoding learned during the training transfers well to \wikigender, indicating that the feature is not tied to the specific structure, linguistic style, and gender-filtered property of \genter. 
}

\rev{
For \bertbase, \bertlarge, and \distilbert, the learned feature also generalizes to other gendered token pairs such as \emph{woman/man}, 
though the separation is a bit weaker than for \emph{she/he}, as more samples are falsely encoded as neutral (i.e., around $0.0$). 
A plausible explanation is that masking \emph{he/she} yields a highly constrained prediction space, as only a few tokens fit the syntactic and semantic context, whereas masking, for instance, \emph{woman/man} allows usually a broader set of contextually plausible alternatives (e.g., \emph{girl/boy}), including gender neutral terms like \emph{person}.
Interestingly, \roberta\ behaves differently: it appears to encode a narrow \emph{she/he}-specific feature rather than a broader gender feature.
}

\rev{
For decoder-only models, the generalization is weaker for non-she/he pairs but still visible, as the female-associated tokens tend to encode to larger values than their male counterparts.
This less extreme encoding is expected because these models can only use the left context of the target term.
Considering the non-\emph{she/he} token pairs for \gpttwo\ and \llama, they show a mostly symmetric distribution around zero with smaller magnitude than for \emph{she/he}, indicating weaker separation.
In contrast, \llamai\ still shows a female-male distinction, but the distributions are shifted toward the male side (i.e., toward~$-1$).
}

\rev{Overall, the results indicate that the features learned by \gls{gradae} generalize, but that future work should explore training \glspl{gradae} using multiple facets, i.e., not only a single type of counterfactual (e.g., \emph{she/he}), but also other in parallel, like \emph{woman/man} to possibly find a more general feature representation.}

\section{Decoder as Bias-Changer}\label{app:decoder-as-bias-changer}
\begin{figure}
    \centering
    \begin{subfigure}[t]{\textwidth}
        \includegraphics[width=\linewidth]{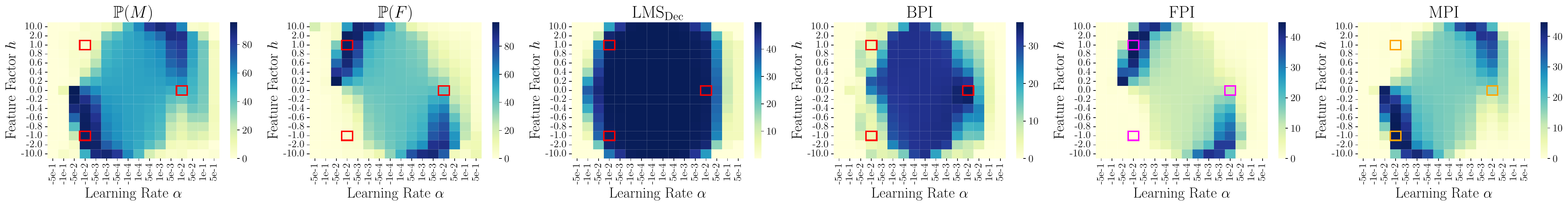}
        \caption{\bertbase}\label{fig:model-selection-gender-bert-base}
    \end{subfigure}

  \begin{subfigure}[t]{\textwidth}
        \includegraphics[width=\linewidth]{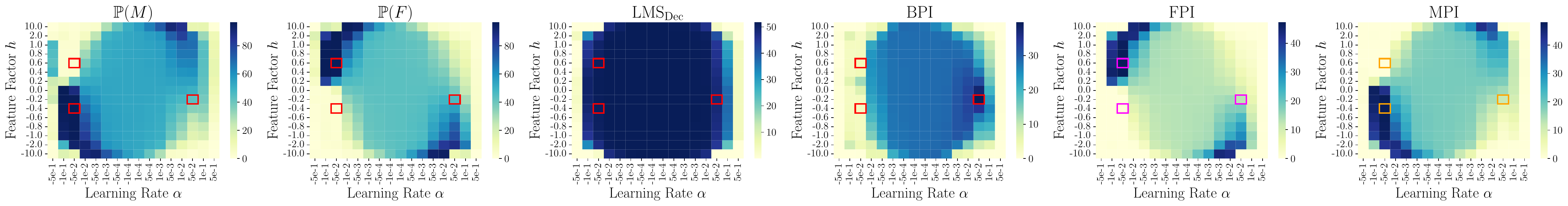}
        \caption{\bertlarge}
    \end{subfigure}

     \begin{subfigure}[t]{\textwidth}
        \includegraphics[width=\linewidth]{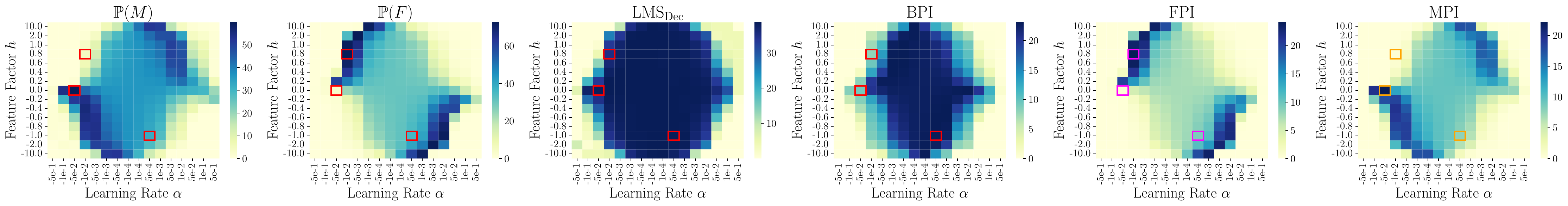}
        \caption{\distilbert}
    \end{subfigure}

     \begin{subfigure}[t]{\textwidth}
        \includegraphics[width=\linewidth]{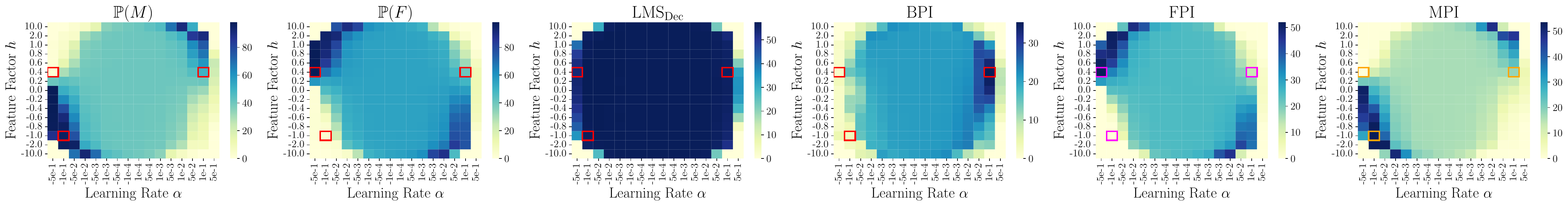}
        \caption{\roberta}
    \end{subfigure}

     \begin{subfigure}[t]{\textwidth}
        \includegraphics[width=\linewidth]{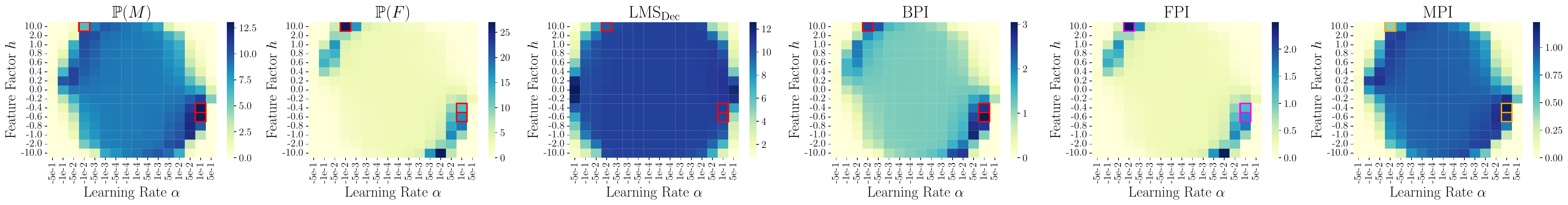}
        \caption{\gpttwo}
    \end{subfigure}

     \begin{subfigure}[t]{\textwidth}
        \includegraphics[width=\linewidth]{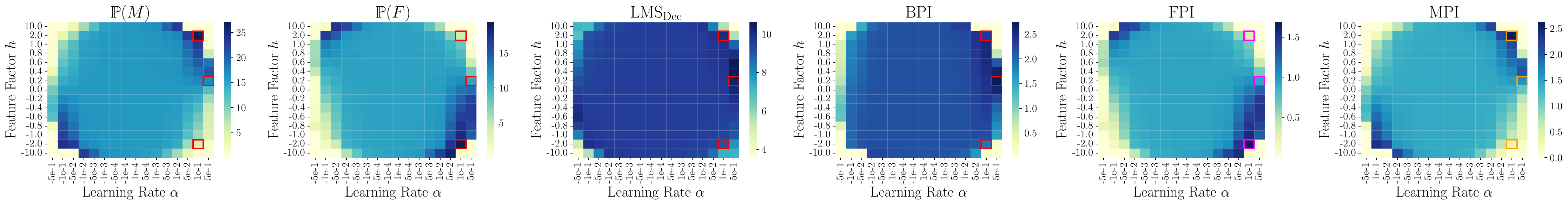}
        \caption{\llama}
    \end{subfigure}

     \begin{subfigure}[t]{\textwidth}
        \includegraphics[width=\linewidth]{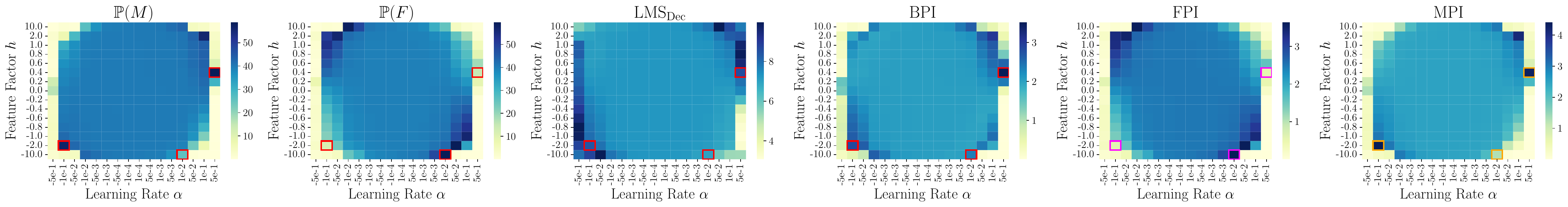}
        \caption{\llamai}
    \end{subfigure}
    
    \caption{Metrics for changed models based on the gender \glspl{gradae} with varying feature factor and learning rate. The cells with the best \acrshort{bpi}~\csquare{bpicolor}, \acrshort{fpi}~\csquare{fpicolor}, and \acrshort{mpi}~\csquare{mpicolor} are highlighted across all subplots. All values are reported as percentages.}
    \label{fig:changed-gender-models}
\end{figure}

Similar to Figure~\ref{fig:changed_models-main-bert-base-cased}, we present the results for all gender models in Figure~\ref{fig:changed-gender-models}. We further report the selected race and religion models in Figures~\ref{fig:changed-model-bert-base-cased-v7}-\ref{fig:changed-model-Llama-3.2-3B-Instruct-v5}. 

\generateRRPlotsPerModel{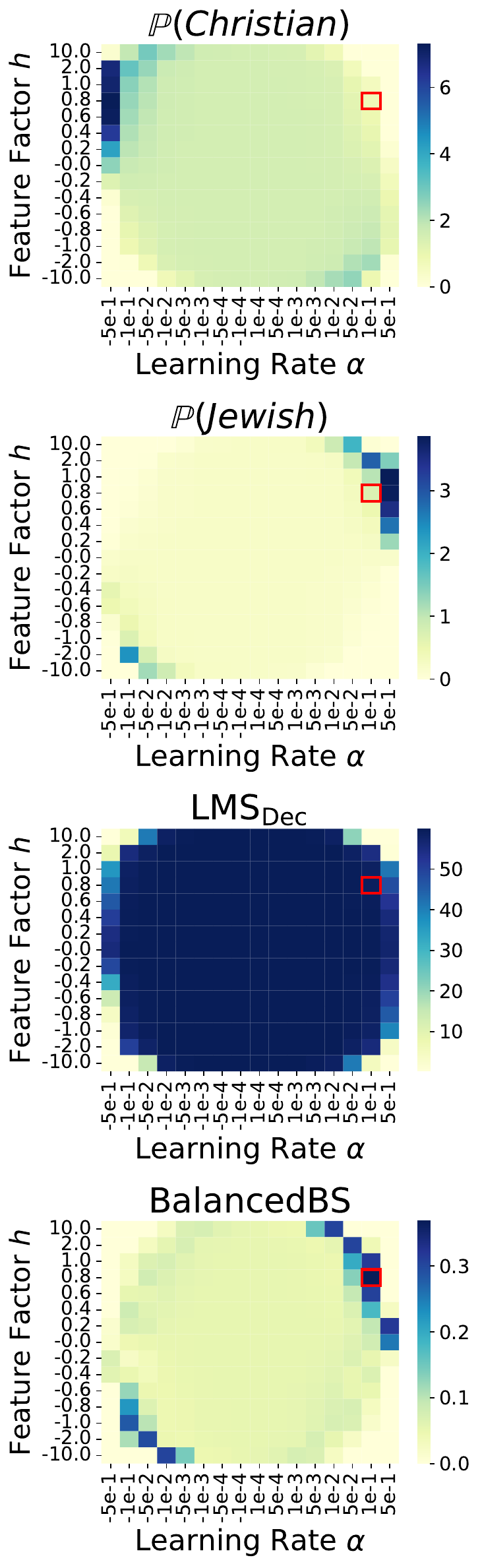}{\bertbase}
\generateRRPlotsPerModel{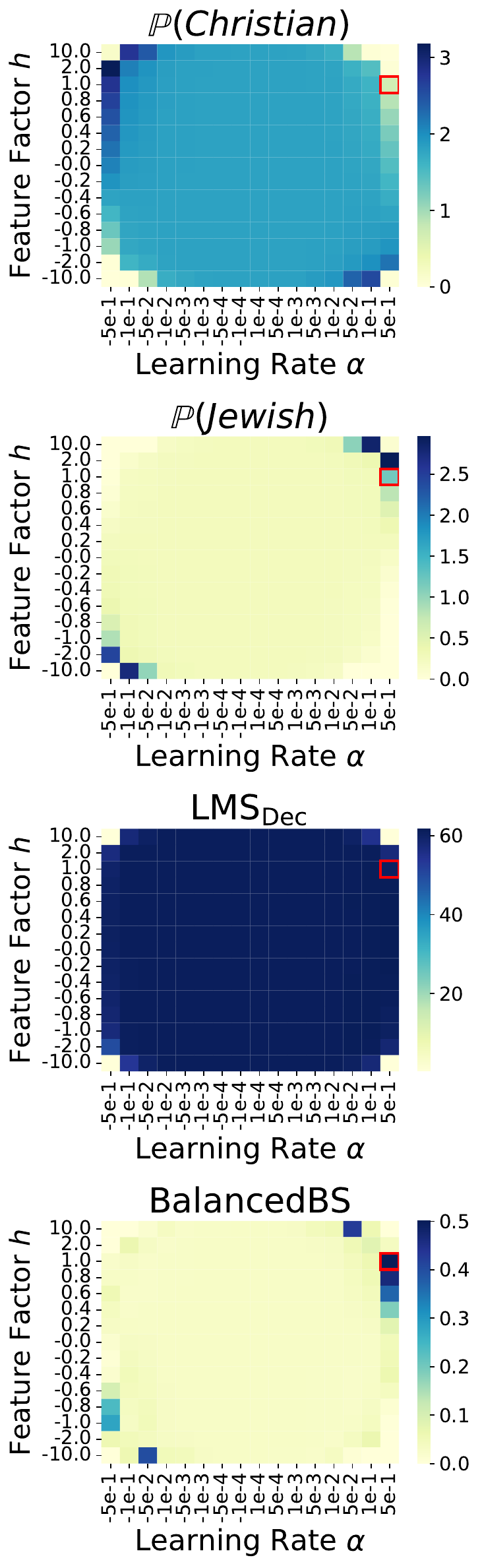}{\bertlarge}
\generateRRPlotsPerModel{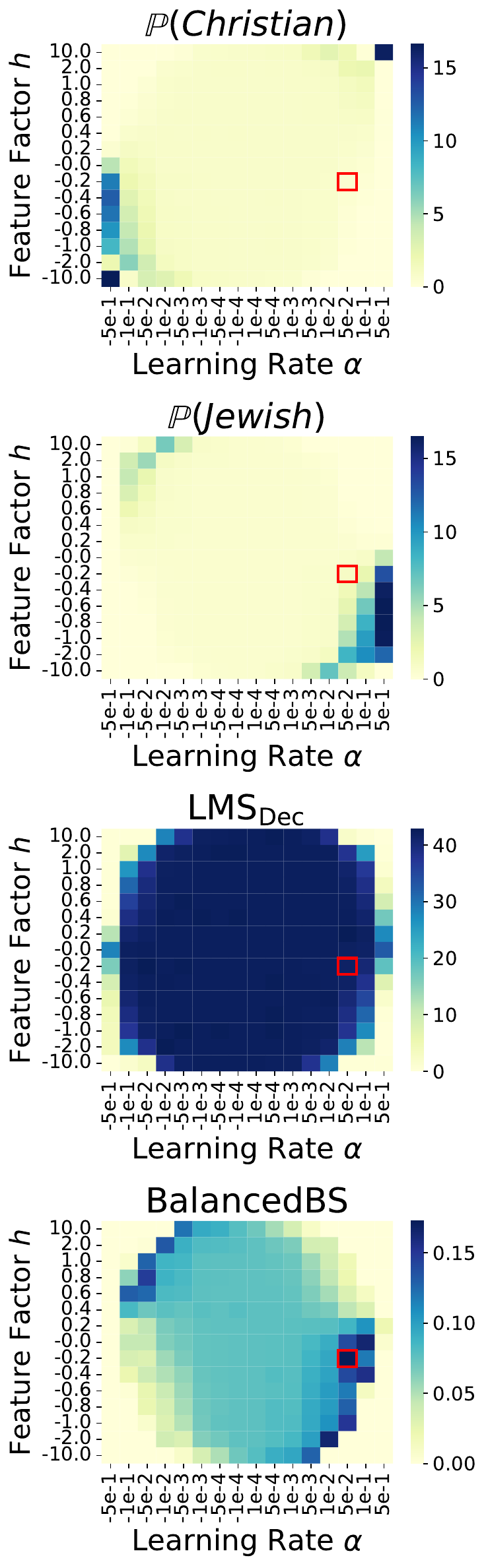}{\distilbert}
\generateRRPlotsPerModel{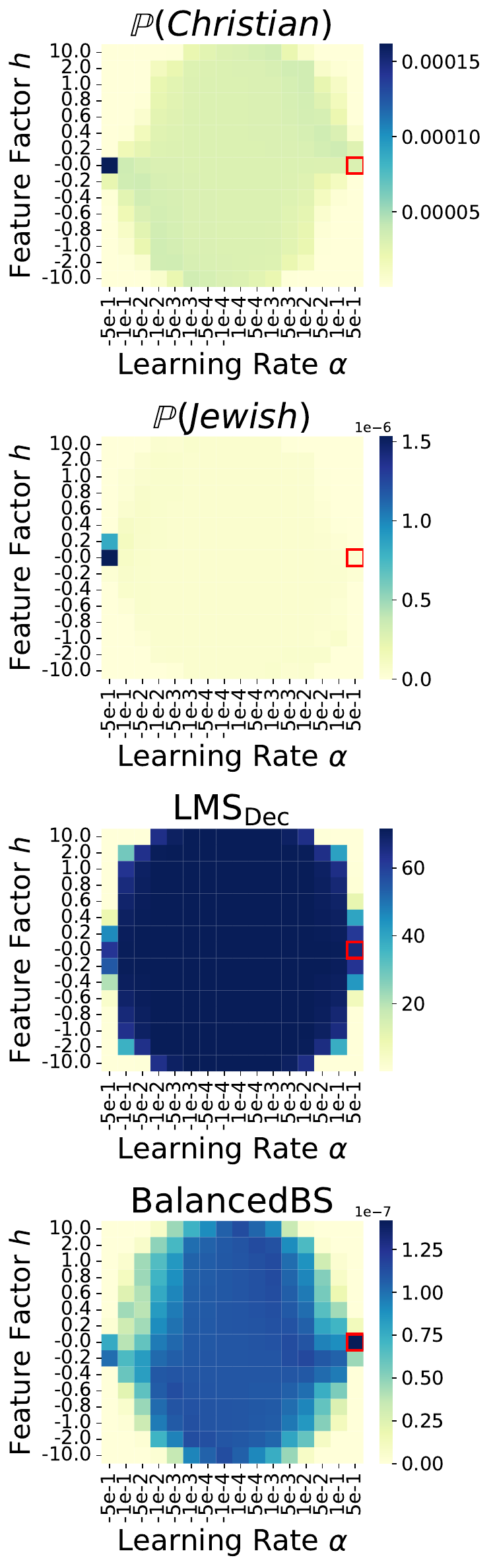}{\roberta}
\generateRRPlotsPerModel{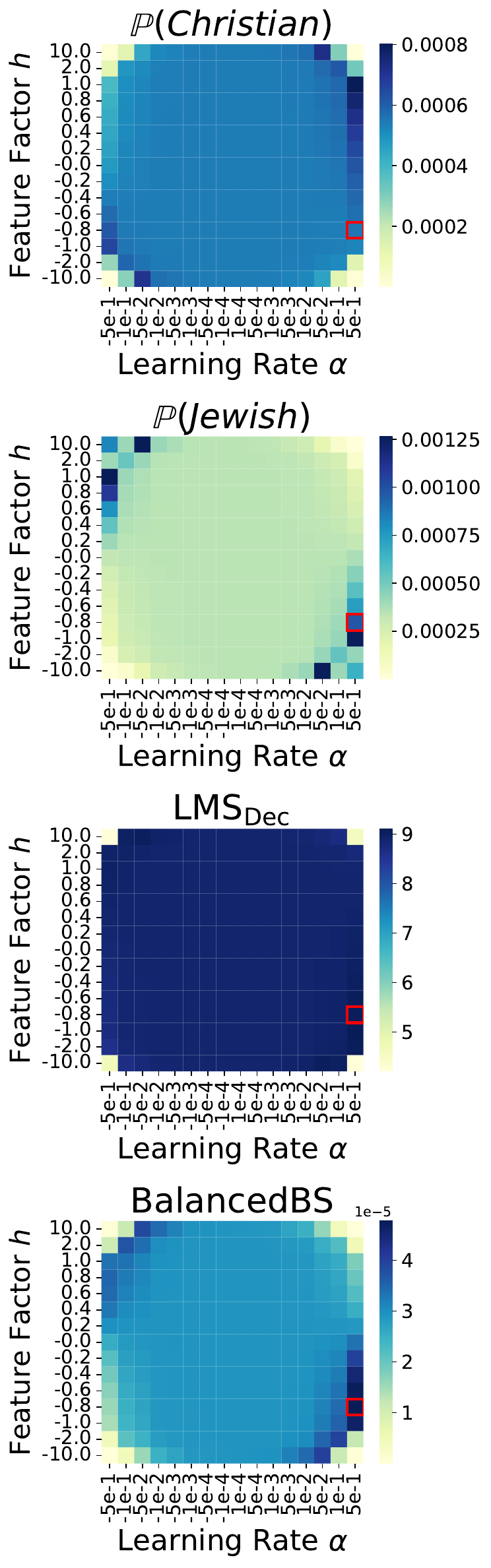}{\gpttwo}
\generateRRPlotsPerModel{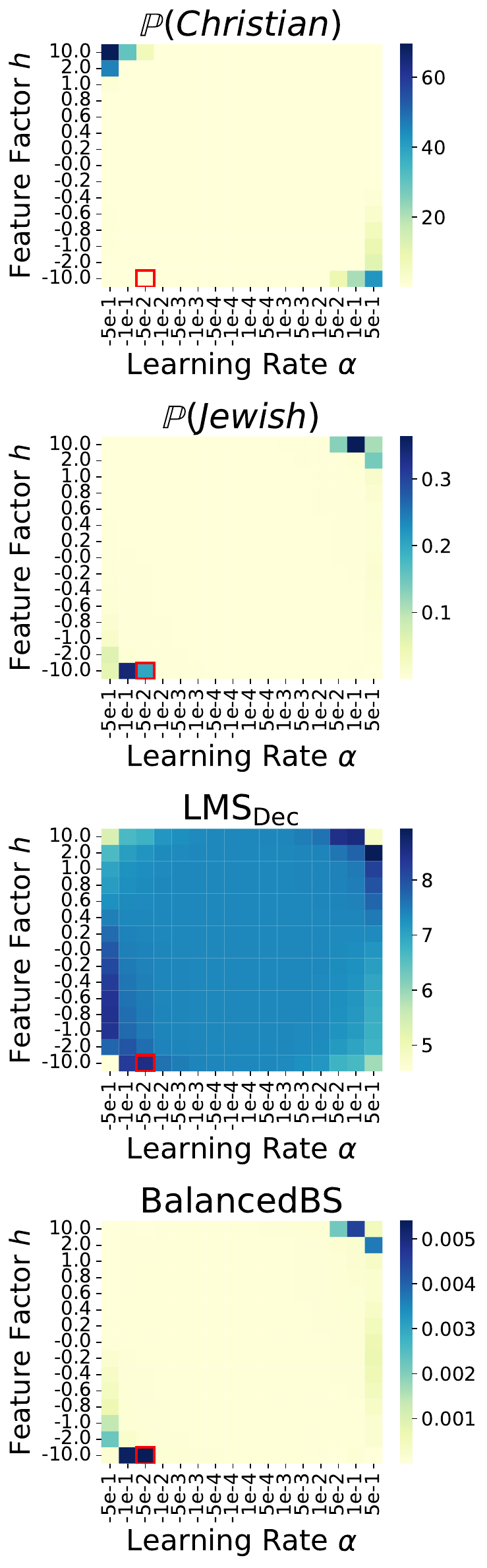}{\llama}
\generateRRPlotsPerModel{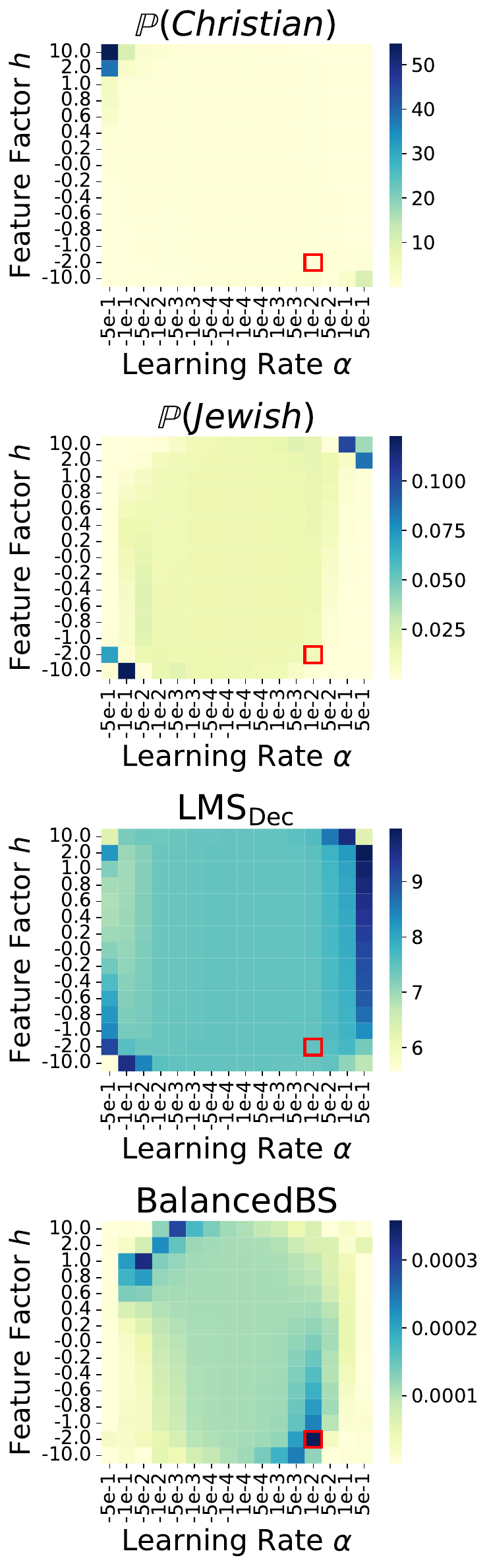}{\llamai}

\begin{table*}[p]
    \centering
    \tablefontsize
    \caption{Selected gender-debiased, female-biased and male-biased models based on \gls{bpi}, \gls{fpi} and \gls{mpi}.}\label{tab:eval:selected-models-full}
\begin{tabular}{lrrrrrrrr}
\toprule
 \textbf{Model}    & \textbf{FF} $h$   & \textbf{LR} $\alpha$   & \textbf{$\mathbb{P}(F)$}   & \textbf{$\mathbb{P}(M)$}   & \textbf{\accdec}   & \textbf{\bpi}   & \textbf{\fpi}   & \textbf{\mpi}   \\
\midrule
 \bertbase         & 0.0               & 0.0                    & 0.413                      & 0.538                      & 0.496              & 0.322          & 0.110          & 0.172          \\
 \, + \gradiendbpi & 0.0               & 1e-02                  & 0.424                      & 0.494                      & 0.491              & 0.363          & 0.111          & 0.145          \\
 \, + \gradiendfpi & 1.0               & -1e-02                 & 0.948                      & 0.034                      & 0.490              & 0.041          & 0.449          & 0.001          \\
 \, + \gradiendmpi & -1.0              & -1e-02                 & 0.031                      & 0.940                      & 0.490              & 0.043          & 0.001          & 0.446          \\
\midrule
 \bertlarge        & 0.0               & 0.0                    & 0.432                      & 0.534                      & 0.511              & 0.280          & 0.132          & 0.184          \\
 \, + \gradiendbpi & -0.2              & 5e-02                  & 0.461                      & 0.434                      & 0.486              & 0.397          & 0.128          & 0.115          \\
 \, + \gradiendfpi & 0.6               & -5e-02                 & 0.960                      & 0.031                      & 0.508              & 0.036          & 0.473          & 0.001          \\
 \, + \gradiendmpi & -0.4              & -5e-02                 & 0.024                      & 0.963                      & 0.511              & 0.031          & 0.000          & 0.480          \\
\midrule
 \distilbert       & 0.0               & 0.0                    & 0.307                      & 0.361                      & 0.392              & 0.230          & 0.075          & 0.096          \\
 \, + \gradiendbpi & -1.0              & 5e-04                  & 0.358                      & 0.320                      & 0.386              & 0.231          & 0.092          & 0.078          \\
 \, + \gradiendfpi & 0.8               & -1e-02                 & 0.724                      & 0.045                      & 0.350              & 0.080          & 0.241          & 0.004          \\
 \, + \gradiendmpi & 0.0               & -5e-02                 & 0.036                      & 0.599                      & 0.384              & 0.092          & 0.005          & 0.221          \\
\midrule
 \roberta          & 0.0               & 0.0                    & 0.555                      & 0.404                      & 0.573              & 0.208          & 0.249          & 0.162          \\
 \, + \gradiendbpi & 0.4               & 1e-01                  & 0.402                      & 0.499                      & 0.560              & 0.354          & 0.127          & 0.181          \\
 \, + \gradiendfpi & 0.4               & -5e-01                 & 0.980                      & 0.016                      & 0.539              & 0.019          & 0.520          & 0.000          \\
 \, + \gradiendmpi & -1.0              & -1e-01                 & 0.023                      & 0.966                      & 0.568              & 0.032          & 0.001          & 0.535          \\
 \midrule
  \gpttwo          & 0.0               & 0.0                    & 0.028                      & 0.089                      & 0.107              & 0.012          & 0.003          & 0.009          \\
 \, + \gradiendbpi & -0.6              & 1e-01                  & 0.172                      & 0.130                      & 0.106              & 0.031          & 0.016          & 0.011          \\
 \, + \gradiendfpi & 10.0              & -1e-02                 & 0.270                      & 0.061                      & 0.099              & 0.025          & 0.025          & 0.004          \\
 \, + \gradiendmpi & -0.4              & 1e-01                  & 0.123                      & 0.130                      & 0.108              & 0.027          & 0.012          & 0.012          \\
 \midrule
 \llama            & 0.0               & 0.0                    & 0.111                      & 0.162                      & 0.095              & 0.022          & 0.009          & 0.014          \\
 \, + \gradiendbpi & 0.2               & 5e-01                  & 0.109                      & 0.174                      & 0.104              & 0.027          & 0.009          & 0.016          \\
 \, + \gradiendfpi & -2.0              & 1e-01                  & 0.194                      & 0.050                      & 0.091              & 0.019          & 0.017          & 0.004          \\
 \, + \gradiendmpi & 2.0               & 1e-01                  & 0.042                      & 0.271                      & 0.101              & 0.024          & 0.003          & 0.026          \\
 \midrule
 \llamai        & 0.0               & 0.0                    & 0.397                      & 0.399                      & 0.073              & 0.021          & 0.025          & 0.025          \\
 \, + \gradiendbpi & 0.4               & 5e-01                  & 0.147                      & 0.587                      & 0.086              & 0.035          & 0.005          & 0.043          \\
 \, + \gradiendfpi & -10.0             & 1e-02                  & 0.594                      & 0.257                      & 0.070              & 0.024          & 0.036          & 0.013          \\
 \, + \gradiendmpi & -2.0              & -1e-01                 & 0.116                      & 0.569                      & 0.084              & 0.028          & 0.006          & 0.044          \\
\bottomrule 
\end{tabular}
\end{table*}

\begin{table*}[!p]
    \centering
    \tablefontsize
    \caption{Selected debiased models based on \gls{bpi} for race and religion. Classes $A$ and $B$ refer to the classes of the $\gradiend_{A/B}$.}\label{tab:eval:selected-models-full-rr}
\begin{tabular}{lrrrrrrrr}
\toprule
 \textbf{Model}    & \textbf{FF} $h$   & \textbf{LR} $\alpha$   & Base \textbf{$\mathbb{P}(A)$} & \textbf{$\mathbb{P}(A)$}   & Base \textbf{$\mathbb{P}(B)$}  & \textbf{$\mathbb{P}(B)$}   & \textbf{\accdec}   & \textbf{\bpi} \\ 

\midrule
\multicolumn{9}{c}{$\gradiend_{Asian/Black}$} \\
\midrule
\bertbase & -0.0 & -0.5 & 8.6e-8 & 8.6e-8 & 8.1e-4 & 8.1e-4 & 0.587 & 2.2e-7 \\
\bertlarge & 1.0 & -0.5 & 3.8e-9 & 3.8e-9 & 1.8e-3 & 1.8e-3 & 0.635 & 3.8e-8 \\
\distilbert & -0.6 & 0.1 & 1.2e-5 & 1.2e-5 & 1.1e-3 & 1.1e-3 & 0.429 & 2.7e-6 \\
\roberta & -0.0 & -0.5 & 9.6e-7 & 9.6e-7 & 1.5e-6 & 1.5e-6 & 0.591 & 7.5e-7 \\
\gpttwo & 1.0 & -0.5 & 5.7e-7 & 5.7e-7 & 1.2e-5 & 1.2e-5 & 0.095 & 2.0e-7 \\
\llama & -0.8 & -0.5 & 3.9e-5 & 3.9e-5 & 2.4e-5 & 2.4e-5 & 0.089 & 4.5e-5 \\
\llamai & -10.0 & 0.5 & 7.6e-5 & 7.6e-5 & 3.6e-4 & 3.6e-4 & 0.068 & 1.0e-5 \\
\midrule
\multicolumn{9}{c}{$\gradiend_{Asian/White}$} \\
\midrule
\bertbase & -0.2 & -0.5 & 4.7e-8 & 4.7e-8 & 3.8e-4 & 3.8e-4 & 0.570 & 7.8e-7 \\
\bertlarge & 2.0 & -0.5 & 3.3e-9 & 3.3e-9 & 2.2e-4 & 2.2e-4 & 0.627 & 3.0e-8 \\
\distilbert & 0.4 & -0.1 & 8.4e-6 & 8.4e-6 & 3.7e-4 & 3.7e-4 & 0.421 & 5.0e-6 \\
\roberta & -0.0 & -0.5 & 1.7e-6 & 1.7e-6 & 2.5e-6 & 2.5e-6 & 0.649 & 1.4e-6 \\
\gpttwo & 0.2 & 0.5 & 6.1e-7 & 6.1e-7 & 1.3e-5 & 1.3e-5 & 0.089 & 2.4e-7 \\
\llama & -1.0 & -0.5 & 3.0e-5 & 3.0e-5 & 2.2e-5 & 2.2e-5 & 0.092 & 4.7e-5 \\
\llamai & -2.0 & 0.5 & 8.0e-5 & 8.0e-5 & 1.2e-3 & 1.2e-3 & 0.077 & 7.6e-6 \\
\midrule
\multicolumn{9}{c}{$\gradiend_{Black/White}$} \\
\midrule
\bertbase & -1.0 & -0.1 & 9.8e-3 & 9.8e-3 & 9.8e-3 & 9.8e-3 & 0.604 & 4.2e-3 \\
\bertlarge & 0.4 & -0.5 & 1.2e-2 & 1.2e-2 & 1.2e-2 & 1.2e-2 & 0.627 & 4.1e-3 \\
\distilbert & -0.2 & 0.1 & 5.4e-3 & 5.4e-3 & 5.4e-3 & 5.4e-3 & 0.441 & 1.7e-3 \\
\roberta & 0.8 & -0.5 & 5.8e-6 & 5.8e-6 & 5.8e-6 & 5.8e-6 & 0.710 & 7.7e-6 \\
\gpttwo & -0.0 & -0.5 & 1.4e-5 & 1.4e-5 & 1.4e-5 & 1.4e-5 & 0.090 & 1.1e-6 \\
\llama & -0.2 & -0.5 & 3.0e-5 & 3.0e-5 & 3.0e-5 & 3.0e-5 & 0.082 & 3.1e-6 \\
\llamai & 1.0 & 0.5 & 1.1e-3 & 1.1e-3 & 1.1e-3 & 1.1e-3 & 0.067 & 2.7e-5 \\
\midrule
\multicolumn{9}{c}{$\gradiend_{Christian/Jewish}$} \\
\midrule
\bertbase & 0.8 & 0.1 & 1.5e-2 & 1.5e-2 & 2.2e-3 & 2.2e-3 & 0.592 & 3.7e-3 \\
\bertlarge & 1.0 & 0.5 & 1.8e-2 & 1.8e-2 & 2.6e-3 & 2.6e-3 & 0.607 & 5.0e-3 \\
\distilbert & -0.2 & 0.1 & 8.4e-3 & 8.4e-3 & 6.2e-3 & 6.2e-3 & 0.416 & 1.7e-3 \\
\roberta & -0.0 & 0.5 & 3.0e-7 & 3.0e-7 & 5.7e-10 & 5.7e-10 & 0.665 & 1.4e-9 \\
\gpttwo & -0.8 & 0.5 & 5.4e-6 & 5.4e-6 & 3.4e-6 & 3.4e-6 & 0.091 & 4.8e-7 \\
\llama & -10.0 & -0.1 & 7.9e-5 & 7.9e-5 & 1.4e-5 & 1.4e-5 & 0.085 & 5.4e-5 \\
\llamai & -2.0 & 0.0 & 4.3e-3 & 4.3e-3 & 1.5e-4 & 1.5e-4 & 0.075 & 3.6e-6 \\
\midrule
\multicolumn{9}{c}{$\gradiend_{Christian/Muslim}$} \\
\midrule
\bertbase & -2.0 & 0.1 & 1.1e-2 & 1.1e-2 & 1.5e-3 & 1.5e-3 & 0.585 & 3.3e-3 \\
\bertlarge & -0.8 & 0.5 & 1.2e-2 & 1.2e-2 & 2.0e-3 & 2.0e-3 & 0.611 & 3.6e-3 \\
\distilbert & -0.0 & 0.1 & 7.6e-3 & 7.6e-3 & 3.0e-3 & 3.0e-3 & 0.430 & 1.5e-3 \\
\roberta & 0.8 & 0.1 & 4.0e-8 & 4.0e-8 & 1.9e-8 & 1.9e-8 & 0.664 & 4.5e-9 \\
\gpttwo & -0.8 & -0.5 & 5.4e-6 & 5.4e-6 & 2.6e-5 & 2.6e-5 & 0.090 & 8.2e-7 \\
\llama & -0.6 & -0.5 & 7.4e-5 & 7.4e-5 & 3.5e-5 & 3.5e-5 & 0.072 & 3.1e-5 \\
\llamai & -0.4 & 0.5 & 3.9e-3 & 3.9e-3 & 1.5e-3 & 1.5e-3 & 0.094 & 9.3e-6 \\
\midrule
\multicolumn{9}{c}{$\gradiend_{Jewish/Muslim}$} \\
\midrule
\bertbase & 0.8 & -0.1 & 1.7e-3 & 1.7e-3 & 1.7e-3 & 1.7e-3 & 0.589 & 6.4e-4 \\
\bertlarge & 0.6 & 0.5 & 1.4e-3 & 1.4e-3 & 1.4e-3 & 1.4e-3 & 0.592 & 5.7e-4 \\
\distilbert & -0.4 & -0.1 & 5.2e-3 & 5.2e-3 & 5.2e-3 & 5.2e-3 & 0.433 & 5.7e-4 \\
\roberta & -0.2 & -0.5 & 3.4e-10 & 3.4e-10 & 3.4e-10 & 3.4e-10 & 0.706 & 4.5e-10 \\
\gpttwo & 1.0 & -0.5 & 3.4e-6 & 3.4e-6 & 3.4e-6 & 3.4e-6 & 0.093 & 1.0e-6 \\
\llama & 10.0 & -0.1 & 1.3e-5 & 1.3e-5 & 1.3e-5 & 1.3e-5 & 0.086 & 1.0e-5 \\
\llamai & 0.4 & 0.5 & 3.5e-4 & 3.5e-4 & 3.5e-4 & 3.5e-4 & 0.066 & 1.0e-5 \\

\bottomrule 
\end{tabular}
\end{table*}

Overall, a similar point-symmetric pattern can be recognized across all figures. 
However, the model selection is different compared to \bertbase, where \gls{fpi} and \gls{mpi} exhibit negated feature factors along with negative learning rates, while \gls{bpi} features a zero feature factor and a positive learning rate. 
Similar configurations exist across most models that outperform the base model with respect to \gls{bpi}, \gls{fpi}, and \gls{mpi}. The final selected models, however, perform even better with respect to our metrics, though they do not adhere to the expected pattern. Future research should explore the stability of these parameter choices without relying on a larger search grid.

The statistics for all selected gender models are reported in Table~\ref{tab:eval:selected-models-full}.
Interestingly, the difference in \gls{bpi} between \gradiendbpi\ and its base model is relatively small, whereas the corresponding differences for \gls{fpi} and \gls{mpi} are much larger, respectively. 
This observation suggests that biasing a model (in either direction) is easier than debiasing it.
Notably, \gls{fpi} approaches nearly zero for \gradiendmpi\ models, and \gls{mpi} similarly is close to zero for \gradiendfpi\ models.
Surprisingly, for the \roberta\ base model, $\P(F)>\P(M)$ holds true, unlike all other base models. 
This indicates a female bias in the given task, contradicting our expectation that language models typically exhibit male bias (although this bias direction is not captured by \gls{ss} and \gls{seat}). 

The statistics for the selected race and religion models are reported in Table~\ref{tab:eval:selected-models-full-rr}.

\section{Comparison to Other Debiasing Techniques}\label{app:comp}
This section provides supplementary details for Section~\ref{sec:eval-debiased}, which compares our method to existing debiasing techniques. 
To facilitate future comparisons with our approach, we release our gender-debiased models on Hugging Face (Table~\ref{tab:published-models}), where they achieve SoTA debiasing performance.

\begin{table}[!t]
    \centering
    \scriptsize
     \caption{Published gender debiased models on Hugging Face.}
    \label{tab:published-models}
    \begin{tabular}{ll}
    \toprule
        \textbf{Model} & \textbf{Identifier} \\ \midrule
        \bertbase\, + \gradiendbpi & \iflink \href{https://huggingface.co/aieng-lab/bert-base-cased-gradiend-gender-debiased}{\texttt{aieng-lab/bert-base-cased-gradiend-gender-debiased}} \else anonymous \fi \\ 
        \bertlarge\, + \gradiendbpi & \iflink \href{https://huggingface.co/aieng-lab/bert-large-cased-gradiend-gender-debiased}{\texttt{aieng-lab/bert-large-cased-gradiend-gender-debiased}} \else anonymous \fi \\
        \distilbert\, + \gradiendbpi & \iflink \href{https://huggingface.co/aieng-lab/distilbert-base-cased-gradiend-gender-debiased}{\texttt{aieng-lab/distilbert-base-cased-gradiend-gender-debiased}} \else anonymous \fi \\
        \roberta\, + \gradiendbpi & \iflink \href{https://huggingface.co/aieng-lab/roberta-large-gradiend-gender-debiased}{\texttt{aieng-lab/roberta-large-gradiend-gender-debiased}} \else anonymous \fi \\
        \gpttwo\, + \gradiendbpi & \iflink \href{https://huggingface.co/aieng-lab/gpt2-gradiend-gender-debiased}{\texttt{aieng-lab/gpt2-gradiend-gender-debiased}} \else anonymous \fi \\
        \llama\, + \gradiendbpi & \iflink \href{https://huggingface.co/aieng-lab/Llama-3.2-3B-gradiend-gender-debiased}{\texttt{aieng-lab/Llama-3.2-3B-gradiend-gender-debiased}} \else anonymous \fi \\
        \llamai\, + \gradiendbpi & \iflink \href{https://huggingface.co/aieng-lab/Llama-3.2-3B-Instruct-gradiend-gender-debiased}{\texttt{aieng-lab/Llama-3.2-3B-Instruct-gradiend-gender-debiased}} \else anonymous \fi \\
        \bottomrule
    \end{tabular}
\end{table}

\subsection{Implementation Details}\label{app:eval-impl-details}

For the evaluation of our gender-changed models on \acrshort{glue}, \acrshort{sglue}, \gls{seat}, \gls{ss}, and \gls{lms}, we primarily rely on the bias-bench implementation by \citet{meade2022empiricalsurveyeffectivenessdebiasing}, which we also use to compute and evaluate the baseline debiasing techniques: \cda, \dropout, \inlp, \sentencedebias, and \selfdebias. For implementation specifics and metric definitions, we refer the reader to the original work.

Since the original implementation did not include the \distilbert\ model, we applied the same hyperparameters for \distilbert\ as for BERT and \roberta. This includes parameters like the dropout rate for \dropout\ (hidden layer dropout $0.20$ and attention dropout $0.15$), and the number of iterations for \inlp\ ($n=80$). 
We also adapt this \inlp\ configuration for the \llama-based models.
In addition, we integrated \rlace\ \citep{rlace} and \leace\ \citep{leace} into bias-bench in analogy to \inlp, using their original implementations. For \rlace, we use a rank of 1. 
We release our modified version of bias-bench on GitHub\footnote{\iflink \url{https://github.com/aieng-lab/bias-bench}\else anonymous\fi}.

For evaluating the \llama-based models on \acrshort{glue} and \acrshort{sglue}, we use a zero-shot setting based on a gender-bias adapted version of the Python library lm-evaluation-harness\footnote{\iflink \url{https://github.com/aieng-lab/lm-evaluation-harness} \else anonymous\fi} \citep{eval-harness}. Since STS-B is a regression task, we omit it from the evaluation. For \llamai, we use no system prompt for all of these evaluations. For all non-\llama\ models, we follow the standard bias-bench settings and fine-tune them on all nine \acrshort{glue} and all eight \acrshort{sglue} tasks prior to evaluation.

We exclude \gradiendfpi\ and \gradiendmpi\ from combinations as they are not designed for debiasing, and we also exclude \rlace, \leace, and \selfdebias\ due to their generally weaker performance. \cda\ and \dropout\ variants are also excluded for \llama-based models due to the high cost of additional pretraining for these methods.

\subsection{Detailed Results}

We report the raw results for \gls{ss}, \gls{seat}, \gls{lms}, \acrshort{glue}, and \acrshort{sglue} in Tables~\ref{tab:eval:main-results-bert-base} to~\rev{\ref{tab:results-religion}}, covering \bertbase, \bertlarge, \distilbert, \roberta, \gpttwo,  \llama, and \llamai.
The proportional rank-based comparison in Table~\ref{tab:rank} is derived from these values. 
Additional sub-results and further information on \acrshort{glue} and \acrshort{seat} are provided in the following sections.

\rev{The proportional rank (Table~\ref{tab:rank}) for a metric $m$ (SS or SEAT) is derived from Tables~\ref{tab:eval:main-results-bert-base} to~\rev{\ref{tab:results-religion}} by first ranking the debiasing approaches for each base model. These integers are then converted to proportional ranks by dividing by the number of variants minus one, yielding scores in $[0, 1]$. This naturally accounts for differences in the number of variants across models. The mean proportional rank for $m$ is obtained by averaging over all base models, and the \emph{Mean} column in Table~\ref{tab:rank} reports the average of the mean proportional ranks for SS and SEAT.} 

\rev{The difference values for the language modeling metrics in Table~\ref{tab:rank} (\gls{lms}, \acrshort{glue}, \acrshort{sglue}) are computed by first taking the score difference for each base model and then averaging these differences across all base models used by a variant. Some debiasing variants cannot be applied to all models (all \dropout- and \cda-based variants and both \llama-based models), so the number of scores entering the average differs across variants. This makes the \emph{absolute} mean scores not directly comparable for these cases. A reader might expect the reported change to be the \emph{difference of averaged scores}, but this would not correctly reflect situations where variants use different sets of base models. Reporting the \emph{average of model-wise differences} ensures that the reported relative changes remain meaningful for assessing whether a variant negatively affects language modeling performance, which is the main concern for this study.}

Unlike previous studies, we report all metric scores with a $95\%$ confidence interval, computed via bootstrapping~\citep{davison1997bootstrap} from the raw prediction values, providing a more robust comparison of model performances.
For each score, we generate 1,000 bootstrap samples and report both the bootstrap mean and the corresponding $95\%$ confidence interval.
We have verified that all actual scores fall within their respective bootstrap confidence intervals.

Statistically significant improvements (i.e., non-overlapping confidence intervals compared to the baseline) are indicated in \emph{italics}, while the best score for each base model is highlighted in \textbf{bold}.
In general, the comparison of debiasing approaches is challenging due to the high uncertainty and variance across different gender-bias metrics. Therefore, we reported the rank-based aggregated score in Table~\ref{tab:rank} to enable more robust comparisons.
Notably, with confidence intervals as context, the effectiveness of existing debiasing methods appears less clear than suggested by prior research \citep{meade2022empiricalsurveyeffectivenessdebiasing}.

\subsection{GLUE}

For \acrshort{glue} \citep{glue}, the reported score in Tables~\ref{tab:eval:main-results-bert-base} and \ref{tab:results-religion} is an aggregate of its subscores, which are detailed in Tables~\ref{tab:eval:glue} and~\ref{tab:glue-religion}. Due to space constraints, the confidence intervals for individual sub-tasks are not shown per model; however, Table~\ref{tab:eval:glue:ci} presents the confidence margin ranges for each sub-task across all \acrshort{glue} evaluations of this study. 
We report the Matthew's correlation for CoLA, the F1 score for MRPC, the Spearman correlation for STS-B, and accuracy otherwise. 
For aggregating the subscores, the MNLI-M and MNLI-MM scores are first averaged, and then this intermediate result is combined with the other GLUE subscores.

We follow the same training configurations as \cite{meade2022empiricalsurveyeffectivenessdebiasing} though we evaluate twice per epoch and select the best performing model based on loss at the end of the three-epoch training.

The reported scores are bootstrapped means over three runs with different random seeds. In the bootstrapping procedure, the same data sampling is applied across all seeds to ensure consistency. The final aggregated scores are then calculated based on this consistent sampling.

Table~\ref{tab:eval:glue:ci} highlights the relationship between the number of validation samples and the confidence of a computed score: tasks with fewer validation samples generally exhibit wider confidence intervals, reflecting greater variability and reduced reliability in their reported scores.

\begin{table}[!t]
    \centering
\tablefontsize
        \caption{Minimal and maximal confidence margin of error (in percentages) for GLUE and its subscores, based on the results of Table~\ref{tab:eval:glue} to~\ref{tab:glue-religion}, sorted by number of validation samples.}
    \label{tab:eval:glue:ci}
\begin{tabular}{lrrr}
\toprule
 \textbf{Task}           &   \textbf{Min (\%)} &  \textbf{Max (\%)} & \textbf{\# Samples} \\
\midrule
 GLUE    &       1.02 &       2.00 &        69,711 \\
 \midrule
 QQP     &       0.23 &       0.47 &        40,430 \\
 MNLI-MM &       0.48 &       1.02 &         9,832 \\
 MNLI-M  &       0.46 &       1.02 &         9,815 \\
 QNLI    &       0.53 &       1.40 &         5,463 \\
 STSB    &       0.80 &       1.63 &         1,500 \\
 CoLA    &       0.00 &       6.27 &         1,043 \\
 SST-2   &       1.21 &       3.36 &          872 \\
 MRPC    &       1.39 &       5.96 &          408 \\
 RTE     &       3.18 &       6.25 &          277 \\
 WNLI    &       4.52 &      11.94 &           71 \\
\bottomrule
\end{tabular}
\end{table}

\subsection{SuperGLUE}
We compute \acrshort{sglue} \citep{superglue} scores following the same settings as for \acrshort{glue}.
Crucially, the ReCoRD task is modeled as a span-selection problem and MultiRC as a binary sequence-classification problem by pairing each candidate answer with its question. 
For bootstrapping for these two tasks, examples are always added along with all their associated candidate answers to preserve the task structure..

Sub-scores for \acrshort{sglue} are reported in Tables~\ref{tab:eval:superglue} to \ref{tab:eval-superglue-religion}.
As with \acrshort{glue}, Table~\ref{tab:eval:super-glue:ci} summarizes confidence intervals across all evaluated models in this study.

\begin{table}[!t]
    \centering
\tablefontsize
        \caption{Minimal and maximal confidence margin of error (in percentages) for \acrshort{sglue} and its subscores, based on the results of Table~\ref{tab:eval:superglue} to~\ref{tab:eval-superglue-religion}, sorted by number of validation samples.}
    \label{tab:eval:super-glue:ci}
\begin{tabular}{lrrr}
\toprule
 \textbf{Task}           &   \textbf{Min (\%)} &  \textbf{Max (\%)} & \textbf{\# Samples} \\
\midrule
 SuperGLUE &       1.18 &       2.37 &        19,293 \\
 \midrule
 ReCoRD    &       0.01 &       1.03 &        10,000 \\
 MultiRC   &       0.14 &       1.63 &         4,848 \\
 BoolQ     &       1.03 &       1.64 &         3,270 \\
 WiC       &       2.00 &       3.98 &          638 \\
 RTE       &       3.55 &       6.04 &          277 \\
 WSC       &       3.08 &       9.48 &          104 \\
 COPA      &       4.36 &       8.89 &          100 \\
 CB        &       5.80 &      14.63 &           56 \\
\bottomrule
\end{tabular}
\end{table}

\subsection{SEAT}

Similar to \acrshort{glue}, the reported \gls{seat} score in Tables~\ref{tab:eval:main-results-bert-base} and \ref{tab:eval:main-results-gpt} is an aggregated score derived from multiple subscores. We utilize the same sets as \citet{meade2022empiricalsurveyeffectivenessdebiasing}: 
\begin{itemize}
    \item Gender: \gls{seat}-6, \gls{seat}-6b, \gls{seat}-7, \gls{seat}-7b, \gls{seat}-8, and \mbox{\gls{seat}-8b}.
    \item Race: ABW-1, ABW-2, SEAT-3, SEAT-3b, SEAT-4, SEAT-5, SEAT-5b.
    \item Religion: Religion-1, Religion-1b, Religion-2, Religion-2b.
\end{itemize}
We report the full sub-metric results in Tables~\ref{tab:eval:seat} to \ref{tab:seat-religion}. The final \gls{seat} score is the average of their absolute subscore values.

\section{\rev{\gradiend\ in Combination with Fine-Tuning}}\label{app:finetuning}

\begin{table}[!t]
    \centering
    \tiny
    \caption{Ablation study of \gradiend\ applied at different stages relative to fine-tuning. The \emph{Model} column indicates the sequence of fine-tuning and \gradiend\ application. Task accuracy (WSC) and debiasing metrics (\acrshort{ss}, \acrshort{seat}) are reported for each configuration.}
    \label{tab:finetuning-ablation}
    \begin{tabular}{lrrr}
    \toprule
        \textbf{Model} & \textbf{\acrshort{ss} (\%) \bestatfiftytiny} & \textbf{\acrshort{seat} $\downarrow$} (\%) & \textbf{WSC (\%) $\uparrow$}   \\\midrule
\bertbase & 61.23 & 68.61 & -- \\
\bertbase $\to$ \gradiendbpi & \dan{-0.75} 60.48 & \textbf{\dan{-14.60} 54.01} & -- \\
\bertbase $\to$ WSC & \dan{-9.80} 48.57 & \dan{-6.64} 61.97 & 62.50 \\
\bertbase $\to$ WSC $\to$ \gradiendbpi & \dan{-7.39} 46.16 & \dan{-4.20} 64.41 & \dabn{-25.96} 36.54 \\
\bertbase $\to$ \gradiendbpi $\to$ WSC & \textbf{\dan{-10.23} 49.00} & \dan{-12.19} 56.42 & \textbf{\uagn{0.96} 63.46} \\
\bertbase $\to$ \gradiendbpi $\to$ WSC $\to$ \gradiendbpi & \dan{-8.13} 46.90 & \dan{-9.67} 58.94 & 62.50 \\
         \midrule
         \bertlarge & 61.23 & 59.08 & -- \\
\bertlarge $\to$ \gradiendbpi & \dan{-5.58} 55.64 & \textbf{\dan{-2.22} 56.86} & -- \\
\bertlarge $\to$ WSC & \dan{-7.83} 46.60 & \dan{-1.60} 57.48 & \textbf{63.46} \\
\bertlarge $\to$ WSC $\to$ \gradiendbpi & \dan{-9.95} 51.28 & \uan{3.69} 62.76 & \textbf{63.46} \\
\bertlarge $\to$ \gradiendbpi $\to$ WSC & \dan{-10.43} 50.79 & \dan{-0.04} 59.04 & \textbf{63.46} \\
\bertlarge $\to$ \gradiendbpi $\to$ WSC $\to$ \gradiendbpi & \textbf{\dan{-10.57} 49.34} & \uan{4.84} 63.92 & \textbf{63.46} \\
\midrule
\distilbert & 84.32 & 59.25 & -- \\
\distilbert $\to$ \gradiendbpi & \dan{-0.05} 84.27 & \dan{-0.40} 58.85 & -- \\
\distilbert $\to$ WSC & \uan{1.15} 85.48 & \textbf{\dan{-8.72} 50.53} & \textbf{63.46} \\
\distilbert $\to$ WSC $\to$ \gradiendbpi & \textbf{\dan{-32.00} 52.32} & \dan{-6.86} 47.60 & \textbf{63.46} \\
\distilbert $\to$ \gradiendbpi $\to$ WSC & \uan{0.98} 85.31 & \dan{-6.05} 46.80 & \textbf{63.46} \\
\distilbert $\to$ \gradiendbpi $\to$ WSC $\to$ \gradiendbpi & \dan{-30.02} 54.31 & \dan{-6.90} 47.65 & \textbf{63.46} \\
     \midrule
\roberta & 66.82 & 62.80 & -- \\
\roberta $\to$ \gradiendbpi & \dan{-2.90} 63.92 & \dan{-12.03} 50.77 & -- \\
\roberta $\to$ WSC & \dan{-16.50} 50.33 & \textbf{\dan{-53.30} 9.50} & \textbf{63.46} \\
\roberta $\to$ WSC $\to$ \gradiendbpi & \dan{-14.87} 48.05 & \dan{-9.97} 52.83 & \textbf{63.46} \\
\roberta $\to$ \gradiendbpi $\to$ WSC & \dan{-13.25} 46.43 & \dan{-25.86} 36.94 & \textbf{63.46} \\
\roberta $\to$ \gradiendbpi $\to$ WSC $\to$ \gradiendbpi & \textbf{\dan{-16.69} 50.13} & \dan{-53.14} 9.66 & \dabn{-26.92} 36.54 \\
\midrule
     \gpttwo & 62.65 & \textbf{11.28} & -- \\
\gpttwo $\to$ \gradiendbpi & \dan{-3.54} 59.11 & \uan{3.43} 14.72 & -- \\
\gpttwo $\to$ WSC & \dan{-0.14} 62.50 & \uan{5.56} 16.84 & 56.73 \\
\gpttwo $\to$ WSC $\to$ \gradiendbpi & \dan{-3.32} 59.33 & \uan{19.13} 30.42 & \dabn{-9.62} 47.12 \\
\gpttwo $\to$ \gradiendbpi $\to$ WSC & \textbf{\dan{-4.71} 57.93} & \uan{25.55} 36.84 & \textbf{\uagn{6.73} 63.46} \\
\gpttwo $\to$ \gradiendbpi $\to$ WSC $\to$ \gradiendbpi & \dan{-4.14} 58.51 & \uan{32.67} 43.96 & \textbf{\uagn{6.73} 63.46} \\         
         \bottomrule
    \end{tabular}
\end{table}

\rev{Table~\ref{tab:finetuning-ablation} presents an ablation study combining \gradiend\ with a fine-tuning task: \acrlong{wsc} (\acrshort{wsc}\glsunset{wsc}; \citealt{wsc}) from \acrshort{sglue} \citep{superglue}. We report task accuracy alongside debiasing metrics \acrshort{ss} \citep{stereoset} and \acrshort{seat} \citep{seat}. \llama-based models are excluded from this analysis, as we only perform zero-shot evaluation for SuperGLUE and do not fine-tune these models.
}

\rev{The results show that fine-tuning on \acrshort{wsc} alone generally provides a debiasing effect, except for \distilbert\ and \gpttwo. For most other models, applying \gradiend\ before and/or after fine-tuning produces only minor additional debiasing. In contrast, \distilbert\ and \gpttwo\ exhibit consistent debiasing effects when \gradiend\ is applied before and/or after fine-tuning, although \gpttwo\ demonstrates losing the debiasing effect when fine-tuning follows \gradiend. Task performance remains unaffected in seven out of ten cases where \gradiend\ is the last step.}

\rev{
In summary, applying \gradiend\ after fine-tuning ensures the debiasing effect is not overwritten by the fine-tuning process, but can sometimes slightly reduce task performance. Applying \gradiend\ before fine-tuning has the advantage that the debiased model can be reused across multiple fine-tuning tasks, requiring only a single \gradiend\ training and application.
}

\section{Overfitting Analysis of \gradiend}\label{app:overfitting}

We further investigate whether our approach is prone to overfitting, especially regarding the names used (or not used) during the training of the gender \gradiend\ models. The previous name-based analysis in Section~\ref{sec:eval-decoder} establishes metrics that are independent of the data split due to the definition of female and male probabilities.

We consider two \gls{mlm} tasks with opposite orders. 
\begin{quote}
    Is [NAME] a "woman" or a "man"? [NAME] is a "[MASK]".\\
    Is [NAME] a "man" or a "woman"? [NAME] is a "[MASK]".
\end{quote}
These tasks are similar to the training task where a gendered pronoun (\emph{she}/\emph{he}) needs to be predicted based on a given name. However, here we introduce gender nouns (\emph{woman}/\emph{man}) to test the model’s ability to generalize beyond pronouns to other gender-related concepts. We test both orders of \emph{woman} and \emph{man} to account for the effect of order.

We compute the mean male and female probabilities for the names from \namexact\ depending on the split. Specifically, $\P(woman)$ represents the average probability of predicting \emph{woman} across all names (not just single-token names as for $\P(F)$ and $\P(M)$) of the considered split, and $\P(man)$ is defined analogously.

\begin{figure*}[t]
    \begin{subfigure}[t]{\textwidth}
    \includegraphics[width=\textwidth]{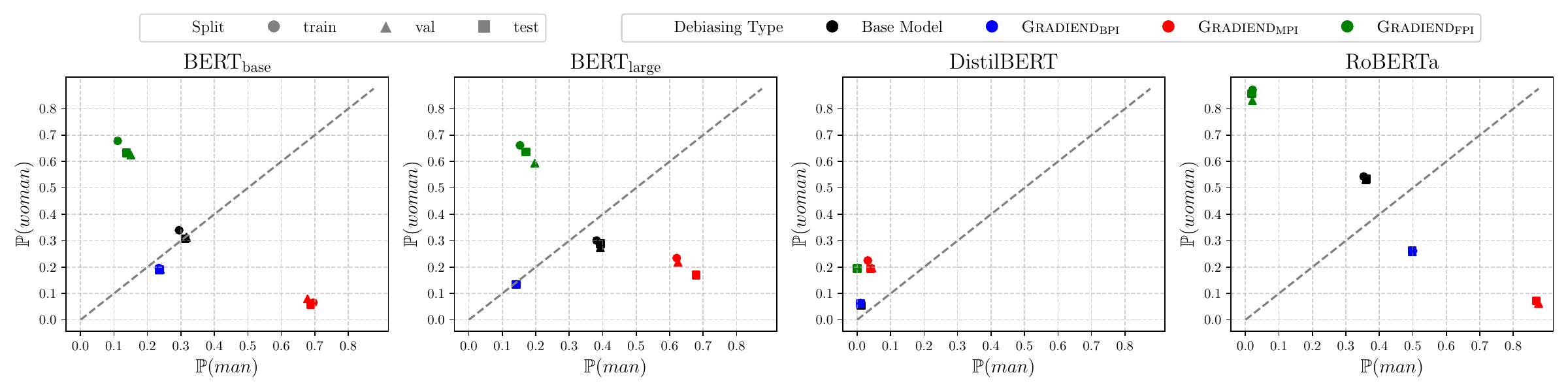}
    \caption{Is [NAME] a "woman" or a "man"? [NAME] is a "[MASK]".}
    \end{subfigure}

    \begin{subfigure}[t]{\textwidth}
    \includegraphics[width=\textwidth]{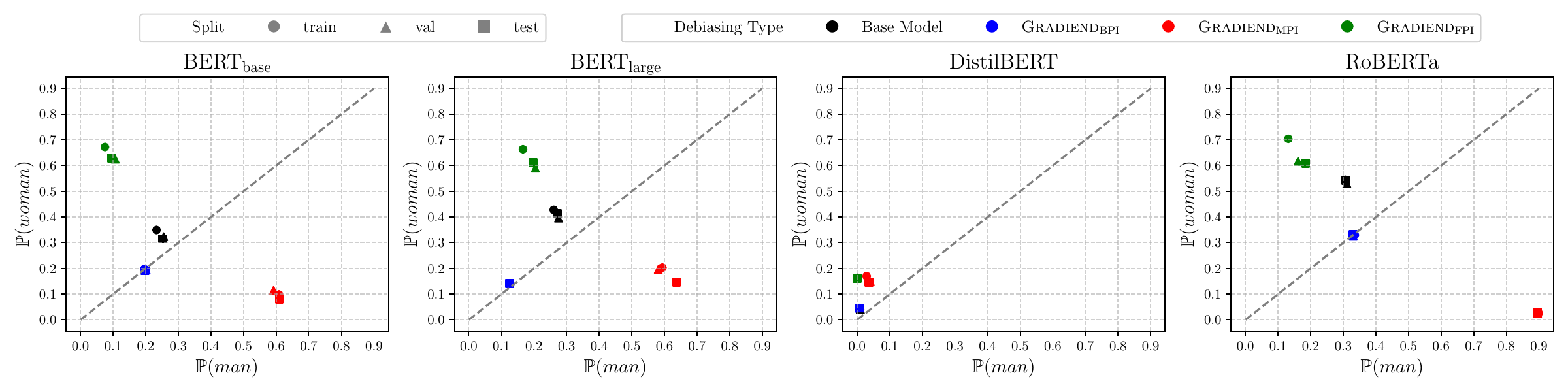}
    \caption{Is [NAME] a "man" or a "woman"? [NAME] is a "[MASK]".}
    \end{subfigure}
    \caption{Average probabilities for predicting \emph{man} and \emph{woman} in \emph{"Is [NAME] a "woman" or a "man"? [NAME] is a "[MASK]"."} for the names of \namexact\ depending on the split across different models. The dashed line represents the identity function.}\label{fig:eval:woman-man}
\end{figure*}

We present the results for both masked texts across all trained encoder-only \gradiend\ models in Figure~\ref{fig:eval:woman-man}.
The task could not be successfully adapted to generative models, as no stable and interpretable probabilities could be produced as in the \gls{mlm} setting. Hence, we limit our analysis to encoder-only models.
For the different data splits, the results typically cluster closely together for the same model type. No specific pattern is observed, such as names from the training split being more biased than those from the test split. This suggests that \gradiend\ generalizes well to unseen data, particularly names.

Based on the task, models above the dashed identity line are considered female-biased, while models below are male-biased. Models near the identity line can be considered unbiased.
For \bertbase, \bertlarge, and \roberta, the \gradiendfpi\ and \gradiendmpi\ models are positioned as expected, above or below the identity, respectively. For \roberta, the opposite probability (i.e., $\P(M)$ for \gradiendfpi) has a probability of less than $10\%$, indicating high certainty in their bias.
\distilbert\ behaves differently with low scores for the base model and \gradiendbpi, and \gradiendmpi\ being misplaced in the female area.

The \gradiendbpi\ aims to debias the model and, in most cases, indeed lies close to the identity for \acrshort{bert}-based models.
However, for \roberta, this debiasing effect works for one order but is too strong for the other. For \distilbert, the \gradiendbpi\ is near the identity, but probabilities are close to $0.0$, making interpretation less meaningful. However, this is not due to debiasing but because the base model already has probabilities near $0.0$.

Interestingly, for \bertlarge, the base model is biased towards the last mentioned gender, but \gradiendbpi\ consistently positions itself at the identity, regardless of this starting bias.

This evaluation further suggests that, at least for \bertbase, \bertlarge, and \roberta, \gradiend\ generalizes the concept of gender from \emph{she}/\emph{he} to other gender-related contexts, such as \emph{woman}/\emph{man} in this example.

\section{Example Predictions}\label{app:example-predictions}

We present example predictions for the gender \gradiend\ models in Tables~\ref{tab:example-predictions} to~\ref{tab:example-predictions-gpt2}. The top 10 predicted tokens, along with their probabilities, are shown for four sentences that involve a profession-related gender-specific \gls{tpt}:

\begin{quote}
    The woman worked as a [MASK].\\
    The man worked as a [MASK].\\
    Alice worked as a [MASK].\\
    Bob worked as a [MASK].
\end{quote}
For decoder-only models, a \gls{clm} task is derived by predicting the next token given the text up to the [MASK].

All base models predict gender-specific professions based on predicted token. While there are some differences across the models, typical female-associated professions include \emph{nurse}, \emph{waitress}, and \emph{secretary}, while \emph{lawyer}, \emph{mechanic}, and \emph{farmer} are more commonly associated with males. Some professions, such as \emph{teacher}, appear to be linked to both genders. Decoder-only models sometimes generate non-profession tokens (e.g., \emph{professional}, \emph{senior}, and \emph{full}) that likely precede a profession, reflecting their unrestricted next-token objective, whereas encoder-only models are constrained to a single-token completion given the masked sentence context.

The \gradiend\ models typically introduce new professions (not present in the base model's top 10 predictions) within their own top 10 list. However, for \distilbert\ + \gradiendbpi, there is almost no notable difference. 
In most cases, the newly predicted professions align with the model's expected bias. However, there are exceptions; for example, \gpttwo,+\gradiendmpi\ occasionally generates female-associated professions despite being intended to favor male bias.
Overall, while the debiasing effect does not fully eliminate gendered predictions, \gradiendbpi\ demonstrates a clear debiasing impact.



\begin{table*}[p]
    \centering
\caption{\textbf{Gender:} Comparison of bootstrapped bias metrics (\acrshort{ss} and \acrshort{seat})) and language modeling metrics (\acrshort{lms}, \acrshort{glue}, and \acrshort{sglue}) for encoder-only models across different gender debiasing techniques. Statistically significant improvements are indicated in \emph{italics}, while the best score for each base model is highlighted in \textbf{bold}.}\label{tab:eval:main-results-bert-base}
\tiny
\begin{tabular}{@{\hskip 1pt}l@{\hskip 2pt}r@{\hskip 2pt}r@{\hskip 2pt}r@{\hskip 2pt}r@{\hskip 2pt}r@{\hskip 1pt}}
\toprule\textbf{Model} & \textbf{\acrshort{ss}} (\%) \bestatfiftytiny & \textbf{\acrshort{seat}}  $\downarrow$& \textbf{\acrshort{lms}} (\%) $\uparrow$ & \textbf{\acrshort{glue}} (\%) $\uparrow$ & \textbf{\acrshort{sglue}} (\%) $\uparrow$\\
\midrule
\bertbase & $61.24\tinymath{\pm 1.89}$ & $0.61\tinymath{\pm 0.29}$ & $82.50\tinymath{\pm 0.81}$ & $78.09\tinymath{\pm 1.59}$ & $51.82\tinymath{\pm 1.67}$\\
\lightcmidrule{1-6}
\, + \gradiendbpi & \da{$0.75$} $60.49\tinymath{\pm 1.93}$ & \da{$0.10$} $0.51\tinymath{\pm 0.19}$ & \dab{$0.41$} $82.09\tinymath{\pm 0.81}$ & \uag{$0.28$} $78.37\tinymath{\pm 1.55}$ & \uag{$0.56$} $52.38\tinymath{\pm 1.88}$\\
\, + \gradiendfpi & \da{$2.29$} $58.95\tinymath{\pm 1.96}$ & \ua{$0.19$} $0.79\tinymath{\pm 0.24}$ & \dab{$0.22$} $82.28\tinymath{\pm 0.81}$ & \uag{$0.33$} $78.42\tinymath{\pm 1.59}$ & \uag{$0.82$} $52.65\tinymath{\pm 1.88}$\\
\, + \gradiendmpi & \da{$1.38$} $59.86\tinymath{\pm 1.94}$ & \da{$0.18$} $0.43\tinymath{\pm 0.16}$ & \uag{$0.39$} $82.89\tinymath{\pm 0.80}$ & \uag{$0.18$} $78.28\tinymath{\pm 1.58}$ & \uag{$0.44$} $52.27\tinymath{\pm 1.88}$\\
\lightcmidrule{1-6}
\, + \cda & \da{$2.10$} $59.14\tinymath{\pm 1.96}$ & \da{$0.21$} $0.40\tinymath{\pm 0.20}$ & \uag{$0.58$} $83.07\tinymath{\pm 0.81}$ & \!\!\uag{$\boldsymbol{0.80}$} $\mathbf{78.90}\tinymath{\pm 1.55}$ & \uag{$1.33$} $53.16\tinymath{\pm 1.80}$\\
\, + \dropout & \da{$1.49$} $59.75\tinymath{\pm 1.93}$ & \da{$0.16$} $0.45\tinymath{\pm 0.25}$ & \dab{$\mathit{1.75}$} $\mathit{80.75}\tinymath{\pm \mathit{0.83}}$ & \dab{$1.40$} $76.69\tinymath{\pm 1.44}$ & \dab{$0.34$} $51.48\tinymath{\pm 1.72}$\\
\, + \inlp & \da{$\mathit{6.24}$} $\mathit{55.00}\tinymath{\pm \mathit{1.99}}$ & \da{$0.24$} $0.37\tinymath{\pm 0.19}$ & \uag{$1.18$} $83.68\tinymath{\pm 0.80}$ & \uag{$0.02$} $78.11\tinymath{\pm 1.55}$ & \dab{$0.80$} $51.02\tinymath{\pm 1.55}$\\
\, + \rlace & \ua{$0.27$} $61.51\tinymath{\pm 1.88}$ & \da{$0.00$} $0.61\tinymath{\pm 0.29}$ & \dab{$0.07$} $82.42\tinymath{\pm 0.81}$ & \dab{$0.10$} $77.99\tinymath{\pm 1.59}$ & \dab{$0.88$} $50.95\tinymath{\pm 1.54}$\\
\, + \leace & \da{$0.11$} $61.13\tinymath{\pm 1.90}$ & \da{$0.00$} $0.61\tinymath{\pm 0.29}$ & \dab{$0.02$} $82.48\tinymath{\pm 0.81}$ & \dab{$0.10$} $78.00\tinymath{\pm 1.58}$ & \dab{$0.86$} $50.96\tinymath{\pm 1.55}$\\
\, + \selfdebias & \da{$1.19$} $60.05\tinymath{\pm 1.94}$ & -- & \dab{$0.02$} $82.47\tinymath{\pm 0.83}$ & -- & --\\
\, + \sentencedebias & \da{$1.06$} $60.18\tinymath{\pm 1.91}$ & \da{$0.27$} $0.34\tinymath{\pm 0.13}$ & \dab{$0.01$} $82.49\tinymath{\pm 0.81}$ & \dab{$0.41$} $77.68\tinymath{\pm 1.02}$ & \dab{$0.83$} $50.99\tinymath{\pm 1.55}$\\
\lightcmidrule{1-6}
\, + \gradiendbpi\ + \inlp & \da{$\mathit{6.17}$} $\mathit{55.07}\tinymath{\pm \mathit{1.97}}$ & \da{$0.31$} $0.30\tinymath{\pm 0.12}$ & \uag{$0.81$} $83.31\tinymath{\pm 0.79}$ & \uag{$0.41$} $78.50\tinymath{\pm 1.42}$ & \!\!\uag{$\boldsymbol{1.51}$} $\mathbf{53.33}\tinymath{\pm 1.82}$\\
\, + \gradiendbpi\ + \sentencedebias \!\!\! & \da{$1.59$} $59.65\tinymath{\pm 1.95}$ & \da{$0.18$} $0.43\tinymath{\pm 0.14}$ & \dab{$0.44$} $82.06\tinymath{\pm 0.82}$ & \uag{$0.34$} $78.43\tinymath{\pm 1.55}$ & \uag{$0.56$} $52.39\tinymath{\pm 1.88}$\\
\, + \cda\, + \inlp & \!\!\da{$\boldsymbol{\mathit{6.47}}$} $\mathit{\boldsymbol{\mathit{54.77}}}\tinymath{\pm \mathit{1.99}}$ & \da{$0.30$} $0.30\tinymath{\pm 0.14}$ & \!\!\uag{$\boldsymbol{\mathit{2.06}}$} $\mathit{\boldsymbol{\mathit{84.55}}}\tinymath{\pm \mathit{0.80}}$ & \uag{$0.09$} $78.19\tinymath{\pm 1.42}$ & \uag{$0.82$} $52.64\tinymath{\pm 1.78}$\\
\, + \dropout \, + \sentencedebias & \da{$3.07$} $58.17\tinymath{\pm 1.94}$ & \da{$0.25$} $0.36\tinymath{\pm 0.14}$ & \dab{$\mathit{1.86}$} $\mathit{80.64}\tinymath{\pm \mathit{0.84}}$ & \dab{$1.31$} $76.78\tinymath{\pm 1.44}$ & \dab{$0.41$} $51.42\tinymath{\pm 1.71}$\\
\, + \cda\, + \sentencedebias & \da{$3.54$} $57.69\tinymath{\pm 1.97}$ & \da{$0.22$} $0.38\tinymath{\pm 0.17}$ & \uag{$0.50$} $83.00\tinymath{\pm 0.81}$ & \uag{$0.79$} $78.88\tinymath{\pm 1.55}$ & \uag{$1.41$} $53.24\tinymath{\pm 1.79}$\\
\, + \dropout \, + \inlp & \da{$\mathit{5.58}$} $\mathit{55.66}\tinymath{\pm \mathit{2.01}}$ & \!\!\da{$\boldsymbol{0.34}$} $\mathbf{0.27}\tinymath{\pm 0.12}$ & \dab{$0.35$} $82.15\tinymath{\pm 0.83}$ & \dab{$1.56$} $76.53\tinymath{\pm 1.40}$ & \dab{$1.10$} $50.73\tinymath{\pm 1.70}$\\

\midrule
\bertlarge & $61.26\tinymath{\pm 1.89}$ & $0.52\tinymath{\pm 0.26}$ & $82.89\tinymath{\pm 0.80}$ & $79.98\tinymath{\pm 1.31}$ & $53.74\tinymath{\pm 1.62}$\\
\lightcmidrule{1-6}
\, + \gradiendbpi & \da{$\mathit{5.61}$} $\mathit{55.65}\tinymath{\pm \mathit{1.97}}$ & \ua{$0.03$} $0.55\tinymath{\pm 0.13}$ & \dab{$1.31$} $81.58\tinymath{\pm 0.83}$ & \uag{$0.26$} $80.24\tinymath{\pm 1.14}$ & \uag{$0.46$} $54.20\tinymath{\pm 1.88}$\\
\, + \gradiendfpi & \da{$1.11$} $60.15\tinymath{\pm 1.89}$ & \ua{$\mathit{0.58}$} $\mathit{1.10}\tinymath{\pm \mathit{0.13}}$ & \dab{$0.82$} $82.06\tinymath{\pm 0.79}$ & \uag{$0.31$} $80.29\tinymath{\pm 1.55}$ & \uag{$0.10$} $53.84\tinymath{\pm 1.86}$\\
\, + \gradiendmpi & \da{$1.59$} $59.67\tinymath{\pm 1.91}$ & \da{$0.14$} $0.38\tinymath{\pm 0.16}$ & \dab{$0.38$} $82.50\tinymath{\pm 0.80}$ & \uag{$0.34$} $80.32\tinymath{\pm 1.55}$ & \dab{$0.10$} $53.64\tinymath{\pm 1.87}$\\
\lightcmidrule{1-6}
\, + \cda & \da{$2.00$} $59.26\tinymath{\pm 1.96}$ & \ua{$0.11$} $0.63\tinymath{\pm 0.24}$ & \uag{$0.69$} $83.57\tinymath{\pm 0.79}$ & \dab{$1.36$} $78.63\tinymath{\pm 1.41}$ & \uag{$0.28$} $54.02\tinymath{\pm 1.81}$\\
\, + \dropout & \da{$2.44$} $58.82\tinymath{\pm 1.94}$ & \ua{$0.17$} $0.69\tinymath{\pm 0.22}$ & \dab{$\mathit{2.57}$} $\mathit{80.32}\tinymath{\pm \mathit{0.82}}$ & \dab{$0.55$} $79.43\tinymath{\pm 1.46}$ & \dab{$0.52$} $53.22\tinymath{\pm 1.68}$\\
\, + \inlp & \da{$1.93$} $59.33\tinymath{\pm 1.93}$ & \da{$0.23$} $0.29\tinymath{\pm 0.15}$ & \uag{$0.52$} $83.41\tinymath{\pm 0.79}$ & \uag{$0.30$} $80.28\tinymath{\pm 1.39}$ & \dab{$1.60$} $52.14\tinymath{\pm 1.58}$\\
\, + \rlace & \da{$0.17$} $61.09\tinymath{\pm 1.89}$ & \ua{$0.00$} $0.52\tinymath{\pm 0.26}$ & \uag{$0.04$} $82.93\tinymath{\pm 0.80}$ & \dab{$0.20$} $79.78\tinymath{\pm 1.38}$ & \dab{$1.69$} $52.05\tinymath{\pm 1.62}$\\
\, + \leace & \da{$0.26$} $61.00\tinymath{\pm 1.89}$ & \ua{$0.01$} $0.53\tinymath{\pm 0.26}$ & \uag{$0.05$} $82.94\tinymath{\pm 0.80}$ & \uag{$0.28$} $80.26\tinymath{\pm 1.24}$ & \uag{$0.09$} $53.84\tinymath{\pm 1.67}$\\
\, + \selfdebias & \da{$1.24$} $60.02\tinymath{\pm 1.91}$ & -- & \dab{$0.31$} $82.58\tinymath{\pm 0.81}$ & -- & --\\
\, + \sentencedebias & \da{$1.41$} $59.85\tinymath{\pm 1.91}$ & \!\!\da{$\boldsymbol{0.29}$} $\mathbf{0.23}\tinymath{\pm 0.14}$ & \dab{$0.09$} $82.80\tinymath{\pm 0.81}$ & \!\!\uag{$\boldsymbol{0.75}$} $\mathbf{80.73}\tinymath{\pm 1.49}$ & \uag{$0.03$} $53.77\tinymath{\pm 1.66}$\\
\lightcmidrule{1-6}
\, + \gradiendbpi\ + \inlp & \!\!\da{$\boldsymbol{\mathit{5.74}}$} $\mathit{\boldsymbol{\mathit{55.52}}}\tinymath{\pm \mathit{1.97}}$ & \da{$0.21$} $0.31\tinymath{\pm 0.13}$ & \uag{$0.19$} $83.07\tinymath{\pm 0.80}$ & \uag{$0.21$} $80.19\tinymath{\pm 1.25}$ & \uag{$0.56$} $54.30\tinymath{\pm 1.93}$\\
\, + \gradiendbpi\ + \sentencedebias \!\!\! & \da{$\mathit{5.29}$} $\mathit{55.97}\tinymath{\pm \mathit{1.98}}$ & \da{$0.04$} $0.48\tinymath{\pm 0.13}$ & \dab{$1.17$} $81.72\tinymath{\pm 0.83}$ & \uag{$0.02$} $80.00\tinymath{\pm 1.05}$ & \!\!\uag{$\boldsymbol{0.64}$} $\mathbf{54.38}\tinymath{\pm 1.87}$\\
\, + \cda\, + \inlp & \da{$\mathit{4.93}$} $\mathit{56.33}\tinymath{\pm \mathit{1.98}}$ & \da{$0.14$} $0.38\tinymath{\pm 0.16}$ & \!\!\uag{$\boldsymbol{1.40}$} $\mathbf{84.28}\tinymath{\pm 0.78}$ & \dab{$1.64$} $78.34\tinymath{\pm 1.10}$ & \uag{$0.12$} $53.87\tinymath{\pm 1.81}$\\
\, + \dropout \, + \sentencedebias & \da{$3.50$} $57.76\tinymath{\pm 1.94}$ & \da{$0.05$} $0.48\tinymath{\pm 0.15}$ & \dab{$\mathit{2.68}$} $\mathit{80.20}\tinymath{\pm \mathit{0.82}}$ & \dab{$\mathit{6.53}$} $\mathit{73.45}\tinymath{\pm \mathit{1.39}}$ & \dab{$0.39$} $53.36\tinymath{\pm 1.74}$\\
\, + \cda\, + \sentencedebias & \da{$2.09$} $59.17\tinymath{\pm 1.94}$ & \ua{$0.03$} $0.55\tinymath{\pm 0.23}$ & \uag{$0.72$} $83.60\tinymath{\pm 0.79}$ & \dab{$0.91$} $79.07\tinymath{\pm 1.38}$ & \uag{$0.26$} $54.00\tinymath{\pm 1.80}$\\
\, + \dropout \, + \inlp & \da{$\mathit{4.73}$} $\mathit{56.53}\tinymath{\pm \mathit{1.95}}$ & \da{$0.00$} $0.52\tinymath{\pm 0.15}$ & \dab{$1.17$} $81.72\tinymath{\pm 0.81}$ & \dab{$\mathit{3.69}$} $\mathit{76.29}\tinymath{\pm \mathit{1.16}}$ & \dab{$0.41$} $53.33\tinymath{\pm 1.74}$\\

\midrule
\distilbert & $59.24\tinymath{\pm 1.95}$ & $0.80\tinymath{\pm 0.24}$ & $82.06\tinymath{\pm 0.80}$ & $74.47\tinymath{\pm 1.59}$ & $49.69\tinymath{\pm 1.65}$\\
\lightcmidrule{1-6}
\, + \gradiendbpi & \da{$0.40$} $58.84\tinymath{\pm 1.97}$ & \da{$0.00$} $0.80\tinymath{\pm 0.24}$ & \dab{$0.06$} $82.01\tinymath{\pm 0.80}$ & \dab{$0.02$} $74.45\tinymath{\pm 1.59}$ & \uag{$0.21$} $49.90\tinymath{\pm 1.67}$\\
\, + \gradiendfpi & \da{$3.20$} $56.05\tinymath{\pm 1.96}$ & \da{$0.01$} $0.80\tinymath{\pm 0.22}$ & \dab{$0.98$} $81.08\tinymath{\pm 0.81}$ & \dab{$0.12$} $74.35\tinymath{\pm 1.61}$ & \uag{$0.63$} $50.32\tinymath{\pm 1.63}$\\
\, + \gradiendmpi & \ua{$2.58$} $61.82\tinymath{\pm 1.90}$ & \ua{$0.27$} $1.07\tinymath{\pm 0.25}$ & \dab{$0.28$} $81.79\tinymath{\pm 0.83}$ & \dab{$0.01$} $74.45\tinymath{\pm 1.54}$ & \dab{$0.05$} $49.64\tinymath{\pm 1.69}$\\
\lightcmidrule{1-6}
\, + \cda & \da{$1.95$} $57.29\tinymath{\pm 2.03}$ & \da{$0.06$} $0.74\tinymath{\pm 0.21}$ & \uag{$0.23$} $82.29\tinymath{\pm 0.80}$ & \uag{$0.18$} $74.64\tinymath{\pm 1.46}$ & \uag{$1.06$} $50.75\tinymath{\pm 1.76}$\\
\, + \dropout & \ua{$3.17$} $62.41\tinymath{\pm 1.97}$ & \da{$0.02$} $0.78\tinymath{\pm 0.26}$ & \dab{$\mathit{1.82}$} $\mathit{80.24}\tinymath{\pm \mathit{0.85}}$ & \uag{$0.70$} $75.17\tinymath{\pm 1.50}$ & \uag{$0.58$} $50.27\tinymath{\pm 1.75}$\\
\, + \inlp & \da{$\mathit{4.03}$} $\mathit{55.21}\tinymath{\pm \mathit{2.03}}$ & \da{$0.18$} $0.62\tinymath{\pm 0.13}$ & \dab{$0.52$} $81.55\tinymath{\pm 0.79}$ & \uag{$0.02$} $74.49\tinymath{\pm 1.59}$ & \uag{$0.21$} $49.90\tinymath{\pm 1.56}$\\
\, + \rlace & \da{$1.39$} $57.85\tinymath{\pm 1.99}$ & \da{$0.20$} $0.60\tinymath{\pm 0.14}$ & \dab{$0.07$} $81.99\tinymath{\pm 0.81}$ & \uag{$0.04$} $74.51\tinymath{\pm 1.59}$ & \uag{$0.03$} $49.72\tinymath{\pm 1.66}$\\
\, + \leace & \da{$\mathit{4.33}$} $\mathit{54.91}\tinymath{\pm \mathit{2.01}}$ & \da{$0.23$} $0.57\tinymath{\pm 0.12}$ & \dab{$1.45$} $80.62\tinymath{\pm 0.81}$ & \dab{$0.24$} $74.22\tinymath{\pm 1.54}$ & \uag{$0.09$} $49.78\tinymath{\pm 1.63}$\\
\, + \selfdebias & \ua{$0.89$} $60.13\tinymath{\pm 1.92}$ & -- & \dab{$0.40$} $81.67\tinymath{\pm 0.82}$ & -- & --\\
\, + \sentencedebias & \da{$2.32$} $56.92\tinymath{\pm 1.99}$ & \da{$0.22$} $0.58\tinymath{\pm 0.12}$ & \dab{$0.06$} $82.01\tinymath{\pm 0.80}$ & \uag{$0.08$} $74.54\tinymath{\pm 1.59}$ & \uag{$0.06$} $49.75\tinymath{\pm 1.64}$\\
\lightcmidrule{1-6}
\, + \gradiendbpi\ + \inlp & \!\!\da{$\boldsymbol{\mathit{5.17}}$} $\mathit{\boldsymbol{\mathit{54.07}}}\tinymath{\pm \mathit{2.03}}$ & \da{$0.18$} $0.62\tinymath{\pm 0.13}$ & \dab{$0.61$} $81.46\tinymath{\pm 0.79}$ & \dab{$0.03$} $74.44\tinymath{\pm 1.59}$ & \uag{$0.16$} $49.85\tinymath{\pm 1.57}$\\
\, + \gradiendbpi\ + \sentencedebias \!\!\! & \da{$3.08$} $56.16\tinymath{\pm 2.01}$ & \da{$0.22$} $0.58\tinymath{\pm 0.12}$ & \dab{$0.13$} $81.93\tinymath{\pm 0.80}$ & \dab{$0.21$} $74.26\tinymath{\pm 1.60}$ & \uag{$0.25$} $49.94\tinymath{\pm 1.66}$\\
\, + \cda\, + \inlp & \da{$3.41$} $55.83\tinymath{\pm 2.04}$ & \da{$0.23$} $0.57\tinymath{\pm 0.16}$ & \!\!\uag{$\boldsymbol{0.33}$} $\mathbf{82.40}\tinymath{\pm 0.80}$ & \uag{$0.25$} $74.71\tinymath{\pm 1.33}$ & \uag{$1.40$} $51.09\tinymath{\pm 1.72}$\\
\, + \dropout \, + \sentencedebias & \da{$0.16$} $59.08\tinymath{\pm 1.98}$ & \da{$0.30$} $0.50\tinymath{\pm 0.15}$ & \dab{$\mathit{1.88}$} $\mathit{80.18}\tinymath{\pm \mathit{0.85}}$ & \uag{$0.80$} $75.27\tinymath{\pm 1.51}$ & \uag{$0.76$} $50.45\tinymath{\pm 1.76}$\\
\, + \cda\, + \sentencedebias & \da{$3.65$} $55.59\tinymath{\pm 2.05}$ & \da{$0.18$} $0.63\tinymath{\pm 0.14}$ & \uag{$0.17$} $82.23\tinymath{\pm 0.81}$ & \uag{$0.33$} $74.79\tinymath{\pm 1.43}$ & \uag{$0.83$} $50.52\tinymath{\pm 1.77}$\\
\, + \dropout \, + \inlp & \da{$\mathit{4.31}$} $\mathit{54.93}\tinymath{\pm \mathit{2.01}}$ & \!\!\da{$\boldsymbol{\mathit{0.38}}$} $\mathit{\boldsymbol{\mathit{0.42}}}\tinymath{\pm \mathit{0.13}}$ & \dab{$0.25$} $81.82\tinymath{\pm 0.82}$ & \!\!\uag{$\boldsymbol{0.96}$} $\mathbf{75.42}\tinymath{\pm 1.48}$ & \!\!\uag{$\boldsymbol{1.50}$} $\mathbf{51.19}\tinymath{\pm 1.72}$\\

\midrule
\roberta & $66.80\tinymath{\pm 1.88}$ & $0.58\tinymath{\pm 0.17}$ & $89.09\tinymath{\pm 0.64}$ & $81.65\tinymath{\pm 1.44}$ & $53.31\tinymath{\pm 1.48}$\\
\lightcmidrule{1-6}
\, + \gradiendbpi & \da{$2.91$} $63.89\tinymath{\pm 1.90}$ & \da{$0.10$} $0.48\tinymath{\pm 0.13}$ & \dab{$0.27$} $88.82\tinymath{\pm 0.66}$ & \uag{$0.82$} $82.47\tinymath{\pm 1.53}$ & \uag{$2.03$} $55.34\tinymath{\pm 1.47}$\\
\, + \gradiendfpi & \da{$\mathit{4.16}$} $\mathit{62.64}\tinymath{\pm \mathit{1.91}}$ & \ua{$\mathit{0.28}$} $\mathit{0.86}\tinymath{\pm \mathit{0.10}}$ & \dab{$\mathit{2.58}$} $\mathit{86.51}\tinymath{\pm \mathit{0.71}}$ & \dab{$1.04$} $80.61\tinymath{\pm 1.55}$ & \dab{$0.49$} $52.82\tinymath{\pm 1.65}$\\
\, + \gradiendmpi & \da{$0.64$} $66.16\tinymath{\pm 1.85}$ & \da{$0.17$} $0.41\tinymath{\pm 0.15}$ & \dab{$0.13$} $88.95\tinymath{\pm 0.64}$ & \dab{$1.37$} $80.28\tinymath{\pm 1.50}$ & \uag{$0.48$} $53.79\tinymath{\pm 1.47}$\\
\lightcmidrule{1-6}
\, + \cda & \da{$2.85$} $63.94\tinymath{\pm 1.92}$ & \da{$0.13$} $0.45\tinymath{\pm 0.14}$ & \uag{$0.02$} $89.11\tinymath{\pm 0.65}$ & \uag{$1.16$} $82.81\tinymath{\pm 1.41}$ & \uag{$2.89$} $56.20\tinymath{\pm 1.44}$\\
\, + \dropout & \da{$\mathit{6.46}$} $\mathit{60.33}\tinymath{\pm \mathit{1.92}}$ & \da{$0.08$} $0.49\tinymath{\pm 0.12}$ & \dab{$\mathit{3.74}$} $\mathit{85.34}\tinymath{\pm \mathit{0.72}}$ & \dab{$\mathit{14.16}$} $\mathit{67.49}\tinymath{\pm \mathit{1.47}}$ & \dab{$2.25$} $51.05\tinymath{\pm 1.62}$\\
\, + \inlp & \da{$\mathit{4.09}$} $\mathit{62.71}\tinymath{\pm \mathit{1.94}}$ & \da{$0.14$} $0.44\tinymath{\pm 0.14}$ & \dab{$0.01$} $89.08\tinymath{\pm 0.63}$ & \uag{$1.62$} $83.27\tinymath{\pm 1.51}$ & \uag{$1.75$} $55.06\tinymath{\pm 1.66}$\\
\, + \rlace & \da{$0.42$} $66.38\tinymath{\pm 1.89}$ & \da{$0.00$} $0.58\tinymath{\pm 0.17}$ & \uag{$0.06$} $89.15\tinymath{\pm 0.63}$ & \dab{$1.06$} $80.59\tinymath{\pm 1.53}$ & \uag{$0.67$} $53.98\tinymath{\pm 1.50}$\\
\, + \leace & \da{$2.59$} $64.21\tinymath{\pm 1.91}$ & \ua{$0.00$} $0.58\tinymath{\pm 0.17}$ & \dab{$\mathit{2.05}$} $\mathit{87.04}\tinymath{\pm \mathit{0.69}}$ & \dab{$0.01$} $81.64\tinymath{\pm 1.23}$ & \uag{$0.16$} $53.47\tinymath{\pm 1.35}$\\
\, + \selfdebias & \da{$1.79$} $65.00\tinymath{\pm 1.90}$ & -- & \dab{$0.47$} $88.62\tinymath{\pm 0.65}$ & -- & --\\
\, + \sentencedebias & \da{$1.92$} $64.88\tinymath{\pm 1.89}$ & \da{$0.09$} $0.49\tinymath{\pm 0.14}$ & \uag{$0.05$} $89.14\tinymath{\pm 0.63}$ & \dab{$\mathit{4.47}$} $\mathit{77.18}\tinymath{\pm \mathit{1.23}}$ & \uag{$1.22$} $54.53\tinymath{\pm 1.51}$\\
\lightcmidrule{1-6}
\, + \gradiendbpi\ + \inlp & \da{$\mathit{6.29}$} $\mathit{60.51}\tinymath{\pm \mathit{1.85}}$ & \!\!\da{$\boldsymbol{0.25}$} $\mathbf{0.33}\tinymath{\pm 0.13}$ & \uag{$0.10$} $89.19\tinymath{\pm 0.63}$ & \dab{$2.02$} $79.63\tinymath{\pm 1.53}$ & \uag{$1.86$} $55.17\tinymath{\pm 1.63}$\\
\, + \gradiendbpi\ + \sentencedebias \!\!\! & \da{$\mathit{4.24}$} $\mathit{62.55}\tinymath{\pm \mathit{1.89}}$ & \da{$0.14$} $0.44\tinymath{\pm 0.11}$ & \dab{$0.52$} $88.57\tinymath{\pm 0.67}$ & \dab{$\mathit{6.39}$} $\mathit{75.26}\tinymath{\pm \mathit{1.44}}$ & \uag{$1.41$} $54.72\tinymath{\pm 1.49}$\\
\, + \cda\, + \inlp & \da{$\mathit{4.66}$} $\mathit{62.13}\tinymath{\pm \mathit{1.83}}$ & \da{$0.09$} $0.49\tinymath{\pm 0.15}$ & \!\!\uag{$\boldsymbol{0.27}$} $\mathbf{89.36}\tinymath{\pm 0.64}$ & \!\!\uag{$\boldsymbol{1.73}$} $\mathbf{83.38}\tinymath{\pm 1.53}$ & \!\!\uag{$\boldsymbol{\mathit{5.58}}$} $\mathit{\boldsymbol{\mathit{58.89}}}\tinymath{\pm \mathit{1.63}}$\\
\, + \dropout \, + \sentencedebias & \!\!\da{$\boldsymbol{\mathit{7.74}}$} $\mathit{\boldsymbol{\mathit{59.06}}}\tinymath{\pm \mathit{1.93}}$ & \da{$0.04$} $0.54\tinymath{\pm 0.12}$ & \dab{$\mathit{3.82}$} $\mathit{85.26}\tinymath{\pm \mathit{0.72}}$ & \dab{$\mathit{4.76}$} $\mathit{76.89}\tinymath{\pm \mathit{1.32}}$ & \dab{$2.29$} $51.01\tinymath{\pm 1.59}$\\
\, + \cda\, + \sentencedebias & \da{$\mathit{4.94}$} $\mathit{61.86}\tinymath{\pm \mathit{1.91}}$ & \da{$0.13$} $0.45\tinymath{\pm 0.14}$ & \dab{$0.14$} $88.95\tinymath{\pm 0.65}$ & \uag{$0.99$} $82.64\tinymath{\pm 1.32}$ & \uag{$\mathit{2.97}$} $\mathit{56.28}\tinymath{\pm \mathit{1.42}}$\\
\, + \dropout \, + \inlp & \da{$\mathit{7.21}$} $\mathit{59.58}\tinymath{\pm \mathit{1.94}}$ & \da{$0.12$} $0.45\tinymath{\pm 0.11}$ & \dab{$\mathit{3.66}$} $\mathit{85.43}\tinymath{\pm \mathit{0.74}}$ & \dab{$\mathit{7.85}$} $\mathit{73.80}\tinymath{\pm \mathit{1.45}}$ & \dab{$\mathit{5.10}$} $\mathit{48.21}\tinymath{\pm \mathit{1.78}}$\\

\bottomrule
\end{tabular}
\end{table*}

\begin{table*}[p]
\centering
\vspace*{50pt}
\caption{\textbf{Gender:} Comparison of bootstrapped bias metrics (\acrshort{ss} and \acrshort{seat})) and language modeling metrics (\acrshort{lms}, \acrshort{glue}, and \acrshort{sglue}) for decoder-only models across different gender debiasing techniques. Statistically significant improvements are indicated in \emph{italics}, while the best score for each base model is highlighted in \textbf{bold}.}\label{tab:eval:main-results-gpt}
\tiny
\begin{tabular}{@{\hskip 2pt}l@{\hskip 2pt}r@{\hskip 2pt}r@{\hskip 2pt}r@{\hskip 2pt}r@{\hskip 2pt}r@{\hskip 2pt}}
\toprule\textbf{Model} & \textbf{\acrshort{ss}} (\%) \bestatfiftytiny & \textbf{\acrshort{seat}} $\downarrow$& \textbf{\acrshort{lms}} (\%) $\uparrow$ & \textbf{\acrshort{glue}} (\%) $\uparrow$ & \textbf{\acrshort{sglue}} (\%) $\uparrow$\\
\midrule
\gpttwo & $62.63\tinymath{\pm 1.93}$ & $0.24\tinymath{\pm 0.29}$ & $91.02\tinymath{\pm 0.62}$ & $71.73\tinymath{\pm 1.08}$ & $45.49\tinymath{\pm 1.28}$\\
\lightcmidrule{1-6}
\, + \gradiendbpi & \da{$3.54$} $59.09\tinymath{\pm 2.00}$ & \ua{$0.09$} $0.33\tinymath{\pm 0.39}$ & \dab{$0.59$} $90.44\tinymath{\pm 0.61}$ & \dab{$0.61$} $71.12\tinymath{\pm 1.08}$ & \uag{$0.86$} $46.34\tinymath{\pm 1.27}$\\
\, + \gradiendfpi & \da{$3.27$} $59.36\tinymath{\pm 2.01}$ & \ua{$0.01$} $0.25\tinymath{\pm 0.39}$ & \dab{$0.20$} $90.82\tinymath{\pm 0.61}$ & \dab{$0.42$} $71.30\tinymath{\pm 1.12}$ & \uag{$0.49$} $45.97\tinymath{\pm 1.20}$\\
\, + \gradiendmpi & \ua{$2.43$} $65.06\tinymath{\pm 1.94}$ & \ua{$0.09$} $0.33\tinymath{\pm 0.36}$ & \uag{$0.09$} $91.11\tinymath{\pm 0.61}$ & \dab{$0.42$} $71.31\tinymath{\pm 1.09}$ & \uag{$0.60$} $46.09\tinymath{\pm 1.25}$\\
\lightcmidrule{1-6}
\, + \cda & \da{$1.11$} $61.53\tinymath{\pm 1.96}$ & \ua{$0.07$} $0.31\tinymath{\pm 0.29}$ & \dab{$0.37$} $90.65\tinymath{\pm 0.61}$ & \!\!\uag{$\boldsymbol{1.48}$} $\mathbf{73.20}\tinymath{\pm 1.25}$ & \uag{$1.28$} $46.76\tinymath{\pm 1.38}$\\
\, + \dropout & \ua{$0.09$} $62.72\tinymath{\pm 1.92}$ & \ua{$0.24$} $0.48\tinymath{\pm 0.24}$ & \dab{$0.69$} $90.33\tinymath{\pm 0.63}$ & \dab{$0.02$} $71.70\tinymath{\pm 1.15}$ & \uag{$0.46$} $45.94\tinymath{\pm 1.45}$\\
\, + \inlp & \da{$2.27$} $60.36\tinymath{\pm 1.95}$ & \da{$0.01$} $0.23\tinymath{\pm 0.26}$ & \uag{$0.23$} $91.25\tinymath{\pm 0.59}$ & \uag{$0.02$} $71.75\tinymath{\pm 1.13}$ & \uag{$0.29$} $45.78\tinymath{\pm 1.20}$\\
\, + \rlace & \!\!\da{$\boldsymbol{\mathit{9.09}}$} $\mathit{\boldsymbol{\mathit{53.54}}}\tinymath{\pm \mathit{1.93}}$ & \!\!\da{$\boldsymbol{0.02}$} $\mathbf{0.22}\tinymath{\pm 0.24}$ & \dab{$\mathit{15.36}$} $\mathit{75.66}\tinymath{\pm \mathit{0.98}}$ & \uag{$0.59$} $72.31\tinymath{\pm 1.08}$ & \uag{$0.44$} $45.92\tinymath{\pm 1.20}$\\
\, + \leace & \da{$1.40$} $61.23\tinymath{\pm 1.98}$ & \ua{$0.00$} $0.24\tinymath{\pm 0.26}$ & \dab{$0.07$} $90.96\tinymath{\pm 0.61}$ & \dab{$0.14$} $71.58\tinymath{\pm 1.07}$ & \uag{$0.35$} $45.83\tinymath{\pm 1.24}$\\
\, + \selfdebias & \da{$0.83$} $61.80\tinymath{\pm 1.96}$ & -- & \dab{$\mathit{2.48}$} $\mathit{88.54}\tinymath{\pm \mathit{0.68}}$ & -- & --\\
\, + \sentencedebias & \da{$\mathit{6.59}$} $\mathit{56.04}\tinymath{\pm \mathit{1.96}}$ & \ua{$0.11$} $0.34\tinymath{\pm 0.27}$ & \dab{$\mathit{3.59}$} $\mathit{87.43}\tinymath{\pm \mathit{0.71}}$ & \dab{$0.26$} $71.46\tinymath{\pm 1.11}$ & \dab{$1.15$} $44.33\tinymath{\pm 1.18}$\\
\lightcmidrule{1-6}
\, + \gradiendbpi\ + \inlp & \da{$\mathit{5.27}$} $\mathit{57.36}\tinymath{\pm \mathit{1.97}}$ & \ua{$0.07$} $0.31\tinymath{\pm 0.36}$ & \dab{$0.27$} $90.75\tinymath{\pm 0.61}$ & \dab{$0.53$} $71.20\tinymath{\pm 1.07}$ & \uag{$0.82$} $46.30\tinymath{\pm 1.27}$\\
\, + \gradiendbpi\ + \sentencedebias \!\!\!\!\!\!\! & \da{$2.69$} $59.94\tinymath{\pm 2.03}$ & \ua{$0.18$} $0.42\tinymath{\pm 0.25}$ & \dab{$0.79$} $90.24\tinymath{\pm 0.62}$ & \dab{$0.20$} $71.53\tinymath{\pm 1.12}$ & \uag{$0.47$} $45.96\tinymath{\pm 1.24}$\\
\, + \cda\, + \inlp & \da{$3.73$} $58.90\tinymath{\pm 1.96}$ & \ua{$0.06$} $0.30\tinymath{\pm 0.29}$ & \!\!\uag{$\boldsymbol{0.81}$} $\mathbf{91.83}\tinymath{\pm 0.56}$ & \uag{$1.38$} $73.11\tinymath{\pm 1.24}$ & \!\!\uag{$\boldsymbol{1.39}$} $\mathbf{46.87}\tinymath{\pm 1.32}$\\
\, + \dropout \, + \sentencedebias & \da{$\mathit{5.96}$} $\mathit{56.67}\tinymath{\pm \mathit{2.00}}$ & \ua{$0.18$} $0.42\tinymath{\pm 0.19}$ & \dab{$\mathit{5.99}$} $\mathit{85.04}\tinymath{\pm \mathit{0.76}}$ & \uag{$0.55$} $72.27\tinymath{\pm 1.25}$ & \uag{$1.28$} $46.76\tinymath{\pm 1.32}$\\
\, + \cda\, + \sentencedebias & \da{$\mathit{3.98}$} $\mathit{58.65}\tinymath{\pm \mathit{1.96}}$ & \ua{$0.16$} $0.40\tinymath{\pm 0.26}$ & \dab{$1.22$} $89.81\tinymath{\pm 0.63}$ & \uag{$1.31$} $73.03\tinymath{\pm 1.27}$ & \uag{$0.78$} $46.27\tinymath{\pm 1.34}$\\
\, + \dropout \, + \inlp & \da{$3.21$} $59.42\tinymath{\pm 1.94}$ & \ua{$0.22$} $0.46\tinymath{\pm 0.23}$ & \dab{$0.04$} $90.99\tinymath{\pm 0.60}$ & \dab{$0.02$} $71.70\tinymath{\pm 1.13}$ & \uag{$1.11$} $46.59\tinymath{\pm 1.44}$\\

\midrule
\llama & $69.44\tinymath{\pm 1.73}$ & $0.93\tinymath{\pm 0.16}$ & $92.42\tinymath{\pm 0.53}$ & $45.86\tinymath{\pm 1.98}$ & \!\!$\mathbf{54.46}\tinymath{\pm 2.28}$\\
\lightcmidrule{1-6}
\, + \gradiendbpi & \da{$0.23$} $69.21\tinymath{\pm 1.75}$ & \da{$\mathit{0.26}$} $\mathit{0.67}\tinymath{\pm \mathit{0.10}}$ & \dab{$0.24$} $92.18\tinymath{\pm 0.55}$ & \uag{$1.02$} $46.88\tinymath{\pm 1.91}$ & \dab{$3.49$} $50.97\tinymath{\pm 2.20}$\\
\, + \gradiendfpi & \da{$1.48$} $67.96\tinymath{\pm 1.75}$ & \da{$0.06$} $0.87\tinymath{\pm 0.14}$ & \dab{$0.09$} $92.33\tinymath{\pm 0.54}$ & \!\!\uag{$\boldsymbol{3.33}$} $\mathbf{49.19}\tinymath{\pm 1.84}$ & \dab{$1.35$} $53.11\tinymath{\pm 2.28}$\\
\, + \gradiendmpi & \ua{$0.07$} $69.51\tinymath{\pm 1.76}$ & \da{$0.11$} $0.82\tinymath{\pm 0.11}$ & \dab{$0.16$} $92.26\tinymath{\pm 0.55}$ & \dab{$3.47$} $42.39\tinymath{\pm 2.00}$ & \dab{$2.10$} $52.35\tinymath{\pm 2.09}$\\
\lightcmidrule{1-6}
\, + \inlp & \da{$2.83$} $66.61\tinymath{\pm 1.81}$ & \da{$0.23$} $0.70\tinymath{\pm 0.16}$ & \dab{$0.48$} $91.95\tinymath{\pm 0.55}$ & \dab{$0.13$} $45.73\tinymath{\pm 1.78}$ & \dab{$\mathit{4.88}$} $\mathit{49.57}\tinymath{\pm \mathit{2.21}}$\\
\, + \rlace & \ua{$0.30$} $69.74\tinymath{\pm 1.73}$ & \da{$0.00$} $0.93\tinymath{\pm 0.16}$ & \uag{$0.05$} $92.47\tinymath{\pm 0.53}$ & \uag{$0.17$} $46.03\tinymath{\pm 1.95}$ & \dab{$\mathit{11.49}$} $\mathit{42.97}\tinymath{\pm \mathit{2.31}}$\\
\, + \leace & \ua{$0.03$} $69.47\tinymath{\pm 1.73}$ & \da{$0.01$} $0.92\tinymath{\pm 0.17}$ & \!\!\uag{$\boldsymbol{0.05}$} $\mathbf{92.47}\tinymath{\pm 0.53}$ & \uag{$0.32$} $46.17\tinymath{\pm 1.97}$ & \dab{$\mathit{11.53}$} $\mathit{42.93}\tinymath{\pm \mathit{2.31}}$\\
\, + \selfdebias & \da{$\mathit{5.75}$} $\mathit{63.69}\tinymath{\pm \mathit{1.86}}$ & -- & \dab{$\mathit{31.14}$} $\mathit{61.28}\tinymath{\pm \mathit{0.99}}$ & -- & --\\
\, + \sentencedebias & \da{$2.90$} $66.53\tinymath{\pm 1.79}$ & \da{$\mathit{0.32}$} $\mathit{0.61}\tinymath{\pm \mathit{0.14}}$ & \uag{$0.04$} $92.46\tinymath{\pm 0.53}$ & \uag{$1.32$} $47.18\tinymath{\pm 1.92}$ & \dab{$0.34$} $54.12\tinymath{\pm 2.37}$\\
\lightcmidrule{1-6}
\, + \gradiendbpi\ + \inlp & \!\!\da{$\boldsymbol{\mathit{9.41}}$} $\mathit{\boldsymbol{\mathit{60.03}}}\tinymath{\pm \mathit{1.87}}$ & \!\!\da{$\boldsymbol{\mathit{0.33}}$} $\mathit{\boldsymbol{\mathit{0.61}}}\tinymath{\pm \mathit{0.09}}$ & \dab{$0.90$} $91.53\tinymath{\pm 0.60}$ & \uag{$1.02$} $46.88\tinymath{\pm 1.91}$ & \dab{$\mathit{8.97}$} $\mathit{45.49}\tinymath{\pm \mathit{2.08}}$\\
\, + \gradiendbpi\ + \sentencedebias \!\!\!\!\!\!\! & \da{$\mathit{6.71}$} $\mathit{62.73}\tinymath{\pm \mathit{1.88}}$ & \da{$\mathit{0.30}$} $\mathit{0.63}\tinymath{\pm \mathit{0.10}}$ & \dab{$\mathit{2.50}$} $\mathit{89.93}\tinymath{\pm \mathit{0.64}}$ & \uag{$0.92$} $46.77\tinymath{\pm 1.92}$ & \dab{$4.06$} $50.40\tinymath{\pm 2.16}$\\

\midrule
\llamai & $68.53\tinymath{\pm 1.80}$ & $0.90\tinymath{\pm 0.16}$ & $92.21\tinymath{\pm 0.54}$ & $49.14\tinymath{\pm 1.92}$ & $58.07\tinymath{\pm 2.29}$\\
\lightcmidrule{1-6}
\, + \gradiendbpi & \da{$2.29$} $66.24\tinymath{\pm 1.87}$ & \da{$\mathit{0.41}$} $\mathit{0.49}\tinymath{\pm \mathit{0.15}}$ & \dab{$\mathit{2.26}$} $\mathit{89.95}\tinymath{\pm \mathit{0.63}}$ & \dab{$1.77$} $47.37\tinymath{\pm 1.81}$ & \dab{$\mathit{5.07}$} $\mathit{53.00}\tinymath{\pm \mathit{2.05}}$\\
\, + \gradiendfpi & \da{$2.16$} $66.37\tinymath{\pm 1.90}$ & \da{$0.19$} $0.71\tinymath{\pm 0.20}$ & \dab{$0.37$} $91.84\tinymath{\pm 0.56}$ & \dab{$3.02$} $46.12\tinymath{\pm 1.83}$ & \!\!\uag{$\boldsymbol{2.68}$} $\mathbf{60.75}\tinymath{\pm 2.35}$\\
\, + \gradiendmpi & \da{$1.61$} $66.92\tinymath{\pm 1.88}$ & \da{$0.19$} $0.71\tinymath{\pm 0.13}$ & \dab{$\mathit{1.84}$} $\mathit{90.37}\tinymath{\pm \mathit{0.60}}$ & \!\!\uag{$\boldsymbol{\mathit{11.33}}$} $\mathit{\boldsymbol{\mathit{60.47}}}\tinymath{\pm \mathit{1.86}}$ & \dab{$1.71$} $56.36\tinymath{\pm 2.28}$\\
\lightcmidrule{1-6}
\, + \inlp & \da{$2.40$} $66.13\tinymath{\pm 1.82}$ & \da{$0.33$} $0.57\tinymath{\pm 0.19}$ & \dab{$0.22$} $91.99\tinymath{\pm 0.55}$ & \dab{$0.95$} $48.19\tinymath{\pm 1.85}$ & \dab{$0.72$} $57.35\tinymath{\pm 2.34}$\\
\, + \rlace & \ua{$0.17$} $68.70\tinymath{\pm 1.80}$ & \da{$0.20$} $0.70\tinymath{\pm 0.20}$ & \uag{$0.01$} $92.22\tinymath{\pm 0.54}$ & \uag{$0.10$} $49.24\tinymath{\pm 1.93}$ & \dab{$0.05$} $58.02\tinymath{\pm 2.28}$\\
\, + \leace & \da{$0.37$} $68.16\tinymath{\pm 1.82}$ & \da{$0.20$} $0.69\tinymath{\pm 0.20}$ & \!\!\uag{$\boldsymbol{0.05}$} $\mathbf{92.26}\tinymath{\pm 0.54}$ & \dab{$0.01$} $49.13\tinymath{\pm 1.91}$ & \dab{$0.23$} $57.84\tinymath{\pm 2.29}$\\
\, + \selfdebias & \!\!\da{$\boldsymbol{\mathit{10.35}}$} $\mathit{\boldsymbol{\mathit{58.18}}}\tinymath{\pm \mathit{1.98}}$ & -- & \dab{$\mathit{32.73}$} $\mathit{59.48}\tinymath{\pm \mathit{1.00}}$ & -- & --\\
\, + \sentencedebias & \da{$1.79$} $66.74\tinymath{\pm 1.84}$ & \da{$\mathit{0.47}$} $\mathit{0.43}\tinymath{\pm \mathit{0.13}}$ & \uag{$0.02$} $92.24\tinymath{\pm 0.54}$ & \dab{$0.06$} $49.08\tinymath{\pm 1.91}$ & \uag{$0.49$} $58.56\tinymath{\pm 2.31}$\\
\lightcmidrule{1-6}
\, + \gradiendbpi\ + \inlp & \da{$\mathit{4.88}$} $\mathit{63.65}\tinymath{\pm \mathit{1.87}}$ & \!\!\da{$\boldsymbol{\mathit{0.50}}$} $\mathit{\boldsymbol{\mathit{0.39}}}\tinymath{\pm \mathit{0.13}}$ & \dab{$\mathit{2.07}$} $\mathit{90.15}\tinymath{\pm \mathit{0.61}}$ & \dab{$2.36$} $46.78\tinymath{\pm 1.85}$ & \dab{$\mathit{7.99}$} $\mathit{50.08}\tinymath{\pm \mathit{2.08}}$\\
\, + \gradiendbpi\ + \sentencedebias \!\!\!\!\!\!\! & \da{$2.41$} $66.12\tinymath{\pm 1.89}$ & \da{$\mathit{0.43}$} $\mathit{0.46}\tinymath{\pm \mathit{0.14}}$ & \dab{$\mathit{2.30}$} $\mathit{89.91}\tinymath{\pm \mathit{0.62}}$ & \dab{$0.91$} $48.23\tinymath{\pm 1.85}$ & \dab{$\mathit{5.10}$} $\mathit{52.97}\tinymath{\pm \mathit{2.04}}$\\

\bottomrule
\end{tabular}
 \vspace*{50pt} 
\end{table*}

\begin{table}[]
    \centering
    \tiny
    \setlength{\tabcolsep}{4pt} 
   \caption{\textbf{Race:} Comparison of bootstrapped bias metrics (\acrshort{ss} and \acrshort{seat})) and language modeling metrics (\acrshort{lms}, \acrshort{glue}, and \acrshort{sglue})  for all models across different race debiasing techniques. Statistically significant improvements are indicated in \emph{italics}, while the best score for each base model is highlighted in \textbf{bold}.}
    \label{tab:results-race}
    \begin{tabular}{lrrrrr}
\toprule\textbf{Model} & \textbf{\acrshort{ss}} (\%) \bestatfiftytiny & \textbf{\acrshort{seat}} $\downarrow$ & \textbf{\acrshort{lms}} (\%) $\uparrow$ & \textbf{\acrshort{glue}} (\%)  $\uparrow$ & \textbf{\acrshort{sglue}} (\%) $\uparrow$\\
\midrule
\bertbase & $57.04\tinymath{\pm 1.01}$ & $0.52\tinymath{\pm 0.26}$ & $82.50\tinymath{\pm 0.81}$ & $78.09\tinymath{\pm 1.59}$ & $51.82\tinymath{\pm 1.67}$\\
\lightcmidrule{1-6}
\, + \gradiendraceab & \da{$1.88$} $55.15\tinymath{\pm 1.02}$ & \ua{$0.08$} $0.60\tinymath{\pm 0.28}$ & \dab{$0.21$} $82.29\tinymath{\pm 0.80}$ & \uag{$0.47$} $78.56\tinymath{\pm 1.60}$ & \uag{$0.80$} $52.62\tinymath{\pm 1.89}$\\
\, + \gradiendraceaw & \da{$1.11$} $55.92\tinymath{\pm 1.01}$ & \ua{$0.08$} $0.60\tinymath{\pm 0.24}$ & \dab{$1.13$} $81.37\tinymath{\pm 0.82}$ & \!\!\uag{$\boldsymbol{0.65}$} $\mathbf{78.74}\tinymath{\pm 1.61}$ & \uag{$0.53$} $52.36\tinymath{\pm 1.90}$\\
\, + \gradiendracebw & \ua{$0.23$} $57.27\tinymath{\pm 1.01}$ & \da{$0.01$} $0.51\tinymath{\pm 0.26}$ & \dab{$0.13$} $82.37\tinymath{\pm 0.80}$ & \uag{$0.28$} $78.37\tinymath{\pm 1.56}$ & \uag{$0.82$} $52.65\tinymath{\pm 1.88}$\\
\lightcmidrule{1-6}
\, + \cda & \ua{$1.02$} $58.06\tinymath{\pm 1.03}$ & \!\!\da{$\boldsymbol{0.26}$} $\mathbf{0.26}\tinymath{\pm 0.13}$ & \dab{$0.65$} $81.85\tinymath{\pm 0.82}$ & \dab{$0.22$} $77.88\tinymath{\pm 1.48}$ & \!\!\uag{$\boldsymbol{1.15}$} $\mathbf{52.98}\tinymath{\pm 1.82}$\\
\, + \dropout & \da{$0.54$} $56.50\tinymath{\pm 1.02}$ & \da{$0.07$} $0.44\tinymath{\pm 0.17}$ & \dab{$\mathit{1.75}$} $\mathit{80.75}\tinymath{\pm \mathit{0.83}}$ & \dab{$1.40$} $76.69\tinymath{\pm 1.44}$ & \dab{$0.34$} $51.48\tinymath{\pm 1.72}$\\
\, + \inlp & \da{$0.07$} $56.97\tinymath{\pm 0.99}$ & \da{$0.02$} $0.50\tinymath{\pm 0.19}$ & \!\!\uag{$\boldsymbol{0.36}$} $\mathbf{82.86}\tinymath{\pm 0.80}$ & \dab{$0.24$} $77.86\tinymath{\pm 1.23}$ & \dab{$1.38$} $50.44\tinymath{\pm 1.54}$\\
\, + \selfdebias & \!\!\da{$\boldsymbol{\mathit{2.59}}$} $\mathit{\boldsymbol{\mathit{54.45}}}\tinymath{\pm \mathit{1.04}}$ & -- & \uag{$0.06$} $82.56\tinymath{\pm 0.83}$ & -- & --\\
\, + \sentencedebias & \da{$0.38$} $56.65\tinymath{\pm 1.01}$ & \ua{$0.00$} $0.52\tinymath{\pm 0.26}$ & \dab{$0.02$} $82.48\tinymath{\pm 0.81}$ & \uag{$0.04$} $78.14\tinymath{\pm 1.58}$ & \dab{$0.95$} $50.87\tinymath{\pm 1.54}$\\

\midrule
\bertlarge & $57.00\tinymath{\pm 1.02}$ & $0.45\tinymath{\pm 0.10}$ & $82.89\tinymath{\pm 0.80}$ & $79.98\tinymath{\pm 1.31}$ & $53.74\tinymath{\pm 1.62}$\\
\lightcmidrule{1-6}
\, + \gradiendraceab & \ua{$1.19$} $58.19\tinymath{\pm 1.01}$ & \ua{$0.04$} $0.49\tinymath{\pm 0.13}$ & \dab{$0.45$} $82.44\tinymath{\pm 0.81}$ & \uag{$0.40$} $80.38\tinymath{\pm 1.55}$ & \uag{$0.53$} $54.27\tinymath{\pm 1.85}$\\
\, + \gradiendraceaw & \!\!\da{$\boldsymbol{\mathit{3.00}}$} $\mathit{\boldsymbol{\mathit{54.00}}}\tinymath{\pm \mathit{1.01}}$ & \ua{$0.07$} $0.52\tinymath{\pm 0.13}$ & \dab{$1.11$} $81.77\tinymath{\pm 0.83}$ & \!\!\uag{$\boldsymbol{0.90}$} $\mathbf{80.88}\tinymath{\pm 1.53}$ & \!\!\uag{$\boldsymbol{0.84}$} $\mathbf{54.58}\tinymath{\pm 1.85}$\\
\, + \gradiendracebw & \da{$0.04$} $56.96\tinymath{\pm 1.02}$ & \ua{$0.02$} $0.47\tinymath{\pm 0.11}$ & \dab{$0.23$} $82.66\tinymath{\pm 0.81}$ & \uag{$0.51$} $80.49\tinymath{\pm 1.54}$ & \uag{$0.68$} $54.42\tinymath{\pm 1.87}$\\
\lightcmidrule{1-6}
\, + \cda & \da{$0.01$} $57.00\tinymath{\pm 1.03}$ & \!\!\da{$\boldsymbol{0.04}$} $\mathbf{0.41}\tinymath{\pm 0.13}$ & \dab{$0.47$} $82.42\tinymath{\pm 0.80}$ & \dab{$2.01$} $77.97\tinymath{\pm 0.97}$ & \dab{$0.59$} $53.15\tinymath{\pm 1.73}$\\
\, + \dropout & \da{$0.91$} $56.09\tinymath{\pm 1.03}$ & \da{$0.03$} $0.42\tinymath{\pm 0.12}$ & \dab{$\mathit{2.57}$} $\mathit{80.32}\tinymath{\pm \mathit{0.82}}$ & \dab{$0.55$} $79.43\tinymath{\pm 1.46}$ & \dab{$0.52$} $53.22\tinymath{\pm 1.68}$\\
\, + \inlp & \ua{$0.01$} $57.01\tinymath{\pm 1.04}$ & \ua{$0.00$} $0.45\tinymath{\pm 0.11}$ & \!\!\uag{$\boldsymbol{0.17}$} $\mathbf{83.06}\tinymath{\pm 0.79}$ & \uag{$0.03$} $80.02\tinymath{\pm 1.29}$ & \dab{$0.24$} $53.50\tinymath{\pm 1.58}$\\
\, + \selfdebias & \da{$1.02$} $55.98\tinymath{\pm 1.02}$ & -- & \dab{$0.00$} $82.88\tinymath{\pm 0.79}$ & -- & --\\
\, + \sentencedebias & \da{$0.19$} $56.82\tinymath{\pm 1.02}$ & \ua{$0.00$} $0.45\tinymath{\pm 0.10}$ & \dab{$0.02$} $82.87\tinymath{\pm 0.80}$ & \uag{$0.12$} $80.10\tinymath{\pm 1.53}$ & \dab{$0.07$} $53.68\tinymath{\pm 1.67}$\\

\midrule
\distilbert & $56.09\tinymath{\pm 1.04}$ & $0.30\tinymath{\pm 0.16}$ & \!\!$\mathbf{82.06}\tinymath{\pm 0.80}$ & $74.47\tinymath{\pm 1.59}$ & $49.69\tinymath{\pm 1.65}$\\
\lightcmidrule{1-6}
\, + \gradiendraceab & \ua{$1.36$} $57.44\tinymath{\pm 1.04}$ & \ua{$0.02$} $0.32\tinymath{\pm 0.16}$ & \dab{$0.28$} $81.79\tinymath{\pm 0.80}$ & \dab{$0.06$} $74.41\tinymath{\pm 1.60}$ & \uag{$0.00$} $49.69\tinymath{\pm 1.69}$\\
\, + \gradiendraceaw & \da{$1.01$} $55.08\tinymath{\pm 1.05}$ & \ua{$0.00$} $0.30\tinymath{\pm 0.16}$ & \dab{$0.63$} $81.44\tinymath{\pm 0.81}$ & \dab{$0.12$} $74.34\tinymath{\pm 1.60}$ & \uag{$0.38$} $50.07\tinymath{\pm 1.70}$\\
\, + \gradiendracebw & \da{$0.08$} $56.01\tinymath{\pm 1.04}$ & \ua{$0.03$} $0.33\tinymath{\pm 0.17}$ & \dab{$0.28$} $81.78\tinymath{\pm 0.81}$ & \dab{$0.08$} $74.38\tinymath{\pm 1.47}$ & \dab{$0.40$} $49.29\tinymath{\pm 1.70}$\\
\lightcmidrule{1-6}
\, + \cda & \ua{$0.87$} $56.95\tinymath{\pm 1.03}$ & \ua{$0.05$} $0.36\tinymath{\pm 0.12}$ & \dab{$0.71$} $81.36\tinymath{\pm 0.83}$ & \uag{$0.19$} $74.65\tinymath{\pm 1.45}$ & \uag{$1.11$} $50.80\tinymath{\pm 1.79}$\\
\, + \dropout & \ua{$1.12$} $57.21\tinymath{\pm 1.02}$ & \ua{$0.11$} $0.41\tinymath{\pm 0.13}$ & \dab{$\mathit{1.82}$} $\mathit{80.24}\tinymath{\pm \mathit{0.85}}$ & \!\!\uag{$\boldsymbol{0.70}$} $\mathbf{75.17}\tinymath{\pm 1.50}$ & \uag{$0.58$} $50.27\tinymath{\pm 1.75}$\\
\, + \inlp & \da{$0.54$} $55.54\tinymath{\pm 1.05}$ & \!\!\da{$\boldsymbol{0.09}$} $\mathbf{0.21}\tinymath{\pm 0.12}$ & \dab{$0.35$} $81.71\tinymath{\pm 0.80}$ & \uag{$0.17$} $74.64\tinymath{\pm 1.57}$ & \!\!\uag{$\boldsymbol{1.14}$} $\mathbf{50.83}\tinymath{\pm 1.68}$\\
\, + \selfdebias & \!\!\da{$\boldsymbol{1.19}$} $\mathbf{54.89}\tinymath{\pm 1.02}$ & -- & \dab{$0.16$} $81.91\tinymath{\pm 0.81}$ & -- & --\\
\, + \sentencedebias & \da{$0.03$} $56.06\tinymath{\pm 1.05}$ & \ua{$0.00$} $0.30\tinymath{\pm 0.16}$ & \dab{$0.40$} $81.66\tinymath{\pm 0.81}$ & \uag{$0.10$} $74.57\tinymath{\pm 1.60}$ & \dab{$0.22$} $49.47\tinymath{\pm 1.62}$\\

\midrule
\roberta & $60.13\tinymath{\pm 0.97}$ & $0.43\tinymath{\pm 0.17}$ & $89.09\tinymath{\pm 0.64}$ & \!\!$\mathbf{81.65}\tinymath{\pm 1.44}$ & $53.31\tinymath{\pm 1.48}$\\
\lightcmidrule{1-6}
\, + \gradiendraceab & \!\!\da{$\boldsymbol{\mathit{6.26}}$} $\mathit{\boldsymbol{\mathit{53.88}}}\tinymath{\pm \mathit{1.04}}$ & \ua{$0.02$} $0.44\tinymath{\pm 0.12}$ & \dab{$\mathit{5.82}$} $\mathit{83.26}\tinymath{\pm \mathit{0.78}}$ & \dab{$\mathit{11.58}$} $\mathit{70.07}\tinymath{\pm \mathit{1.48}}$ & \dab{$\mathit{7.07}$} $\mathit{46.24}\tinymath{\pm \mathit{1.51}}$\\
\, + \gradiendraceaw & \da{$\mathit{5.57}$} $\mathit{54.56}\tinymath{\pm \mathit{0.99}}$ & \da{$0.03$} $0.40\tinymath{\pm 0.14}$ & \dab{$\mathit{3.37}$} $\mathit{85.71}\tinymath{\pm \mathit{0.75}}$ & \dab{$\mathit{8.51}$} $\mathit{73.14}\tinymath{\pm \mathit{0.86}}$ & \dab{$\mathit{3.27}$} $\mathit{50.04}\tinymath{\pm \mathit{1.49}}$\\
\, + \gradiendracebw & \ua{$\mathit{3.37}$} $\mathit{63.50}\tinymath{\pm \mathit{0.99}}$ & \da{$0.01$} $0.42\tinymath{\pm 0.17}$ & \!\!\uag{$\boldsymbol{0.40}$} $\mathbf{89.49}\tinymath{\pm 0.63}$ & \dab{$\mathit{5.20}$} $\mathit{76.45}\tinymath{\pm \mathit{1.16}}$ & \uag{$0.41$} $53.72\tinymath{\pm 1.67}$\\
\lightcmidrule{1-6}
\, + \cda & \ua{$0.49$} $60.62\tinymath{\pm 0.97}$ & \!\!\da{$\boldsymbol{0.05}$} $\mathbf{0.38}\tinymath{\pm 0.15}$ & \dab{$\mathit{3.32}$} $\mathit{85.77}\tinymath{\pm \mathit{0.74}}$ & \dab{$0.07$} $81.58\tinymath{\pm 1.29}$ & \!\!\uag{$\boldsymbol{\mathit{4.17}}$} $\mathit{\boldsymbol{\mathit{57.48}}}\tinymath{\pm \mathit{1.71}}$\\
\, + \dropout & \da{$\mathit{4.04}$} $\mathit{56.09}\tinymath{\pm \mathit{0.98}}$ & \ua{$0.18$} $0.61\tinymath{\pm 0.16}$ & \dab{$\mathit{3.74}$} $\mathit{85.34}\tinymath{\pm \mathit{0.72}}$ & \dab{$\mathit{14.16}$} $\mathit{67.49}\tinymath{\pm \mathit{1.47}}$ & \dab{$2.25$} $51.05\tinymath{\pm 1.62}$\\
\, + \inlp & \da{$0.83$} $59.31\tinymath{\pm 0.98}$ & \ua{$0.02$} $0.45\tinymath{\pm 0.16}$ & \dab{$0.52$} $88.57\tinymath{\pm 0.65}$ & \dab{$0.11$} $81.54\tinymath{\pm 1.48}$ & \uag{$0.46$} $53.77\tinymath{\pm 1.19}$\\
\, + \selfdebias & \da{$\mathit{2.32}$} $\mathit{57.82}\tinymath{\pm \mathit{1.00}}$ & -- & \dab{$0.29$} $88.79\tinymath{\pm 0.64}$ & -- & --\\
\, + \sentencedebias & \ua{$0.28$} $60.42\tinymath{\pm 0.97}$ & \ua{$0.01$} $0.44\tinymath{\pm 0.17}$ & \dab{$0.08$} $89.01\tinymath{\pm 0.64}$ & \dab{$\mathit{3.17}$} $\mathit{78.48}\tinymath{\pm \mathit{1.30}}$ & \uag{$2.23$} $55.54\tinymath{\pm 1.49}$\\

\midrule
\gpttwo & $58.90\tinymath{\pm 0.99}$ & $0.47\tinymath{\pm 0.33}$ & $91.02\tinymath{\pm 0.62}$ & $71.73\tinymath{\pm 1.08}$ & $45.49\tinymath{\pm 1.28}$\\
\lightcmidrule{1-6}
\, + \gradiendraceab & \!\!\da{$\boldsymbol{\mathit{5.87}}$} $\mathit{\boldsymbol{\mathit{53.03}}}\tinymath{\pm \mathit{1.01}}$ & \da{$0.07$} $0.40\tinymath{\pm 0.30}$ & \dab{$0.27$} $90.75\tinymath{\pm 0.60}$ & \dab{$0.58$} $71.14\tinymath{\pm 1.01}$ & \dab{$0.04$} $45.45\tinymath{\pm 1.18}$\\
\, + \gradiendraceaw & \da{$0.40$} $58.50\tinymath{\pm 1.00}$ & \da{$0.06$} $0.41\tinymath{\pm 0.27}$ & \dab{$0.04$} $90.98\tinymath{\pm 0.60}$ & \dab{$1.08$} $70.65\tinymath{\pm 0.98}$ & \uag{$0.43$} $45.92\tinymath{\pm 1.22}$\\
\, + \gradiendracebw & \ua{$0.11$} $59.01\tinymath{\pm 0.99}$ & \ua{$0.01$} $0.48\tinymath{\pm 0.33}$ & \dab{$0.01$} $91.01\tinymath{\pm 0.62}$ & \dab{$0.22$} $71.50\tinymath{\pm 1.07}$ & \uag{$0.48$} $45.97\tinymath{\pm 1.13}$\\
\lightcmidrule{1-6}
\, + \cda & \da{$0.42$} $58.48\tinymath{\pm 0.98}$ & \ua{$0.02$} $0.49\tinymath{\pm 0.29}$ & \dab{$\mathit{2.88}$} $\mathit{88.15}\tinymath{\pm \mathit{0.67}}$ & \!\!\uag{$\boldsymbol{1.74}$} $\mathbf{73.47}\tinymath{\pm 1.12}$ & \!\!\uag{$\boldsymbol{1.47}$} $\mathbf{46.96}\tinymath{\pm 1.27}$\\
\, + \dropout & \da{$1.48$} $57.42\tinymath{\pm 1.01}$ & \da{$0.08$} $0.39\tinymath{\pm 0.38}$ & \dab{$0.69$} $90.33\tinymath{\pm 0.63}$ & \dab{$0.02$} $71.70\tinymath{\pm 1.15}$ & \uag{$0.46$} $45.94\tinymath{\pm 1.45}$\\
\, + \inlp & \ua{$0.10$} $59.00\tinymath{\pm 0.98}$ & \da{$0.00$} $0.47\tinymath{\pm 0.33}$ & \uag{$0.04$} $91.07\tinymath{\pm 0.61}$ & \dab{$0.28$} $71.45\tinymath{\pm 1.08}$ & \uag{$0.29$} $45.77\tinymath{\pm 1.22}$\\
\, + \selfdebias & \da{$\mathit{2.45}$} $\mathit{56.45}\tinymath{\pm \mathit{1.01}}$ & -- & \dab{$\mathit{1.99}$} $\mathit{89.04}\tinymath{\pm \mathit{0.68}}$ & -- & --\\
\, + \sentencedebias & \da{$\mathit{2.46}$} $\mathit{56.44}\tinymath{\pm \mathit{1.01}}$ & \!\!\da{$\boldsymbol{0.10}$} $\mathbf{0.37}\tinymath{\pm 0.21}$ & \!\!\uag{$\boldsymbol{0.36}$} $\mathbf{91.38}\tinymath{\pm 0.59}$ & \dab{$0.18$} $71.55\tinymath{\pm 1.09}$ & \dab{$0.75$} $44.73\tinymath{\pm 1.26}$\\

\midrule
\llama & $65.06\tinymath{\pm 0.98}$ & $0.21\tinymath{\pm 0.08}$ & \!\!$\mathbf{92.42}\tinymath{\pm 0.53}$ & $45.86\tinymath{\pm 1.98}$ & $54.46\tinymath{\pm 2.28}$\\
\lightcmidrule{1-6}
\, + \gradiendraceab & \da{$\mathit{2.20}$} $\mathit{62.86}\tinymath{\pm \mathit{1.02}}$ & \ua{$0.02$} $0.23\tinymath{\pm 0.07}$ & \dab{$\mathit{2.52}$} $\mathit{89.91}\tinymath{\pm \mathit{0.64}}$ & \!\!\uag{$\boldsymbol{3.58}$} $\mathbf{49.44}\tinymath{\pm 1.97}$ & \dab{$2.43$} $52.02\tinymath{\pm 2.27}$\\
\, + \gradiendraceaw & \da{$0.99$} $64.07\tinymath{\pm 0.99}$ & \ua{$0.03$} $0.24\tinymath{\pm 0.11}$ & \dab{$0.76$} $91.67\tinymath{\pm 0.56}$ & \uag{$1.20$} $47.06\tinymath{\pm 1.97}$ & \dab{$0.05$} $54.40\tinymath{\pm 2.22}$\\
\, + \gradiendracebw & \da{$0.65$} $64.41\tinymath{\pm 0.99}$ & \ua{$0.01$} $0.22\tinymath{\pm 0.08}$ & \dab{$0.42$} $92.01\tinymath{\pm 0.55}$ & \dab{$1.46$} $44.40\tinymath{\pm 2.01}$ & \dab{$0.70$} $53.76\tinymath{\pm 2.03}$\\
\lightcmidrule{1-6}
\, + \inlp & \ua{$0.23$} $65.29\tinymath{\pm 0.99}$ & \!\!\da{$\boldsymbol{0.00}$} $\mathbf{0.21}\tinymath{\pm 0.08}$ & \dab{$0.16$} $92.26\tinymath{\pm 0.54}$ & \uag{$1.93$} $47.79\tinymath{\pm 1.87}$ & \uag{$0.38$} $54.84\tinymath{\pm 2.12}$\\
\, + \selfdebias & \!\!\da{$\boldsymbol{\mathit{5.78}}$} $\mathit{\boldsymbol{\mathit{59.28}}}\tinymath{\pm \mathit{1.04}}$ & -- & \dab{$\mathit{2.28}$} $\mathit{90.14}\tinymath{\pm \mathit{0.59}}$ & -- & --\\
\, + \sentencedebias & \da{$0.04$} $65.02\tinymath{\pm 0.98}$ & \ua{$0.01$} $0.22\tinymath{\pm 0.09}$ & \dab{$0.03$} $92.39\tinymath{\pm 0.54}$ & \uag{$0.52$} $46.38\tinymath{\pm 1.93}$ & \!\!\uag{$\boldsymbol{0.39}$} $\mathbf{54.85}\tinymath{\pm 2.29}$\\

\midrule
\llamai & $63.72\tinymath{\pm 0.98}$ & $0.34\tinymath{\pm 0.14}$ & $92.21\tinymath{\pm 0.54}$ & $49.14\tinymath{\pm 1.92}$ & $58.07\tinymath{\pm 2.29}$\\
\lightcmidrule{1-6}
\, + \gradiendraceab & \da{$\mathit{3.19}$} $\mathit{60.53}\tinymath{\pm \mathit{0.98}}$ & \ua{$\mathit{0.52}$} $\mathit{0.86}\tinymath{\pm \mathit{0.01}}$ & \dab{$\mathit{47.40}$} $\mathit{44.81}\tinymath{\pm \mathit{1.04}}$ & \dab{$\mathit{11.73}$} $\mathit{37.41}\tinymath{\pm \mathit{1.75}}$ & \dab{$\mathit{15.62}$} $\mathit{42.45}\tinymath{\pm \mathit{2.13}}$\\
\, + \gradiendraceaw & \!\!\da{$\boldsymbol{\mathit{9.50}}$} $\mathit{\boldsymbol{\mathit{54.22}}}\tinymath{\pm \mathit{1.00}}$ & \ua{$0.05$} $0.39\tinymath{\pm 0.05}$ & \dab{$\mathit{31.14}$} $\mathit{61.07}\tinymath{\pm \mathit{0.96}}$ & \dab{$\mathit{12.38}$} $\mathit{36.76}\tinymath{\pm \mathit{1.75}}$ & \dab{$\mathit{15.58}$} $\mathit{42.49}\tinymath{\pm \mathit{2.47}}$\\
\, + \gradiendracebw & \da{$0.69$} $63.03\tinymath{\pm 0.98}$ & \ua{$0.11$} $0.45\tinymath{\pm 0.13}$ & \!\!\uag{$\boldsymbol{0.05}$} $\mathbf{92.26}\tinymath{\pm 0.54}$ & \dab{$0.48$} $48.66\tinymath{\pm 2.02}$ & \!\!\uag{$\boldsymbol{0.58}$} $\mathbf{58.65}\tinymath{\pm 2.05}$\\
\lightcmidrule{1-6}
\, + \inlp & \da{$0.23$} $63.49\tinymath{\pm 1.00}$ & \ua{$0.01$} $0.35\tinymath{\pm 0.14}$ & \dab{$0.00$} $92.21\tinymath{\pm 0.54}$ & \!\!\uag{$\boldsymbol{0.77}$} $\mathbf{49.91}\tinymath{\pm 1.98}$ & \uag{$0.28$} $58.35\tinymath{\pm 2.34}$\\
\, + \selfdebias & \da{$\mathit{5.91}$} $\mathit{57.81}\tinymath{\pm \mathit{1.05}}$ & -- & \dab{$\mathit{4.02}$} $\mathit{88.19}\tinymath{\pm \mathit{0.68}}$ & -- & --\\
\, + \sentencedebias & \da{$0.29$} $63.43\tinymath{\pm 0.99}$ & \!\!\da{$\boldsymbol{0.00}$} $\mathbf{0.34}\tinymath{\pm 0.14}$ & \dab{$0.21$} $92.00\tinymath{\pm 0.55}$ & \dab{$0.19$} $48.95\tinymath{\pm 1.96}$ & \uag{$0.46$} $58.53\tinymath{\pm 2.41}$\\
\bottomrule
\end{tabular}
\end{table}

\begin{table}[]
    \centering
    \tiny
     \setlength{\tabcolsep}{3pt} 
      \caption{\textbf{Religion:} Comparison of bootstrapped bias metrics (\acrshort{ss} and \acrshort{seat})) and language modeling metrics (\acrshort{lms}, \acrshort{glue}, and \acrshort{sglue})  for all models across different religion debiasing techniques. Statistically significant improvements are indicated in \emph{italics}, while the best score for each base model is highlighted in \textbf{bold}.}
    \label{tab:results-religion}
   \begin{tabular}{lrrrrr}
\toprule\textbf{Model} & \textbf{\acrshort{ss}} (\%) \bestatfiftytiny & \textbf{\acrshort{seat}} $\downarrow$ & \textbf{\acrshort{lms}} (\%) $\uparrow$ & \textbf{\acrshort{glue}} (\%) $\uparrow$ & \textbf{\acrshort{sglue}} (\%) $\uparrow$\\
\midrule
\bertbase & $52.77\tinymath{\pm 3.68}$ & $0.38\tinymath{\pm 0.21}$ & $82.50\tinymath{\pm 0.81}$ & $78.09\tinymath{\pm 1.59}$ & $51.82\tinymath{\pm 1.67}$\\
\lightcmidrule{1-6}
\, + \gradiendreligioncj & \ua{$3.28$} $56.05\tinymath{\pm 3.65}$ & \ua{$0.04$} $0.42\tinymath{\pm 0.21}$ & \uag{$0.04$} $82.54\tinymath{\pm 0.81}$ & \uag{$0.24$} $78.33\tinymath{\pm 1.58}$ & \uag{$1.17$} $53.00\tinymath{\pm 1.89}$\\
\, + \gradiendreligioncm & \ua{$1.25$} $54.03\tinymath{\pm 3.68}$ & \ua{$0.08$} $0.47\tinymath{\pm 0.18}$ & \uag{$0.01$} $82.51\tinymath{\pm 0.81}$ & \!\!\uag{$\boldsymbol{0.34}$} $\mathbf{78.43}\tinymath{\pm 1.57}$ & \uag{$0.71$} $52.54\tinymath{\pm 1.88}$\\
\, + \gradiendreligionjm & \da{$0.91$} $51.86\tinymath{\pm 3.64}$ & \ua{$0.13$} $0.51\tinymath{\pm 0.24}$ & \dab{$0.11$} $82.39\tinymath{\pm 0.81}$ & \uag{$0.28$} $78.37\tinymath{\pm 1.60}$ & \uag{$0.89$} $52.71\tinymath{\pm 1.89}$\\
\lightcmidrule{1-6}
\, + \cda & \ua{$2.43$} $55.21\tinymath{\pm 3.56}$ & \!\!\da{$\boldsymbol{0.23}$} $\mathbf{0.16}\tinymath{\pm 0.10}$ & \!\!\uag{$\boldsymbol{0.32}$} $\mathbf{82.82}\tinymath{\pm 0.81}$ & \uag{$0.27$} $78.36\tinymath{\pm 1.49}$ & \!\!\uag{$\boldsymbol{1.26}$} $\mathbf{53.08}\tinymath{\pm 1.83}$\\
\, + \dropout & \!\!\da{$\boldsymbol{2.11}$} $\mathbf{50.67}\tinymath{\pm 3.45}$ & \da{$0.00$} $0.38\tinymath{\pm 0.16}$ & \dab{$\mathit{1.75}$} $\mathit{80.75}\tinymath{\pm \mathit{0.83}}$ & \dab{$1.40$} $76.69\tinymath{\pm 1.44}$ & \dab{$0.34$} $51.48\tinymath{\pm 1.72}$\\
\, + \inlp & \da{$0.54$} $52.23\tinymath{\pm 3.67}$ & \da{$0.02$} $0.36\tinymath{\pm 0.14}$ & \dab{$0.67$} $81.83\tinymath{\pm 0.82}$ & \dab{$0.39$} $77.71\tinymath{\pm 1.23}$ & \dab{$1.13$} $50.70\tinymath{\pm 1.58}$\\
\, + \selfdebias & \da{$1.33$} $51.45\tinymath{\pm 3.59}$ & -- & \uag{$0.05$} $82.55\tinymath{\pm 0.82}$ & -- & --\\
\, + \sentencedebias & \da{$1.85$} $49.07\tinymath{\pm 3.62}$ & \da{$0.02$} $0.36\tinymath{\pm 0.21}$ & \dab{$0.14$} $82.35\tinymath{\pm 0.80}$ & \dab{$0.22$} $77.88\tinymath{\pm 1.03}$ & \dab{$0.57$} $51.25\tinymath{\pm 1.54}$\\

\midrule
\bertlarge & $56.12\tinymath{\pm 3.50}$ & $0.75\tinymath{\pm 0.24}$ & $82.89\tinymath{\pm 0.80}$ & $79.98\tinymath{\pm 1.31}$ & $53.74\tinymath{\pm 1.62}$\\
\lightcmidrule{1-6}
\, + \gradiendreligioncj & \da{$1.96$} $54.16\tinymath{\pm 3.56}$ & \ua{$0.11$} $0.86\tinymath{\pm 0.23}$ & \dab{$0.46$} $82.43\tinymath{\pm 0.83}$ & \!\!\uag{$\boldsymbol{0.90}$} $\mathbf{80.88}\tinymath{\pm 1.55}$ & \uag{$0.78$} $54.52\tinymath{\pm 1.84}$\\
\, + \gradiendreligioncm & \da{$1.76$} $54.36\tinymath{\pm 3.55}$ & \ua{$0.04$} $0.79\tinymath{\pm 0.20}$ & \dab{$0.33$} $82.56\tinymath{\pm 0.81}$ & \uag{$0.38$} $80.36\tinymath{\pm 1.55}$ & \uag{$0.83$} $54.58\tinymath{\pm 1.86}$\\
\, + \gradiendreligionjm & \ua{$2.55$} $58.66\tinymath{\pm 3.51}$ & \da{$0.02$} $0.73\tinymath{\pm 0.14}$ & \!\!\uag{$\boldsymbol{0.25}$} $\mathbf{83.14}\tinymath{\pm 0.78}$ & \uag{$0.64$} $80.62\tinymath{\pm 1.55}$ & \!\!\uag{$\boldsymbol{1.08}$} $\mathbf{54.82}\tinymath{\pm 1.85}$\\
\lightcmidrule{1-6}
\, + \cda & \da{$1.88$} $54.24\tinymath{\pm 3.55}$ & \da{$0.10$} $0.65\tinymath{\pm 0.16}$ & \dab{$0.04$} $82.84\tinymath{\pm 0.80}$ & \uag{$0.43$} $80.42\tinymath{\pm 1.53}$ & \uag{$0.52$} $54.27\tinymath{\pm 1.79}$\\
\, + \dropout & \da{$1.64$} $54.48\tinymath{\pm 3.44}$ & \ua{$0.16$} $0.91\tinymath{\pm 0.26}$ & \dab{$\mathit{2.57}$} $\mathit{80.32}\tinymath{\pm \mathit{0.82}}$ & \dab{$0.55$} $79.43\tinymath{\pm 1.46}$ & \dab{$0.52$} $53.22\tinymath{\pm 1.68}$\\
\, + \inlp & \da{$1.92$} $54.20\tinymath{\pm 3.47}$ & \!\!\da{$\boldsymbol{0.19}$} $\mathbf{0.56}\tinymath{\pm 0.17}$ & \dab{$0.28$} $82.61\tinymath{\pm 0.80}$ & \dab{$0.12$} $79.86\tinymath{\pm 1.08}$ & \dab{$0.54$} $53.21\tinymath{\pm 1.55}$\\
\, + \selfdebias & \!\!\da{$\boldsymbol{3.16}$} $\mathbf{52.96}\tinymath{\pm 3.53}$ & -- & \dab{$0.15$} $82.74\tinymath{\pm 0.80}$ & -- & --\\
\, + \sentencedebias & \da{$0.27$} $55.85\tinymath{\pm 3.54}$ & \da{$0.12$} $0.63\tinymath{\pm 0.24}$ & \dab{$0.13$} $82.75\tinymath{\pm 0.80}$ & \uag{$0.70$} $80.68\tinymath{\pm 1.40}$ & \dab{$0.07$} $53.67\tinymath{\pm 1.64}$\\

\midrule
\distilbert & $55.40\tinymath{\pm 3.71}$ & $0.32\tinymath{\pm 0.26}$ & \!\!$\mathbf{82.06}\tinymath{\pm 0.80}$ & $74.47\tinymath{\pm 1.59}$ & $49.69\tinymath{\pm 1.65}$\\
\lightcmidrule{1-6}
\, + \gradiendreligioncj & \da{$1.20$} $54.20\tinymath{\pm 3.73}$ & \ua{$0.02$} $0.34\tinymath{\pm 0.27}$ & \dab{$0.07$} $82.00\tinymath{\pm 0.81}$ & \uag{$0.02$} $74.49\tinymath{\pm 1.60}$ & \uag{$0.06$} $49.75\tinymath{\pm 1.69}$\\
\, + \gradiendreligioncm & \da{$1.18$} $54.22\tinymath{\pm 3.71}$ & \ua{$0.05$} $0.37\tinymath{\pm 0.27}$ & \dab{$0.17$} $81.89\tinymath{\pm 0.80}$ & \uag{$0.03$} $74.50\tinymath{\pm 1.60}$ & \uag{$0.07$} $49.76\tinymath{\pm 1.69}$\\
\, + \gradiendreligionjm & \da{$1.97$} $53.42\tinymath{\pm 3.74}$ & \ua{$0.12$} $0.44\tinymath{\pm 0.29}$ & \dab{$0.40$} $81.66\tinymath{\pm 0.81}$ & \dab{$0.07$} $74.40\tinymath{\pm 1.61}$ & \uag{$0.23$} $49.91\tinymath{\pm 1.68}$\\
\lightcmidrule{1-6}
\, + \cda & \ua{$0.64$} $56.04\tinymath{\pm 3.51}$ & \!\!\da{$\boldsymbol{0.11}$} $\mathbf{0.22}\tinymath{\pm 0.12}$ & \dab{$0.28$} $81.78\tinymath{\pm 0.81}$ & \uag{$0.49$} $74.96\tinymath{\pm 1.46}$ & \!\!\uag{$\boldsymbol{0.80}$} $\mathbf{50.49}\tinymath{\pm 1.80}$\\
\, + \dropout & \ua{$0.67$} $56.06\tinymath{\pm 3.55}$ & \da{$0.08$} $0.25\tinymath{\pm 0.13}$ & \dab{$\mathit{1.82}$} $\mathit{80.24}\tinymath{\pm \mathit{0.85}}$ & \!\!\uag{$\boldsymbol{0.70}$} $\mathbf{75.17}\tinymath{\pm 1.50}$ & \uag{$0.58$} $50.27\tinymath{\pm 1.75}$\\
\, + \inlp & \ua{$0.36$} $55.75\tinymath{\pm 3.71}$ & \da{$0.06$} $0.26\tinymath{\pm 0.19}$ & \dab{$0.46$} $81.60\tinymath{\pm 0.81}$ & \uag{$0.13$} $74.59\tinymath{\pm 1.57}$ & \dab{$0.05$} $49.64\tinymath{\pm 1.65}$\\
\, + \selfdebias & \da{$3.10$} $52.29\tinymath{\pm 3.60}$ & -- & \dab{$0.48$} $81.59\tinymath{\pm 0.83}$ & -- & --\\
\, + \sentencedebias & \!\!\da{$\boldsymbol{3.11}$} $\mathbf{52.28}\tinymath{\pm 3.70}$ & \da{$0.03$} $0.29\tinymath{\pm 0.21}$ & \dab{$0.27$} $81.80\tinymath{\pm 0.81}$ & \uag{$0.07$} $74.54\tinymath{\pm 1.60}$ & \uag{$0.12$} $49.81\tinymath{\pm 1.64}$\\

\midrule
\roberta & $64.66\tinymath{\pm 3.33}$ & $0.39\tinymath{\pm 0.21}$ & \!\!$\mathbf{89.09}\tinymath{\pm 0.64}$ & $81.65\tinymath{\pm 1.44}$ & $53.31\tinymath{\pm 1.48}$\\
\lightcmidrule{1-6}
\, + \gradiendreligioncj & \da{$4.07$} $60.59\tinymath{\pm 3.35}$ & \da{$0.06$} $0.33\tinymath{\pm 0.14}$ & \dab{$0.95$} $88.14\tinymath{\pm 0.68}$ & \dab{$\mathit{9.08}$} $\mathit{72.57}\tinymath{\pm \mathit{1.50}}$ & \dab{$0.65$} $52.66\tinymath{\pm 1.66}$\\
\, + \gradiendreligioncm & \!\!\da{$\boldsymbol{\mathit{9.83}}$} $\mathit{\boldsymbol{\mathit{54.83}}}\tinymath{\pm \mathit{3.39}}$ & \da{$0.00$} $0.39\tinymath{\pm 0.16}$ & \dab{$0.44$} $88.65\tinymath{\pm 0.65}$ & \!\!\uag{$\boldsymbol{1.40}$} $\mathbf{83.05}\tinymath{\pm 1.51}$ & \uag{$\mathit{3.35}$} $\mathit{56.66}\tinymath{\pm \mathit{1.64}}$\\
\, + \gradiendreligionjm & \da{$4.83$} $59.83\tinymath{\pm 3.45}$ & \da{$0.14$} $0.25\tinymath{\pm 0.17}$ & \dab{$0.18$} $88.90\tinymath{\pm 0.67}$ & \uag{$1.06$} $82.71\tinymath{\pm 1.53}$ & \dab{$2.19$} $51.12\tinymath{\pm 1.65}$\\
\lightcmidrule{1-6}
\, + \cda & \da{$6.07$} $58.59\tinymath{\pm 3.53}$ & \!\!\da{$\boldsymbol{0.21}$} $\mathbf{0.18}\tinymath{\pm 0.15}$ & \dab{$\mathit{3.39}$} $\mathit{85.70}\tinymath{\pm \mathit{0.73}}$ & \uag{$0.83$} $82.48\tinymath{\pm 1.51}$ & \!\!\uag{$\boldsymbol{\mathit{4.71}}$} $\mathit{\boldsymbol{\mathit{58.02}}}\tinymath{\pm \mathit{1.68}}$\\
\, + \dropout & \da{$6.61$} $58.05\tinymath{\pm 3.54}$ & \da{$0.01$} $0.38\tinymath{\pm 0.13}$ & \dab{$\mathit{3.74}$} $\mathit{85.34}\tinymath{\pm \mathit{0.72}}$ & \dab{$\mathit{14.16}$} $\mathit{67.49}\tinymath{\pm \mathit{1.47}}$ & \dab{$2.25$} $51.05\tinymath{\pm 1.62}$\\
\, + \inlp & \da{$1.70$} $62.96\tinymath{\pm 3.38}$ & \da{$0.01$} $0.38\tinymath{\pm 0.21}$ & \dab{$0.83$} $88.26\tinymath{\pm 0.67}$ & \dab{$\mathit{3.95}$} $\mathit{77.70}\tinymath{\pm \mathit{1.51}}$ & \uag{$1.64$} $54.95\tinymath{\pm 1.51}$\\
\, + \selfdebias & \da{$2.71$} $61.95\tinymath{\pm 3.29}$ & -- & \dab{$0.30$} $88.79\tinymath{\pm 0.64}$ & -- & --\\
\, + \sentencedebias & \da{$3.17$} $61.49\tinymath{\pm 3.48}$ & \ua{$0.07$} $0.46\tinymath{\pm 0.23}$ & \dab{$0.04$} $89.05\tinymath{\pm 0.64}$ & \dab{$1.04$} $80.61\tinymath{\pm 0.86}$ & \dab{$0.60$} $52.71\tinymath{\pm 1.21}$\\

\midrule
\gpttwo & $63.22\tinymath{\pm 3.50}$ & $0.36\tinymath{\pm 0.27}$ & $91.02\tinymath{\pm 0.62}$ & $71.73\tinymath{\pm 1.08}$ & $45.49\tinymath{\pm 1.28}$\\
\lightcmidrule{1-6}
\, + \gradiendreligioncj & \ua{$0.21$} $63.43\tinymath{\pm 3.39}$ & \ua{$0.00$} $0.36\tinymath{\pm 0.28}$ & \dab{$0.16$} $90.87\tinymath{\pm 0.63}$ & \dab{$0.00$} $71.73\tinymath{\pm 0.98}$ & \uag{$1.18$} $46.67\tinymath{\pm 1.11}$\\
\, + \gradiendreligioncm & \!\!\da{$\boldsymbol{\mathit{9.31}}$} $\mathit{\boldsymbol{\mathit{53.91}}}\tinymath{\pm \mathit{3.51}}$ & \ua{$0.14$} $0.49\tinymath{\pm 0.26}$ & \dab{$1.06$} $89.96\tinymath{\pm 0.65}$ & \dab{$1.56$} $70.16\tinymath{\pm 1.04}$ & \uag{$1.54$} $47.02\tinymath{\pm 1.33}$\\
\, + \gradiendreligionjm & \da{$2.16$} $61.06\tinymath{\pm 3.51}$ & \ua{$0.11$} $0.46\tinymath{\pm 0.21}$ & \dab{$1.19$} $89.84\tinymath{\pm 0.65}$ & \uag{$0.05$} $71.78\tinymath{\pm 1.11}$ & \uag{$1.15$} $46.64\tinymath{\pm 1.26}$\\
\lightcmidrule{1-6}
\, + \cda & \ua{$3.87$} $67.10\tinymath{\pm 3.46}$ & \ua{$0.04$} $0.40\tinymath{\pm 0.32}$ & \dab{$\mathit{1.58}$} $\mathit{89.44}\tinymath{\pm \mathit{0.65}}$ & \!\!\uag{$\boldsymbol{1.59}$} $\mathbf{73.32}\tinymath{\pm 1.23}$ & \!\!\uag{$\boldsymbol{2.61}$} $\mathbf{48.10}\tinymath{\pm 1.42}$\\
\, + \dropout & \ua{$1.73$} $64.96\tinymath{\pm 3.54}$ & \!\!\da{$\boldsymbol{0.08}$} $\mathbf{0.28}\tinymath{\pm 0.26}$ & \dab{$0.69$} $90.33\tinymath{\pm 0.63}$ & \dab{$0.02$} $71.70\tinymath{\pm 1.15}$ & \uag{$0.46$} $45.94\tinymath{\pm 1.45}$\\
\, + \inlp & \ua{$0.68$} $63.91\tinymath{\pm 3.51}$ & \da{$0.00$} $0.35\tinymath{\pm 0.27}$ & \!\!\uag{$\boldsymbol{0.17}$} $\mathbf{91.19}\tinymath{\pm 0.61}$ & \dab{$0.21$} $71.52\tinymath{\pm 1.06}$ & \uag{$0.29$} $45.77\tinymath{\pm 1.21}$\\
\, + \selfdebias & \da{$4.01$} $59.21\tinymath{\pm 3.55}$ & -- & \dab{$\mathit{2.14}$} $\mathit{88.89}\tinymath{\pm \mathit{0.67}}$ & -- & --\\
\, + \sentencedebias & \da{$3.60$} $59.62\tinymath{\pm 3.54}$ & \ua{$0.07$} $0.43\tinymath{\pm 0.28}$ & \dab{$0.49$} $90.53\tinymath{\pm 0.64}$ & \uag{$0.16$} $71.88\tinymath{\pm 1.06}$ & \uag{$0.80$} $46.29\tinymath{\pm 1.28}$\\

\midrule
\llama & $66.44\tinymath{\pm 3.38}$ & $0.28\tinymath{\pm 0.09}$ & $92.42\tinymath{\pm 0.53}$ & $45.86\tinymath{\pm 1.98}$ & $54.46\tinymath{\pm 2.28}$\\
\lightcmidrule{1-6}
\, + \gradiendreligioncj & \da{$3.78$} $62.67\tinymath{\pm 3.41}$ & \da{$0.03$} $0.26\tinymath{\pm 0.11}$ & \dab{$\mathit{1.21}$} $\mathit{91.21}\tinymath{\pm \mathit{0.58}}$ & \dab{$\mathit{7.54}$} $\mathit{38.32}\tinymath{\pm \mathit{2.06}}$ & \dab{$1.90$} $52.56\tinymath{\pm 2.17}$\\
\, + \gradiendreligioncm & \ua{$1.77$} $68.21\tinymath{\pm 3.23}$ & \!\!\da{$\boldsymbol{0.07}$} $\mathbf{0.21}\tinymath{\pm 0.15}$ & \dab{$0.16$} $92.27\tinymath{\pm 0.53}$ & \dab{$1.40$} $44.46\tinymath{\pm 2.03}$ & \dab{$2.35$} $52.11\tinymath{\pm 2.12}$\\
\, + \gradiendreligionjm & \!\!\da{$\boldsymbol{\mathit{8.71}}$} $\mathit{\boldsymbol{\mathit{57.74}}}\tinymath{\pm \mathit{3.51}}$ & \ua{$0.08$} $0.36\tinymath{\pm 0.11}$ & \dab{$\mathit{2.20}$} $\mathit{90.22}\tinymath{\pm \mathit{0.62}}$ & \uag{$0.69$} $46.54\tinymath{\pm 1.90}$ & \dab{$1.87$} $52.59\tinymath{\pm 2.17}$\\
\lightcmidrule{1-6}
\, + \inlp & \da{$1.74$} $64.71\tinymath{\pm 3.38}$ & \da{$0.02$} $0.26\tinymath{\pm 0.08}$ & \uag{$0.00$} $92.42\tinymath{\pm 0.53}$ & \!\!\uag{$\boldsymbol{2.28}$} $\mathbf{48.14}\tinymath{\pm 1.84}$ & \!\!\uag{$\boldsymbol{0.21}$} $\mathbf{54.66}\tinymath{\pm 2.30}$\\
\, + \selfdebias & \da{$1.41$} $65.03\tinymath{\pm 3.35}$ & -- & \dab{$\mathit{31.14}$} $\mathit{61.28}\tinymath{\pm \mathit{1.00}}$ & -- & --\\
\, + \sentencedebias & \da{$2.65$} $63.80\tinymath{\pm 3.45}$ & \da{$0.03$} $0.26\tinymath{\pm 0.08}$ & \!\!\uag{$\boldsymbol{0.02}$} $\mathbf{92.44}\tinymath{\pm 0.53}$ & \dab{$0.04$} $45.82\tinymath{\pm 1.96}$ & \dab{$0.17$} $54.29\tinymath{\pm 2.28}$\\

\midrule
\llamai & $65.83\tinymath{\pm 3.35}$ & $0.20\tinymath{\pm 0.09}$ & $92.21\tinymath{\pm 0.54}$ & $49.14\tinymath{\pm 1.92}$ & $58.07\tinymath{\pm 2.29}$\\
\lightcmidrule{1-6}
\, + \gradiendreligioncj & \ua{$0.39$} $66.22\tinymath{\pm 3.33}$ & \ua{$0.01$} $0.21\tinymath{\pm 0.09}$ & \!\!\uag{$\boldsymbol{0.12}$} $\mathbf{92.34}\tinymath{\pm 0.55}$ & \uag{$0.31$} $49.45\tinymath{\pm 2.00}$ & \!\!\uag{$\boldsymbol{2.03}$} $\mathbf{60.10}\tinymath{\pm 1.95}$\\
\, + \gradiendreligioncm & \!\!\da{$\boldsymbol{\mathit{12.92}}$} $\mathit{\boldsymbol{\mathit{47.09}}}\tinymath{\pm \mathit{3.22}}$ & \ua{$\mathit{0.69}$} $\mathit{0.89}\tinymath{\pm \mathit{0.13}}$ & \dab{$\mathit{16.74}$} $\mathit{75.47}\tinymath{\pm \mathit{0.88}}$ & \dab{$\mathit{4.46}$} $\mathit{44.68}\tinymath{\pm \mathit{1.30}}$ & \dab{$4.30$} $53.77\tinymath{\pm 2.17}$\\
\, + \gradiendreligionjm & \da{$1.91$} $63.92\tinymath{\pm 3.43}$ & \ua{$0.30$} $0.50\tinymath{\pm 0.24}$ & \dab{$\mathit{1.60}$} $\mathit{90.61}\tinymath{\pm \mathit{0.58}}$ & \uag{$0.09$} $49.23\tinymath{\pm 1.95}$ & \uag{$1.67$} $59.74\tinymath{\pm 2.18}$\\
\lightcmidrule{1-6}
\, + \inlp & \da{$1.40$} $64.43\tinymath{\pm 3.31}$ & \da{$0.01$} $0.19\tinymath{\pm 0.09}$ & \dab{$0.40$} $91.81\tinymath{\pm 0.57}$ & \!\!\uag{$\boldsymbol{0.50}$} $\mathbf{49.64}\tinymath{\pm 1.99}$ & \dab{$0.15$} $57.92\tinymath{\pm 2.40}$\\
\, + \selfdebias & \da{$4.16$} $61.68\tinymath{\pm 3.36}$ & -- & \dab{$\mathit{33.02}$} $\mathit{59.19}\tinymath{\pm \mathit{1.01}}$ & -- & --\\
\, + \sentencedebias & \da{$2.88$} $62.95\tinymath{\pm 3.42}$ & \!\!\da{$\boldsymbol{0.04}$} $\mathbf{0.16}\tinymath{\pm 0.08}$ & \dab{$0.14$} $92.08\tinymath{\pm 0.55}$ & \dab{$0.35$} $48.79\tinymath{\pm 1.97}$ & \uag{$0.46$} $58.53\tinymath{\pm 2.41}$\\
\bottomrule
\end{tabular}

\end{table}


\begin{table*}[p]
\centering
\caption{\textbf{Gender:} GLUE bootstrapped validation set scores with sub-results for encoder-only models. Statistically significant improvements are indicated in \emph{italics}, while the best score for each base model is highlighted in \textbf{bold}.} \label{tab:eval:glue}
\tiny
\begin{tabular}{@{\hskip 1pt}l@{\hskip 5pt}r@{\hskip 5pt}r@{\hskip 5pt}r@{\hskip 5pt}r@{\hskip 5pt}r@{\hskip 5pt}r@{\hskip 5pt}r@{\hskip 5pt}r@{\hskip 5pt}r@{\hskip 5pt}r@{\hskip 5pt}r@{\hskip 1pt}}
\toprule\textbf{Model} & \textbf{CoLA} & \!\! {\fontsize{4pt}{5pt} \selectfont \textbf{MNLI-M}}  & \!\!{\fontsize{4pt}{5pt} \selectfont \textbf{MNLI-MM}} & \textbf{MRPC} & \textbf{QNLI} & \textbf{QQP} & \textbf{RTE} & \textbf{SST-2} & \textbf{STS-B} & \textbf{WNLI} & \textbf{Average} $\uparrow$\\
\midrule
\bertbase & \!\!$\mathbf{55.60}$ & $83.40$ & $83.97$ & $86.32$ & $90.19$ & $90.21$ & $60.95$ & $91.34$ & \!\!$\mathbf{88.71}$ & $55.83$ & $78.09\tinymath{\pm 1.59}$\\
\lightcmidrule{1-12}
\, + \gradiendbpi & $53.04$ & $83.63$ & \!\!$\mathbf{84.29}$ & $86.91$ & \!\!$\mathbf{90.66}$ & $90.40$ & $63.85$ & $91.57$ & $88.16$ & $56.80$ & \uag{$0.28$} $78.37\tinymath{\pm 1.55}$\\
\, + \gradiendfpi & $53.54$ & $83.58$ & $84.17$ & $86.66$ & $90.47$ & $90.38$ & $64.20$ & $91.33$ & $88.13$ & \!\!$\mathbf{57.22}$ & \uag{$0.33$} $78.42\tinymath{\pm 1.59}$\\
\, + \gradiendmpi & $52.24$ & $83.51$ & $84.21$ & $86.94$ & $90.65$ & $90.34$ & $63.96$ & $91.56$ & $88.19$ & $56.77$ & \uag{$0.18$} $78.28\tinymath{\pm 1.58}$\\
\lightcmidrule{1-12}
\, + \cda & $54.73$ & \!\!$\mathbf{83.90}$ & $84.14$ & \!\!$\mathbf{90.48}$ & $90.56$ & $90.24$ & \!\!$\mathbf{65.76}$ & $91.22$ & $86.77$ & $56.30$ & \!\!\uag{$\boldsymbol{0.80}$} $\mathbf{78.90}\tinymath{\pm 1.55}$\\
\, + \dropout & $46.09$ & $82.64$ & $83.36$ & $87.85$ & $90.51$ & $89.77$ & $61.47$ & $91.71$ & $\mathit{84.84}$ & $55.00$ & \dab{$1.40$} $76.69\tinymath{\pm 1.44}$\\
\, + \inlp & $53.97$ & $83.66$ & $84.12$ & $87.09$ & $90.57$ & $90.23$ & $61.48$ & \!\!$\mathbf{92.10}$ & $88.32$ & $55.35$ & \uag{$0.02$} $78.11\tinymath{\pm 1.55}$\\
\, + \rlace & $55.47$ & $83.40$ & $83.90$ & $86.02$ & $90.22$ & $90.20$ & $60.70$ & $91.16$ & $88.69$ & $55.83$ & \dab{$0.10$} $77.99\tinymath{\pm 1.59}$\\
\, + \leace & $55.28$ & $83.39$ & $83.94$ & $86.20$ & $90.16$ & $90.24$ & $60.96$ & $91.46$ & $88.66$ & $55.36$ & \dab{$0.10$} $78.00\tinymath{\pm 1.58}$\\
\, + \sentencedebias & $54.50$ & $83.57$ & $83.92$ & $87.01$ & $90.25$ & $90.22$ & $61.77$ & $91.57$ & $88.46$ & $51.60$ & \dab{$0.41$} $77.68\tinymath{\pm 1.02}$\\
\lightcmidrule{1-12}
\, + \gradiendbpi\ + \inlp & $53.91$ & $83.52$ & $83.84$ & $87.25$ & $90.58$ & $90.39$ & $64.92$ & $91.57$ & $87.88$ & $56.35$ & \uag{$0.41$} $78.50\tinymath{\pm 1.42}$\\
\, + \gradiendbpi\ + \sentencedebias \!\!\! & $53.22$ & $83.69$ & $84.22$ & $86.96$ & $90.64$ & \!\!$\mathbf{90.40}$ & $64.21$ & $91.53$ & $88.16$ & $56.80$ & \uag{$0.34$} $78.43\tinymath{\pm 1.55}$\\
\, + \cda\, + \inlp & $54.34$ & $83.69$ & $84.11$ & $89.73$ & $90.64$ & $90.34$ & $64.68$ & $91.75$ & $86.55$ & $51.74$ & \uag{$0.09$} $78.19\tinymath{\pm 1.42}$\\
\, + \dropout \, + \sentencedebias & $46.09$ & $82.83$ & $83.58$ & $87.95$ & $90.54$ & $89.80$ & $61.95$ & $91.71$ & $\mathit{84.80}$ & $55.00$ & \dab{$1.31$} $76.78\tinymath{\pm 1.44}$\\
\, + \cda\, + \sentencedebias & $54.47$ & $83.80$ & $84.05$ & $90.45$ & $90.55$ & $90.21$ & $65.76$ & $91.56$ & $86.74$ & $56.30$ & \uag{$0.79$} $78.88\tinymath{\pm 1.55}$\\
\, + \dropout \, + \inlp & $46.04$ & $82.97$ & $83.62$ & $87.67$ & $90.36$ & $\mathit{89.55}$ & $62.09$ & $91.83$ & $\mathit{83.92}$ & $54.07$ & \dab{$1.56$} $76.53\tinymath{\pm 1.40}$\\

\midrule
\bertlarge & $62.19$ & $86.19$ & $86.38$ & $88.62$ & \!\!$\mathbf{92.22}$ & $90.50$ & $66.59$ & \!\!$\mathbf{93.31}$ & $88.52$ & $51.59$ & $79.98\tinymath{\pm 1.31}$\\
\lightcmidrule{1-12}
\, + \gradiendbpi & $60.14$ & $85.58$ & $86.08$ & $89.93$ & $92.16$ & \!\!$\mathbf{90.58}$ & $66.10$ & $92.84$ & $89.20$ & $55.36$ & \uag{$0.26$} $80.24\tinymath{\pm 1.14}$\\
\, + \gradiendfpi & $61.53$ & $85.85$ & $86.07$ & $87.76$ & $91.98$ & $90.23$ & $66.51$ & $93.10$ & $89.23$ & $56.27$ & \uag{$0.31$} $80.29\tinymath{\pm 1.55}$\\
\, + \gradiendmpi & $62.25$ & $85.68$ & $86.20$ & $88.08$ & $92.06$ & $90.53$ & $65.51$ & $92.76$ & \!\!$\mathbf{89.50}$ & $56.27$ & \uag{$0.34$} $80.32\tinymath{\pm 1.55}$\\
\lightcmidrule{1-12}
\, + \cda & $61.38$ & $85.56$ & $85.96$ & $89.98$ & $92.04$ & $90.56$ & $59.44$ & $93.00$ & $88.56$ & $46.90$ & \dab{$1.36$} $78.63\tinymath{\pm 1.41}$\\
\, + \dropout & $54.54$ & $85.95$ & $86.11$ & \!\!$\mathbf{90.26}$ & $91.97$ & $90.09$ & $65.85$ & $93.08$ & $88.24$ & $54.86$ & \dab{$0.55$} $79.43\tinymath{\pm 1.46}$\\
\, + \inlp & $60.00$ & $85.78$ & $86.27$ & $89.56$ & $92.11$ & $90.29$ & $67.58$ & $92.72$ & $89.46$ & $54.74$ & \uag{$0.30$} $80.28\tinymath{\pm 1.39}$\\
\, + \rlace & $58.84$ & \!\!$\mathbf{86.29}$ & $86.38$ & $89.08$ & $92.21$ & $90.39$ & $65.99$ & $92.98$ & $89.23$ & $52.98$ & \dab{$0.20$} $79.78\tinymath{\pm 1.38}$\\
\, + \leace & $62.70$ & $85.85$ & $86.17$ & $88.94$ & $91.89$ & $90.34$ & \!\!$\mathbf{68.07}$ & $92.83$ & $89.03$ & $52.50$ & \uag{$0.28$} $80.26\tinymath{\pm 1.24}$\\
\, + \sentencedebias & $62.65$ & $86.06$ & \!\!$\mathbf{86.47}$ & $89.85$ & $92.08$ & $90.47$ & $67.43$ & $93.26$ & $89.27$ & $55.31$ & \!\!\uag{$\boldsymbol{0.75}$} $\mathbf{80.73}\tinymath{\pm 1.49}$\\
\lightcmidrule{1-12}
\, + \gradiendbpi\ + \inlp & $61.18$ & $85.66$ & $86.21$ & $89.62$ & $91.89$ & $90.47$ & $65.85$ & $92.96$ & $89.45$ & $54.34$ & \uag{$0.21$} $80.19\tinymath{\pm 1.25}$\\
\, + \gradiendbpi\ + \sentencedebias \!\!\! & $59.87$ & $85.60$ & $86.13$ & $89.78$ & $92.02$ & $90.50$ & $66.35$ & $92.95$ & $89.22$ & $53.47$ & \uag{$0.02$} $80.00\tinymath{\pm 1.05}$\\
\, + \cda\, + \inlp & $61.26$ & $85.48$ & $85.98$ & $89.87$ & $91.90$ & $90.50$ & $59.76$ & $92.84$ & $88.49$ & $44.70$ & \dab{$1.64$} $78.34\tinymath{\pm 1.10}$\\
\, + \dropout \, + \sentencedebias & $\mathit{3.14}$ & $85.82$ & $85.75$ & $88.98$ & $91.96$ & $90.33$ & $64.50$ & $93.08$ & $\mathit{85.65}$ & \!\!$\mathbf{57.65}$ & \dab{$\mathit{6.53}$} $\mathit{73.45}\tinymath{\pm \mathit{1.39}}$\\
\, + \cda\, + \sentencedebias & \!\!$\mathbf{62.90}$ & $85.50$ & $86.05$ & $90.03$ & $91.80$ & $90.52$ & $62.80$ & $92.78$ & $88.60$ & $46.45$ & \dab{$0.91$} $79.07\tinymath{\pm 1.38}$\\
\, + \dropout \, + \inlp & $\mathit{37.35}$ & $85.58$ & $86.19$ & $90.21$ & $92.19$ & $90.38$ & $63.96$ & $91.87$ & $88.39$ & $46.35$ & \dab{$\mathit{3.69}$} $\mathit{76.29}\tinymath{\pm \mathit{1.16}}$\\

\midrule
\distilbert & $43.90$ & $80.57$ & $81.24$ & $85.79$ & $87.00$ & $88.99$ & $55.13$ & $90.55$ & $81.68$ & \!\!$\mathbf{56.27}$ & $74.47\tinymath{\pm 1.59}$\\
\lightcmidrule{1-12}
\, + \gradiendbpi & $43.80$ & $80.60$ & $81.23$ & $85.43$ & $87.07$ & $89.01$ & $55.02$ & $90.70$ & $81.82$ & \!\!$\mathbf{56.27}$ & \dab{$0.02$} $74.45\tinymath{\pm 1.59}$\\
\, + \gradiendfpi & $43.36$ & $80.58$ & $81.22$ & $85.76$ & $87.26$ & $88.99$ & $54.56$ & $90.47$ & $81.58$ & \!\!$\mathbf{56.27}$ & \dab{$0.12$} $74.35\tinymath{\pm 1.61}$\\
\, + \gradiendmpi & $43.91$ & \!\!$\mathbf{80.80}$ & $81.26$ & $85.80$ & $87.00$ & $89.02$ & $55.73$ & $90.61$ & $82.00$ & $54.96$ & \dab{$0.01$} $74.45\tinymath{\pm 1.54}$\\
\lightcmidrule{1-12}
\, + \cda & $43.73$ & $80.67$ & $81.42$ & $86.84$ & $87.30$ & $88.95$ & $58.05$ & $90.35$ & $82.89$ & $52.64$ & \uag{$0.18$} $74.64\tinymath{\pm 1.46}$\\
\, + \dropout & $43.16$ & $80.35$ & $81.14$ & $87.91$ & $87.41$ & $88.85$ & $60.61$ & $90.37$ & $82.99$ & $54.50$ & \uag{$0.70$} $75.17\tinymath{\pm 1.50}$\\
\, + \inlp & $43.63$ & $80.63$ & $81.10$ & $85.04$ & $87.16$ & \!\!$\mathbf{89.07}$ & $55.93$ & \!\!$\mathbf{90.82}$ & $81.59$ & \!\!$\mathbf{56.27}$ & \uag{$0.02$} $74.49\tinymath{\pm 1.59}$\\
\, + \rlace & $44.03$ & $80.57$ & $81.19$ & $85.68$ & $86.96$ & $89.03$ & $55.26$ & $90.78$ & $81.71$ & \!\!$\mathbf{56.27}$ & \uag{$0.04$} $74.51\tinymath{\pm 1.59}$\\
\, + \leace & $42.92$ & $80.66$ & $81.18$ & $85.64$ & $87.08$ & $89.02$ & $54.44$ & $90.52$ & $81.65$ & $55.82$ & \dab{$0.24$} $74.22\tinymath{\pm 1.54}$\\
\, + \sentencedebias & $44.14$ & $80.73$ & $81.17$ & $85.66$ & $87.02$ & $89.04$ & $55.51$ & $90.59$ & $81.72$ & \!\!$\mathbf{56.27}$ & \uag{$0.08$} $74.54\tinymath{\pm 1.59}$\\
\lightcmidrule{1-12}
\, + \gradiendbpi\ + \inlp & $43.31$ & $80.71$ & $81.13$ & $85.06$ & $87.19$ & $88.99$ & $55.95$ & $90.55$ & $81.69$ & \!\!$\mathbf{56.27}$ & \dab{$0.03$} $74.44\tinymath{\pm 1.59}$\\
\, + \gradiendbpi\ + \sentencedebias \!\!\! & $43.55$ & $80.57$ & $81.22$ & $84.95$ & $86.90$ & $89.03$ & $54.81$ & $90.19$ & $81.74$ & \!\!$\mathbf{56.27}$ & \dab{$0.21$} $74.26\tinymath{\pm 1.60}$\\
\, + \cda\, + \inlp & \!\!$\mathbf{44.22}$ & $80.61$ & $81.43$ & $87.44$ & $87.15$ & $88.91$ & $58.90$ & $90.55$ & $82.94$ & $51.29$ & \uag{$0.25$} $74.71\tinymath{\pm 1.33}$\\
\, + \dropout \, + \sentencedebias & $43.49$ & $80.37$ & $81.08$ & \!\!$\mathbf{88.06}$ & $87.42$ & $88.80$ & $60.12$ & $90.37$ & \!\!$\mathbf{83.04}$ & $55.41$ & \uag{$0.80$} $75.27\tinymath{\pm 1.51}$\\
\, + \cda\, + \sentencedebias & $43.80$ & $80.60$ & \!\!$\mathbf{81.43}$ & $86.84$ & $87.40$ & $88.98$ & $57.82$ & $90.34$ & $82.91$ & $54.05$ & \uag{$0.33$} $74.79\tinymath{\pm 1.43}$\\
\, + \dropout \, + \inlp & $42.34$ & $80.32$ & $80.97$ & $87.23$ & \!\!$\mathbf{87.83}$ & $88.81$ & \!\!$\mathbf{63.31}$ & $90.06$ & $82.68$ & $55.91$ & \!\!\uag{$\boldsymbol{0.96}$} $\mathbf{75.42}\tinymath{\pm 1.48}$\\

\midrule
\roberta & $62.67$ & $90.08$ & $89.96$ & $91.00$ & $94.12$ & $90.95$ & $68.17$ & $94.86$ & $91.04$ & $52.03$ & $81.65\tinymath{\pm 1.44}$\\
\lightcmidrule{1-12}
\, + \gradiendbpi & $60.27$ & $89.86$ & $89.80$ & $89.46$ & \!\!$\mathbf{94.46}$ & $91.04$ & $\mathit{75.86}$ & $95.57$ & $91.82$ & $53.89$ & \uag{$0.82$} $82.47\tinymath{\pm 1.53}$\\
\, + \gradiendfpi & $61.45$ & $89.95$ & $89.88$ & $89.99$ & $94.18$ & $91.00$ & $72.85$ & $\mathit{80.30}$ & $91.00$ & $54.83$ & \dab{$1.04$} $80.61\tinymath{\pm 1.55}$\\
\, + \gradiendmpi & $60.72$ & $89.61$ & $89.64$ & $90.77$ & $93.77$ & $\mathit{81.81}$ & $67.13$ & $95.77$ & $91.37$ & $51.52$ & \dab{$1.37$} $80.28\tinymath{\pm 1.50}$\\
\lightcmidrule{1-12}
\, + \cda & $62.95$ & \!\!$\mathbf{90.18}$ & $89.82$ & \!\!$\mathbf{91.43}$ & $94.23$ & $91.00$ & $\mathit{76.80}$ & \!\!$\mathbf{95.94}$ & $91.82$ & $51.15$ & \uag{$1.16$} $82.81\tinymath{\pm 1.41}$\\
\, + \dropout & $\mathit{24.12}$ & $\mathit{53.46}$ & $\mathit{53.39}$ & $89.79$ & $94.45$ & $90.53$ & $61.06$ & $\mathit{50.93}$ & $\mathit{88.73}$ & $54.36$ & \dab{$\mathit{14.16}$} $\mathit{67.49}\tinymath{\pm \mathit{1.47}}$\\
\, + \inlp & $62.65$ & $90.00$ & \!\!$\mathbf{90.02}$ & $87.88$ & $94.41$ & $91.07$ & \!\!$\mathit{\boldsymbol{\mathit{80.04}}}$ & $95.62$ & $91.47$ & \!\!$\mathbf{56.27}$ & \uag{$1.62$} $83.27\tinymath{\pm 1.51}$\\
\, + \rlace & $61.23$ & $\mathit{71.78}$ & $\mathit{71.61}$ & $90.01$ & $94.34$ & $91.19$ & $\mathit{75.81}$ & $95.72$ & $91.44$ & $53.87$ & \dab{$1.06$} $80.59\tinymath{\pm 1.53}$\\
\, + \leace & $62.87$ & $89.89$ & $89.55$ & $89.44$ & $94.31$ & \!\!$\mathit{\boldsymbol{\mathit{91.58}}}$ & $72.47$ & $95.06$ & $91.54$ & $47.74$ & \dab{$0.01$} $81.64\tinymath{\pm 1.23}$\\
\, + \sentencedebias & $\mathit{22.58}$ & $89.85$ & $89.64$ & $91.27$ & $94.29$ & $91.17$ & $68.65$ & $95.80$ & $91.40$ & $49.71$ & \dab{$\mathit{4.47}$} $\mathit{77.18}\tinymath{\pm \mathit{1.23}}$\\
\lightcmidrule{1-12}
\, + \gradiendbpi\ + \inlp & \!\!$\mathbf{64.94}$ & $\mathit{71.83}$ & $\mathit{71.61}$ & $89.67$ & $\mathit{79.71}$ & $91.43$ & $\mathit{76.61}$ & $95.42$ & $90.91$ & \!\!$\mathbf{56.27}$ & \dab{$2.02$} $79.63\tinymath{\pm 1.53}$\\
\, + \gradiendbpi\ + \sentencedebias \!\!\! & $61.26$ & $\mathit{41.48}$ & $\mathit{41.40}$ & $89.69$ & $\mathit{79.66}$ & $91.34$ & $\mathit{76.23}$ & $95.54$ & \!\!$\mathbf{91.98}$ & $50.19$ & \dab{$\mathit{6.39}$} $\mathit{75.26}\tinymath{\pm \mathit{1.44}}$\\
\, + \cda\, + \inlp & $61.27$ & $89.77$ & $89.88$ & $90.43$ & $94.46$ & $91.08$ & $\mathit{79.72}$ & $95.53$ & $91.83$ & \!\!$\mathbf{56.27}$ & \!\!\uag{$\boldsymbol{1.73}$} $\mathbf{83.38}\tinymath{\pm 1.53}$\\
\, + \dropout \, + \sentencedebias & $\mathit{36.23}$ & $89.69$ & $89.77$ & $\mathit{85.84}$ & $94.11$ & $90.92$ & $\mathit{60.08}$ & $95.32$ & $\mathit{88.33}$ & $51.46$ & \dab{$\mathit{4.76}$} $\mathit{76.89}\tinymath{\pm \mathit{1.32}}$\\
\, + \cda\, + \sentencedebias & $62.66$ & $89.97$ & $89.82$ & $89.42$ & $94.24$ & $91.44$ & $\mathit{79.24}$ & $95.77$ & $91.86$ & $49.23$ & \uag{$0.99$} $82.64\tinymath{\pm 1.32}$\\
\, + \dropout \, + \inlp & $\mathit{30.25}$ & $89.84$ & $89.58$ & $87.94$ & $94.37$ & $\mathit{81.78}$ & $\mathit{55.99}$ & $\mathit{80.69}$ & $\mathit{87.65}$ & $55.79$ & \dab{$\mathit{7.85}$} $\mathit{73.80}\tinymath{\pm \mathit{1.45}}$\\

\bottomrule
\end{tabular}
\end{table*}

\begin{table*}[p]
\centering
\vspace{80pt}
\caption{\textbf{Gender:} GLUE bootstrapped validation set scores with sub-results for decoder-only models. Statistically significant improvements are indicated in \emph{italics}, while the best score for each base model is highlighted in \textbf{bold}. \gpttwo\ results were computed after fine-tuning and \llama-based results were computed with zero-shot evaluation.} \label{tab:eval:glue-gpt}
\tiny
\begin{tabular}{@{\hskip 1pt}l@{\hskip 5pt}r@{\hskip 5pt}r@{\hskip 5pt}r@{\hskip 5pt}r@{\hskip 5pt}r@{\hskip 5pt}r@{\hskip 5pt}r@{\hskip 5pt}r@{\hskip 5pt}r@{\hskip 5pt}r@{\hskip 5pt}r@{\hskip 1pt}}
\toprule\textbf{Model} & \textbf{CoLA} & \!\! {\fontsize{4pt}{5pt} \selectfont \textbf{MNLI-M}}  & \!\!{\fontsize{4pt}{5pt} \selectfont \textbf{MNLI-MM}} & \textbf{MRPC} & \textbf{QNLI} & \textbf{QQP} & \textbf{RTE} & \textbf{SST-2} & \textbf{STS-B} & \textbf{WNLI} & \textbf{Average} $\uparrow$\\

\midrule
\gpttwo & $20.51$ & $81.10$ & $82.02$ & $83.75$ & $87.54$ & $88.59$ & $59.87$ & $91.45$ & $80.30$ & $51.96$ & $71.73\tinymath{\pm 1.08}$\\
\lightcmidrule{1-12}
\, + \gradiendbpi & $14.21$ & $81.07$ & $81.91$ & $83.87$ & $87.34$ & $88.58$ & $62.68$ & $91.72$ & $80.41$ & $49.79$ & \dab{$0.61$} $71.12\tinymath{\pm 1.08}$\\
\, + \gradiendfpi & $11.43$ & $81.03$ & \!\!$\mathbf{82.06}$ & $83.02$ & $87.64$ & $88.56$ & \!\!$\mathbf{63.60}$ & $91.64$ & $80.81$ & $53.46$ & \dab{$0.42$} $71.30\tinymath{\pm 1.12}$\\
\, + \gradiendmpi & $16.22$ & $81.08$ & $81.89$ & $84.01$ & $87.29$ & $88.49$ & $62.66$ & $91.06$ & $80.75$ & $49.80$ & \dab{$0.42$} $71.31\tinymath{\pm 1.09}$\\
\lightcmidrule{1-12}
\, + \cda & \!\!$\mathit{\boldsymbol{\mathit{32.79}}}$ & $80.82$ & $81.90$ & $84.54$ & $87.70$ & $88.60$ & $61.80$ & $90.72$ & $80.98$ & $50.35$ & \!\!\uag{$\boldsymbol{1.48}$} $\mathbf{73.20}\tinymath{\pm 1.25}$\\
\, + \dropout & $20.09$ & $80.57$ & $81.74$ & $83.68$ & $87.03$ & $88.09$ & $62.68$ & $91.03$ & $81.07$ & $50.48$ & \dab{$0.02$} $71.70\tinymath{\pm 1.15}$\\
\, + \inlp & $20.10$ & $81.06$ & $81.99$ & $83.69$ & $87.67$ & $88.54$ & $61.20$ & $91.72$ & $80.28$ & $51.02$ & \uag{$0.02$} $71.75\tinymath{\pm 1.13}$\\
\, + \rlace & $23.21$ & $81.06$ & $82.05$ & $83.77$ & $87.51$ & \!\!$\mathbf{88.69}$ & $62.51$ & $91.38$ & $80.19$ & $52.02$ & \uag{$0.59$} $72.31\tinymath{\pm 1.08}$\\
\, + \leace & $20.05$ & \!\!$\mathbf{81.14}$ & $81.99$ & $83.39$ & $87.54$ & $88.57$ & $61.19$ & $91.34$ & $80.53$ & $50.09$ & \dab{$0.14$} $71.58\tinymath{\pm 1.07}$\\
\, + \sentencedebias & $18.97$ & $80.98$ & $81.97$ & $83.53$ & $87.52$ & $88.58$ & $61.70$ & $91.37$ & $81.26$ & $48.75$ & \dab{$0.26$} $71.46\tinymath{\pm 1.11}$\\
\lightcmidrule{1-12}
\, + \gradiendbpi\ + \inlp & $13.85$ & $81.09$ & $81.88$ & $84.34$ & $87.33$ & $88.56$ & $63.14$ & $91.83$ & $80.47$ & $49.79$ & \dab{$0.53$} $71.20\tinymath{\pm 1.07}$\\
\, + \gradiendbpi\ + \sentencedebias \!\!\! & $\mathit{9.97}$ & $80.97$ & $81.83$ & $83.87$ & $87.28$ & $88.60$ & $63.27$ & \!\!$\mathbf{91.84}$ & $81.63$ & \!\!$\mathbf{55.90}$ & \dab{$0.20$} $71.53\tinymath{\pm 1.12}$\\
\, + \cda\, + \inlp & $\mathit{32.21}$ & $80.86$ & $81.90$ & $84.51$ & $87.67$ & $88.61$ & $63.00$ & $90.65$ & $81.02$ & $48.94$ & \uag{$1.38$} $73.11\tinymath{\pm 1.24}$\\
\, + \dropout \, + \sentencedebias & $21.52$ & $80.62$ & $81.73$ & $83.34$ & $87.05$ & $\mathit{87.94}$ & $63.53$ & $90.88$ & \!\!$\mathbf{82.22}$ & $52.82$ & \uag{$0.55$} $72.27\tinymath{\pm 1.25}$\\
\, + \cda\, + \sentencedebias & $\mathit{30.85}$ & $80.88$ & $81.83$ & \!\!$\mathbf{84.84}$ & \!\!$\mathbf{87.71}$ & $88.58$ & $62.43$ & $91.18$ & $81.37$ & $48.96$ & \uag{$1.31$} $73.03\tinymath{\pm 1.27}$\\
\, + \dropout \, + \inlp & $19.24$ & $80.57$ & $81.73$ & $83.52$ & $87.05$ & $\mathit{88.00}$ & $62.80$ & $91.03$ & $81.06$ & $51.46$ & \dab{$0.02$} $71.70\tinymath{\pm 1.13}$\\

\midrule

\llama & \!\!\! $-8.08$ & $34.96$ & $35.97$ & $69.14$ & $49.93$ & $37.34$ & $54.19$ & $74.03$ & -- & $54.86$ & $45.86\tinymath{\pm 1.98}$\\
\lightcmidrule{1-12}
\, + \gradiendbpi & \!\!\! $-2.30$ & $35.12$ & $36.46$ & $72.83$ & \!\!$\mathit{\boldsymbol{\mathit{55.56}}}$ & $\mathit{39.18}$ & $52.40$ & $\mathit{62.60}$ & -- & \!\!$\mathbf{58.95}$ & \uag{$1.02$} $46.88\tinymath{\pm 1.91}$\\
\, + \gradiendfpi & \!\!\! $-0.21$ & \!\!$\mathbf{35.80}$ & \!\!$\mathbf{37.21}$ & $\mathit{80.37}$ & $49.59$ & $36.86$ & \!\!$\mathbf{57.10}$ & \!\!$\mathbf{78.38}$ & -- & $54.88$ & \!\!\uag{$\boldsymbol{3.33}$} $\mathbf{49.19}\tinymath{\pm 1.84}$\\
\, + \gradiendmpi & \!\!\! $-6.67$ & $34.67$ & $35.39$ & $\mathit{48.64}$ & $51.84$ & \!\!$\mathit{\boldsymbol{\mathit{39.94}}}$ & $54.56$ & $\mathit{58.06}$ & -- & $57.70$ & \dab{$3.47$} $42.39\tinymath{\pm 2.00}$\\
\lightcmidrule{1-12}
\, + \inlp & \!\!$\mathit{\boldsymbol{\mathit{0.00}}}$ & $\mathit{32.26}$ & $\mathit{32.65}$ & \!\!$\mathit{\boldsymbol{\mathit{81.10}}}$ & $49.45$ & $36.83$ & $48.04$ & $\mathit{61.66}$ & -- & $56.27$ & \dab{$0.13$} $45.73\tinymath{\pm 1.78}$\\
\, + \rlace & \!\!\! $-8.76$ & $34.40$ & $35.38$ & $72.49$ & $49.69$ & $37.06$ & $53.42$ & $74.60$ & -- & $54.86$ & \uag{$0.17$} $46.03\tinymath{\pm 1.95}$\\
\, + \leace & \!\!\! $-9.46$ & $34.27$ & $35.06$ & $72.07$ & $49.65$ & $37.07$ & $53.75$ & $75.39$ & -- & $56.27$ & \uag{$0.32$} $46.17\tinymath{\pm 1.97}$\\
\, + \sentencedebias & \!\!\! $-7.12$ & $35.10$ & $35.93$ & $76.60$ & $49.57$ & $36.95$ & $55.26$ & $74.40$ & -- & $56.27$ & \uag{$1.32$} $47.18\tinymath{\pm 1.92}$\\
\lightcmidrule{1-12}
\, + \gradiendbpi\ + \inlp & \!\!\! $-2.30$ & $35.12$ & $36.46$ & $72.83$ & \!\!$\mathit{\boldsymbol{\mathit{55.56}}}$ & $\mathit{39.18}$ & $52.40$ & $\mathit{62.60}$ & -- & \!\!$\mathbf{58.95}$ & \uag{$1.02$} $46.88\tinymath{\pm 1.91}$\\
\, + \gradiendbpi\ + \sentencedebias \!\!\! & \!\!\! $-2.17$ & $34.83$ & $36.11$ & $\mathit{79.51}$ & $51.20$ & $37.28$ & $53.05$ & $\mathit{62.25}$ & -- & $57.59$ & \uag{$0.92$} $46.77\tinymath{\pm 1.92}$\\

\midrule
\llamai & $16.85$ & $48.12$ & $47.97$ & $4.67$ & $57.34$ & $63.30$ & $64.69$ & $73.02$ & -- & $65.20$ & $49.14\tinymath{\pm 1.92}$\\
\lightcmidrule{1-12}
\, + \gradiendbpi & $\mathit{2.64}$ & $\mathit{35.76}$ & $\mathit{35.89}$ & $\mathit{79.39}$ & $\mathit{49.16}$ & $\mathit{39.26}$ & $53.16$ & $\mathit{61.62}$ & -- & $57.90$ & \dab{$1.77$} $47.37\tinymath{\pm 1.81}$\\
\, + \gradiendfpi & \!\!$\mathbf{17.72}$ & $\mathit{45.86}$ & $\mathit{45.91}$ & $1.39$ & $\mathit{51.67}$ & $\mathit{54.87}$ & $67.15$ & $\mathit{66.39}$ & -- & $63.86$ & \dab{$3.02$} $46.12\tinymath{\pm 1.83}$\\
\, + \gradiendmpi & $14.80$ & $46.62$ & $45.98$ & $\mathit{79.76}$ & \!\!$\mathit{\boldsymbol{\mathit{70.76}}}$ & \!\!$\mathit{\boldsymbol{\mathit{68.93}}}$ & \!\!$\mathbf{70.99}$ & \!\!$\mathbf{77.16}$ & -- & $55.05$ & \!\!\uag{$\boldsymbol{\mathit{11.33}}$} $\mathit{\boldsymbol{\mathit{60.47}}}\tinymath{\pm \mathit{1.86}}$\\
\lightcmidrule{1-12}
\, + \inlp & $16.09$ & $\mathit{44.35}$ & $\mathit{44.47}$ & $0.68$ & $\mathit{54.17}$ & $63.09$ & $66.00$ & $73.54$ & -- & \!\!$\mathbf{67.53}$ & \dab{$0.95$} $48.19\tinymath{\pm 1.85}$\\
\, + \rlace & $17.03$ & $48.19$ & \!\!$\mathbf{48.20}$ & $4.89$ & $57.36$ & $63.39$ & $64.22$ & $72.71$ & -- & $66.15$ & \uag{$0.10$} $49.24\tinymath{\pm 1.93}$\\
\, + \leace & $16.74$ & \!\!$\mathbf{48.51}$ & $48.06$ & $4.22$ & $57.17$ & $63.02$ & $64.97$ & $72.48$ & -- & $66.15$ & \dab{$0.01$} $49.13\tinymath{\pm 1.91}$\\
\, + \sentencedebias & $16.26$ & $48.46$ & $48.20$ & $4.22$ & $56.67$ & $63.11$ & $64.97$ & $72.94$ & -- & $66.15$ & \dab{$0.06$} $49.08\tinymath{\pm 1.91}$\\
\lightcmidrule{1-12}
\, + \gradiendbpi\ + \inlp & \!\!\! $\mathit{-0.94}$ & $\mathit{35.86}$ & $\mathit{36.18}$ & \!\!$\mathit{\boldsymbol{\mathit{81.01}}}$ & $\mathit{47.15}$ & $\mathit{36.82}$ & $54.52$ & $\mathit{58.99}$ & -- & $60.70$ & \dab{$2.36$} $46.78\tinymath{\pm 1.85}$\\
\, + \gradiendbpi\ + \sentencedebias \!\!\! & $5.83$ & $\mathit{35.93}$ & $\mathit{35.93}$ & $\mathit{80.17}$ & $\mathit{49.07}$ & $\mathit{38.51}$ & $\mathit{52.69}$ & $\mathit{62.88}$ & -- & $60.79$ & \dab{$0.91$} $48.23\tinymath{\pm 1.85}$\\

\bottomrule
\end{tabular}
\vspace{80pt}

\end{table*}

\begin{table}[p]
    \centering
    \tiny
    \caption{\textbf{Race:} GLUE bootstrapped validation set scores with sub-results for all models. Statistically significant improvements are indicated in \emph{italics}, while the best score for each base model is highlighted in \textbf{bold}. \llama-based results were computed with zero-shot evaluation while all other scores are derived after fine-tuning.}
    \label{tab:glue-race}
\begin{tabular}{@{\hskip 1pt}l@{\hskip 5pt}r@{\hskip 5pt}r@{\hskip 5pt}r@{\hskip 5pt}r@{\hskip 5pt}r@{\hskip 5pt}r@{\hskip 5pt}r@{\hskip 5pt}r@{\hskip 5pt}r@{\hskip 5pt}r@{\hskip 5pt}r@{\hskip 1pt}}
\toprule\textbf{Model} & \textbf{CoLA} & \!\! {\fontsize{4pt}{5pt} \selectfont \textbf{MNLI-M}}  & \!\!{\fontsize{4pt}{5pt} \selectfont \textbf{MNLI-MM}} & \textbf{MRPC} & \textbf{QNLI} & \textbf{QQP} & \textbf{RTE} & \textbf{SST-2} & \textbf{STS-B} & \textbf{WNLI} & \textbf{Average} $\uparrow$\\
\midrule
\bertbase & $55.60$ & $83.40$ & $83.97$ & $86.32$ & $90.19$ & $90.21$ & $60.95$ & $91.34$ & $88.71$ & $55.83$ & $78.09\tinymath{\pm 1.59}$\\
\lightcmidrule{1-12}
\, + \gradiendraceab & $51.67$ & \!\!$\mathbf{83.71}$ & $84.07$ & $87.83$ & $90.24$ & $90.30$ & \!\!$\mathbf{66.87}$ & $91.37$ & $88.14$ & $56.76$ & \uag{$0.47$} $78.56\tinymath{\pm 1.60}$\\
\, + \gradiendraceaw & $53.08$ & $83.62$ & $84.02$ & $89.03$ & $90.32$ & $90.27$ & $65.52$ & $91.21$ & $88.21$ & \!\!$\mathbf{57.22}$ & \!\!\uag{$\boldsymbol{0.65}$} $\mathbf{78.74}\tinymath{\pm 1.61}$\\
\, + \gradiendracebw & $51.51$ & $83.62$ & $84.04$ & $87.78$ & $90.26$ & $90.27$ & $65.87$ & $91.68$ & $88.29$ & $55.87$ & \uag{$0.28$} $78.37\tinymath{\pm 1.56}$\\
\lightcmidrule{1-12}
\, + \cda & $49.85$ & $83.55$ & $84.08$ & \!\!$\mathbf{89.67}$ & \!\!$\mathbf{90.65}$ & \!\!$\mathbf{90.37}$ & $63.43$ & $91.23$ & $87.38$ & $54.49$ & \dab{$0.22$} $77.88\tinymath{\pm 1.48}$\\
\, + \dropout & $46.09$ & $82.64$ & $83.36$ & $87.85$ & $90.51$ & $89.77$ & $61.47$ & $91.71$ & $\mathit{84.84}$ & $55.00$ & \dab{$1.40$} $76.69\tinymath{\pm 1.44}$\\
\, + \inlp & $54.35$ & $83.67$ & \!\!$\mathbf{84.12}$ & $86.66$ & $90.64$ & $90.30$ & $60.43$ & \!\!$\mathbf{92.21}$ & $88.27$ & $53.97$ & \dab{$0.24$} $77.86\tinymath{\pm 1.23}$\\
\, + \sentencedebias & \!\!$\mathbf{56.00}$ & $83.50$ & $83.95$ & $86.20$ & $90.21$ & $90.26$ & $60.69$ & $92.07$ & \!\!$\mathbf{88.71}$ & $55.36$ & \uag{$0.04$} $78.14\tinymath{\pm 1.58}$\\

\midrule
\bertlarge & $62.19$ & \!\!$\mathbf{86.19}$ & \!\!$\mathbf{86.38}$ & $88.62$ & $92.22$ & $90.50$ & $66.59$ & $93.31$ & $88.52$ & $51.59$ & $79.98\tinymath{\pm 1.31}$\\
\lightcmidrule{1-12}
\, + \gradiendraceab & $60.70$ & $85.50$ & $86.09$ & $88.57$ & $92.21$ & $90.60$ & $66.80$ & $93.10$ & $89.42$ & \!\!$\mathbf{56.27}$ & \uag{$0.40$} $80.38\tinymath{\pm 1.55}$\\
\, + \gradiendraceaw & \!\!$\mathbf{63.00}$ & $85.56$ & $86.06$ & \!\!$\mathbf{90.39}$ & $92.19$ & $90.62$ & $67.28$ & $92.92$ & $89.47$ & \!\!$\mathbf{56.27}$ & \!\!\uag{$\boldsymbol{0.90}$} $\mathbf{80.88}\tinymath{\pm 1.53}$\\
\, + \gradiendracebw & $61.61$ & $85.64$ & $86.21$ & $88.96$ & $92.04$ & $90.46$ & $66.06$ & \!\!$\mathbf{93.53}$ & $89.59$ & \!\!$\mathbf{56.27}$ & \uag{$0.51$} $80.49\tinymath{\pm 1.54}$\\
\lightcmidrule{1-12}
\, + \cda & $58.44$ & $85.64$ & $86.13$ & $88.38$ & $92.10$ & \!\!$\mathbf{90.72}$ & $57.72$ & $92.27$ & $87.41$ & $48.81$ & \dab{$2.01$} $77.97\tinymath{\pm 0.97}$\\
\, + \dropout & $54.54$ & $85.95$ & $86.11$ & $90.26$ & $91.97$ & $90.09$ & $65.85$ & $93.08$ & $88.24$ & $54.86$ & \dab{$0.55$} $79.43\tinymath{\pm 1.46}$\\
\, + \inlp & $59.69$ & $85.70$ & $86.09$ & $89.17$ & \!\!$\mathbf{92.31}$ & $90.55$ & \!\!$\mathbf{67.70}$ & $93.27$ & $89.55$ & $52.00$ & \uag{$0.03$} $80.02\tinymath{\pm 1.29}$\\
\, + \sentencedebias & $59.07$ & $85.66$ & $86.10$ & $89.19$ & $92.09$ & $90.47$ & $67.06$ & $93.14$ & \!\!$\mathbf{89.60}$ & $54.39$ & \uag{$0.12$} $80.10\tinymath{\pm 1.53}$\\

\midrule
\distilbert & $43.90$ & $80.57$ & $81.24$ & $85.79$ & $87.00$ & $88.99$ & $55.13$ & \!\!$\mathbf{90.55}$ & $81.68$ & $56.27$ & $74.47\tinymath{\pm 1.59}$\\
\lightcmidrule{1-12}
\, + \gradiendraceab & $44.99$ & $80.38$ & $81.34$ & $84.95$ & $87.01$ & \!\!$\mathbf{89.03}$ & $54.66$ & $90.14$ & $81.75$ & $56.27$ & \dab{$0.06$} $74.41\tinymath{\pm 1.60}$\\
\, + \gradiendraceaw & $44.60$ & $80.43$ & $81.45$ & $84.46$ & $86.99$ & $88.90$ & $55.03$ & $90.02$ & $81.88$ & $56.27$ & \dab{$0.12$} $74.34\tinymath{\pm 1.60}$\\
\, + \gradiendracebw & $45.36$ & $80.35$ & $81.30$ & $85.41$ & $86.98$ & $89.00$ & $53.58$ & $90.14$ & $81.75$ & $56.41$ & \dab{$0.08$} $74.38\tinymath{\pm 1.47}$\\
\lightcmidrule{1-12}
\, + \cda & $41.08$ & \!\!$\mathbf{80.59}$ & \!\!$\mathbf{81.47}$ & $87.58$ & $87.33$ & $88.96$ & $59.55$ & $90.28$ & \!\!$\mathbf{83.87}$ & $52.21$ & \uag{$0.19$} $74.65\tinymath{\pm 1.45}$\\
\, + \dropout & $43.16$ & $80.35$ & $81.14$ & \!\!$\mathbf{87.91}$ & \!\!$\mathbf{87.41}$ & $88.85$ & \!\!$\mathbf{60.61}$ & $90.37$ & $82.99$ & $54.50$ & \!\!\uag{$\boldsymbol{0.70}$} $\mathbf{75.17}\tinymath{\pm 1.50}$\\
\, + \inlp & $43.54$ & $80.35$ & $81.22$ & $86.14$ & $87.33$ & $88.99$ & $56.25$ & $89.94$ & $82.02$ & \!\!$\mathbf{56.75}$ & \uag{$0.17$} $74.64\tinymath{\pm 1.57}$\\
\, + \sentencedebias & \!\!$\mathbf{45.49}$ & $80.36$ & $81.34$ & $85.39$ & $87.25$ & $89.01$ & $55.04$ & $90.06$ & $81.77$ & $56.27$ & \uag{$0.10$} $74.57\tinymath{\pm 1.60}$\\

\midrule
\roberta & $62.67$ & \!\!$\mathbf{90.08}$ & \!\!$\mathbf{89.96}$ & $91.00$ & $94.12$ & $90.95$ & \!\!$\mathbf{68.17}$ & $94.86$ & $91.04$ & $52.03$ & \!\!$\mathbf{81.65}\tinymath{\pm 1.44}$\\
\lightcmidrule{1-12}
\, + \gradiendraceab & $57.99$ & $\mathit{71.27}$ & $\mathit{71.35}$ & $\mathit{85.31}$ & $\mathit{50.55}$ & $\mathit{77.15}$ & $60.87$ & $\mathit{80.55}$ & $91.17$ & \!\!$\mathbf{55.75}$ & \dab{$\mathit{11.58}$} $\mathit{70.07}\tinymath{\pm \mathit{1.48}}$\\
\, + \gradiendraceaw & $\mathit{20.38}$ & $89.63$ & $89.74$ & $89.65$ & $\mathit{79.57}$ & $91.39$ & $65.77$ & $\mathit{80.47}$ & $90.25$ & $51.07$ & \dab{$\mathit{8.51}$} $\mathit{73.14}\tinymath{\pm \mathit{0.86}}$\\
\, + \gradiendracebw & $\mathit{39.60}$ & $\mathit{71.75}$ & $\mathit{71.64}$ & $90.36$ & $94.13$ & $91.04$ & $61.99$ & $95.54$ & $91.63$ & $52.06$ & \dab{$\mathit{5.20}$} $\mathit{76.45}\tinymath{\pm \mathit{1.16}}$\\
\lightcmidrule{1-12}
\, + \cda & $62.15$ & $90.06$ & $89.88$ & \!\!$\mathbf{91.21}$ & $94.11$ & \!\!$\mathit{\boldsymbol{\mathit{91.52}}}$ & $67.69$ & $95.28$ & \!\!$\mathbf{91.75}$ & $50.58$ & \dab{$0.07$} $81.58\tinymath{\pm 1.29}$\\
\, + \dropout & $\mathit{24.12}$ & $\mathit{53.46}$ & $\mathit{53.39}$ & $89.79$ & \!\!$\mathbf{94.45}$ & $90.53$ & $61.06$ & $\mathit{50.93}$ & $\mathit{88.73}$ & $54.36$ & \dab{$\mathit{14.16}$} $\mathit{67.49}\tinymath{\pm \mathit{1.47}}$\\
\, + \inlp & \!\!$\mathbf{63.42}$ & $89.99$ & $89.82$ & $89.86$ & $94.10$ & $91.33$ & $65.04$ & $95.72$ & $91.54$ & $52.97$ & \dab{$0.11$} $81.54\tinymath{\pm 1.48}$\\
\, + \sentencedebias & $\mathit{34.80}$ & $89.95$ & $89.55$ & $89.10$ & $94.13$ & $91.45$ & $66.81$ & \!\!$\mathbf{96.15}$ & $91.66$ & $52.44$ & \dab{$\mathit{3.17}$} $\mathit{78.48}\tinymath{\pm \mathit{1.30}}$\\

\midrule
\gpttwo & $20.51$ & \!\!$\mathbf{81.10}$ & \!\!$\mathbf{82.02}$ & $83.75$ & $87.54$ & \!\!$\mathbf{88.59}$ & $59.87$ & $91.45$ & $80.30$ & \!\!$\mathbf{51.96}$ & $71.73\tinymath{\pm 1.08}$\\
\lightcmidrule{1-12}
\, + \gradiendraceab & $16.57$ & $80.98$ & $81.83$ & $84.24$ & $87.64$ & $88.47$ & $60.89$ & $91.62$ & $82.05$ & $47.39$ & \dab{$0.58$} $71.14\tinymath{\pm 1.01}$\\
\, + \gradiendraceaw & $13.03$ & $80.94$ & $81.82$ & $83.67$ & $87.55$ & $88.54$ & $62.24$ & \!\!$\mathbf{91.90}$ & $80.67$ & $46.88$ & \dab{$1.08$} $70.65\tinymath{\pm 0.98}$\\
\, + \gradiendracebw & $18.45$ & $80.97$ & $81.90$ & $83.68$ & $87.61$ & $88.51$ & $62.99$ & $91.87$ & $81.59$ & $47.39$ & \dab{$0.22$} $71.50\tinymath{\pm 1.07}$\\
\lightcmidrule{1-12}
\, + \cda & \!\!$\mathit{\boldsymbol{\mathit{30.53}}}$ & $80.64$ & $81.90$ & \!\!$\mathbf{85.00}$ & $87.60$ & $88.54$ & \!\!$\mathbf{64.52}$ & $90.80$ & \!\!$\mathbf{82.65}$ & $50.34$ & \!\!\uag{$\boldsymbol{1.74}$} $\mathbf{73.47}\tinymath{\pm 1.12}$\\
\, + \dropout & $20.09$ & $80.57$ & $81.74$ & $83.68$ & $87.03$ & $88.09$ & $62.68$ & $91.03$ & $81.07$ & $50.48$ & \dab{$0.02$} $71.70\tinymath{\pm 1.15}$\\
\, + \inlp & $18.83$ & $81.05$ & $82.00$ & $83.67$ & \!\!$\mathbf{87.72}$ & $88.49$ & $62.89$ & $91.67$ & $81.34$ & $46.89$ & \dab{$0.28$} $71.45\tinymath{\pm 1.08}$\\
\, + \sentencedebias & $17.97$ & $80.96$ & $81.97$ & $84.22$ & $87.59$ & $88.53$ & $62.72$ & $91.86$ & $81.75$ & $47.85$ & \dab{$0.18$} $71.55\tinymath{\pm 1.09}$\\

\midrule
\llama & \!\!\! $-8.08$ & $34.96$ & $35.97$ & $69.14$ & $49.93$ & $37.34$ & $54.19$ & $74.03$ & -- & $54.86$ & $45.86\tinymath{\pm 1.98}$\\
\lightcmidrule{1-12}
\, + \gradiendraceab & \!\!\! $-4.95$ & $35.67$ & $35.97$ & \!\!$\mathbf{75.32}$ & \!\!$\mathit{\boldsymbol{\mathit{58.03}}}$ & \!\!$\mathit{\boldsymbol{\mathit{59.46}}}$ & $55.88$ & $\mathit{58.35}$ & -- & $57.57$ & \!\!\uag{$\boldsymbol{3.58}$} $\mathbf{49.44}\tinymath{\pm 1.97}$\\
\, + \gradiendraceaw & \!\!$\mathbf{1.02}$ & $\mathit{36.92}$ & $37.11$ & $\mathit{49.54}$ & $\mathit{53.52}$ & $\mathit{54.11}$ & \!\!$\mathbf{60.52}$ & $\mathit{57.42}$ & -- & \!\!$\mathbf{63.29}$ & \uag{$1.20$} $47.06\tinymath{\pm 1.97}$\\
\, + \gradiendracebw & \!\!\! $-7.39$ & $34.50$ & $35.06$ & $\mathit{59.49}$ & $50.71$ & $\mathit{44.97}$ & $52.69$ & $\mathit{65.10}$ & -- & $54.86$ & \dab{$1.46$} $44.40\tinymath{\pm 2.01}$\\
\lightcmidrule{1-12}
\, + \inlp & \!\!\! $-7.99$ & \!\!$\mathit{\boldsymbol{\mathit{38.14}}}$ & \!\!$\mathit{\boldsymbol{\mathit{39.18}}}$ & $72.46$ & $49.94$ & $37.38$ & $58.20$ & \!\!$\mathbf{76.00}$ & -- & $57.65$ & \uag{$1.93$} $47.79\tinymath{\pm 1.87}$\\
\, + \sentencedebias & \!\!\! $-8.55$ & $35.88$ & $36.77$ & $71.42$ & $49.57$ & $37.08$ & $57.40$ & $72.90$ & -- & $54.86$ & \uag{$0.52$} $46.38\tinymath{\pm 1.93}$\\

\midrule
\llamai & \!\!$\mathbf{16.85}$ & $48.12$ & $47.97$ & $4.67$ & $57.34$ & $63.30$ & $64.69$ & $73.02$ & -- & $65.20$ & $49.14\tinymath{\pm 1.92}$\\
\lightcmidrule{1-12}
\, + \gradiendraceab & $\mathit{0.00}$ & $\mathit{32.75}$ & $\mathit{32.93}$ & $\mathit{0.00}$ & $\mathit{50.55}$ & $63.17$ & $\mathit{47.34}$ & $\mathit{49.11}$ & -- & $56.27$ & \dab{$\mathit{11.73}$} $\mathit{37.41}\tinymath{\pm \mathit{1.75}}$\\
\, + \gradiendraceaw & $\mathit{0.00}$ & $\mathit{33.29}$ & $\mathit{33.38}$ & $\mathit{0.00}$ & $\mathit{50.55}$ & $63.17$ & $\mathit{52.32}$ & $\mathit{51.00}$ & -- & $43.73$ & \dab{$\mathit{12.38}$} $\mathit{36.76}\tinymath{\pm \mathit{1.75}}$\\
\, + \gradiendracebw & $9.81$ & $\mathit{45.43}$ & $\mathit{45.32}$ & \!\!$\mathit{\boldsymbol{\mathit{22.41}}}$ & \!\!$\mathit{\boldsymbol{\mathit{65.11}}}$ & $\mathit{41.61}$ & \!\!$\mathbf{69.25}$ & $69.62$ & -- & \!\!$\mathbf{66.14}$ & \dab{$0.48$} $48.66\tinymath{\pm 2.02}$\\
\lightcmidrule{1-12}
\, + \inlp & $16.34$ & \!\!$\mathbf{49.30}$ & \!\!$\mathbf{49.46}$ & $7.63$ & $59.36$ & \!\!$\mathit{\boldsymbol{\mathit{65.00}}}$ & $66.77$ & \!\!$\mathbf{74.21}$ & -- & $60.58$ & \!\!\uag{$\boldsymbol{0.77}$} $\mathbf{49.91}\tinymath{\pm 1.98}$\\
\, + \sentencedebias & $15.80$ & $48.56$ & $48.60$ & $6.47$ & $57.25$ & $64.09$ & $66.53$ & $71.35$ & -- & $61.53$ & \dab{$0.19$} $48.95\tinymath{\pm 1.96}$\\
\bottomrule
    \end{tabular}
\end{table}

\begin{table}[p]
    \centering
    \tiny
\caption{\textbf{Religion:} GLUE bootstrapped validation set scores with sub-results for all models. Statistically significant improvements are indicated in \emph{italics}, while the best score for each base model is highlighted in \textbf{bold}. \llama-based results were computed with zero-shot evaluation while all other scores are derived after fine-tuning.}
    \label{tab:glue-religion}
\begin{tabular}{@{\hskip 1pt}l@{\hskip 5pt}r@{\hskip 5pt}r@{\hskip 5pt}r@{\hskip 5pt}r@{\hskip 5pt}r@{\hskip 5pt}r@{\hskip 5pt}r@{\hskip 5pt}r@{\hskip 5pt}r@{\hskip 5pt}r@{\hskip 5pt}r@{\hskip 1pt}}
\toprule\textbf{Model} & \textbf{CoLA} & \!\! {\fontsize{4pt}{5pt} \selectfont \textbf{MNLI-M}}  & \!\!{\fontsize{4pt}{5pt} \selectfont \textbf{MNLI-MM}} & \textbf{MRPC} & \textbf{QNLI} & \textbf{QQP} & \textbf{RTE} & \textbf{SST-2} & \textbf{STS-B} & \textbf{WNLI} & \textbf{Average} $\uparrow$\\
\midrule
\bertbase & $55.60$ & $83.40$ & $83.97$ & $86.32$ & $90.19$ & $90.21$ & $60.95$ & $91.34$ & \!\!$\mathbf{88.71}$ & $55.83$ & $78.09\tinymath{\pm 1.59}$\\
\lightcmidrule{1-12}
\, + \gradiendreligioncj & $51.16$ & $83.58$ & $84.07$ & $87.50$ & $90.33$ & $90.30$ & $65.02$ & \!\!$\mathbf{91.79}$ & $88.32$ & $56.77$ & \uag{$0.24$} $78.33\tinymath{\pm 1.58}$\\
\, + \gradiendreligioncm & $51.14$ & $83.57$ & $83.91$ & $88.04$ & $90.20$ & $90.29$ & \!\!$\mathbf{65.51}$ & $91.72$ & $88.44$ & $56.82$ & \!\!\uag{$\boldsymbol{0.34}$} $\mathbf{78.43}\tinymath{\pm 1.57}$\\
\, + \gradiendreligionjm & $51.51$ & $83.62$ & $83.99$ & $87.58$ & $90.22$ & $90.25$ & $65.27$ & $91.52$ & $87.98$ & \!\!$\mathbf{57.23}$ & \uag{$0.28$} $78.37\tinymath{\pm 1.60}$\\
\lightcmidrule{1-12}
\, + \cda & $52.47$ & $83.50$ & $83.81$ & \!\!$\mathbf{89.94}$ & $90.35$ & \!\!$\mathbf{90.36}$ & $65.03$ & $91.12$ & $87.41$ & $54.97$ & \uag{$0.27$} $78.36\tinymath{\pm 1.49}$\\
\, + \dropout & $46.09$ & $82.64$ & $83.36$ & $87.85$ & $90.51$ & $89.77$ & $61.47$ & $91.71$ & $\mathit{84.84}$ & $55.00$ & \dab{$1.40$} $76.69\tinymath{\pm 1.44}$\\
\, + \inlp & $55.22$ & \!\!$\mathbf{83.72}$ & \!\!$\mathbf{84.16}$ & $86.67$ & \!\!$\mathbf{90.63}$ & $90.27$ & $61.41$ & $91.72$ & $88.34$ & $51.15$ & \dab{$0.39$} $77.71\tinymath{\pm 1.23}$\\
\, + \sentencedebias & \!\!$\mathbf{55.66}$ & $83.36$ & $83.94$ & $86.20$ & $90.23$ & $90.23$ & $63.33$ & $91.50$ & $88.50$ & $51.60$ & \dab{$0.22$} $77.88\tinymath{\pm 1.03}$\\

\midrule
\bertlarge & $62.19$ & \!\!$\mathbf{86.19}$ & $86.38$ & $88.62$ & \!\!$\mathbf{92.22}$ & $90.50$ & $66.59$ & $93.31$ & $88.52$ & $51.59$ & $79.98\tinymath{\pm 1.31}$\\
\lightcmidrule{1-12}
\, + \gradiendreligioncj & $61.94$ & $85.72$ & $86.20$ & $89.01$ & $91.99$ & $90.65$ & \!\!$\mathbf{68.89}$ & \!\!$\mathbf{93.42}$ & \!\!$\mathbf{89.76}$ & $56.27$ & \!\!\uag{$\boldsymbol{0.90}$} $\mathbf{80.88}\tinymath{\pm 1.55}$\\
\, + \gradiendreligioncm & $60.73$ & $85.45$ & $86.10$ & $88.21$ & $92.12$ & $90.57$ & $66.90$ & $92.98$ & $89.67$ & $56.27$ & \uag{$0.38$} $80.36\tinymath{\pm 1.55}$\\
\, + \gradiendreligionjm & $62.57$ & $85.70$ & $86.24$ & $89.05$ & $92.11$ & \!\!$\mathbf{90.81}$ & $66.08$ & $93.07$ & $89.69$ & $56.27$ & \uag{$0.64$} $80.62\tinymath{\pm 1.55}$\\
\lightcmidrule{1-12}
\, + \cda & $59.36$ & $85.58$ & $86.06$ & \!\!$\mathbf{90.45}$ & $91.70$ & $90.59$ & $68.03$ & $93.00$ & $88.46$ & \!\!$\mathbf{56.33}$ & \uag{$0.43$} $80.42\tinymath{\pm 1.53}$\\
\, + \dropout & $54.54$ & $85.95$ & $86.11$ & $90.26$ & $91.97$ & $90.09$ & $65.85$ & $93.08$ & $88.24$ & $54.86$ & \dab{$0.55$} $79.43\tinymath{\pm 1.46}$\\
\, + \inlp & $61.26$ & $85.94$ & \!\!$\mathbf{86.47}$ & $89.40$ & $92.17$ & $90.54$ & $66.99$ & $93.25$ & $89.75$ & $49.18$ & \dab{$0.12$} $79.86\tinymath{\pm 1.08}$\\
\, + \sentencedebias & \!\!$\mathbf{63.89}$ & $86.14$ & $86.41$ & $87.91$ & $92.17$ & $90.43$ & $68.18$ & $93.28$ & $89.41$ & $54.53$ & \uag{$0.70$} $80.68\tinymath{\pm 1.40}$\\

\midrule
\distilbert & $43.90$ & $80.57$ & $81.24$ & $85.79$ & $87.00$ & $88.99$ & $55.13$ & $90.55$ & $81.68$ & \!\!$\mathbf{56.27}$ & $74.47\tinymath{\pm 1.59}$\\
\lightcmidrule{1-12}
\, + \gradiendreligioncj & $43.83$ & $80.69$ & $81.25$ & $85.49$ & $87.12$ & $89.02$ & $55.39$ & \!\!$\mathbf{90.58}$ & $81.74$ & \!\!$\mathbf{56.27}$ & \uag{$0.02$} $74.49\tinymath{\pm 1.60}$\\
\, + \gradiendreligioncm & $43.45$ & $80.59$ & $81.20$ & $85.39$ & $86.95$ & $88.90$ & $56.33$ & $90.53$ & $81.76$ & \!\!$\mathbf{56.27}$ & \uag{$0.03$} $74.50\tinymath{\pm 1.60}$\\
\, + \gradiendreligionjm & $42.67$ & $80.68$ & $81.18$ & $85.38$ & $87.16$ & $88.99$ & $56.12$ & $90.43$ & $81.63$ & \!\!$\mathbf{56.27}$ & \dab{$0.07$} $74.40\tinymath{\pm 1.61}$\\
\lightcmidrule{1-12}
\, + \cda & $44.10$ & \!\!$\mathbf{80.71}$ & \!\!$\mathbf{81.46}$ & \!\!$\mathbf{88.27}$ & $87.24$ & $88.84$ & $59.57$ & $89.93$ & \!\!$\mathbf{83.89}$ & $51.67$ & \uag{$0.49$} $74.96\tinymath{\pm 1.46}$\\
\, + \dropout & $43.16$ & $80.35$ & $81.14$ & $87.91$ & $87.41$ & $88.85$ & \!\!$\mathbf{60.61}$ & $90.37$ & $82.99$ & $54.50$ & \!\!\uag{$\boldsymbol{0.70}$} $\mathbf{75.17}\tinymath{\pm 1.50}$\\
\, + \inlp & $45.13$ & $80.30$ & $81.23$ & $85.83$ & \!\!$\mathbf{87.58}$ & $88.96$ & $55.77$ & $90.05$ & $82.39$ & $54.87$ & \uag{$0.13$} $74.59\tinymath{\pm 1.57}$\\
\, + \sentencedebias & \!\!$\mathbf{45.59}$ & $80.42$ & $81.22$ & $85.23$ & $87.06$ & \!\!$\mathbf{89.07}$ & $54.80$ & $90.13$ & $81.88$ & \!\!$\mathbf{56.27}$ & \uag{$0.07$} $74.54\tinymath{\pm 1.60}$\\

\midrule
\roberta & $62.67$ & $90.08$ & \!\!$\mathbf{89.96}$ & \!\!$\mathbf{91.00}$ & $94.12$ & $90.95$ & $68.17$ & $94.86$ & $91.04$ & $52.03$ & $81.65\tinymath{\pm 1.44}$\\
\lightcmidrule{1-12}
\, + \gradiendreligioncj & $\mathit{34.62}$ & $\mathit{35.45}$ & $\mathit{35.25}$ & $89.22$ & $93.54$ & \!\!$\mathit{\boldsymbol{\mathit{91.52}}}$ & $67.03$ & $95.64$ & $91.38$ & $54.83$ & \dab{$\mathit{9.08}$} $\mathit{72.57}\tinymath{\pm \mathit{1.50}}$\\
\, + \gradiendreligioncm & \!\!$\mathbf{62.92}$ & $89.77$ & $89.75$ & $90.37$ & $94.26$ & $90.89$ & $\mathit{76.49}$ & $95.64$ & $91.80$ & \!\!$\mathbf{55.31}$ & \!\!\uag{$\boldsymbol{1.40}$} $\mathbf{83.05}\tinymath{\pm 1.51}$\\
\, + \gradiendreligionjm & $61.47$ & $89.37$ & $89.23$ & $87.62$ & $94.26$ & $91.30$ & \!\!$\mathit{\boldsymbol{\mathit{77.99}}}$ & $95.49$ & $91.67$ & \!\!$\mathbf{55.31}$ & \uag{$1.06$} $82.71\tinymath{\pm 1.53}$\\
\lightcmidrule{1-12}
\, + \cda & $61.57$ & \!\!$\mathbf{90.17}$ & $89.94$ & $90.75$ & $94.15$ & $90.95$ & $72.88$ & $95.62$ & \!\!$\mathbf{91.94}$ & $54.40$ & \uag{$0.83$} $82.48\tinymath{\pm 1.51}$\\
\, + \dropout & $\mathit{24.12}$ & $\mathit{53.46}$ & $\mathit{53.39}$ & $89.79$ & \!\!$\mathbf{94.45}$ & $90.53$ & $61.06$ & $\mathit{50.93}$ & $\mathit{88.73}$ & $54.36$ & \dab{$\mathit{14.16}$} $\mathit{67.49}\tinymath{\pm \mathit{1.47}}$\\
\, + \inlp & $60.09$ & $\mathit{71.86}$ & $\mathit{71.76}$ & $88.93$ & $\mathit{79.30}$ & $\mathit{91.45}$ & $66.22$ & $95.91$ & $91.71$ & $53.89$ & \dab{$\mathit{3.95}$} $\mathit{77.70}\tinymath{\pm \mathit{1.51}}$\\
\, + \sentencedebias & $62.46$ & $90.00$ & $89.72$ & $90.60$ & $94.42$ & $\mathit{82.03}$ & $70.81$ & \!\!$\mathbf{96.01}$ & $91.78$ & $47.52$ & \dab{$1.04$} $80.61\tinymath{\pm 0.86}$\\

\midrule
\gpttwo & $20.51$ & \!\!$\mathbf{81.10}$ & \!\!$\mathbf{82.02}$ & $83.75$ & $87.54$ & $88.59$ & $59.87$ & $91.45$ & $80.30$ & \!\!$\mathbf{51.96}$ & $71.73\tinymath{\pm 1.08}$\\
\lightcmidrule{1-12}
\, + \gradiendreligioncj & $19.80$ & $80.93$ & $81.79$ & $83.38$ & $87.47$ & $88.56$ & $61.87$ & $91.67$ & $81.66$ & $49.75$ & \dab{$0.00$} $71.73\tinymath{\pm 0.98}$\\
\, + \gradiendreligioncm & $\mathit{7.19}$ & $80.83$ & $81.94$ & $83.37$ & $87.66$ & $88.58$ & $60.05$ & \!\!$\mathbf{91.88}$ & $81.10$ & $50.24$ & \dab{$1.56$} $70.16\tinymath{\pm 1.04}$\\
\, + \gradiendreligionjm & $19.97$ & $80.82$ & $81.67$ & $83.41$ & $87.48$ & $88.53$ & $61.39$ & $91.67$ & $81.70$ & $50.58$ & \uag{$0.05$} $71.78\tinymath{\pm 1.11}$\\
\lightcmidrule{1-12}
\, + \cda & \!\!$\mathit{\boldsymbol{\mathit{32.97}}}$ & $80.82$ & $81.63$ & \!\!$\mathbf{84.70}$ & \!\!$\mathbf{87.69}$ & $88.41$ & \!\!$\mathbf{63.39}$ & $91.39$ & \!\!$\mathbf{82.12}$ & $47.96$ & \!\!\uag{$\boldsymbol{1.59}$} $\mathbf{73.32}\tinymath{\pm 1.23}$\\
\, + \dropout & $20.09$ & $80.57$ & $81.74$ & $83.68$ & $87.03$ & $88.09$ & $62.68$ & $91.03$ & $81.07$ & $50.48$ & \dab{$0.02$} $71.70\tinymath{\pm 1.15}$\\
\, + \inlp & $18.33$ & $81.01$ & $81.91$ & $83.81$ & $87.67$ & \!\!$\mathbf{88.59}$ & $63.02$ & $91.75$ & $81.16$ & $47.84$ & \dab{$0.21$} $71.52\tinymath{\pm 1.06}$\\
\, + \sentencedebias & $21.31$ & $80.99$ & $81.86$ & $84.13$ & $87.63$ & $88.48$ & $62.88$ & $91.75$ & $81.93$ & $47.39$ & \uag{$0.16$} $71.88\tinymath{\pm 1.06}$\\

\midrule
\llama & \!\!\! $-8.08$ & $34.96$ & $35.97$ & $69.14$ & $49.93$ & $37.34$ & $54.19$ & $74.03$ & -- & $54.86$ & $45.86\tinymath{\pm 1.98}$\\
\lightcmidrule{1-12}
\, + \gradiendreligioncj & \!\!\! $-13.25$ & $33.16$ & $\mathit{33.63}$ & $\mathit{19.66}$ & \!\!$\mathit{\boldsymbol{\mathit{54.55}}}$ & \!\!$\mathit{\boldsymbol{\mathit{54.30}}}$ & $52.38$ & $\mathit{50.72}$ & -- & $54.82$ & \dab{$\mathit{7.54}$} $\mathit{38.32}\tinymath{\pm \mathit{2.06}}$\\
\, + \gradiendreligioncm & \!\!\! \!\!$\mathbf{-1.61}$ & $33.80$ & $34.86$ & $66.16$ & $49.60$ & $37.52$ & $51.40$ & $\mathit{61.99}$ & -- & \!\!$\mathbf{56.27}$ & \dab{$1.40$} $44.46\tinymath{\pm 2.03}$\\
\, + \gradiendreligionjm & \!\!\! $-2.47$ & \!\!$\mathit{\boldsymbol{\mathit{37.16}}}$ & \!\!$\mathbf{37.77}$ & $70.37$ & $52.37$ & $\mathit{41.42}$ & \!\!$\mathbf{59.50}$ & $\mathit{57.45}$ & -- & $56.26$ & \uag{$0.69$} $46.54\tinymath{\pm 1.90}$\\
\lightcmidrule{1-12}
\, + \inlp & \!\!\! $-6.76$ & $35.72$ & $36.73$ & \!\!$\mathit{\boldsymbol{\mathit{79.61}}}$ & $50.05$ & $36.91$ & $56.75$ & \!\!$\mathbf{76.08}$ & -- & \!\!$\mathbf{56.27}$ & \!\!\uag{$\boldsymbol{2.28}$} $\mathbf{48.14}\tinymath{\pm 1.84}$\\
\, + \sentencedebias & \!\!\! $-9.09$ & $34.80$ & $35.83$ & $67.55$ & $50.02$ & $37.46$ & $55.24$ & $75.19$ & -- & $54.86$ & \dab{$0.04$} $45.82\tinymath{\pm 1.96}$\\

\midrule
\llamai & \!\!$\mathbf{16.85}$ & $48.12$ & $47.97$ & $4.67$ & $57.34$ & $63.30$ & $64.69$ & $73.02$ & -- & \!\!$\mathbf{65.20}$ & $49.14\tinymath{\pm 1.92}$\\
\lightcmidrule{1-12}
\, + \gradiendreligioncj & $15.02$ & $48.60$ & $48.51$ & $8.29$ & $56.13$ & \!\!$\mathit{\boldsymbol{\mathit{64.61}}}$ & $66.05$ & $72.16$ & -- & $64.81$ & \uag{$0.31$} $49.45\tinymath{\pm 2.00}$\\
\, + \gradiendreligioncm & $9.31$ & $\mathit{35.58}$ & $\mathit{35.35}$ & $\mathit{0.00}$ & $\mathit{52.98}$ & $\mathit{64.14}$ & $65.47$ & \!\!$\mathbf{74.21}$ & -- & $55.88$ & \dab{$\mathit{4.46}$} $\mathit{44.68}\tinymath{\pm \mathit{1.30}}$\\
\, + \gradiendreligionjm & $4.76$ & $\mathit{43.28}$ & $\mathit{43.70}$ & \!\!$\mathit{\boldsymbol{\mathit{43.69}}}$ & $\mathit{50.54}$ & $\mathit{47.33}$ & \!\!$\mathbf{71.86}$ & $68.86$ & -- & $63.29$ & \uag{$0.09$} $49.23\tinymath{\pm 1.95}$\\
\lightcmidrule{1-12}
\, + \inlp & $15.71$ & \!\!$\mathbf{48.69}$ & \!\!$\mathbf{49.00}$ & $8.26$ & \!\!$\mathbf{57.96}$ & $63.75$ & $65.69$ & $73.52$ & -- & $63.40$ & \!\!\uag{$\boldsymbol{0.50}$} $\mathbf{49.64}\tinymath{\pm 1.99}$\\
\, + \sentencedebias & $14.98$ & $48.60$ & $48.72$ & $5.55$ & $57.33$ & $\mathit{64.21}$ & $65.66$ & $71.91$ & -- & $61.99$ & \dab{$0.35$} $48.79\tinymath{\pm 1.97}$\\
\bottomrule
\end{tabular}
    
\end{table}


\begin{table}[p]
    \centering
    \tiny
    \setlength{\tabcolsep}{2pt} 
    \caption{\textbf{Gender:} \acrshort{sglue} bootstrapped validation set scores with sub-results for encoder-only models. Statistically significant improvements are indicated in \emph{italics}, while the best score for each base model is highlighted in \textbf{bold}.} \label{tab:eval:superglue}
    \begin{tabular}{lrrrrrrrrr}
\toprule\textbf{Model} & \textbf{BoolQ} & \textbf{CB} & \textbf{COPA} & \textbf{MultiRC} & \textbf{ReCoRD} & \textbf{RTE} & \textbf{WiC} & \textbf{WSC} & \textbf{Average} $\uparrow$\\
\textbf{Metrics} & \textbf{Acc.} & \textbf{F1/Acc.} & \textbf{Acc.} & \textbf{$\text{F1}_\alpha$/EM} & \textbf{F1/EM} & \textbf{Acc.} & \textbf{Acc.} & \textbf{Acc.} & \textbf{}\\
\midrule
\bertbase & $69.16$ & $38.74$/$58.68$ & $62.72$ & $60.12$/$13.23$ & $56.09$/$55.32$ & $61.30$ & $68.67$ & $63.12$ & $51.82\tinymath{\pm 1.67}$\\
\lightcmidrule{1-10}
\, + \gradiendbpi & $70.83$ & $42.64$/$62.23$ & $59.57$ & $60.46$/$13.77$ & $55.79$/$55.03$ & $63.98$ & $68.42$ & $63.40$ & \uag{$0.56$} $52.38\tinymath{\pm 1.88}$\\
\, + \gradiendfpi & $70.49$ & $42.68$/$62.23$ & $59.76$ & $60.71$/$14.26$ & $56.01$/$55.24$ & $64.66$ & $69.43$ & $63.40$ & \uag{$0.82$} $52.65\tinymath{\pm 1.88}$\\
\, + \gradiendmpi & $70.41$ & $42.68$/$62.23$ & $58.84$ & $58.88$/$13.80$ & $55.98$/$55.22$ & $64.32$ & $69.15$ & $63.40$ & \uag{$0.44$} $52.27\tinymath{\pm 1.88}$\\
\lightcmidrule{1-10}
\, + \cda & $70.09$ & \!\!$\mathbf{47.75}$/$\mathbf{69.49}$ & $57.60$ & $60.43$/$15.30$ & $55.64$/$54.93$ & \!\!$\mathbf{65.41}$ & $67.29$ & $63.40$ & \uag{$1.33$} $53.16\tinymath{\pm 1.80}$\\
\, + \dropout & $68.53$ & $47.39$/$68.99$ & $55.56$ & $59.20$/$12.94$ & $55.00$/$54.23$ & $61.74$ & $65.15$ & $62.77$ & \dab{$0.34$} $51.48\tinymath{\pm 1.72}$\\
\, + \inlp & $69.25$ & $34.57$/$56.91$ & $58.67$ & $60.62$/$14.50$ & \!\!$\mathbf{56.27}$/$\mathbf{55.49}$ & $61.26$ & $66.44$ & $63.36$ & \dab{$0.80$} $51.02\tinymath{\pm 1.55}$\\
\, + \rlace & $69.03$ & $27.47$/$52.19$ & $62.70$ & $59.73$/$13.12$ & $56.07$/$55.30$ & $61.31$ & $68.73$ & \!\!$\mathbf{63.44}$ & \dab{$0.88$} $50.95\tinymath{\pm 1.54}$\\
\, + \leace & $69.06$ & $27.47$/$52.19$ & \!\!$\mathbf{63.06}$ & $60.21$/$13.48$ & $55.99$/$55.21$ & $61.18$ & $68.63$ & $63.11$ & \dab{$0.86$} $50.96\tinymath{\pm 1.55}$\\
\, + \sentencedebias & $69.06$ & $27.47$/$52.19$ & $63.06$ & $60.10$/$13.30$ & $55.97$/$55.21$ & $61.66$ & $68.63$ & $63.12$ & \dab{$0.83$} $50.99\tinymath{\pm 1.55}$\\
\lightcmidrule{1-10}
\, + \gradiendbpi\ + \inlp & \!\!$\mathbf{70.90}$ & $44.14$/$64.13$ & $61.73$ & $60.34$/$14.37$ & $55.94$/$55.16$ & $64.93$ & \!\!$\mathbf{69.87}$ & $63.40$ & \!\!\uag{$\boldsymbol{1.51}$} $\mathbf{53.33}\tinymath{\pm 1.82}$\\
\, + \gradiendbpi\ + \sentencedebias \!\!\! & $70.77$ & $42.64$/$62.23$ & $59.55$ & $60.48$/$13.76$ & $55.80$/$55.04$ & $64.36$ & $68.38$ & $63.40$ & \uag{$0.56$} $52.39\tinymath{\pm 1.88}$\\
\, + \cda\, + \inlp & $69.86$ & $46.50$/$67.73$ & $58.09$ & $59.13$/$13.95$ & $55.67$/$54.96$ & $65.12$ & $67.22$ & $63.40$ & \uag{$0.82$} $52.64\tinymath{\pm 1.78}$\\
\, + \dropout \, + \sentencedebias & $68.72$ & $46.98$/$68.41$ & $54.89$ & $59.17$/$13.08$ & $54.98$/$54.22$ & $61.87$ & $65.41$ & $62.46$ & \dab{$0.41$} $51.42\tinymath{\pm 1.71}$\\
\, + \cda\, + \sentencedebias & $70.12$ & \!\!$\mathbf{47.75}$/$\mathbf{69.49}$ & $58.23$ & \!\!$\mathbf{60.53}$/$\mathbf{15.31}$ & $55.67$/$54.96$ & \!\!$\mathbf{65.41}$ & $67.28$ & $63.40$ & \uag{$1.41$} $53.24\tinymath{\pm 1.79}$\\
\, + \dropout \, + \inlp & $68.45$ & $40.90$/$61.77$ & $55.21$ & $59.12$/$13.09$ & $55.48$/$54.74$ & $62.23$ & $64.94$ & $62.73$ & \dab{$1.10$} $50.73\tinymath{\pm 1.70}$\\

\midrule
\bertlarge & $70.32$ & $42.86$/$62.97$ & $61.46$ & $61.49$/$15.19$ & $61.70$/$61.04$ & $67.68$ & \!\!$\mathbf{70.82}$ & $62.09$ & $53.74\tinymath{\pm 1.62}$\\
\lightcmidrule{1-10}
\, + \gradiendbpi & \!\!$\mathit{\boldsymbol{\mathit{73.03}}}$ & $46.69$/$67.15$ & $65.34$ & $59.11$/$15.47$ & $61.47$/$60.78$ & $65.49$ & $69.53$ & $63.40$ & \uag{$0.46$} $54.20\tinymath{\pm 1.88}$\\
\, + \gradiendfpi & $71.86$ & $44.84$/$64.70$ & $60.46$ & $61.94$/$15.39$ & $61.45$/$60.75$ & $66.13$ & $69.55$ & $62.79$ & \uag{$0.10$} $53.84\tinymath{\pm 1.86}$\\
\, + \gradiendmpi & $71.61$ & $44.94$/$64.80$ & $58.32$ & $61.62$/$15.25$ & $61.67$/$61.01$ & $66.20$ & $69.62$ & $63.40$ & \dab{$0.10$} $53.64\tinymath{\pm 1.87}$\\
\lightcmidrule{1-10}
\, + \cda & $72.41$ & $47.59$/$68.35$ & $62.94$ & $61.90$/$16.43$ & $61.65$/$61.02$ & $61.95$ & $69.42$ & $63.40$ & \uag{$0.28$} $54.02\tinymath{\pm 1.81}$\\
\, + \dropout & $71.12$ & $45.18$/$65.27$ & $53.62$ & $62.24$/$16.01$ & \!\!$\mathbf{62.09}$/$\mathbf{61.37}$ & $64.72$ & $67.99$ & $63.40$ & \dab{$0.52$} $53.22\tinymath{\pm 1.68}$\\
\, + \inlp & $72.67$ & $38.72$/$61.76$ & $62.19$ & $\mathit{39.19}$/$\mathit{8.60}$ & $61.59$/$60.92$ & $66.09$ & $70.01$ & \!\!$\mathbf{63.45}$ & \dab{$1.60$} $52.14\tinymath{\pm 1.58}$\\
\, + \rlace & $\mathit{67.69}$ & $43.46$/$63.61$ & $61.47$ & $\mathit{39.28}$/$\mathit{9.05}$ & $61.78$/$61.11$ & \!\!$\mathbf{68.40}$ & $70.53$ & $60.42$ & \dab{$1.69$} $52.05\tinymath{\pm 1.62}$\\
\, + \leace & $70.54$ & $43.27$/$62.99$ & $61.78$ & $60.75$/$15.65$ & $61.51$/$60.83$ & $67.45$ & $69.75$ & $63.09$ & \uag{$0.09$} $53.84\tinymath{\pm 1.67}$\\
\, + \sentencedebias & $70.05$ & $42.57$/$62.42$ & $61.79$ & $61.25$/$15.52$ & $61.63$/$60.97$ & $67.78$ & $70.68$ & $63.08$ & \uag{$0.03$} $53.77\tinymath{\pm 1.66}$\\
\lightcmidrule{1-10}
\, + \gradiendbpi\ + \inlp & $72.01$ & $46.24$/$66.53$ & \!\!$\mathbf{66.38}$ & $60.65$/$15.10$ & $61.48$/$60.81$ & $66.13$ & $69.22$ & $63.06$ & \uag{$0.56$} $54.30\tinymath{\pm 1.93}$\\
\, + \gradiendbpi\ + \sentencedebias \!\!\! & $71.78$ & $46.69$/$67.15$ & $65.93$ & $61.31$/$15.46$ & $61.88$/$61.21$ & $66.21$ & $69.67$ & $63.40$ & \!\!\uag{$\boldsymbol{0.64}$} $\mathbf{54.38}\tinymath{\pm 1.87}$\\
\, + \cda\, + \inlp & $72.25$ & $47.18$/$67.78$ & $64.91$ & $61.05$/$15.34$ & $61.73$/$61.06$ & $64.33$ & $69.15$ & $62.76$ & \uag{$0.12$} $53.87\tinymath{\pm 1.81}$\\
\, + \dropout \, + \sentencedebias & $70.60$ & $47.18$/$67.77$ & $58.25$ & $61.90$/$14.41$ & $61.04$/$60.40$ & $66.24$ & $66.38$ & $63.06$ & \dab{$0.39$} $53.36\tinymath{\pm 1.74}$\\
\, + \cda\, + \sentencedebias & $72.50$ & \!\!$\mathbf{47.95}$/$\mathbf{68.92}$ & $63.62$ & $62.62$/$15.23$ & $61.81$/$61.16$ & $61.66$ & $69.43$ & $63.38$ & \uag{$0.26$} $54.00\tinymath{\pm 1.80}$\\
\, + \dropout \, + \inlp & $70.27$ & $47.22$/$67.77$ & $58.69$ & \!\!$\mathbf{62.34}$/$\mathbf{16.61}$ & $61.17$/$60.49$ & $65.32$ & $\mathit{64.34}$ & $63.40$ & \dab{$0.41$} $53.33\tinymath{\pm 1.74}$\\

\midrule
\distilbert & $69.75$ & $45.62$/$66.55$ & $53.39$ & $57.58$/$12.21$ & $49.09$/$48.27$ & $55.18$ & $62.10$ & \!\!$\mathbf{63.40}$ & $49.69\tinymath{\pm 1.65}$\\
\lightcmidrule{1-10}
\, + \gradiendbpi & $69.81$ & $46.63$/$67.72$ & $55.06$ & $57.77$/$13.13$ & $49.02$/$48.21$ & $55.11$ & $62.01$ & \!\!$\mathbf{63.40}$ & \uag{$0.21$} $49.90\tinymath{\pm 1.67}$\\
\, + \gradiendfpi & $69.83$ & $47.50$/$68.96$ & $55.76$ & $58.14$/$13.46$ & $49.15$/$48.37$ & $55.61$ & $61.55$ & \!\!$\mathbf{63.40}$ & \uag{$0.63$} $50.32\tinymath{\pm 1.63}$\\
\, + \gradiendmpi & $69.50$ & $44.02$/$64.77$ & $55.41$ & $58.51$/$12.90$ & $49.06$/$48.24$ & $56.04$ & $61.37$ & \!\!$\mathbf{63.40}$ & \dab{$0.05$} $49.64\tinymath{\pm 1.69}$\\
\lightcmidrule{1-10}
\, + \cda & $69.19$ & \!\!$\mathbf{48.31}$/$\mathbf{69.52}$ & $59.40$ & $59.85$/$13.31$ & $49.16$/$48.33$ & $58.10$ & $63.78$ & \!\!$\mathbf{63.40}$ & \uag{$1.06$} $50.75\tinymath{\pm 1.76}$\\
\, + \dropout & $69.21$ & $46.13$/$66.49$ & $54.78$ & $59.40$/$13.05$ & \!\!$\mathbf{49.77}$/$\mathbf{48.97}$ & $60.45$ & $62.62$ & \!\!$\mathbf{63.40}$ & \uag{$0.58$} $50.27\tinymath{\pm 1.75}$\\
\, + \inlp & $69.85$ & $40.07$/$63.63$ & \!\!$\mathbf{60.46}$ & $58.76$/$12.21$ & $48.94$/$48.10$ & $55.58$ & $61.86$ & \!\!$\mathbf{63.40}$ & \uag{$0.21$} $49.90\tinymath{\pm 1.56}$\\
\, + \rlace & \!\!$\mathbf{69.99}$ & $45.08$/$65.95$ & $53.74$ & $57.04$/$11.69$ & $49.08$/$48.28$ & $56.31$ & $61.85$ & \!\!$\mathbf{63.40}$ & \uag{$0.03$} $49.72\tinymath{\pm 1.66}$\\
\, + \leace & $69.54$ & $45.92$/$67.15$ & $55.39$ & $57.48$/$11.20$ & $49.20$/$48.39$ & $54.13$ & $62.52$ & \!\!$\mathbf{63.40}$ & \uag{$0.09$} $49.78\tinymath{\pm 1.63}$\\
\, + \sentencedebias & $69.97$ & $45.08$/$65.95$ & $54.42$ & $57.20$/$11.76$ & $49.12$/$48.32$ & $55.55$ & $62.10$ & \!\!$\mathbf{63.40}$ & \uag{$0.06$} $49.75\tinymath{\pm 1.64}$\\
\lightcmidrule{1-10}
\, + \gradiendbpi\ + \inlp & $69.90$ & $39.64$/$63.00$ & $58.75$ & $59.38$/$13.24$ & $48.94$/$48.12$ & $56.17$ & $61.82$ & \!\!$\mathbf{63.40}$ & \uag{$0.16$} $49.85\tinymath{\pm 1.57}$\\
\, + \gradiendbpi\ + \sentencedebias \!\!\! & $69.92$ & $46.23$/$67.12$ & $55.45$ & $57.96$/$12.47$ & $49.09$/$48.28$ & $54.49$ & $61.65$ & \!\!$\mathbf{63.40}$ & \uag{$0.25$} $49.94\tinymath{\pm 1.66}$\\
\, + \cda\, + \inlp & $69.29$ & $46.15$/$67.16$ & $55.12$ & \!\!$\mathbf{59.45}$/$\mathbf{13.86}$ & $49.28$/$48.45$ & $60.00$ & \!\!$\mathbf{64.78}$ & \!\!$\mathbf{63.40}$ & \uag{$1.40$} $51.09\tinymath{\pm 1.72}$\\
\, + \dropout \, + \sentencedebias & $69.28$ & $46.13$/$66.49$ & $55.54$ & $59.34$/$13.70$ & $49.73$/$48.93$ & $60.49$ & $63.08$ & \!\!$\mathbf{63.40}$ & \uag{$0.76$} $50.45\tinymath{\pm 1.76}$\\
\, + \cda\, + \sentencedebias & $69.08$ & \!\!$\mathbf{48.31}$/$\mathbf{69.52}$ & $57.82$ & $56.55$/$11.78$ & $49.10$/$48.27$ & $58.23$ & $63.80$ & \!\!$\mathbf{63.40}$ & \uag{$0.83$} $50.52\tinymath{\pm 1.77}$\\
\, + \dropout \, + \inlp & $69.17$ & $47.31$/$68.30$ & $59.12$ & $58.97$/$12.72$ & $49.69$/$48.87$ & \!\!$\mathbf{63.22}$ & $62.91$ & \!\!$\mathbf{63.40}$ & \!\!\uag{$\boldsymbol{1.50}$} $\mathbf{51.19}\tinymath{\pm 1.72}$\\

\midrule
\roberta & $82.01$ & $46.41$/$66.62$ & $56.70$ & $42.49$/$12.60$ & $72.14$/$71.46$ & $75.36$ & $56.83$ & $53.49$ & $53.31\tinymath{\pm 1.48}$\\
\lightcmidrule{1-10}
\, + \gradiendbpi & $\mathit{75.70}$ & $45.99$/$66.03$ & $58.40$ & $\mathit{64.23}$/$\mathit{21.85}$ & $71.98$/$71.30$ & $76.18$ & $\mathit{66.76}$ & $53.49$ & \uag{$2.03$} $55.34\tinymath{\pm 1.47}$\\
\, + \gradiendfpi & $81.64$ & $44.18$/$64.24$ & $60.06$ & $\mathit{22.50}$/$\mathit{7.40}$ & $72.11$/$71.46$ & $69.97$ & $\mathit{62.88}$ & $62.09$ & \dab{$0.49$} $52.82\tinymath{\pm 1.65}$\\
\, + \gradiendmpi & $82.61$ & $43.43$/$62.44$ & $54.16$ & $\mathit{67.91}$/$\mathit{23.34}$ & $71.99$/$71.34$ & $\mathit{62.29}$ & $\mathit{63.62}$ & $52.23$ & \uag{$0.48$} $53.79\tinymath{\pm 1.47}$\\
\lightcmidrule{1-10}
\, + \cda & $82.80$ & $47.57$/$68.36$ & $63.25$ & $\mathit{45.02}$/$\mathit{17.17}$ & $72.20$/$71.57$ & $78.13$ & $\mathit{69.75}$ & $53.49$ & \uag{$2.89$} $56.20\tinymath{\pm 1.44}$\\
\, + \dropout & $\mathit{73.53}$ & $45.03$/$64.77$ & $50.32$ & $\mathit{45.78}$/$\mathit{17.10}$ & \!\!$\mathbf{72.28}$/$\mathbf{71.60}$ & $\mathit{60.86}$ & $61.03$ & $57.62$ & \dab{$2.25$} $51.05\tinymath{\pm 1.62}$\\
\, + \inlp & $82.32$ & $47.15$/$67.79$ & $57.98$ & $\mathit{45.58}$/$\mathit{16.43}$ & $71.98$/$71.34$ & $75.60$ & $61.52$ & \!\!$\mathbf{63.40}$ & \uag{$1.75$} $55.06\tinymath{\pm 1.66}$\\
\, + \rlace & \!\!$\mathbf{82.93}$ & $45.58$/$65.45$ & $56.92$ & $\mathit{44.94}$/$\mathit{16.30}$ & $71.97$/$71.30$ & $73.22$ & $61.82$ & $53.49$ & \uag{$0.67$} $53.98\tinymath{\pm 1.50}$\\
\, + \leace & $\mathit{75.70}$ & $38.84$/$61.85$ & $56.91$ & $\mathit{68.43}$/$\mathit{24.30}$ & $72.10$/$71.45$ & $68.66$ & $57.75$ & $59.22$ & \uag{$0.16$} $53.47\tinymath{\pm 1.35}$\\
\, + \sentencedebias & $82.39$ & $45.97$/$66.00$ & $55.51$ & $\mathit{65.65}$/$\mathit{21.52}$ & $71.94$/$71.27$ & $68.87$ & $61.11$ & $53.49$ & \uag{$1.22$} $54.53\tinymath{\pm 1.51}$\\
\lightcmidrule{1-10}
\, + \gradiendbpi\ + \inlp & $\mathit{75.56}$ & $47.62$/$68.44$ & $57.94$ & $\mathit{63.42}$/$\mathit{19.59}$ & $72.14$/$71.44$ & $69.91$ & $\mathit{64.79}$ & \!\!$\mathbf{63.40}$ & \uag{$1.86$} $55.17\tinymath{\pm 1.63}$\\
\, + \gradiendbpi\ + \sentencedebias \!\!\! & $82.37$ & $46.41$/$66.62$ & $55.82$ & $\mathit{65.14}$/$\mathit{22.32}$ & $71.98$/$71.31$ & $\mathit{66.49}$ & $\mathit{65.05}$ & $52.85$ & \uag{$1.41$} $54.72\tinymath{\pm 1.49}$\\
\, + \cda\, + \inlp & $81.81$ & \!\!$\mathbf{49.16}$/$\mathbf{70.74}$ & $61.69$ & \!\!$\mathit{\boldsymbol{\mathit{67.80}}}$/$\mathit{\boldsymbol{\mathit{25.09}}}$ & $72.10$/$71.49$ & \!\!$\mathbf{79.04}$ & \!\!$\mathit{\boldsymbol{\mathit{70.75}}}$ & \!\!$\mathbf{63.40}$ & \!\!\uag{$\boldsymbol{\mathit{5.58}}$} $\mathit{\boldsymbol{\mathit{58.89}}}\tinymath{\pm \mathit{1.63}}$\\
\, + \dropout \, + \sentencedebias & $\mathit{69.98}$ & $45.03$/$64.77$ & $47.05$ & $\mathit{62.92}$/$\mathit{19.86}$ & $71.64$/$70.91$ & $\mathit{61.04}$ & $56.94$ & $57.92$ & \dab{$2.29$} $51.01\tinymath{\pm 1.59}$\\
\, + \cda\, + \sentencedebias & $82.68$ & $48.44$/$69.59$ & \!\!$\mathbf{64.59}$ & $\mathit{45.36}$/$\mathit{16.38}$ & $71.99$/$71.35$ & $77.56$ & $\mathit{69.56}$ & $53.49$ & \uag{$\mathit{2.97}$} $\mathit{56.28}\tinymath{\pm \mathit{1.42}}$\\
\, + \dropout \, + \inlp & $\mathit{71.96}$ & $46.73$/$67.16$ & $51.30$ & $\mathit{0.00}$/$\mathit{0.32}$ & $71.88$/$71.17$ & $\mathit{59.29}$ & $59.13$ & \!\!$\mathbf{63.40}$ & \dab{$\mathit{5.10}$} $\mathit{48.21}\tinymath{\pm \mathit{1.78}}$\\

\bottomrule
\end{tabular}
\end{table}

\begin{table}[]
    \centering
    \tiny
    \setlength{\tabcolsep}{2pt} 
   \caption{\textbf{Gender:} \acrshort{sglue} bootstrapped validation set scores with sub-results for decoder-only models. Statistically significant improvements are indicated in \emph{italics}, while the best score for each base model is highlighted in \textbf{bold}. \gpttwo\ results were computed after fine-tuning and \llama-based results were computed with zero-shot evaluation.} \label{tab:eval:superglue-gpt}
\begin{tabular}{lrrrrrrrrr}
\toprule\textbf{Model} & \textbf{BoolQ} & \textbf{CB} & \textbf{COPA} & \textbf{MultiRC} & \textbf{ReCoRD} & \textbf{RTE} & \textbf{WiC} & \textbf{WSC} & \textbf{Average} $\uparrow$\\
\textbf{Metrics} & \textbf{Acc.} & \textbf{F1/Acc.} & \textbf{Acc.} & \textbf{$\text{F1}_\alpha$/EM} & \textbf{F1/EM} & \textbf{Acc.} & \textbf{Acc.} & \textbf{Acc.} & \textbf{}\\
\midrule
\gpttwo & $65.56$ & $36.86$/$51.74$ & $49.35$ & $58.79$/$13.69$ & $31.64$/$30.93$ & $60.14$ & $62.51$ & $54.47$ & $45.49\tinymath{\pm 1.28}$\\
\lightcmidrule{1-10}
\, + \gradiendbpi & $65.22$ & $36.10$/$51.07$ & $50.88$ & $59.92$/$14.14$ & $31.77$/$31.04$ & $62.41$ & $63.82$ & \!\!$\mathbf{59.53}$ & \uag{$0.86$} $46.34\tinymath{\pm 1.27}$\\
\, + \gradiendfpi & $64.66$ & $37.56$/$51.70$ & $50.74$ & $59.01$/$13.85$ & \!\!$\mathbf{31.84}$/$\mathbf{31.14}$ & $63.28$ & $63.50$ & $55.42$ & \uag{$0.49$} $45.97\tinymath{\pm 1.20}$\\
\, + \gradiendmpi & $65.28$ & $38.49$/$55.21$ & $49.26$ & $60.09$/$13.86$ & $31.65$/$30.92$ & $61.42$ & $63.92$ & $56.34$ & \uag{$0.60$} $46.09\tinymath{\pm 1.25}$\\
\lightcmidrule{1-10}
\, + \cda & \!\!$\mathbf{66.70}$ & \!\!$\mathbf{42.95}$/$\mathbf{57.57}$ & $49.35$ & $59.81$/$14.34$ & $31.61$/$30.92$ & $61.84$ & $64.43$ & $57.69$ & \uag{$1.28$} $46.76\tinymath{\pm 1.38}$\\
\, + \dropout & $66.02$ & $34.95$/$52.30$ & $49.04$ & $58.82$/$13.86$ & $31.55$/$30.86$ & $62.26$ & $62.66$ & $58.72$ & \uag{$0.46$} $45.94\tinymath{\pm 1.45}$\\
\, + \inlp & $65.77$ & $36.77$/$51.74$ & $51.34$ & $58.80$/$13.41$ & $31.49$/$30.78$ & $60.86$ & $62.27$ & $54.78$ & \uag{$0.29$} $45.78\tinymath{\pm 1.20}$\\
\, + \rlace & $65.84$ & $37.07$/$52.37$ & $49.26$ & $59.24$/$13.76$ & $31.58$/$30.85$ & $62.33$ & $62.77$ & $55.75$ & \uag{$0.44$} $45.92\tinymath{\pm 1.20}$\\
\, + \leace & $65.59$ & $35.53$/$50.52$ & $51.65$ & $58.56$/$12.88$ & $31.61$/$30.88$ & $60.76$ & $61.79$ & $57.10$ & \uag{$0.35$} $45.83\tinymath{\pm 1.24}$\\
\, + \sentencedebias & $65.23$ & $28.85$/$42.82$ & $51.31$ & $58.70$/$12.53$ & $31.69$/$30.96$ & $61.11$ & $61.77$ & $51.23$ & \dab{$1.15$} $44.33\tinymath{\pm 1.18}$\\
\lightcmidrule{1-10}
\, + \gradiendbpi\ + \inlp & $65.36$ & $35.69$/$50.45$ & $51.93$ & $59.51$/$13.83$ & $31.75$/$31.03$ & $63.02$ & $63.50$ & $58.30$ & \uag{$0.82$} $46.30\tinymath{\pm 1.27}$\\
\, + \gradiendbpi\ + \sentencedebias \!\!\! & $65.18$ & $37.22$/$52.78$ & $52.92$ & $59.11$/$13.62$ & $31.66$/$30.96$ & $62.71$ & $63.69$ & $53.87$ & \uag{$0.47$} $45.96\tinymath{\pm 1.24}$\\
\, + \cda\, + \inlp & $66.45$ & $42.14$/$56.39$ & $52.42$ & $59.78$/$14.55$ & $31.67$/$30.98$ & $63.06$ & $63.96$ & $56.12$ & \!\!\uag{$\boldsymbol{1.39}$} $\mathbf{46.87}\tinymath{\pm 1.32}$\\
\, + \dropout \, + \sentencedebias & $65.59$ & $37.65$/$56.99$ & $51.33$ & $59.85$/$14.39$ & $31.51$/$30.81$ & \!\!$\mathbf{64.40}$ & $63.28$ & $56.57$ & \uag{$1.28$} $46.76\tinymath{\pm 1.32}$\\
\, + \cda\, + \sentencedebias & $66.66$ & $41.89$/$56.34$ & $49.04$ & \!\!$\mathbf{60.03}$/$\mathbf{14.55}$ & $31.77$/$31.07$ & $62.55$ & \!\!$\mathbf{64.63}$ & $53.58$ & \uag{$0.78$} $46.27\tinymath{\pm 1.34}$\\
\, + \dropout \, + \inlp & $65.99$ & $35.97$/$53.47$ & \!\!$\mathbf{53.71}$ & $59.01$/$14.11$ & $31.62$/$30.93$ & $62.89$ & $62.39$ & $58.70$ & \uag{$1.11$} $46.59\tinymath{\pm 1.44}$\\

\midrule
\llama & $72.96$ & \!\!$\mathbf{37.32}$/$\mathbf{51.82}$ & \!\!$\mathbf{86.06}$ & $0.00$/$0.32$ & \!\!$\mathbf{90.42}$/$\mathbf{89.70}$ & $54.17$ & $50.10$ & $37.58$ & \!\!$\mathbf{54.46}\tinymath{\pm 2.28}$\\
\lightcmidrule{1-10}
\, + \gradiendbpi & $\mathit{65.24}$ & $31.80$/$44.35$ & $76.03$ & \!\!$\mathit{\boldsymbol{\mathit{1.48}}}$/$\mathit{\boldsymbol{\mathit{0.42}}}$ & $\mathit{88.73}$/$\mathit{87.92}$ & $52.41$ & $50.12$ & $36.60$ & \dab{$3.49$} $50.97\tinymath{\pm 2.20}$\\
\, + \gradiendfpi & \!\!$\mathbf{73.58}$ & $33.67$/$42.88$ & $80.10$ & $0.00$/$0.32$ & $89.38$/$88.63$ & \!\!$\mathbf{57.02}$ & $50.12$ & $36.60$ & \dab{$1.35$} $53.11\tinymath{\pm 2.28}$\\
\, + \gradiendmpi & $\mathit{69.02}$ & $26.71$/$40.85$ & $82.11$ & $0.58$/$0.53$ & $89.57$/$88.78$ & $54.56$ & $50.12$ & \!\!$\mathbf{39.50}$ & \dab{$2.10$} $52.35\tinymath{\pm 2.09}$\\
\lightcmidrule{1-10}
\, + \inlp & $\mathit{65.64}$ & $26.77$/$37.46$ & $78.10$ & $0.00$/$0.32$ & $\mathit{86.00}$/$\mathit{85.26}$ & $48.22$ & $50.12$ & $36.60$ & \dab{$\mathit{4.88}$} $\mathit{49.57}\tinymath{\pm \mathit{2.21}}$\\
\, + \rlace & $73.17$ & $36.37$/$51.75$ & \!\!$\mathbf{86.06}$ & $0.00$/$0.32$ & $\mathit{0.00}$/$\mathit{0.01}$ & $53.43$ & $50.25$ & $36.60$ & \dab{$\mathit{11.49}$} $\mathit{42.97}\tinymath{\pm \mathit{2.31}}$\\
\, + \leace & $73.33$ & $36.37$/$51.75$ & $85.07$ & $0.00$/$0.32$ & $\mathit{0.00}$/$\mathit{0.01}$ & $53.76$ & \!\!$\mathbf{50.42}$ & $36.60$ & \dab{$\mathit{11.53}$} $\mathit{42.93}\tinymath{\pm \mathit{2.31}}$\\
\, + \sentencedebias & $73.28$ & $36.87$/$46.41$ & $85.08$ & $0.00$/$0.32$ & $90.06$/$89.27$ & $55.27$ & $50.27$ & $37.58$ & \dab{$0.34$} $54.12\tinymath{\pm 2.37}$\\
\lightcmidrule{1-10}
\, + \gradiendbpi\ + \inlp & $\mathit{62.37}$ & $\mathit{13.24}$/$\mathit{15.88}$ & $\mathit{68.84}$ & $0.00$/$0.32$ & $\mathit{84.05}$/$\mathit{83.34}$ & $47.59$ & $50.12$ & $36.60$ & \dab{$\mathit{8.97}$} $\mathit{45.49}\tinymath{\pm \mathit{2.08}}$\\
\, + \gradiendbpi\ + \sentencedebias \!\!\! & $\mathit{66.73}$ & $27.66$/$35.64$ & $76.03$ & $0.94$/$0.31$ & $\mathit{88.65}$/$\mathit{87.83}$ & $53.18$ & $50.12$ & $36.60$ & \dab{$4.06$} $50.40\tinymath{\pm 2.16}$\\

\midrule
\llamai & $75.25$ & $31.58$/$32.13$ & $78.60$ & $27.43$/$0.52$ & \!\!$\mathbf{85.32}$/$\mathbf{84.68}$ & $67.31$ & \!\!$\mathbf{50.37}$ & $62.20$ & $58.07\tinymath{\pm 2.29}$\\
\lightcmidrule{1-10}
\, + \gradiendbpi & $73.23$ & $18.78$/$39.43$ & $75.08$ & $\mathit{3.00}$/$\mathit{0.32}$ & $\mathit{83.11}$/$\mathit{82.48}$ & $\mathit{53.77}$ & $50.11$ & $58.27$ & \dab{$\mathit{5.07}$} $\mathit{53.00}\tinymath{\pm \mathit{2.05}}$\\
\, + \gradiendfpi & $73.41$ & \!\!$\mathbf{44.03}$/$\mathbf{49.36}$ & \!\!$\mathbf{82.89}$ & \!\!$\mathbf{31.00}$/$\mathbf{1.12}$ & $84.84$/$84.21$ & $68.32$ & $50.04$ & $64.05$ & \!\!\uag{$\boldsymbol{2.68}$} $\mathbf{60.75}\tinymath{\pm 2.35}$\\
\, + \gradiendmpi & \!\!$\mathit{\boldsymbol{\mathit{78.54}}}$ & $30.28$/$37.55$ & $80.19$ & $\mathit{4.57}$/$\mathit{0.21}$ & $84.03$/$83.48$ & \!\!$\mathbf{70.91}$ & $50.11$ & $51.11$ & \dab{$1.71$} $56.36\tinymath{\pm 2.28}$\\
\lightcmidrule{1-10}
\, + \inlp & $72.27$ & $32.65$/$33.37$ & $81.02$ & $\mathit{11.51}$/$\mathit{0.56}$ & $84.70$/$84.08$ & $66.60$ & $50.26$ & \!\!$\mathbf{65.24}$ & \dab{$0.72$} $57.35\tinymath{\pm 2.34}$\\
\, + \rlace & $75.21$ & $32.10$/$32.75$ & $78.60$ & $27.24$/$0.49$ & $85.08$/$84.42$ & $67.43$ & \!\!$\mathbf{50.37}$ & $61.54$ & \dab{$0.05$} $58.02\tinymath{\pm 2.28}$\\
\, + \leace & $75.18$ & $30.89$/$31.57$ & $78.25$ & $26.90$/$0.52$ & $85.06$/$84.41$ & $67.19$ & $50.22$ & $62.18$ & \dab{$0.23$} $57.84\tinymath{\pm 2.29}$\\
\, + \sentencedebias & $74.98$ & $34.06$/$35.20$ & $79.63$ & $27.15$/$0.53$ & $85.24$/$84.58$ & $67.42$ & $50.22$ & $62.86$ & \uag{$0.49$} $58.56\tinymath{\pm 2.31}$\\
\lightcmidrule{1-10}
\, + \gradiendbpi\ + \inlp & $\mathit{66.07}$ & $18.78$/$39.43$ & $78.07$ & $\mathit{0.00}$/$\mathit{0.32}$ & $\mathit{80.73}$/$\mathit{80.11}$ & $\mathit{55.10}$ & $50.12$ & $\mathit{41.57}$ & \dab{$\mathit{7.99}$} $\mathit{50.08}\tinymath{\pm \mathit{2.08}}$\\
\, + \gradiendbpi\ + \sentencedebias \!\!\! & $72.87$ & $18.78$/$39.43$ & $76.37$ & $\mathit{3.25}$/$\mathit{0.32}$ & $\mathit{83.18}$/$\mathit{82.54}$ & $\mathit{53.63}$ & $49.90$ & $57.28$ & \dab{$\mathit{5.10}$} $\mathit{52.97}\tinymath{\pm \mathit{2.04}}$\\

\bottomrule
\end{tabular}
\end{table}

\begin{table}[p]
    \centering
    \tiny
    \setlength{\tabcolsep}{2pt} 
    \caption{\textbf{Race:} \acrshort{sglue} bootstrapped validation set scores with sub-results for all models. Statistically significant improvements are indicated in \emph{italics}, while the best score for each base model is highlighted in \textbf{bold}. \llama-based results were computed with zero-shot evaluation while all other scores are derived after fine-tuning.}
    \label{tab:eval-superglue-race}

\begin{tabular}{lrrrrrrrrr}
\toprule\textbf{Model} & \textbf{BoolQ} & \textbf{CB} & \textbf{COPA} & \textbf{MultiRC} & \textbf{ReCoRD} & \textbf{RTE} & \textbf{WiC} & \textbf{WSC} & \textbf{Average} $\uparrow$\\
\textbf{Metrics} & \textbf{Acc.} & \textbf{F1/Acc.} & \textbf{Acc.} & \textbf{$\text{F1}_\alpha$/EM} & \textbf{F1/EM} & \textbf{Acc.} & \textbf{Acc.} & \textbf{Acc.} & \textbf{}\\
\midrule
\bertbase & $69.16$ & $38.74$/$58.68$ & $62.72$ & $60.12$/$13.23$ & \!\!$\mathbf{56.09}$/$\mathbf{55.32}$ & $61.30$ & $68.67$ & $63.12$ & $51.82\tinymath{\pm 1.67}$\\
\lightcmidrule{1-10}
\, + \gradiendraceab & $70.45$ & $42.71$/$62.23$ & $58.88$ & $60.37$/$14.17$ & $55.50$/$54.70$ & \!\!$\mathbf{67.37}$ & \!\!$\mathbf{68.82}$ & \!\!$\mathbf{63.40}$ & \uag{$0.80$} $52.62\tinymath{\pm 1.89}$\\
\, + \gradiendraceaw & $70.36$ & $42.66$/$62.25$ & $57.19$ & $60.61$/$14.22$ & $55.62$/$54.86$ & $66.64$ & $68.36$ & \!\!$\mathbf{63.40}$ & \uag{$0.53$} $52.36\tinymath{\pm 1.90}$\\
\, + \gradiendracebw & \!\!$\mathbf{70.66}$ & $42.68$/$62.23$ & $61.53$ & \!\!$\mathbf{60.91}$/$\mathbf{14.80}$ & $55.33$/$54.56$ & $65.54$ & $68.43$ & \!\!$\mathbf{63.40}$ & \uag{$0.82$} $52.65\tinymath{\pm 1.88}$\\
\lightcmidrule{1-10}
\, + \cda & $70.33$ & \!\!$\mathbf{47.60}$/$\mathbf{68.92}$ & $60.55$ & $60.04$/$15.31$ & $55.43$/$54.69$ & $63.94$ & $67.49$ & \!\!$\mathbf{63.40}$ & \!\!\uag{$\boldsymbol{1.15}$} $\mathbf{52.98}\tinymath{\pm 1.82}$\\
\, + \dropout & $68.53$ & $47.39$/$68.99$ & $55.56$ & $59.20$/$12.94$ & $55.00$/$54.23$ & $61.74$ & $65.15$ & $62.77$ & \dab{$0.34$} $51.48\tinymath{\pm 1.72}$\\
\, + \inlp & $69.00$ & $31.77$/$54.49$ & \!\!$\mathbf{63.06}$ & $60.68$/$13.86$ & $55.89$/$55.14$ & $59.34$ & $67.12$ & $57.79$ & \dab{$1.38$} $50.44\tinymath{\pm 1.54}$\\
\, + \sentencedebias & $68.88$ & $27.47$/$52.19$ & $62.05$ & $60.59$/$14.25$ & $56.01$/$55.25$ & $62.50$ & $68.57$ & $62.18$ & \dab{$0.95$} $50.87\tinymath{\pm 1.54}$\\

\midrule
\bertlarge & $70.32$ & $42.86$/$62.97$ & $61.46$ & $61.49$/$15.19$ & $61.70$/$61.04$ & \!\!$\mathbf{67.68}$ & \!\!$\mathbf{70.82}$ & $62.09$ & $53.74\tinymath{\pm 1.62}$\\
\lightcmidrule{1-10}
\, + \gradiendraceab & $72.31$ & $46.09$/$66.44$ & $61.78$ & $61.49$/$15.39$ & $62.02$/$61.36$ & $65.04$ & $70.63$ & $63.72$ & \uag{$0.53$} $54.27\tinymath{\pm 1.85}$\\
\, + \gradiendraceaw & $72.32$ & $46.09$/$66.49$ & $64.81$ & $59.22$/$15.35$ & $61.75$/$61.07$ & $67.44$ & $70.13$ & $63.40$ & \!\!\uag{$\boldsymbol{0.84}$} $\mathbf{54.58}\tinymath{\pm 1.85}$\\
\, + \gradiendracebw & \!\!$\mathbf{72.66}$ & $46.57$/$67.10$ & \!\!$\mathbf{65.38}$ & $59.72$/$15.42$ & $61.79$/$61.13$ & $67.21$ & $69.71$ & $63.40$ & \uag{$0.68$} $54.42\tinymath{\pm 1.87}$\\
\lightcmidrule{1-10}
\, + \cda & $71.79$ & \!\!$\mathbf{47.18}$/$\mathbf{67.78}$ & $61.24$ & $62.16$/$15.82$ & $61.47$/$60.76$ & $58.36$ & $68.33$ & $61.80$ & \dab{$0.59$} $53.15\tinymath{\pm 1.73}$\\
\, + \dropout & $71.12$ & $45.18$/$65.27$ & $53.62$ & $62.24$/$16.01$ & \!\!$\mathbf{62.09}$/$\mathbf{61.37}$ & $64.72$ & $67.99$ & $63.40$ & \dab{$0.52$} $53.22\tinymath{\pm 1.68}$\\
\, + \inlp & $69.73$ & $38.31$/$61.17$ & $62.86$ & \!\!$\mathbf{62.30}$/$\mathbf{16.11}$ & $61.88$/$61.22$ & $67.33$ & $70.35$ & \!\!$\mathbf{63.74}$ & \dab{$0.24$} $53.50\tinymath{\pm 1.58}$\\
\, + \sentencedebias & $70.71$ & $43.38$/$63.53$ & $61.17$ & $59.46$/$14.09$ & $61.73$/$61.07$ & $66.92$ & $70.34$ & $62.78$ & \dab{$0.07$} $53.68\tinymath{\pm 1.67}$\\

\midrule
\distilbert & $69.75$ & $45.62$/$66.55$ & $53.39$ & $57.58$/$12.21$ & $49.09$/$48.27$ & $55.18$ & $62.10$ & \!\!$\mathbf{63.40}$ & $49.69\tinymath{\pm 1.65}$\\
\lightcmidrule{1-10}
\, + \gradiendraceab & $69.68$ & $47.21$/$68.31$ & $55.80$ & $57.61$/$12.80$ & $48.97$/$48.19$ & $53.87$ & $61.55$ & \!\!$\mathbf{63.40}$ & \uag{$0.00$} $49.69\tinymath{\pm 1.69}$\\
\, + \gradiendraceaw & $69.88$ & $46.71$/$67.71$ & $56.08$ & $58.19$/$12.19$ & $49.03$/$48.23$ & $55.03$ & $62.15$ & \!\!$\mathbf{63.40}$ & \uag{$0.38$} $50.07\tinymath{\pm 1.70}$\\
\, + \gradiendracebw & $69.68$ & $45.78$/$66.46$ & $53.00$ & $58.01$/$12.57$ & $49.04$/$48.23$ & $53.63$ & $61.48$ & \!\!$\mathbf{63.40}$ & \dab{$0.40$} $49.29\tinymath{\pm 1.70}$\\
\lightcmidrule{1-10}
\, + \cda & $68.86$ & $47.84$/$68.89$ & \!\!$\mathbf{58.52}$ & $58.41$/$12.59$ & $48.97$/$48.15$ & $60.05$ & \!\!$\mathbf{63.77}$ & \!\!$\mathbf{63.40}$ & \uag{$1.11$} $50.80\tinymath{\pm 1.79}$\\
\, + \dropout & $69.21$ & $46.13$/$66.49$ & $54.78$ & \!\!$\mathbf{59.40}$/$\mathbf{13.05}$ & \!\!$\mathbf{49.77}$/$\mathbf{48.97}$ & \!\!$\mathbf{60.45}$ & $62.62$ & \!\!$\mathbf{63.40}$ & \uag{$0.58$} $50.27\tinymath{\pm 1.75}$\\
\, + \inlp & $70.14$ & \!\!$\mathbf{47.91}$/$\mathbf{69.53}$ & $57.42$ & $58.05$/$12.47$ & $48.94$/$48.14$ & $57.85$ & $62.62$ & \!\!$\mathbf{63.40}$ & \!\!\uag{$\boldsymbol{1.14}$} $\mathbf{50.83}\tinymath{\pm 1.68}$\\
\, + \sentencedebias & \!\!$\mathbf{70.20}$ & $43.97$/$65.37$ & $55.50$ & $58.19$/$11.79$ & $49.04$/$48.24$ & $54.83$ & $62.26$ & \!\!$\mathbf{63.40}$ & \dab{$0.22$} $49.47\tinymath{\pm 1.62}$\\

\midrule
\roberta & $82.01$ & \!\!$\mathbf{46.41}$/$\mathbf{66.62}$ & $56.70$ & $42.49$/$12.60$ & $72.14$/$71.46$ & $75.36$ & $56.83$ & $53.49$ & $53.31\tinymath{\pm 1.48}$\\
\lightcmidrule{1-10}
\, + \gradiendraceab & $\mathit{62.22}$ & $38.82$/$61.80$ & $52.67$ & $\mathit{0.00}$/$\mathit{0.32}$ & $71.97$/$71.29$ & $\mathit{52.71}$ & $60.02$ & $61.45$ & \dab{$\mathit{7.07}$} $\mathit{46.24}\tinymath{\pm \mathit{1.51}}$\\
\, + \gradiendraceaw & $\mathit{67.59}$ & $38.46$/$61.25$ & $55.07$ & $\mathit{38.48}$/$\mathit{8.78}$ & $72.30$/$71.65$ & $\mathit{60.54}$ & $58.05$ & $61.45$ & \dab{$\mathit{3.27}$} $\mathit{50.04}\tinymath{\pm \mathit{1.49}}$\\
\, + \gradiendracebw & $82.29$ & $44.76$/$64.30$ & $63.42$ & $43.32$/$15.49$ & $72.08$/$71.42$ & $\mathit{62.88}$ & $61.70$ & $61.45$ & \uag{$0.41$} $53.72\tinymath{\pm 1.67}$\\
\lightcmidrule{1-10}
\, + \cda & $81.76$ & $46.34$/$66.58$ & \!\!$\mathit{\boldsymbol{\mathit{70.41}}}$ & $\mathit{44.94}$/$\mathit{17.11}$ & $71.88$/$71.22$ & \!\!$\mathbf{77.44}$ & \!\!$\mathit{\boldsymbol{\mathit{67.93}}}$ & \!\!$\mathbf{63.10}$ & \!\!\uag{$\boldsymbol{\mathit{4.17}}$} $\mathit{\boldsymbol{\mathit{57.48}}}\tinymath{\pm \mathit{1.71}}$\\
\, + \dropout & $\mathit{73.53}$ & $45.03$/$64.77$ & $50.32$ & $\mathit{45.78}$/$\mathit{17.10}$ & $72.28$/$71.60$ & $\mathit{60.86}$ & $61.03$ & $57.62$ & \dab{$2.25$} $51.05\tinymath{\pm 1.62}$\\
\, + \inlp & $82.53$ & $38.64$/$60.68$ & $53.43$ & $\mathit{44.66}$/$\mathit{16.56}$ & $72.27$/$71.61$ & $74.99$ & $\mathit{66.26}$ & $55.41$ & \uag{$0.46$} $53.77\tinymath{\pm 1.19}$\\
\, + \sentencedebias & \!\!$\mathbf{83.03}$ & $45.99$/$66.04$ & $57.50$ & \!\!$\mathit{\boldsymbol{\mathit{68.15}}}$/$\mathit{\boldsymbol{\mathit{25.50}}}$ & \!\!$\mathbf{72.33}$/$\mathbf{71.69}$ & $76.86$ & $55.56$ & $55.06$ & \uag{$2.23$} $55.54\tinymath{\pm 1.49}$\\

\midrule
\gpttwo & $65.56$ & $36.86$/$51.74$ & $49.35$ & $58.79$/$13.69$ & $31.64$/$30.93$ & $60.14$ & $62.51$ & $54.47$ & $45.49\tinymath{\pm 1.28}$\\
\lightcmidrule{1-10}
\, + \gradiendraceab & $65.36$ & $36.15$/$51.05$ & $49.65$ & $59.09$/$13.58$ & $31.62$/$30.92$ & $61.78$ & $62.70$ & $53.52$ & \dab{$0.04$} $45.45\tinymath{\pm 1.18}$\\
\, + \gradiendraceaw & $64.74$ & $37.93$/$54.61$ & $50.05$ & $59.06$/$13.57$ & $31.43$/$30.77$ & $62.83$ & $61.31$ & $54.39$ & \uag{$0.43$} $45.92\tinymath{\pm 1.22}$\\
\, + \gradiendracebw & $65.18$ & $37.79$/$52.82$ & \!\!$\mathbf{52.36}$ & $58.97$/$13.01$ & $31.49$/$30.81$ & $63.00$ & $62.18$ & $53.19$ & \uag{$0.48$} $45.97\tinymath{\pm 1.13}$\\
\lightcmidrule{1-10}
\, + \cda & \!\!$\mathbf{66.07}$ & \!\!$\mathbf{40.92}$/$\mathbf{55.24}$ & $52.06$ & \!\!$\mathbf{59.62}$/$\mathbf{15.46}$ & \!\!$\mathbf{31.88}$/$\mathbf{31.19}$ & \!\!$\mathbf{66.13}$ & \!\!$\mathbf{63.33}$ & $56.39$ & \!\!\uag{$\boldsymbol{1.47}$} $\mathbf{46.96}\tinymath{\pm 1.27}$\\
\, + \dropout & $66.02$ & $34.95$/$52.30$ & $49.04$ & $58.82$/$13.86$ & $31.55$/$30.86$ & $62.26$ & $62.66$ & \!\!$\mathbf{58.72}$ & \uag{$0.46$} $45.94\tinymath{\pm 1.45}$\\
\, + \inlp & $65.50$ & $36.37$/$51.14$ & $52.31$ & $59.04$/$13.31$ & $31.59$/$30.87$ & $60.38$ & $62.51$ & $54.80$ & \uag{$0.29$} $45.77\tinymath{\pm 1.22}$\\
\, + \sentencedebias & $65.40$ & $31.49$/$45.23$ & $51.05$ & $59.64$/$12.81$ & $31.48$/$30.75$ & $61.37$ & $61.94$ & $53.20$ & \dab{$0.75$} $44.73\tinymath{\pm 1.26}$\\

\midrule
\llama & $72.96$ & $37.32$/$51.82$ & \!\!$\mathbf{86.06}$ & $0.00$/$0.32$ & \!\!$\mathbf{90.42}$/$\mathbf{89.70}$ & $54.17$ & $50.10$ & $37.58$ & $54.46\tinymath{\pm 2.28}$\\
\lightcmidrule{1-10}
\, + \gradiendraceab & $\mathit{57.24}$ & $31.46$/$37.52$ & $80.96$ & $\mathit{2.11}$/$\mathit{0.32}$ & $\mathit{87.97}$/$\mathit{87.04}$ & $55.92$ & $47.86$ & $51.00$ & \dab{$2.43$} $52.02\tinymath{\pm 2.27}$\\
\, + \gradiendraceaw & $\mathit{58.42}$ & $26.02$/$35.68$ & $79.02$ & \!\!$\mathit{\boldsymbol{\mathit{2.79}}}$/$\mathit{\boldsymbol{\mathit{0.32}}}$ & $\mathit{86.98}$/$\mathit{86.09}$ & \!\!$\mathbf{60.69}$ & $48.10$ & \!\!$\mathit{\boldsymbol{\mathit{70.07}}}$ & \dab{$0.05$} $54.40\tinymath{\pm 2.22}$\\
\, + \gradiendracebw & $\mathit{69.79}$ & $33.49$/$55.30$ & $85.07$ & $0.10$/$0.32$ & $90.22$/$89.43$ & $52.66$ & $49.61$ & $38.52$ & \dab{$0.70$} $53.76\tinymath{\pm 2.03}$\\
\lightcmidrule{1-10}
\, + \inlp & \!\!$\mathbf{73.44}$ & $35.98$/$53.69$ & $86.03$ & $0.00$/$0.32$ & $89.82$/$89.03$ & $58.13$ & $50.10$ & $36.60$ & \uag{$0.38$} $54.84\tinymath{\pm 2.12}$\\
\, + \sentencedebias & $73.08$ & \!\!$\mathbf{38.45}$/$\mathbf{51.77}$ & \!\!$\mathbf{86.06}$ & $0.00$/$0.32$ & $90.30$/$89.55$ & $57.40$ & \!\!$\mathbf{50.42}$ & $36.60$ & \!\!\uag{$\boldsymbol{0.39}$} $\mathbf{54.85}\tinymath{\pm 2.29}$\\

\midrule
\llamai & $75.25$ & $31.58$/$32.13$ & \!\!$\mathbf{78.60}$ & $27.43$/$0.52$ & $85.32$/$84.68$ & $67.31$ & $50.37$ & $62.20$ & $58.07\tinymath{\pm 2.29}$\\
\lightcmidrule{1-10}
\, + \gradiendraceab & $\mathit{37.78}$ & $22.13$/$49.86$ & $\mathit{57.05}$ & \!\!$\mathit{\boldsymbol{\mathit{59.91}}}$/$\mathit{\boldsymbol{\mathit{0.82}}}$ & $\mathit{18.26}$/$\mathit{17.62}$ & $\mathit{47.17}$ & $49.88$ & $63.40$ & \dab{$\mathit{15.62}$} $\mathit{42.45}\tinymath{\pm \mathit{2.13}}$\\
\, + \gradiendraceaw & $\mathit{59.25}$ & $24.31$/$28.79$ & $\mathit{56.89}$ & $30.12$/$2.11$ & $\mathit{45.54}$/$\mathit{44.89}$ & $\mathit{49.58}$ & $48.76$ & $\mathit{37.55}$ & \dab{$\mathit{15.58}$} $\mathit{42.49}\tinymath{\pm \mathit{2.47}}$\\
\, + \gradiendracebw & \!\!$\mathit{\boldsymbol{\mathit{78.19}}}$ & \!\!$\mathit{\boldsymbol{\mathit{45.38}}}$/$\mathit{\boldsymbol{\mathit{64.22}}}$ & $78.17$ & $\mathit{4.28}$/$\mathit{0.32}$ & \!\!$\mathit{\boldsymbol{\mathit{87.30}}}$/$\mathit{\boldsymbol{\mathit{86.52}}}$ & \!\!$\mathbf{71.20}$ & $52.48$ & $45.17$ & \!\!\uag{$\boldsymbol{0.58}$} $\mathbf{58.65}\tinymath{\pm 2.05}$\\
\lightcmidrule{1-10}
\, + \inlp & $76.28$ & $35.27$/$35.79$ & $78.25$ & $29.33$/$1.15$ & $85.31$/$84.65$ & $66.31$ & \!\!$\mathbf{52.60}$ & $57.63$ & \uag{$0.28$} $58.35\tinymath{\pm 2.34}$\\
\, + \sentencedebias & $75.02$ & $33.85$/$33.94$ & $77.26$ & $28.30$/$0.80$ & $85.27$/$84.64$ & $66.06$ & $51.46$ & \!\!$\mathbf{65.06}$ & \uag{$0.46$} $58.53\tinymath{\pm 2.41}$\\
\bottomrule
\end{tabular}

\end{table}

\begin{table}[p]
    \centering
    \tiny
    \setlength{\tabcolsep}{2pt} 
    \caption{\textbf{Religion:} \acrshort{sglue} bootstrapped validation set scores with sub-results for all models. Statistically significant improvements are indicated in \emph{italics}, while the best score for each base model is highlighted in \textbf{bold}. \llama-based results were computed with zero-shot evaluation while all other scores are derived after fine-tuning.}
    \label{tab:eval-superglue-religion}

    \begin{tabular}{lrrrrrrrrr}
\toprule\textbf{Model} & \textbf{BoolQ} & \textbf{CB} & \textbf{COPA} & \textbf{MultiRC} & \textbf{ReCoRD} & \textbf{RTE} & \textbf{WiC} & \textbf{WSC} & \textbf{Average} $\uparrow$\\
\textbf{Metrics} & \textbf{Acc.} & \textbf{F1/Acc.} & \textbf{Acc.} & \textbf{$\text{F1}_\alpha$/EM} & \textbf{F1/EM} & \textbf{Acc.} & \textbf{Acc.} & \textbf{Acc.} & \textbf{}\\
\midrule

\bertbase & $69.16$ & $38.74$/$58.68$ & $62.72$ & $60.12$/$13.23$ & $56.09$/$55.32$ & $61.30$ & $68.67$ & $63.12$ & $51.82\tinymath{\pm 1.67}$\\
\lightcmidrule{1-10}
\, + \gradiendreligioncj & \!\!$\mathbf{70.91}$ & $42.68$/$62.23$ & $61.50$ & $60.22$/$14.43$ & $55.74$/$55.01$ & $64.66$ & \!\!$\mathbf{70.13}$ & \!\!$\mathbf{63.40}$ & \uag{$1.17$} $53.00\tinymath{\pm 1.89}$\\
\, + \gradiendreligioncm & $70.82$ & $42.15$/$61.64$ & $60.80$ & $59.37$/$14.18$ & $55.74$/$54.98$ & \!\!$\mathbf{65.27}$ & $67.93$ & \!\!$\mathbf{63.40}$ & \uag{$0.71$} $52.54\tinymath{\pm 1.88}$\\
\, + \gradiendreligionjm & $70.72$ & $43.19$/$62.83$ & $60.84$ & \!\!$\mathbf{60.79}$/$\mathbf{13.93}$ & $55.04$/$54.29$ & $64.43$ & $69.72$ & \!\!$\mathbf{63.40}$ & \uag{$0.89$} $52.71\tinymath{\pm 1.89}$\\
\lightcmidrule{1-10}
\, + \cda & $70.41$ & \!\!$\mathbf{47.79}$/$\mathbf{69.51}$ & $60.50$ & $59.24$/$14.32$ & $55.58$/$54.85$ & $63.96$ & $67.81$ & \!\!$\mathbf{63.40}$ & \!\!\uag{$\boldsymbol{1.26}$} $\mathbf{53.08}\tinymath{\pm 1.83}$\\
\, + \dropout & $68.53$ & $47.39$/$68.99$ & $55.56$ & $59.20$/$12.94$ & $55.00$/$54.23$ & $61.74$ & $65.15$ & $62.77$ & \dab{$0.34$} $51.48\tinymath{\pm 1.72}$\\
\, + \inlp & $68.85$ & $28.43$/$52.77$ & $62.08$ & $60.62$/$13.66$ & \!\!$\mathbf{56.09}$/$\mathbf{55.33}$ & $62.38$ & $67.96$ & $61.29$ & \dab{$1.13$} $50.70\tinymath{\pm 1.58}$\\
\, + \sentencedebias & $69.02$ & $27.47$/$52.19$ & \!\!$\mathbf{64.40}$ & $60.29$/$13.53$ & $55.99$/$55.22$ & $62.74$ & $68.68$ & $63.06$ & \dab{$0.57$} $51.25\tinymath{\pm 1.54}$\\

\midrule
\bertlarge & $70.32$ & $42.86$/$62.97$ & $61.46$ & $61.49$/$15.19$ & $61.70$/$61.04$ & $67.68$ & \!\!$\mathbf{70.82}$ & $62.09$ & $53.74\tinymath{\pm 1.62}$\\
\lightcmidrule{1-10}
\, + \gradiendreligioncj & $72.20$ & $46.17$/$66.51$ & $63.74$ & $61.06$/$16.34$ & $61.81$/$61.16$ & $67.67$ & $70.49$ & \!\!$\mathbf{63.40}$ & \uag{$0.78$} $54.52\tinymath{\pm 1.84}$\\
\, + \gradiendreligioncm & $72.22$ & $46.05$/$66.44$ & $62.73$ & $61.88$/$16.92$ & $61.81$/$61.13$ & \!\!$\mathbf{67.80}$ & $70.06$ & \!\!$\mathbf{63.40}$ & \uag{$0.83$} $54.58\tinymath{\pm 1.86}$\\
\, + \gradiendreligionjm & $72.03$ & $45.74$/$65.91$ & $64.83$ & \!\!$\mathbf{62.81}$/$\mathbf{16.24}$ & $61.95$/$61.27$ & $66.71$ & $70.25$ & \!\!$\mathbf{63.40}$ & \!\!\uag{$\boldsymbol{1.08}$} $\mathbf{54.82}\tinymath{\pm 1.85}$\\
\lightcmidrule{1-10}
\, + \cda & \!\!$\mathbf{72.71}$ & \!\!$\mathbf{46.78}$/$\mathbf{67.19}$ & \!\!$\mathbf{65.90}$ & $61.05$/$15.03$ & $61.79$/$61.11$ & $61.75$ & $69.30$ & $63.10$ & \uag{$0.52$} $54.27\tinymath{\pm 1.79}$\\
\, + \dropout & $71.12$ & $45.18$/$65.27$ & $53.62$ & $62.24$/$16.01$ & \!\!$\mathbf{62.09}$/$\mathbf{61.37}$ & $64.72$ & $67.99$ & \!\!$\mathbf{63.40}$ & \dab{$0.52$} $53.22\tinymath{\pm 1.68}$\\
\, + \inlp & $70.80$ & $38.31$/$61.17$ & $60.08$ & $61.26$/$16.30$ & $61.69$/$61.02$ & $67.32$ & $70.44$ & $63.12$ & \dab{$0.54$} $53.21\tinymath{\pm 1.55}$\\
\, + \sentencedebias & $70.90$ & $43.74$/$63.55$ & $62.14$ & $62.66$/$16.15$ & $61.44$/$60.73$ & $67.46$ & $70.24$ & $58.33$ & \dab{$0.07$} $53.67\tinymath{\pm 1.64}$\\

\midrule
\distilbert & $69.75$ & $45.62$/$66.55$ & $53.39$ & $57.58$/$12.21$ & $49.09$/$48.27$ & $55.18$ & $62.10$ & \!\!$\mathbf{63.40}$ & $49.69\tinymath{\pm 1.65}$\\
\lightcmidrule{1-10}
\, + \gradiendreligioncj & $69.75$ & $45.64$/$66.52$ & $53.38$ & $57.62$/$12.09$ & $49.14$/$48.36$ & $55.73$ & $61.50$ & \!\!$\mathbf{63.40}$ & \uag{$0.06$} $49.75\tinymath{\pm 1.69}$\\
\, + \gradiendreligioncm & $69.75$ & $44.73$/$65.35$ & \!\!$\mathbf{55.35}$ & $57.17$/$11.97$ & $49.14$/$48.35$ & $55.93$ & $61.82$ & \!\!$\mathbf{63.40}$ & \uag{$0.07$} $49.76\tinymath{\pm 1.69}$\\
\, + \gradiendreligionjm & $69.72$ & $45.64$/$66.52$ & $54.41$ & $58.19$/$12.61$ & $49.11$/$48.32$ & $55.94$ & $61.64$ & \!\!$\mathbf{63.40}$ & \uag{$0.23$} $49.91\tinymath{\pm 1.68}$\\
\lightcmidrule{1-10}
\, + \cda & $69.06$ & \!\!$\mathbf{48.37}$/$\mathbf{69.57}$ & $53.40$ & $58.92$/$13.45$ & $48.99$/$48.13$ & \!\!$\mathbf{60.49}$ & \!\!$\mathbf{63.25}$ & \!\!$\mathbf{63.40}$ & \!\!\uag{$\boldsymbol{0.80}$} $\mathbf{50.49}\tinymath{\pm 1.80}$\\
\, + \dropout & $69.21$ & $46.13$/$66.49$ & $54.78$ & \!\!$\mathbf{59.40}$/$\mathbf{13.05}$ & \!\!$\mathbf{49.77}$/$\mathbf{48.97}$ & $60.45$ & $62.62$ & \!\!$\mathbf{63.40}$ & \uag{$0.58$} $50.27\tinymath{\pm 1.75}$\\
\, + \inlp & $69.42$ & $44.32$/$64.73$ & $55.28$ & $57.18$/$11.72$ & $49.02$/$48.21$ & $57.52$ & $62.55$ & \!\!$\mathbf{63.40}$ & \dab{$0.05$} $49.64\tinymath{\pm 1.65}$\\
\, + \sentencedebias & \!\!$\mathbf{70.11}$ & $45.62$/$66.55$ & $54.73$ & $58.43$/$12.24$ & $49.02$/$48.23$ & $55.66$ & $61.89$ & \!\!$\mathbf{63.40}$ & \uag{$0.12$} $49.81\tinymath{\pm 1.64}$\\

\midrule
\roberta & \!\!$\mathbf{82.01}$ & $46.41$/$66.62$ & $56.70$ & $42.49$/$12.60$ & $72.14$/$71.46$ & $75.36$ & $56.83$ & $53.49$ & $53.31\tinymath{\pm 1.48}$\\
\lightcmidrule{1-10}
\, + \gradiendreligioncj & $\mathit{74.98}$ & $43.45$/$62.51$ & $55.70$ & $\mathit{45.24}$/$\mathit{17.07}$ & $72.31$/$71.64$ & $67.72$ & $59.24$ & $61.76$ & \dab{$0.65$} $52.66\tinymath{\pm 1.66}$\\
\, + \gradiendreligioncm & $81.92$ & $45.54$/$65.44$ & $57.63$ & \!\!$\mathit{\boldsymbol{\mathit{67.66}}}$/$\mathit{\boldsymbol{\mathit{22.48}}}$ & \!\!$\mathbf{72.57}$/$\mathbf{71.90}$ & \!\!$\mathbf{77.31}$ & $\mathit{63.17}$ & $62.07$ & \uag{$\mathit{3.35}$} $\mathit{56.66}\tinymath{\pm \mathit{1.64}}$\\
\, + \gradiendreligionjm & $\mathit{79.04}$ & $40.34$/$58.98$ & $50.38$ & $\mathit{22.29}$/$\mathit{9.14}$ & $72.22$/$71.60$ & $\mathit{62.92}$ & $\mathit{65.91}$ & $61.76$ & \dab{$2.19$} $51.12\tinymath{\pm 1.65}$\\
\lightcmidrule{1-10}
\, + \cda & $\mathit{75.96}$ & \!\!$\mathbf{48.02}$/$\mathbf{68.95}$ & \!\!$\mathit{\boldsymbol{\mathit{70.99}}}$ & $\mathit{65.41}$/$\mathit{22.63}$ & $71.99$/$71.33$ & $74.79$ & \!\!$\mathit{\boldsymbol{\mathit{67.57}}}$ & \!\!$\mathbf{62.42}$ & \!\!\uag{$\boldsymbol{\mathit{4.71}}$} $\mathit{\boldsymbol{\mathit{58.02}}}\tinymath{\pm \mathit{1.68}}$\\
\, + \dropout & $\mathit{73.53}$ & $45.03$/$64.77$ & $50.32$ & $\mathit{45.78}$/$\mathit{17.10}$ & $72.28$/$71.60$ & $\mathit{60.86}$ & $61.03$ & $57.62$ & \dab{$2.25$} $51.05\tinymath{\pm 1.62}$\\
\, + \inlp & $\mathit{75.72}$ & $45.99$/$66.04$ & $54.17$ & $\mathit{66.07}$/$\mathit{22.54}$ & $71.95$/$71.32$ & $77.07$ & $\mathit{64.48}$ & $53.49$ & \uag{$1.64$} $54.95\tinymath{\pm 1.51}$\\
\, + \sentencedebias & $\mathit{75.28}$ & $46.13$/$67.20$ & $57.48$ & $\mathit{45.58}$/$\mathit{14.63}$ & $72.14$/$71.47$ & $68.39$ & $\mathit{64.98}$ & $53.49$ & \dab{$0.60$} $52.71\tinymath{\pm 1.21}$\\

\midrule
\gpttwo & $65.56$ & $36.86$/$51.74$ & $49.35$ & $58.79$/$13.69$ & \!\!$\mathbf{31.64}$/$\mathbf{30.93}$ & $60.14$ & $62.51$ & $54.47$ & $45.49\tinymath{\pm 1.28}$\\
\lightcmidrule{1-10}
\, + \gradiendreligioncj & $66.02$ & $36.89$/$51.69$ & $52.39$ & \!\!$\mathbf{61.03}$/$\mathbf{13.64}$ & $31.43$/$30.75$ & $64.63$ & $63.42$ & $55.17$ & \uag{$1.18$} $46.67\tinymath{\pm 1.11}$\\
\, + \gradiendreligioncm & $65.37$ & $42.78$/$55.85$ & $52.66$ & $60.48$/$13.16$ & $31.63$/$30.94$ & $63.96$ & $61.88$ & $57.05$ & \uag{$1.54$} $47.02\tinymath{\pm 1.33}$\\
\, + \gradiendreligionjm & \!\!$\mathbf{66.31}$ & $37.60$/$51.10$ & $51.38$ & $60.12$/$14.32$ & $31.32$/$30.63$ & \!\!$\mathbf{65.84}$ & $63.53$ & $55.42$ & \uag{$1.15$} $46.64\tinymath{\pm 1.26}$\\
\lightcmidrule{1-10}
\, + \cda & $66.16$ & \!\!$\mathbf{46.73}$/$\mathbf{58.76}$ & \!\!$\mathbf{54.91}$ & $59.52$/$14.34$ & $31.54$/$30.85$ & $65.67$ & \!\!$\mathbf{63.69}$ & \!\!$\mathbf{59.67}$ & \!\!\uag{$\boldsymbol{2.61}$} $\mathbf{48.10}\tinymath{\pm 1.42}$\\
\, + \dropout & $66.02$ & $34.95$/$52.30$ & $49.04$ & $58.82$/$13.86$ & $31.55$/$30.86$ & $62.26$ & $62.66$ & $58.72$ & \uag{$0.46$} $45.94\tinymath{\pm 1.45}$\\
\, + \inlp & $65.52$ & $36.77$/$51.74$ & $52.66$ & $58.68$/$13.92$ & $31.61$/$30.89$ & $60.26$ & $62.36$ & $54.16$ & \uag{$0.29$} $45.77\tinymath{\pm 1.21}$\\
\, + \sentencedebias & $65.35$ & $42.15$/$57.65$ & $53.99$ & $57.11$/$11.91$ & $31.62$/$30.88$ & $62.54$ & $61.72$ & $53.23$ & \uag{$0.80$} $46.29\tinymath{\pm 1.28}$\\

\midrule
\llama & $72.96$ & $37.32$/$51.82$ & $86.06$ & $0.00$/$0.32$ & \!\!$\mathbf{90.42}$/$\mathbf{89.70}$ & $54.17$ & $50.10$ & $37.58$ & $54.46\tinymath{\pm 2.28}$\\
\lightcmidrule{1-10}
\, + \gradiendreligioncj & $\mathit{61.19}$ & \!\!$\mathbf{38.95}$/$\mathbf{60.67}$ & $81.00$ & \!\!$\mathbf{0.37}$/$\mathbf{0.53}$ & $89.38$/$88.55$ & $52.32$ & $50.13$ & $36.60$ & \dab{$1.90$} $52.56\tinymath{\pm 2.17}$\\
\, + \gradiendreligioncm & $74.34$ & $24.92$/$44.53$ & $80.13$ & $0.10$/$0.42$ & $89.87$/$89.16$ & $51.18$ & $50.12$ & $36.60$ & \dab{$2.35$} $52.11\tinymath{\pm 2.12}$\\
\, + \gradiendreligionjm & $\mathit{68.66}$ & $30.37$/$42.72$ & $79.06$ & $0.10$/$0.42$ & $\mathit{88.71}$/$\mathit{87.87}$ & \!\!$\mathbf{59.61}$ & $47.75$ & \!\!$\mathbf{40.54}$ & \dab{$1.87$} $52.59\tinymath{\pm 2.17}$\\
\lightcmidrule{1-10}
\, + \inlp & \!\!$\mathbf{74.54}$ & $39.56$/$51.89$ & $83.10$ & $0.00$/$0.32$ & $90.31$/$89.58$ & $56.66$ & \!\!$\mathbf{50.59}$ & $36.60$ & \!\!\uag{$\boldsymbol{0.21}$} $\mathbf{54.66}\tinymath{\pm 2.30}$\\
\, + \sentencedebias & $72.80$ & $35.03$/$50.01$ & \!\!$\mathbf{86.11}$ & $0.00$/$0.32$ & $90.12$/$89.40$ & $55.25$ & $50.10$ & $37.58$ & \dab{$0.17$} $54.29\tinymath{\pm 2.28}$\\

\midrule
\llamai & $75.25$ & $31.58$/$32.13$ & $78.60$ & $27.43$/$0.52$ & $85.32$/$84.68$ & $67.31$ & $50.37$ & $62.20$ & $58.07\tinymath{\pm 2.29}$\\
\lightcmidrule{1-10}
\, + \gradiendreligioncj & \!\!$\mathit{\boldsymbol{\mathit{78.50}}}$ & \!\!$\mathit{\boldsymbol{\mathit{52.18}}}$/$\mathit{\boldsymbol{\mathit{73.15}}}$ & $80.10$ & $\mathit{7.82}$/$\mathit{0.21}$ & \!\!$\mathit{\boldsymbol{\mathit{87.42}}}$/$\mathit{\boldsymbol{\mathit{86.64}}}$ & \!\!$\mathbf{74.80}$ & $51.37$ & $\mathit{42.32}$ & \!\!\uag{$\boldsymbol{2.03}$} $\mathbf{60.10}\tinymath{\pm 1.95}$\\
\, + \gradiendreligioncm & $75.56$ & $37.38$/$55.66$ & $70.97$ & $\mathit{12.57}$/$\mathit{0.10}$ & $\mathit{73.80}$/$\mathit{73.05}$ & $69.67$ & $50.12$ & $\mathit{37.53}$ & \dab{$4.30$} $53.77\tinymath{\pm 2.17}$\\
\, + \gradiendreligionjm & $74.37$ & $\mathit{43.66}$/$\mathit{64.14}$ & \!\!$\mathbf{82.28}$ & $\mathit{7.93}$/$\mathit{0.63}$ & $84.72$/$84.03$ & $69.01$ & $51.05$ & $58.68$ & \uag{$1.67$} $59.74\tinymath{\pm 2.18}$\\
\lightcmidrule{1-10}
\, + \inlp & $75.36$ & $33.47$/$33.95$ & $77.26$ & $\mathit{22.80}$/$\mathit{0.42}$ & $85.55$/$84.91$ & $66.31$ & \!\!$\mathbf{52.43}$ & $61.44$ & \dab{$0.15$} $57.92\tinymath{\pm 2.40}$\\
\, + \sentencedebias & $75.06$ & $33.85$/$33.94$ & $77.26$ & \!\!$\mathbf{28.99}$/$\mathbf{0.64}$ & $85.35$/$84.72$ & $65.93$ & $51.82$ & \!\!$\mathbf{64.40}$ & \uag{$0.46$} $58.53\tinymath{\pm 2.41}$\\
\bottomrule
\end{tabular}

\end{table}


\begin{table*}[p]
    \centering
    \caption{\textbf{Gender:} SEAT bootstrapped effect sizes for encoder-only models. Statistically significant improvements are indicated in \emph{italics}, while the best score for each base model is highlighted in \textbf{bold}.}\label{tab:eval:seat}
\fontsize{5.5pt}{6pt}\selectfont
\begin{tabular}{@{\hskip 1pt}l@{\hskip 6pt}r@{\hskip 6pt}r@{\hskip 6pt}r@{\hskip 6pt}r@{\hskip 6pt}r@{\hskip 6pt}r@{\hskip 6pt}r@{\hskip 1pt}}
\toprule\textbf{Model} & \textbf{SEAT-6} \bestatzerotiny & \textbf{SEAT-6b} \bestatzerotiny & \textbf{SEAT-7} \bestatzerotiny & \textbf{SEAT-7b} \bestatzerotiny & \textbf{SEAT-8} \bestatzerotiny & \textbf{SEAT-8b} \bestatzerotiny & \textbf{Absolute Average} $\downarrow$\\
\midrule
\bertbase & $0.84\tinymath{\pm 0.29}$ & $0.20\tinymath{\pm 0.25}$ & $0.57\tinymath{\pm 0.62}$ & $1.03\tinymath{\pm 0.49}$ & $0.54\tinymath{\pm 0.48}$ & $0.45\tinymath{\pm 0.52}$ &  $0.61\tinymath{\pm 0.29}$\\
\lightcmidrule{1-8}
\, + \gradiendbpi & $0.59\tinymath{\pm 0.21}$ & \!\!\! \!\!$\mathbf{-0.03}\tinymath{\pm 0.15}$ & $0.18\tinymath{\pm 0.57}$ & $0.92\tinymath{\pm 0.45}$ & $0.53\tinymath{\pm 0.46}$ & $0.64\tinymath{\pm 0.36}$ & \da{$0.10$} $0.51\tinymath{\pm 0.19}$\\
\, + \gradiendfpi & $0.84\tinymath{\pm 0.46}$ & $0.27\tinymath{\pm 0.40}$ & $1.05\tinymath{\pm 0.34}$ & $1.18\tinymath{\pm 0.31}$ & $0.72\tinymath{\pm 0.32}$ & $0.68\tinymath{\pm 0.34}$ & \ua{$0.19$} $0.79\tinymath{\pm 0.24}$\\
\, + \gradiendmpi & $0.95\tinymath{\pm 0.27}$ & $0.43\tinymath{\pm 0.32}$ & $0.24\tinymath{\pm 0.54}$ & $0.33\tinymath{\pm 0.51}$ & \!\!\! $-0.05\tinymath{\pm 0.46}$ & \!\!\! $-0.35\tinymath{\pm 0.40}$ & \da{$0.18$} $0.43\tinymath{\pm 0.16}$\\
\lightcmidrule{1-8}
\, + \cda & $0.48\tinymath{\pm 0.18}$ & \!\!\! $-0.05\tinymath{\pm 0.13}$ & $0.07\tinymath{\pm 0.61}$ & $0.74\tinymath{\pm 0.50}$ & $0.38\tinymath{\pm 0.50}$ & $0.47\tinymath{\pm 0.44}$ & \da{$0.21$} $0.40\tinymath{\pm 0.20}$\\
\, + \dropout & $\mathit{0.16}\tinymath{\pm \mathit{0.37}}$ & $0.17\tinymath{\pm 0.27}$ & $0.24\tinymath{\pm 0.60}$ & $0.81\tinymath{\pm 0.42}$ & $0.67\tinymath{\pm 0.47}$ & $0.53\tinymath{\pm 0.43}$ & \da{$0.16$} $0.45\tinymath{\pm 0.25}$\\
\bottomrule
\end{tabular}
\end{table*}

\begin{table*}[p]
    \centering
    \vspace{70pt}
    \caption{\textbf{Gender:} SEAT bootstrapped effect sizes for decoder-only models. Statistically significant improvements are indicated in \emph{italics}, while the best score for each base model is highlighted in \textbf{bold}.}\label{tab:eval:seat-gpt}
\fontsize{5.5pt}{6pt}\selectfont
\begin{tabular}{@{\hskip 1pt}l@{\hskip 6pt}r@{\hskip 6pt}r@{\hskip 6pt}r@{\hskip 6pt}r@{\hskip 6pt}r@{\hskip 6pt}r@{\hskip 6pt}r@{\hskip 1pt}}
\toprule\textbf{Model} & \textbf{SEAT-6} \bestatzerotiny & \textbf{SEAT-6b} \bestatzerotiny & \textbf{SEAT-7} \bestatzerotiny & \textbf{SEAT-7b} \bestatzerotiny & \textbf{SEAT-8} \bestatzerotiny & \textbf{SEAT-8b} \bestatzerotiny & \textbf{Absolute Average} $\downarrow$\\
\midrule
\gpttwo & $0.27\tinymath{\pm 0.61}$ & $0.03\tinymath{\pm 0.47}$ & $0.03\tinymath{\pm 0.39}$ & $0.14\tinymath{\pm 0.61}$ & \!\!\! $-0.07\tinymath{\pm 0.76}$ & \!\!\! $-0.14\tinymath{\pm 0.72}$ &  $0.24\tinymath{\pm 0.29}$\\
\lightcmidrule{1-8}
\, + \gradiendbpi & $0.34\tinymath{\pm 0.56}$ & $0.02\tinymath{\pm 0.50}$ & $0.42\tinymath{\pm 0.61}$ & $0.51\tinymath{\pm 0.63}$ & \!\!\! $-0.05\tinymath{\pm 0.63}$ & $0.18\tinymath{\pm 0.71}$ & \ua{$0.09$} $0.33\tinymath{\pm 0.39}$\\
\, + \gradiendfpi & \!\!$\mathbf{0.20}\tinymath{\pm 0.57}$ & \!\!$\mathbf{0.00}\tinymath{\pm 0.41}$ & $0.14\tinymath{\pm 0.54}$ & $0.39\tinymath{\pm 0.81}$ & $0.05\tinymath{\pm 0.67}$ & $0.19\tinymath{\pm 0.65}$ & \ua{$0.01$} $0.25\tinymath{\pm 0.39}$\\
\, + \gradiendmpi & $0.31\tinymath{\pm 0.60}$ & \!\!\! $-0.02\tinymath{\pm 0.55}$ & $0.42\tinymath{\pm 0.59}$ & $0.48\tinymath{\pm 0.60}$ & \!\!\! $-0.09\tinymath{\pm 0.65}$ & $0.11\tinymath{\pm 0.69}$ & \ua{$0.09$} $0.33\tinymath{\pm 0.36}$\\
\lightcmidrule{1-8}
\, + \cda & $0.34\tinymath{\pm 0.45}$ & $0.03\tinymath{\pm 0.28}$ & $0.14\tinymath{\pm 0.55}$ & $0.43\tinymath{\pm 0.70}$ & $0.28\tinymath{\pm 0.90}$ & $0.02\tinymath{\pm 0.77}$ & \ua{$0.07$} $0.31\tinymath{\pm 0.29}$\\
\, + \dropout & $0.33\tinymath{\pm 0.47}$ & $0.03\tinymath{\pm 0.34}$ & $0.66\tinymath{\pm 0.73}$ & $0.85\tinymath{\pm 0.59}$ & $0.45\tinymath{\pm 0.61}$ & $0.22\tinymath{\pm 0.79}$ & \ua{$0.24$} $0.48\tinymath{\pm 0.24}$\\
\bottomrule
\end{tabular}
    \vspace{70pt}
\end{table*}

\begin{table}[]
    \centering
    \tiny
    \setlength{\tabcolsep}{2pt} 
\caption{\textbf{Race:} SEAT bootstrapped effect sizes for all models. Statistically significant improvements are indicated in \emph{italics}, while the best score for each base model is highlighted in \textbf{bold}.}
    \label{tab:seat-race}
    \begin{tabular}{lrrrrrrrr}
\toprule\textbf{Model} & \tiny{\textbf{ABW1}} \bestatzerotiny & \tiny{\textbf{ABW2}} \bestatzerotiny & \tiny{\textbf{SEAT-3}} \bestatzerotiny & \textbf{SEAT-3b} \bestatzerotiny & \textbf{SEAT-4} \bestatzerotiny & \textbf{SEAT-5} \bestatzerotiny & \textbf{SEAT-5b} \bestatzerotiny & \textbf{Absolute Average} \bestatzerotiny\\
\midrule
\bertbase & $0.53\tinymath{\pm 0.78}$ & $0.50\tinymath{\pm 0.23}$ & $0.72\tinymath{\pm 0.36}$ & \!\!\! $-0.15\tinymath{\pm 0.48}$ & $0.68\tinymath{\pm 0.48}$ & $0.77\tinymath{\pm 0.56}$ & $0.04\tinymath{\pm 0.33}$ & $51.64\tinymath{\pm 25.85}$\\
\lightcmidrule{1-9}
\, + \gradiendraceab & $0.82\tinymath{\pm 0.65}$ & $0.40\tinymath{\pm 0.28}$ & $0.83\tinymath{\pm 0.37}$ & \!\!$\mathbf{0.02}\tinymath{\pm 0.49}$ & $0.85\tinymath{\pm 0.54}$ & $0.97\tinymath{\pm 0.53}$ & $0.06\tinymath{\pm 0.29}$ & \ua{$8.31$} $59.96\tinymath{\pm 27.62}$\\
\, + \gradiendraceaw & $1.08\tinymath{\pm 0.44}$ & \!\!$\mathbf{0.29}\tinymath{\pm 0.28}$ & $0.76\tinymath{\pm 0.40}$ & \!\!\! $-0.09\tinymath{\pm 0.50}$ & $0.74\tinymath{\pm 0.56}$ & $0.97\tinymath{\pm 0.50}$ & $0.06\tinymath{\pm 0.32}$ & \ua{$8.29$} $59.94\tinymath{\pm 24.04}$\\
\, + \gradiendracebw & $0.56\tinymath{\pm 0.73}$ & $0.47\tinymath{\pm 0.23}$ & $0.70\tinymath{\pm 0.37}$ & \!\!\! $-0.20\tinymath{\pm 0.49}$ & $0.65\tinymath{\pm 0.49}$ & $0.75\tinymath{\pm 0.55}$ & \!\!$\mathbf{0.01}\tinymath{\pm 0.34}$ & \da{$0.62$} $51.02\tinymath{\pm 25.56}$\\
\lightcmidrule{1-9}
\, + \cda & $0.44\tinymath{\pm 0.46}$ & $0.35\tinymath{\pm 0.16}$ & \!\!$\mathbf{0.13}\tinymath{\pm 0.46}$ & \!\!\! $-0.19\tinymath{\pm 0.34}$ & \!\!\! \!\!$\mathbf{-0.04}\tinymath{\pm 0.51}$ & \!\!$\mathbf{0.22}\tinymath{\pm 0.39}$ & $0.03\tinymath{\pm 0.30}$ & \!\!\da{$\boldsymbol{25.74}$} $\mathbf{25.90}\tinymath{\pm 12.84}$\\
\bottomrule
\end{tabular}
\end{table}

\begin{table}[]
    \centering
     \tiny
   \caption{\textbf{Religion:} SEAT bootstrapped effect sizes for all models. Statistically significant improvements are indicated in \emph{italics}, while the best score for each base model is highlighted in \textbf{bold}.}
    \label{tab:seat-religion}
   \begin{tabular}{lrrrrr}
\toprule\textbf{Model} & \textbf{SEAT-REL1} \bestatzerotiny & \textbf{SEAT-REL1b} \bestatzerotiny & \textbf{SEAT-REL2} \bestatzerotiny & \textbf{SEAT-REL2b} \bestatzerotiny & \textbf{Absolute Average} \bestatzerotiny\\
\midrule
\bertbase & $0.18\tinymath{\pm 0.40}$ & \!\!\! $-0.10\tinymath{\pm 0.31}$ & $0.73\tinymath{\pm 0.40}$ & $0.43\tinymath{\pm 0.40}$ & $38.29\tinymath{\pm 21.39}$\\
\lightcmidrule{1-6}
\, + \gradiendreligioncj & $0.22\tinymath{\pm 0.41}$ & \!\!\! $-0.15\tinymath{\pm 0.32}$ & $0.81\tinymath{\pm 0.38}$ & $0.44\tinymath{\pm 0.40}$ & \ua{$3.79$} $42.07\tinymath{\pm 20.57}$\\
\, + \gradiendreligioncm & \!\!\! $-0.29\tinymath{\pm 0.42}$ & \!\!\! $-0.46\tinymath{\pm 0.28}$ & $0.72\tinymath{\pm 0.41}$ & $0.37\tinymath{\pm 0.42}$ & \ua{$8.48$} $46.77\tinymath{\pm 17.84}$\\
\, + \gradiendreligionjm & $0.61\tinymath{\pm 0.31}$ & $0.12\tinymath{\pm 0.30}$ & $0.82\tinymath{\pm 0.35}$ & $0.45\tinymath{\pm 0.37}$ & \ua{$12.74$} $51.03\tinymath{\pm 23.95}$\\
\lightcmidrule{1-6}
\, + \cda & \!\!\! $-0.16\tinymath{\pm 0.25}$ & \!\!\! $-0.08\tinymath{\pm 0.22}$ & \!\!$\mathbf{0.21}\tinymath{\pm 0.35}$ & \!\!\! \!\!$\mathbf{-0.00}\tinymath{\pm 0.27}$ & \!\!\da{$\boldsymbol{22.76}$} $\mathbf{15.53}\tinymath{\pm 9.86}$\\
\, + \dropout & \!\!\! $-0.21\tinymath{\pm 0.43}$ & \!\!\! $-0.20\tinymath{\pm 0.34}$ & $0.66\tinymath{\pm 0.36}$ & $0.39\tinymath{\pm 0.37}$ & \da{$0.10$} $38.18\tinymath{\pm 15.73}$\\
\, + \inlp & \!\!\! \!\!$\mathbf{-0.05}\tinymath{\pm 0.40}$ & \!\!\! $-0.32\tinymath{\pm 0.29}$ & $0.61\tinymath{\pm 0.32}$ & $0.36\tinymath{\pm 0.35}$ & \da{$1.86$} $36.43\tinymath{\pm 14.46}$\\
\bottomrule
\end{tabular}
\end{table}


\begin{table*}[p]
    \centering
    \small
        \caption{Example predictions for \bertbase\ and its gender \gradiend\ models. Predictions of the \gradiend\ models that were not retrieved by the base model as one of the top 10 results, are highlighted in \textbf{bold}.}
    \label{tab:example-predictions}
\begin{tabular}{rllll}
    \toprule
    \textbf{Index} & \textbf{\bertbase} & \, + \textbf{\gradiendbpi} & \, + \textbf{\gradiendfpi} & \, + \textbf{\gradiendmpi} \\
    \midrule
    \multicolumn{5}{c}{The woman worked as a [MASK].}
\\\midrule
1 & nurse (16.9\%) & nurse (10.2\%) & waitress (34.4\%) & nurse (13.6\%)\\
2 & waitress (15.0\%) & waitress (7.1\%) & nurse (22.0\%) & waitress (9.5\%)\\
3 & maid (5.6\%) & \textbf{waiter (3.6\%)} & maid (9.4\%) & maid (4.6\%)\\
4 & housekeeper (4.8\%) & maid (3.2\%) & housekeeper (9.0\%) & housekeeper (4.5\%)\\
5 & cook (3.0\%) & \textbf{doctor (2.4\%)} & model (3.5\%) & cook (3.3\%)\\
6 & secretary (2.9\%) & cook (2.3\%) & secretary (2.8\%) & secretary (3.1\%)\\
\bottomrule
\end{tabular}

\end{table*}

\begin{table*}[p]
    \centering
        \caption{Example predictions for \bertlarge\ and its gender \gradiend\ models. Predictions of the \gradiend\ models that were not retrieved by the base model as one of the top 10 results, are highlighted in \textbf{bold}.}
    \label{tab:example-predictions-bert-large}
\begin{tabular}{rllll}
    \toprule
    \small
    \textbf{Index} & \textbf{\bertlarge} & \, + \textbf{\gradiendbpi} & \, + \textbf{\gradiendfpi} & \, + \textbf{\gradiendmpi} \\
    \midrule
    \multicolumn{5}{c}{The woman worked as a [MASK].}
\\\midrule
 1 & nurse (25.7\%) & nurse (5.0\%) & nurse (41.6\%) & nurse (34.4\%)\\
2 & waitress (16.7\%) & teacher (4.4\%) & waitress (19.7\%) & waitress (24.5\%)\\
3 & teacher (4.6\%) & doctor (3.5\%) & secretary (11.8\%) & secretary (10.3\%)\\
4 & secretary (3.6\%) & \textbf{cop (2.5\%)} & librarian (5.8\%) & librarian (3.8\%)\\
5 & maid (3.3\%) & waitress (2.1\%) & \textbf{cleaner (2.8\%)} & maid (3.1\%)\\
6 & prostitute (3.0\%) & \textbf{model (1.6\%)} & maid (2.3\%) & housekeeper (2.6\%)\\
\bottomrule
\end{tabular}
\end{table*}

\begin{table*}[p]
    \centering
    \small
        \caption{Example predictions for \distilbert\ and its gender \gradiend\ models. Predictions of the \gradiend\ models that were not retrieved by the base model as one of the top 10 results, are highlighted in \textbf{bold}.}
    \label{tab:example-predictions-distilbert}
\begin{tabular}{rllll}
    \toprule
    \textbf{Index} & \textbf{\distilbert} & \, + \textbf{\gradiendbpi} & \, + \textbf{\gradiendfpi} & \, + \textbf{\gradiendmpi} \\
    \midrule
    \multicolumn{5}{c}{The woman worked as a [MASK].}
\\\midrule
 1 & nurse (25.0\%) & nurse (25.8\%) & nurse (40.8\%) & nurse (40.7\%)\\
2 & maid (8.1\%) & maid (8.5\%) & maid (21.5\%) & maid (18.5\%)\\
3 & prostitute (7.5\%) & prostitute (7.7\%) & waitress (13.9\%) & waitress (11.7\%)\\
4 & waitress (7.0\%) & waitress (7.4\%) & prostitute (6.3\%) & prostitute (7.7\%)\\
5 & teacher (5.5\%) & teacher (5.3\%) & housekeeper (4.3\%) & housekeeper (4.9\%)\\
6 & housekeeper (4.4\%) & housekeeper (4.6\%) & \textbf{woman (3.1\%)} & \textbf{woman (2.2\%)}\\
\bottomrule
\end{tabular}

\end{table*}

\begin{table*}[p]
    \centering
    \small
        \caption{Example predictions for \roberta\ and its gender \gradiend\ models. Predictions of the \gradiend\ models that were not retrieved by the base model as one of the top 10 results, are highlighted in \textbf{bold}.}
    \label{tab:example-predictions-roberta}
\begin{tabular}{rllll}
    \toprule
    \textbf{Index} & \textbf{\roberta} & \, + \textbf{\gradiendbpi} & \, + \textbf{\gradiendfpi} & \, + \textbf{\gradiendmpi} \\
   \midrule
    \multicolumn{5}{c}{The woman worked as a [MASK].}
\\\midrule
 1 &  nurse (28.2\%) &  nurse (19.9\%) &  waitress (90.9\%) &  nurse (31.2\%)\\
2 &  waitress (10.9\%) &  teacher (13.2\%) &  secretary (3.1\%) &  waitress (15.4\%)\\
3 &  teacher (10.4\%) &  waitress (4.9\%) &  bartender (2.0\%) &  secretary (14.6\%)\\
4 &  cleaner (5.9\%) &  cleaner (3.3\%) &  nurse (1.1\%) &  cleaner (7.7\%)\\
5 &  secretary (5.8\%) &  secretary (2.7\%) & \textbf{clerk (0.6\%)} &  teacher (6.3\%)\\
6 &  bartender (3.0\%) &  bartender (2.0\%) & \textbf{server (0.3\%)} &  cook (3.2\%)\\
\bottomrule
\end{tabular}

\end{table*}

\begin{table*}[p]
    \centering
        \caption{Example predictions for \gpttwo\ and its gender \gradiend\ models. Predictions of the \gradiend\ models that were not retrieved by the base model as one of the top 10 results, are highlighted in \textbf{bold}.}
    \label{tab:example-predictions-gpt2}

\begin{tabular}{rllll}
    \toprule
    \small
    \textbf{Index} & \textbf{\gpttwo} & \, + \textbf{\gradiendbpi} & \, + \textbf{\gradiendfpi} & \, + \textbf{\gradiendmpi} \\
    \midrule
    \multicolumn{5}{c}{The woman worked as a [MASK]}
\\\midrule
 1 &  waitress (29.2\%) &  waitress (28.8\%) &  waitress (39.8\%) &  waitress (27.9\%)\\
2 &  maid (15.6\%) &  nurse (19.1\%) &  nurse (18.9\%) &  prostitute (18.5\%)\\
3 &  nurse (13.9\%) &  prostitute (18.6\%) &  maid (8.7\%) &  nurse (17.5\%)\\
4 &  reception (8.0\%) &  maid (9.8\%) &  reception (8.5\%) &  maid (10.8\%)\\
5 &  security (7.1\%) & \textbf{babys (5.2\%)} &  prostitute (6.4\%) &  bartender (5.0\%)\\
6 &  prostitute (6.2\%) & \textbf{model (4.6\%)} & \textbf{babys (4.2\%)} & \textbf{babys (4.5\%)}\\
\bottomrule    
\end{tabular}
\end{table*}

\begin{table*}[p]
    \centering
    \small
        \caption{Example predictions for \llama\ and its gender \gradiend\ models. Predictions of the \gradiend\ models that were not retrieved by the base model as one of the top 10 results, are highlighted in \textbf{bold}.}
    \label{tab:example-predictions-llama}

\begin{tabular}{rllll}
    \toprule
    \textbf{Index} & \textbf{\llama} & \, + \textbf{\gradiendbpi} & \, + \textbf{\gradiendfpi} & \, + \textbf{\gradiendmpi} \\
    \midrule
    \multicolumn{5}{c}{The woman worked as a [MASK]}
\\\midrule
 1 &  waitress (16.7\%) &  waitress (16.2\%) &  nurse (17.7\%) &  waitress (15.5\%)\\
2 &  nurse (16.4\%) &  nurse (16.2\%) &  waitress (14.8\%) &  nurse (15.4\%)\\
3 &  reception (12.2\%) &  teacher (13.0\%) &  cashier (13.7\%) &  teacher (11.7\%)\\
4 &  secretary (10.8\%) &  secretary (8.7\%) &  nanny (9.5\%) &  reception (10.5\%)\\
5 &  nanny (8.0\%) & \textbf{bartender (8.5\%)} & \textbf{caregiver (9.3\%)} &  secretary (10.2\%)\\
6 &  teacher (8.0\%) & \textbf{prostitute (8.3\%)} &  reception (8.2\%) & \textbf{prostitute (9.2\%)}\\
\bottomrule    
\end{tabular}
\end{table*}

\begin{table*}[p]
    \centering
    \small
        \caption{Example predictions for \llamai\ and its gender \gradiend\ models. Predictions of the \gradiend\ models that were not retrieved by the base model as one of the top 10 results, are highlighted in \textbf{bold}.}
    \label{tab:example-predictions-llamai}

\begin{tabular}{rllll}
    \toprule
    \textbf{Index} & \textbf{\llamai} & \, + \textbf{\gradiendbpi} & \, + \textbf{\gradiendfpi} & \, + \textbf{\gradiendmpi} \\
    \midrule
    \multicolumn{5}{c}{The woman worked as a [MASK]}
\\\midrule
 1 &  nurse (28.9\%) &  waitress (31.3\%) &  nurse (29.9\%) &  nurse (22.7\%)\\
2 &  waitress (24.0\%) &  nurse (19.2\%) &  waitress (24.0\%) &  waitress (16.7\%)\\
3 &  librarian (8.6\%) &  bartender (7.5\%) &  secretary (8.2\%) &  librarian (13.8\%)\\
4 &  secretary (8.2\%) & \textbf{model (7.4\%)} &  librarian (7.9\%) &  bartender (10.6\%)\\
5 &  reception (6.4\%) & \textbf{server (7.1\%)} &  freelance (5.8\%) &  secretary (10.5\%)\\
6 &  bartender (6.0\%) & \textbf{flight (6.4\%)} &  reception (5.6\%) &  reception (8.1\%)\\
\bottomrule
\end{tabular}
\end{table*}

\end{document}